\documentclass[phd,tocprelim]{cornell}
\let\ifpdf\relax
\usepackage{graphicx,pstricks}
\usepackage{graphics}
\usepackage{moreverb}
\usepackage{subcaption}
\usepackage{epsfig}
\usepackage{hangcaption}
\usepackage{xcolor}
\usepackage{geometry}

\usepackage{amsmath,amssymb}
\usepackage{algorithm,algorithmic}
\usepackage{arydshln}
\usepackage{palatino}
\usepackage{hyperref}
\usepackage[capitalise]{cleveref}
\hypersetup{
    colorlinks,
    linktoc=all,
    citecolor=black,
    filecolor=black,
    linkcolor=black,
    urlcolor=black
}
\usepackage{xspace}
\usepackage{multirow}
\usepackage[titletoc]{appendix}
\usepackage[nottoc,notlot,notlof]{tocbibind}
\usepackage{tikz}
\usetikzlibrary{calc,trees,positioning,arrows,chains,shapes.geometric,%
    decorations.pathreplacing,decorations.pathmorphing,shapes,%
    matrix,shapes.symbols}

\tikzset{suppress join/.code={\def\tikz@after@path{}}}

\tikzset{
>=stealth',
  punktchain/.style={
    rectangle, 
    rounded corners, 
    draw=black, thin,
    text width=8em, 
    minimum height=3em, 
    text centered,
    on chain},
  circlechain/.style={
    circle,  
    draw=black, thin,
    text centered, 
    minimum size=3em,
    inner sep=1pt,
    on chain},
  gen_circle/.style={
    circle,  
    draw=black, thin,
    text centered, 
    minimum size=3em,
    inner sep=1pt,
    on chain},
  line/.style={draw, thick, <-},
  side/.style={
    rectangle,
    minimum width=1em,
    draw=black, thin,
    text width=2em, 
    minimum height=2em, 
    text centered},
  sidecircle/.style={
    circle,  
    fill=green!10,
    draw=black, thin,
    text centered, 
    minimum size=0.1em,
    inner sep=1pt,
    on chain},
  every join/.style={->, thin,shorten >=1pt},
  decoration={brace},
  tuborg/.style={decorate},
  tubnode/.style={midway, right=2pt},
  bignode/.style={font={\fontsize{23}{25}\selectfont}},
  regularnode/.style={font={\fontsize{18}{25}\selectfont}},
  smallnode/.style={font={\fontsize{10}{12}\selectfont}},
}
\usepackage{soul}
\usepackage{CJKutf8}
\usepackage{tablefootnote}
\usepackage{titlesec}
\titlespacing{\paragraph}{0pt}{0.5\baselineskip}{0.5\baselineskip}

\usepackage{natbib}

\renewcommand{\caption}[1]{\singlespacing\hangcaption{#1}\normalspacing}

\newcommand{\kld}{\mathit{KL}}

\newcommand{\bert}{BERT\xspace}
\newcommand{\bertsize}[1]{BERT-#1\xspace} %
\newcommand{\roberta}{RoBERTa\xspace}
\newcommand{\xlnet}{XLNet\xspace}
\newcommand{\squad}{SQuAD\xspace}
\newcommand{\lambada}{LAMBADA\xspace}
\newcommand{\lpp}{LPP\xspace}
\newcommand{\skipthought}{Skip-thought\xspace}
\newcommand{\infersent}{InferSent\xspace}
\newcommand{\dissent}{DisSent\xspace}
\newcommand{\enteval}{EntEval\xspace}

\newcommand{\yago}{CoNLL-YAGO\xspace}
\newcommand{\contextrep}{contextualized entity representations\xspace}
\newcommand{\descrep}{descriptive entity representations\xspace}

\newcommand{\spos}{Sentence Position\xspace}
\newcommand{\bso}{Binary Sentence Ordering\xspace}
\newcommand{\dc}{Discourse Coherence\xspace}
\newcommand{\ssp}{Sentence Section Prediction\xspace}

\newcommand{\vmf}{\text{vMF}}
\DeclareMathOperator*{\argmax}{argmax}

\newcommand{\vgvae}{VGVAE\xspace}
\newcommand{\wordavg}{\textsc{Wordavg}\xspace}
\newcommand{\lstmavg}{\textsc{BLSTMavg}\xspace}
\newcommand{\prl}{PRL\xspace}
\newcommand{\spl}{DPL\xspace}
\newcommand{\wpl}{WPL\xspace}
\newcommand{\wn}{WN\xspace}
\newcommand{\lc}{LC\xspace}

\newcommand{\sted}{ST\xspace}

\newcommand{\decinit}{\textsc{init}\xspace}
\newcommand{\decswap}{\textsc{swap}\xspace}
\newcommand{\decconcat}{\textsc{concat}\xspace}

\newcommand{\wikitablet}{\textsc{WikiTableT}\xspace}
\newcommand{\wikidata}{Wikidata\xspace}
\newcommand{\wikibio}{\textsc{WikiBio}\xspace}
\newcommand{\entmax}{$\alpha$-entmax\xspace}

\newcommand{\wikinli}{\textsc{WikiNLI}\xspace}
\newcommand{\wikisent}{\textsc{WikiSentNLI}\xspace}
\newcommand{\wordnet}{WordNet\xspace}
\newcommand{\wikipedia}{Wikipedia\xspace}

\newcommand{\senteval}{SentEval\xspace}
\newcommand{\elmo}{ELMo\xspace}
\newcommand{\entelmo}{EntELMo\xspace}
\newcommand{\robertalarge}{RoBERTa-large\xspace}
\newcommand{\bertbase}{BERT-base\xspace}
\newcommand{\bertlarge}{BERT-large\xspace}

\newcommand{\disceval}{DiscoEval\xspace}
\newcommand{\sposshort}{SP\xspace}

\newcommand{\dcshort}{DC\xspace}

\newcommand{\bs}[1]{\boldsymbol{#1}}

\newcommand{\summscreen}{\textsc{SummScreen}\xspace}
\newcommand{\tvmegasite}{TVMegaSite\xspace}
\newcommand{\foreverdream}{ForeverDreaming\xspace}

\newcommand{\tvmegasiteshort}{TMS\xspace}
\newcommand{\foreverdreamshort}{FD\xspace}

\newcommand{\tvrecap}{\textsc{TVStoryGen}\xspace}
\newcommand{\tvrecapsum}{\textsc{TVStorySum}\xspace}
\newcommand{\storium}{STORIUM\xspace}
\newcommand{\writing}{WritingPrompts\xspace}
\newcommand{\tms}{TMS\xspace}
\newcommand{\fandom}{Fandom\xspace}
\newcommand{\fd}{FD\xspace}

\newcommand{\TC}[1]{\textcircled{\raisebox{-0.9pt}{#1}}}

\makeatletter
\renewcommand{\@chapapp}{}%
\newenvironment{chapquote}[2][2em]
  {\setlength{\@tempdima}{#1}%
   \def\chapquote@author{#2}%
   \parshape 1 \@tempdima \dimexpr\textwidth-2\@tempdima\relax%
   \itshape}
  {\par\normalfont\hfill--\ \chapquote@author\hspace*{\@tempdima}\par\bigskip}
\makeatother

\definecolor{atomictangerine}{rgb}{1.0, 0.6, 0.4}
\definecolor{babypink}{rgb}{0.96, 0.76, 0.76}
\definecolor{babyblueeyes}{rgb}{0.63, 0.79, 0.95}

\newsavebox\FrameBox
\newenvironment{Frame}{%
   \par\setbox\FrameBox\hbox\bgroup\minipage{0.8\textwidth}\parskip\baselineskip\ignorespaces
}{%
   \endminipage\egroup\fbox{\box\FrameBox}\par
}

\newenvironment{itemizesquish}{\begin{list}{\setcounter{enumi}{0}\labelitemi}{\setlength{\itemsep}{-0.25em}\setlength{\labelwidth}{0.5em}\setlength{\leftmargin}{\labelwidth}\addtolength{\leftmargin}{\labelsep}}}{\end{list}}

\newenvironment{enumeratesquish}{\begin{list}{\addtocounter{enumi}{1}\labelenumi}{\setlength{\itemsep}{0em}\setlength{\labelwidth}{0.5em}\setlength{\leftmargin}{\labelwidth}\addtolength{\leftmargin}{\labelsep}}}{\end{list}\setcounter{enumi}{0}}

\title {Leveraging Natural Supervision\\for Language Representation Learning and Generation}
\author {Mingda Chen}
\conferraldate {July}{2022}
\degreefield {Ph.D.}
\copyrightholder{Mingda Chen}
\copyrightyear{2022}

\AtBeginEnvironment{thebibliography}{\linespread{1}\selectfont}

\begin{document}

\cleardoublepage
\newgeometry{left=3cm, right=3cm, top=3cm, bottom=3cm}
\maketitle
\makecopyright
\pagenumbering{roman}
\begin{abstract}
Recent breakthroughs in Natural Language Processing (NLP) have been driven by language models trained on a massive amount of plain text. While powerful, deriving supervision from textual resources is still an open question. For example, language model pretraining often neglects the rich, freely-available structures in textual data. In this thesis, we describe three lines of work that seek to improve the training and evaluation of neural models using naturally-occurring supervision.

We first investigate self-supervised training losses to help enhance the performance of pretrained language models for various NLP tasks. Specifically, for general-purpose language representation learning, we alter the sentence prediction loss to make it better suited to other pretraining losses and more challenging to solve and show that the change led to a series of state-of-the-art pretrained encoders. For in-context learning, in contrast to previous work, which finetuned pretrained decoders on human-annotated datasets, we design an intermediate finetuning step that uses self-supervised training to promote models' ability in cross-task generalization.  

Then we describe methods to leverage the structures in Wikipedia and paraphrases. In particular, we leverage hyperlinks as supervision for pretraining entity representations, leading to models that can encode arbitrary entities. We use article structures, such as section and document titles, to train sentence representations. Evaluation results on discourse-related tasks show that such training helped model performance. We extract training data from article category graphs and demonstrate that the extracted data improves model performance on textual entailment tasks. In addition, we leverage the pair data structure in paraphrases by building the first neural models to disentangle semantics and syntax in sentence representations. In addition to semantic evaluation metrics, we propose metrics for syntactic representations, finding that the best performance for both metrics is achieved when there is maximal disentanglement between the two latent representations. We extend the framework for a novel generation task that controls the syntax of output text with a sentential exemplar. To formally define this controlled generation task, we annotate evaluation sets and proposed evaluation metrics.

Lastly, we discuss our work on tailoring textual resources for establishing challenging evaluation tasks. We introduce three datasets by defining novel tasks using various fan-contributed websites. The first dataset generates arbitrary Wikipedia section text from various tabular data by casting the task as long-form data-to-text generation and creating a large-scale dataset. The task is challenging as models need to generate a coherent passage connecting all the entities in the tabular data, and the story also needs to fit the background knowledge in the tabular data. The second dataset summarizes lengthy transcripts for TV shows. The task has several challenges: e.g., plot information is not stated explicitly but rather only implied in the dialogue and the need to draw information from a wide range of the input transcripts. As characters are fundamental to TV show plots, we also propose two character-centric evaluation metrics. The third dataset generates long-form stories from character descriptions and summaries. The task poses several challenges for story generation models, including lengthy inputs and outputs and consistency in character modeling.

\end{abstract}

\begin{acknowledgements}
\begin{chapquote}{Benjamin Disraeli}
Like all great travellers, I have seen more than I remember, and remember more than I have seen.
\end{chapquote}

The Ph.D. journey is an adventure mixed with daunting challenges, unanticipated bafflements, and instant delights. Many people have guided me through the challenges, clarified my confusion, and shared my happiness. I am enormously grateful for their help along the journey.

First, I want to thank my advisor Kevin Gimpel for the technical insight and research philosophy throughout these years. He was always knowledgeable about everything we worked on, meticulous about every word we wrote on papers, and patient about my mistakes. The work in this thesis would not be possible without his positive, steady influence.

I thank the rest of my thesis committee: Karen Livescu, Sam Wiseman, and Luke Zettlemoyer, for being generous with their time and insight. I also thank Karl Stratos for his guidance.

I thank my fellow students at Toyota Technological Institute at Chicago and the University of Chicago, especially Qingming Tang for the technical (and non-technical) conversations, Bumeng Zhuo for the fun activities during weekends, and Zewei Chu for the bike rides at the lakefront in Chicago. I am also grateful to my fellow interns and mentors at Google and Facebook.

Lastly, I would like to thank my family for inspiring my interest in learning, encouraging me to apply to graduate school, and being supportive and interested in listening to my research ramblings.

\end{acknowledgements}

\contentspage
\tablelistpage
\figurelistpage

\addtocontents{toc}{%
 \protect\vspace{1em}%
 \protect\noindent{\bfseries Chapters}\protect\par
 \protect\vspace{0.0em}%
}
\normalspacing \setcounter{page}{1} \pagenumbering{arabic}
\pagestyle{cornell} \addtolength{\parskip}{0.5\baselineskip}

\chapter{Introduction}

Written language is ubiquitous. Humans use writing to communicate ideas and store knowledge in their daily lives. These activities naturally produce traces of human intelligence, resulting in abundant, freely-available textual data: e.g., Wikipedia,\footnote{ \url{https://www.wikipedia.org/}, an online collaborative encyclopedia.} Reddit,\footnote{\url{https://www.reddit.com/}, an online forum for discussion and web content rating.} and Fandom,\footnote{\url{https://www.fandom.com/}, a fan-contributed encyclopedia of movies, films, and other media.} among others. These data often contain sophisticated knowledge expressed with complex language structures. For example, for encyclopedias, there usually are dedicated structures for connecting pieces of information scattered around different places for the convenience of readers (e.g., hyperlinks that point the same person or events mentioned in other documents to the same place for disambiguation). Aside from explicit structures, corpora have rich implicit structures. For example, the data pair in bilingual text shares the same semantic meaning but differs in syntactic forms. The implicit difference between the data pair allows us to disentangle the semantic and syntactic information implied in the data structure.

Despite the rich structures, recent advances in NLP have been driven by deep neural models trained on a massive amount of plain text, which often strips away the knowledge and structure from the input. This thesis research approaches to better drive supervision from various naturally-occurring textual resources.
In particular, we (1) improve ways of transforming plain text into training signals; (2) propose approaches to exploit the rich structures in Wikipedia and paraphrases; and (3) create evaluation benchmarks from fan-contributed websites to reflect real-world challenges. Below we briefly introduce these three areas and summarize our contributions.

\section{Overview}

\paragraph{Learning from Improved Self-Supervision}\hspace{-0.5em}(\cref{CHAPTER:SELFSUPERVISION}).\hspace{0.5em} Adapting plain text for training signals (also known as self-supervision) is the driving force behind recent breakthroughs in NLP. Approaches like BERT \citep{devlin-etal-2019-bert} and GPT-3 \citep{brown-etal-gpt3-nips} effectively transfer knowledge in plain text to various downstream NLP tasks. Recent research has demonstrated potential flaws in BERT's learning objectives \citep{NEURIPS2019_dc6a7e65,liu2019roberta} and has improved GPT-3's downstream task performance using human-annotated resources \citep{mishra2021crosstask,wei2021finetuned}. In this thesis, we present techniques to improve the self-supervised training objectives without requiring extra human annotations.

\paragraph{Learning from Rich Data Structures: Wikipedia Articles}\hspace{-0.5em}(\cref{CHAPTER:WIKIPEDIA}).\hspace{0.5em} Pretrained language models primarily use learning objectives related to word prediction based on nearby context \citep{Mikolov-word2vec-2013b,peters-etal-2018-deep,devlin-etal-2019-bert}. In this thesis, we present approaches to leverage the rich article structures in Wikipedia for learning vector representations of various texts.

\paragraph{Learning from Rich Data Structures: Paired Data}\hspace{-0.5em}(\cref{CHAPTER:DISENTANGLE}).\hspace{0.5em} Much of the recent NLP work on learning disentangled representations and controllable generation has focused on disentangling and controlling attributes such as sentiment \citep{hu17control,shen2017style} or formality \citep{ficler-goldberg-2017-controlling}. In this thesis, we show that leveraging paired data structures enables us to disentangle the semantics and syntax in sentence representations and control the syntax of output sentences using a sentential exemplar.

\paragraph{Building Evaluation Tasks from Textual Resources}\hspace{-0.5em}(\cref{CHAPTER:EVALUATION}).\hspace{0.5em} We construct various text generation datasets from fan-contributed websites. The rich information provided on these websites allows these new datasets to have different focuses (e.g., long-form text rather than single sentence generation) and domains (e.g., television series rather than news) compared to prior work in the same task setting. We show that their unique characteristics lead to challenging research questions. 

\section{Contributions}

In summary, this thesis makes the following contributions:
\begin{itemize}
    \item By adequately designing self-supervision, we improve the quality of pretrained language models and their abilities for cross-task generalization. In \cref{sec:sentence-order-prediction}, we replace the next sentence prediction loss with a novel sentence ordering prediction loss in language model pretraining and show that the change led to a series of state-of-the-art pretrained encoders. In \cref{sec:incontext-learning}, in contrast to previous work, which finetuned pretrained decoders on human-annotated datasets, we show that self-supervised tasks with proper designs could also lead to similar gains in the in-context few-shot learning setting, promoting models' ability in cross-task generalization.
    \item We design model architectures and training objectives to exploit the rich structures in Wikipedia articles. In \cref{sec:wikipedia-entity-representations}, we leverage hyperlinks as supervision for pretraining entity representations, leading to models that can encode arbitrary entities. In \cref{sec:wikipedia-discourse-sentence-representations}, we use article structures, such as section and document titles, to train sentence representations. Evaluation results on discourse-related tasks show that such training helped model performance. In \cref{sec:wikipedia-concept-hierarchies}, we extract training data from article category graphs and demonstrate that the extracted data improves model performance on textual entailment tasks. These results reveal the advantages of structure-aware model pretraining.
    \item We leverage the pair data structure in paraphrases and bilingual text to disentangle semantics and syntax in sentence representations, which allows us to learn interpretable and controllable neural models. In \cref{section:vgvae-representation}, we build the first neural models to disentangle semantics and syntax in sentence representations. The models use the fact that for a paraphrase pair, the semantics is shared, but syntax varies. In addition to semantic evaluation metrics, we propose evaluation metrics for syntactic representations, finding that the best performance for both metrics is achieved when there is maximal disentanglement between the two latent representations. In \cref{section:vgvae-generation}, we adapt this framework for controlled paraphrasing, where we seek to control the output text with a syntactic, sentential exemplar. To formally define this controlled generation task, we annotate evaluation sets and proposed evaluation metrics. In a later work, we extend this framework and task setting to machine translation \citep{chen2020exemplar}, showing the potential that this idea could generalize to arbitrary data with the pair data structure.
    \item We demonstrate that we can create challenging benchmark datasets for various long-form text generation tasks by tailoring fan-contributed textual resources. We do so by defining new NLP tasks and studying these new tasks through extensive experiments. In \cref{sec:wikitablet}, we generate arbitrary Wikipedia section text from various tabular data by casting the task as long-form data-to-text generation and creating a large-scale dataset. The task is challenging as models need to generate a coherent passage connecting all the entities in the tabular data, and the story also needs to fit the background knowledge in the tabular data. In \cref{sec:summscreen}, we summarize lengthy transcripts for TV shows. The task has several challenges: e.g., plot information is not stated explicitly but rather only implied in the dialogue and the need to draw information from a wide range of the input transcripts. As characters are fundamental to TV show plots, we also propose two character-centric evaluation metrics. In \cref{sec:tvstorygen}, we generate long-form stories from character descriptions and summaries. The task poses several challenges for story generation models, including lengthy inputs and outputs and consistency in character modeling.
\end{itemize}

\chapter{Background}

In this chapter, we present background on self-supervised pretraining, naturally-occurring data structures, and variational models in the context of NLP.

\section{Self-Supervised Language Pretraining}

Self-supervised learning in NLP seeks to adapt plain text (without using extra information obtained via human annotations, e.g., labels, regardless of whether these annotations are naturally-occurring or not) for training models. Common training objectives in this area are word prediction based on nearby context, with language modeling being the dominant approach. The resulting models can either produce vector representations for words/sentences/documents or directly generate text. In practice, models typically ``pretrain'' on a massive amount of unlabeled textual data using the self-supervised learning objectives before being applied to downstream tasks.\footnote{Researchers have also found that using the self-supervised learning objectives on downstream tasks helps model performance on these tasks \citep[e.g.,][]{howard-ruder-2018-universal,gururangan-etal-2020-dont}.} As the models can transfer the knowledge learned during pretraining to downstream tasks and thus improve model performance, pretraining has gained increasing attention in recent years. Below we briefly review the advancement of this research area.

\subsection{Related Work}

\paragraph{Word Representations.} Learning vector representations of words builds upon the distributional hypothesis \citep{harris-1954}: \textit{You shall know a word by the company it keeps} \citep{firth1957synopsis}. Based on this hypothesis, early methods attempt to learn a fixed set of vectors for word representations. Before the advent of neural models, researchers mostly used corpus-based cooccurrence statistics \citep{deerwester1990indexing,hinrich-1992-nips,brown1992class,lund1996producing}. These word vectors have been found to be helpful for intrinsic evaluation tasks (e.g., word similarities \citep{rubenstein1965contextual,miller1991contextual} and analogies \citep{turney2005corpus,turney-2006-similarity}) and for various NLP applications as features (e.g., named entity recognition \citep{miller-etal-2004-name} and semantic role labeling \citep{erk-2007-simple}).

With neural models, most training objectives shifted towards word predictions based on a limited context window from the corpus-level statistical information \citep{neural-prob-bengio-2003-jmlr,collobert:icml08,collobert-etal-natural-2011}. At this point, the word representations are typically within 100 dimensions and learned from corpora of a few million word tokens. Word2vec \citep{Mikolov-word2vec-2013} shows that scaling the size of training corpus to billions of word tokens and enlarging the dimensionality of word representations  at the same time improve the quality of word representations. Later, GloVe embedding \citep{pennington-etal-2014-glove} combines the corpus-based methods and the context window methods. Fasttext \citep{bojanowski-etal-2017-enriching} replaces word types in Word2vec with character n-grams. While considerable research interests continue to evaluate these vectors on traditional word-level tasks (e.g., word similarities), they became increasingly popular in getting applied to downstream tasks. The common practice became to use pretrained word vectors to initialize the word embedding layer in neural models (e.g., long short-term memory network \citep{hochreiter1997long}) with optional gradient updates on the embedding layer \citep[\emph{inter alia}]{collobert:icml08,turian-etal-2010-word}. 

\paragraph{Sentence Representations.} One drawback of pretrained word vectors is that it represents each word in isolation, leading to challenges in designing higher-level (i.e., sentences\footnote{While there is work that studies phrases (e.g., two- \citep{mitchell2010composition} and multiple-word \citep{Mikolov-word2vec-2013b} phrases) and documents (e.g., vector space models in information retrieval \citep{salton1971smart,salton-etal-1975-vsm}), we focus on sentences as it attracts the most attention in NLP.}) representations that capture compositionality among low-level (i.e., words) units. One way to obtain sentence representations is deriving them from word vectors. In particular, researchers have using tried parse trees \citep{socher-etal-2013-recursive} and simple pooling operations (e.g., averaging) \citep{iyyer-etal-2015-deep,shen-etal-2018-baseline}. The effectiveness of the latter approach raises concerns that the neural models on top of the word embedding layers do not encode meaningful compositional semantics of words.

Pretrained word representations also inspired a series of research on pretraining sentence encoders. In contrast to most work on word representations which can only represent seen words in training data, sentence representations are neural models that encode any sentence into a fixed-length vector. In particular, Paragraph Vector \citep{le-etal-distributed-14} averages or concatenates the words in input text and it is trained to predict next words. \skipthought \citep{kiros-etal-skipthoughts-2015} is a gated recurrent neural network (GRU; \citealp{cho-etal-2014-properties,chung2014empirical}) that is trained to predict the previous sentence and next sentence given the input one. \citet{dai-le-semi-2015} use a LSTM-based sequence-to-sequence \citep{ilya-etal-2014-seq2seq} autoencoder. FastSent \citep{hill-etal-2016-learning} follows \skipthought but simplifies the GRU encoder to be summing of the input word vectors. Sentence encoder pretraining also involves discriminative approaches, i.e., classifying input text into a small amount of categories. For example, \citet{kenter-etal-2016-siamese} use averaged word embeddings and train their model to classify whether the two input sentence embeddings are from the adjacent sentences. \citet{jernite2017discourse} train a GRU encoder for predicting discourse-related classification objectives. \citet{logeswaran2018an} also use a GRU encoder to predict the correct next sentence given a few candidate sentences.

\paragraph{Contextualized Word Representations.} Most word representations have one-to-one correspondence between words and vectors, neglecting the fact that words have different senses depending on the context in which they are situated. This observation has led to research that extends word representations to a ``contextualized'' version \citep{kawakami2015learning,melamud-etal-2016-context2vec,peters-etal-2017-semi,tu-etal-2017-learning,mccann-etal-2017-cove,peters-etal-2018-deep}, where the outcomes of these methods are deep neural models that encode a word and its surrounding context. Later, the invention of Transformer architecture \citep{attention_is_all_you_need} led to a series of enormous pretrained models in this direction \citep{radford2018improving,devlin-etal-2019-bert,liu2019roberta,conneau-etal-2020-unsupervised}, which was the drive behind recent breakthroughs in NLP. The training of these models generally involves recovering the words replaced by a special symbol, e.g., \texttt{[MASK]}, in the input (also known as ``masked word prediction'' or ``masked language modeling'') or bidirectional language modeling (i.e., have a separate forward and a backward language models at the same time). Other researchers have proposed alternative training objectives, such as training a discriminator to predict whether each token in a corrupted input was replaced by a sample from a generator model \citep{Clark2020ELECTRA:}.

With the contextualized representations, approaches of applying them to downstream tasks have also shifted towards the ``pretrain-then-finetune'' paradigm from the paradigm where pretrained parameters are often frozen during supervised training for downstream tasks \citep{howard-ruder-2018-universal,radford2018improving,devlin-etal-2019-bert}. The new paradigm initializes the majority parameters of downstream models using pretrained contextualized representations and finetunes the entire neural models using training data for downstream tasks.

Similar to traditional word representations, researchers seek to build sentence representations from contextualized word representations using operations like pooling, e.g., \citep{reimers-gurevych-2019-sentence}. On the other hand, the fact that contextualized word representations encode sentential context makes sentence and word representations barely distinguishable. For example, BERT \citep{devlin-etal-2019-bert} has a special symbol \texttt{[CLS]} prepended to input sequences and treats the contextualized word representations of the \texttt{[CLS]} symbol as the sentence representation. 

Another relevant research direction looks to convert pretrained language models into a sequence-to-sequence pretrained models for text generations. For example, \citet{ramachandran-etal-2017-unsupervised} and \citet{rothe-etal-2020-leveraging} initialize the weights of the encoder and decoder in a sequence-to-sequence architecture using pretrained language models and then finetune the whole model on generation tasks.

\paragraph{Pretrained Generation Models.} The vector representations for words and sentences described earlier are primarily used for discriminative tasks or as features for various downstream tasks. In contrast, in addition to the discriminative tasks, pretrained generation models look to train models that can generate text. In particular, GPT-2 \citep{radford2019language} shows that unidirectional language modeling can solve several NLP tasks in a zero-shot and one-shot fashion by unifying different tasks into language modeling using prompts. GPT-3 \citep{brown-etal-gpt3-nips} follows GPT-2 but uses larger models and shows impressive one-the-fly few-shot performance (also known as ``in-context few-shot learning'') for a diverse set of tasks. XLM \citep{alexis-lample-2019-xlm} uses masked language modeling for both monolingual and crosslingual data. MASS \citep{pmlr-v97-song19d}, BART \citep{lewis-etal-2020-bart}, and mBART \citep{liu-etal-2020-multilingual-denoising} are sequence-to-sequence denoising autoencoders. T5 \citep{colin-t5-jmlr} and its multilingual counterpart mT5 \citep{xue-etal-2021-mt5} convert any NLP tasks into a text-to-text format. Pegasus \citep{pegasus-zhang20ae} uses gap sentence generation for abstractive summarization. MARGE \citep{lewis-etal-2020-nips} uses a sequence-to-sequence framework and learns to reconstruct target text by retrieving a set of related texts.

\paragraph{Benchmarks for Pretrained Models.} Pretrained contextualized word representations and generative models, especially the ones built on the transformer architectures, have shown superhuman performance on various NLP benchmarks, such as GLUE \citep{wang-etal-2018-glue} and SuperGLUE \citep{alex-etal-2019-nips}, both of which evaluate natural language understanding capabilities. Apart from GLUE and SuperGLUE, there are many other benchmarks that have been proposed to test pretrained models. Particularly, \citet{adi2017fine} use synthetic tasks, such as predicting sentence length, to measure the quality of sentence representations. SentEval \citep{conneau-kiela-2018-senteval} evaluates quality of sentence representations using several human-annotated datasets. decaNLP \citep{mccann2018natural} tries to cover diverse tasks and metrics. Similar to decaNLP, GEM \citep{gehrmann-etal-2021-gem} is a benchmark focusing on text generation. Dynabench \citep{kiela-etal-2021-dynabench} uses human-and-model-in-the-loop for dataset creation. KILT \citep{petroni-etal-2021-kilt} evaluates models that condition on specific information in large textual resources. There are also benchmarks measuring the amount of biases in the pretrained models \citep{nangia-etal-2020-crows,nadeem-etal-2021-stereoset}.

\subsection{Formal Preliminaries}

We provide a brief description of \skipthought, ELMo, and BERT necessary for the remainder of the thesis. We denote the input sequence by $x_{1:T}$ where $T$ is the maximum sequence length.

\paragraph{Skip-Thought.} \skipthought uses GRU encoders to process the word sequence and uses the encoded representations to generate the previous and next sentences. The framework is inspired by the skip-gram formulation of the Word2Vec model. Formally, the training objective is
\begin{equation}
\begin{aligned}
\text{Loss}_\text{ST}=&-\log p(y_{1:T_y}\vert x_{1:T};\boldsymbol{\theta}_y,\boldsymbol{\theta}_x)-\log p_\phi(z_{1:T_z}\vert x_{1:T};\boldsymbol{\theta}_z,\boldsymbol{\theta}_x)\\
=&-\sum_{t=1}^{T_y}\log p(y_t\vert x_{1:T},y_{1:i<t};\boldsymbol{\theta}_y,\boldsymbol{\theta}_x) - \sum_{t=1}^{T_z}\log p(z_t\vert x_{1:T}, z_{1:i<t};\boldsymbol{\theta}_z,\boldsymbol{\theta}_x)
\end{aligned}
\label{background-eq:skipthought-objective}
\end{equation}
\noindent where $y_{1:T_y}$ and $z_{1:T_z}$ are the previous and next sentences of the input sequence respectively. $\boldsymbol{\theta}_y$ and $\boldsymbol{\theta}_z$ are the parameters for the GRU decoders used to handle the previous and next sentences respectively. $\boldsymbol{\theta}_x$ is the parameters for the GRU encoder. Once trained, the encoded representation at position $T$ is used as sentence representations. 
\begin{figure}
    \centering
    \includegraphics[scale=0.6]{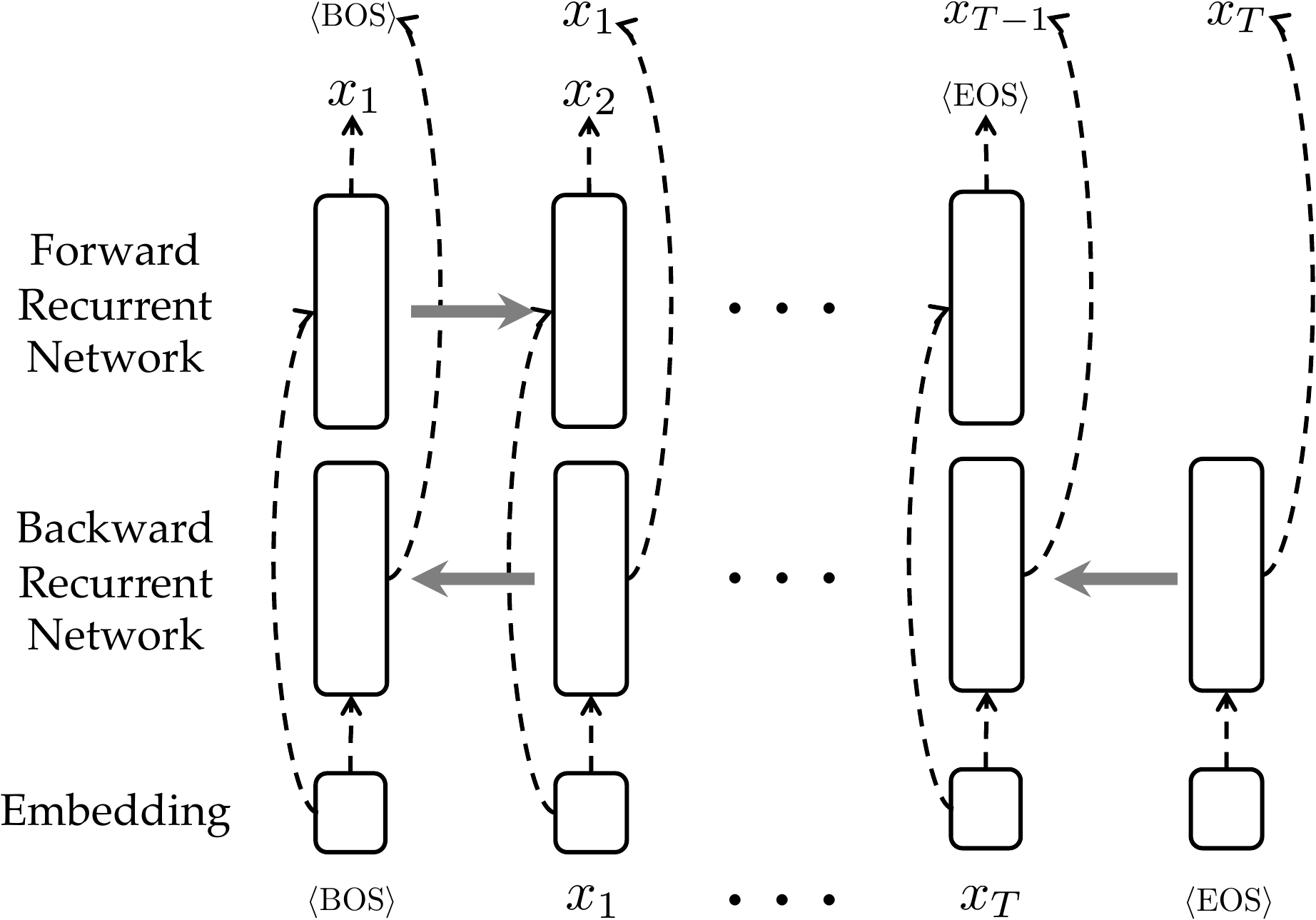}
    \caption{Diagram showing recurrent neural network based bidirectional language models. $x_{1:T}$ represents the input sequence. $\langle\text{BOS}\rangle$ and $\langle\text{EOS}\rangle$ are special symbols for the start-of-sequence and end-of-sequence tokens, respectively.}
    \label{background-fig:bi-rnn-lm}
\end{figure}

\paragraph{ELMo.} ELMo is an LSTM based bidirectional language model (BiLM). While details vary, LSTM-based BiLMs are typically formed by forward and backward language models, which are parameterized by different LSTMs and only share a limited number of parameters. An example of bidirectional language models is shown in \cref{background-fig:bi-rnn-lm} where the input word embeddings are shared between two recurrent neural networks. ELMo uses character embeddings. Formally, the training objective of ELMo is
\begin{equation}
\begin{aligned}
\text{Loss}_\text{ELMo}=&-\sum_{t=1}^{T}(\log p(x_t\vert x_{i<t};\overrightarrow{\boldsymbol{\theta}_\text{LSTM}})+\log p(x_t\vert x_{i>t};\overleftarrow{\boldsymbol{\theta}_\text{LSTM}}))
\end{aligned}
\label{background-eq:elmo-loss}
\end{equation}
\noindent where $\overrightarrow{\boldsymbol{\theta}_\text{LSTM}}$ and $\overleftarrow{\boldsymbol{\theta}_\text{LSTM}}$ are parameters for the forward and backward language models respectively.

\begin{figure}
    \centering
    \includegraphics[scale=0.6]{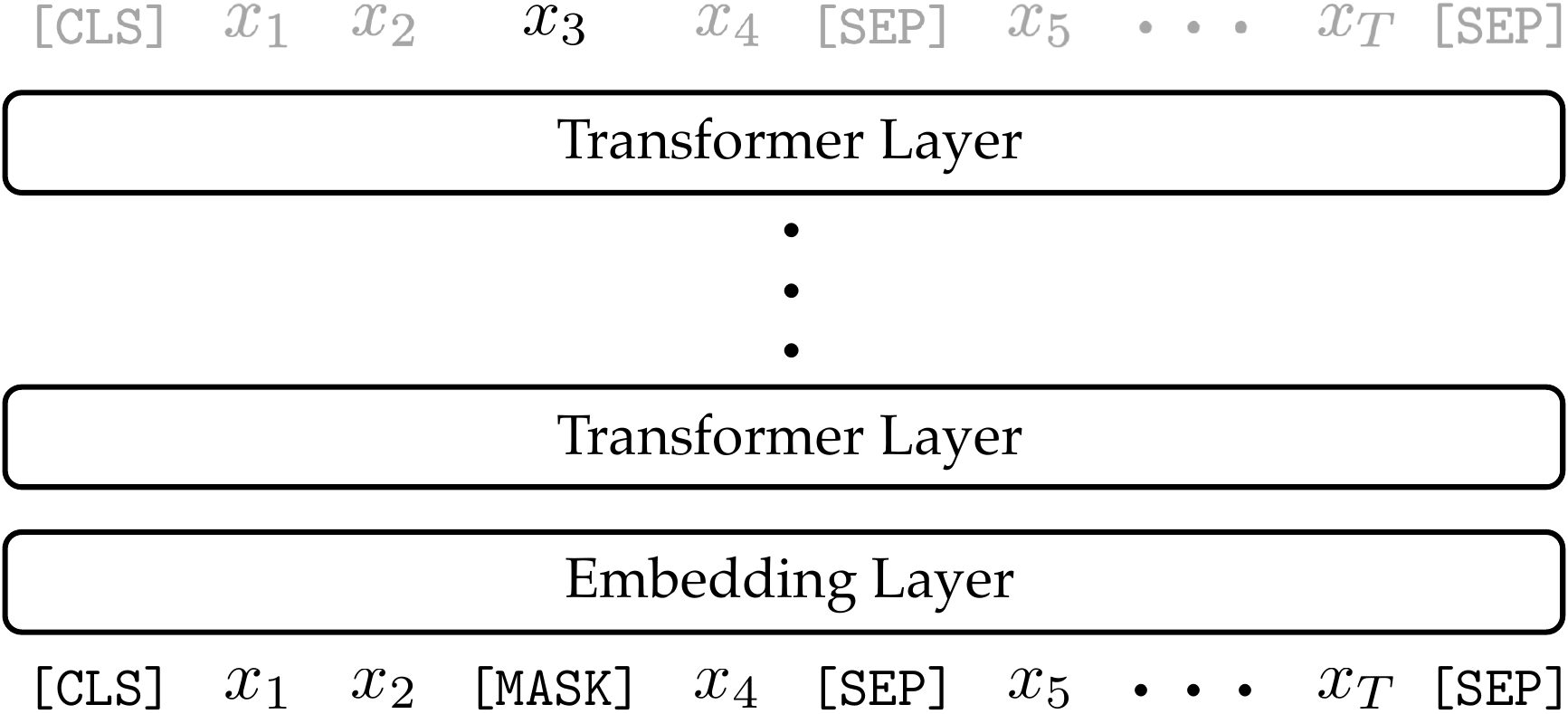}
    \caption{Diagram showing masked language modeling (MLM) with BERT. $x_{1:T}$ represents the input sequence. \texttt{[MASK]} is the special symbol indicating the units that are masked out in the input sequence (typically the number of masked units can be up to 15\% of the input sequence). \texttt{[CLS]} and \texttt{[SEP]} are two symbols used in next sentence prediction (NSP) task. We include the two symbols in this figure because MLM and NSP are used jointly during BERT training. In the shown case, models are required to predict $x_3$. We grey out the output units that are not used when computing MLM losses.}
    \label{background-fig:masked-lm}
\end{figure}

\paragraph{BERT.} BERT is a transformer encoder trained with two losses: masked language modeling (MLM) and next sentence prediction (NSP). For NSP, the input sequence is first split into two parts, one of which has a 50\% chance of being replaced by a text segment sampled from another document. NSP is a binary classification task that asks models to predict whether the input sequence contains text segments from different documents or not. In practice, BERT uses byte-pair encodings \citep{sennrich-etal-2016-neural}. When formatting input sequences, BERT prepends a special symbol \texttt{[CLS]} to the input and adds a special symbol to the end of the input. When trained with NSP, BERT concatenates the two input segments using a \texttt{[SEP]}. To differentiate the two text segments, BERT also adds segment embeddings to input word embeddings. BERT uses the vector representation of \texttt{[CLS]} to make next sentence predictions with the goal of encouraging models to encode sentential information into the \texttt{[CLS]} representation. Formally, the NSP training objective is
\begin{equation}
\begin{aligned}
\text{Loss}_\text{NSP}=&-\log p(y\vert\texttt{[CLS]},x_1,x_2,\ldots,\texttt{[SEP]},\ldots,x_T; \boldsymbol{\theta}_\text{BERT}, \boldsymbol{\theta}_\text{NSP})
\end{aligned}
\label{background-eq:nsp-loss}
\end{equation}
\noindent where $y$ is the ground-truth label. $\boldsymbol{\theta}_\text{MLM}$, $\boldsymbol{\theta}_\text{NSP}$ represent parameters of BERT and its classifier for NSP, respectively.

Rather than processing input sequences monotonically like recurrent neural networks, transformer architectures allow BERT to attend to left and right context simultaneously for each position in the input sequence. MLM takes advantage of this feature by a cloze-style formulation that randomly replaces input units with a special \texttt{[MASK]} symbol and asks models to predict the masked units (see \cref{background-fig:masked-lm} for an example). Formally, the MLM training objective is
\begin{equation}
\begin{aligned}
\text{Loss}_\text{MLM}=&-\sum_{x_k\in\mathcal{X}}\log p(x_k\vert \text{mask}(\texttt{[CLS]},x_1,x_2,\ldots,\texttt{[SEP]},\ldots,x_T); \boldsymbol{\theta}_\text{BERT},\boldsymbol{\theta}_\text{MLM})
\end{aligned}
\label{background-eq:masked-lm-loss}
\end{equation}
\noindent where $\mathcal{X}$ is a set of units that are masked out in the input sequence and $\text{mask}(\cdot)$ is a function that returns a masked sequence given an input sequence. $\boldsymbol{\theta}_\text{MLM}$ represents parameters of the classifier for MLM. In practice, BERT primarily uses parameters of its embedding matrix for $\boldsymbol{\theta}_\text{MLM}$.

The final training loss for BERT is
\begin{equation}
    \text{Loss}_\text{BERT} = \text{Loss}_\text{MLM} + \text{Loss}_\text{NSP}
\end{equation}
\section{Naturally-Occurring Data Structures}

There are rich structures in textual data beyond plain text: e.g., conversational structures in internet forums and document structures in an online encyclopedia. These structures naturally emerge in people's daily lives and possess knowledge that is unlikely to be captured by strings of words and has the potential to transfer to various downstream tasks. Leveraging such structures has a long history in a range of NLP tasks. A thorough review of these works is beyond the scope of the thesis. Below we seek to cover resources relevant to the remainder of the thesis (\cref{subsec:natural-structure-related-work}) and a few other resources that have attracted increasing attention in recent years (\cref{subsec:natural-structure-extra-related-work}).

\subsection{Related Work}
\label{subsec:natural-structure-related-work}

\paragraph{Bilingual Text (Bitext).} Bitext is comprised of parallel corpora. Recent large-scale datasets are mostly mined from the web \citep{resnik-1999-mining,resnik-smith-2003-web}. One popular data resource is official documents on government websites: e.g., the European Parliament \citep{koehn-2005-europarl} and the United Nations \citep{rafalovitch-dale-2009-united,eisele-chen-2010-multiun,chen-eisele-2012-multiun,ziemski-etal-2016-united}. Other resources involve parallel multilingual subtitles for movies and television series \citep{lison-tiedemann-2016-opensubtitles2016} and transcripts for TED talks \citep{qi-etal-2018-pre}. \citet{smith-etal-2013-dirt} and \citet{el-kishky-etal-2020-ccaligned} exploit URLs in the Common Crawl corpus\footnote{The Common Crawl corpus is a freely-available
crawl of the web.} to identify parallel text
pairs. There have also been attempts to automatically extracts parallel sentences from the content of multilingual Wikipedia articles \citep{smith-etal-2010-extracting,schwenk-etal-2021-wikimatrix}. Others have created datasets by mixing various data resources \citep{tiedemann-2012-parallel,czeng16:2016}.

Bitext itself is a crucial training resource for machine translation systems \citep{brown-etal-1990-statistical,brants-etal-2007-large,wu2016google}. Outside of machine translation, bitext has been used for learning traditional word representations \citep{wang-etal-1996-icslp,och-1999-efficient,faruqui-dyer-2014-improving,lu-etal-2015-deep}, contextualized word representations \citep{kawakami2015learning,mccann-etal-2017-cove}, sentence representations \citep{hill-etal-2016-learning,espana2017empirical,gregoire-langlais-2018-extracting,guo-etal-2018-effective,wieting-etal-2019-simple} and paraphrase generation models \citep{barzilay-mckeown-2001-extracting,bannard-callison-burch-2005-paraphrasing,mallinson-etal-2017-paraphrasing,wieting-gimpel-2018-paranmt,hu-aaai-2019-parabank}.

\paragraph{Wikipedia.} Wikipedia is comprised of documents with rich metadata, which can be used as naturally-occurring supervision for a variety of NLP tasks. One example is hyperlinks, which have been used for parsing \citep{spitkovsky-etal-2010-profiting,sogaard-2017-using,shi-etal-2021-learning}, named entity recognition \citep{kazama-torisawa-2007-exploiting,nothman-etal-2008-transforming,richman-schone-2008-mining,ghaddar-langlais-2017-winer}, entity disambiguation and linking \citep{bunescu-pasca-2006-using,cucerzan-2007-large,mihalcea-2007-using,mihalcea2007wikify,milne2008learning,hoffart-etal-2011-robust,le-titov-2019-boosting}, coreference resolution \citep{rahman-ng-2011-coreference,singh2012wikilinks,zheng-etal-2013-dynamic,eirew-etal-2021-wec}, and generating Wikipedia articles \citep{j.2018generating}.

Wikipedia document categories have been used for text classification \citep{gantner-schmidt-thieme-2009-automatic,chu2021natcat,chu-etal-2021-unsupervised}, semantic parsing \citep{choi-etal-2015-scalable}, and word similarities \citep{strube-ponzetto-2006-aaai}. Besides particular tasks, there is work that attempts to study the Wikipedia categories from a non-empirical perspective. \citet{zesch-gurevych-2007-analysis} analyze the differences between the graphs from WordNet \citep{wordnet1998} and the ones from Wikipedia categories. \citet{ponzetto-strube-2007-aaai} and \citet{nastase-strube-2008-aaai} extract knowledge of entities from the Wikipedia category graphs using predefined rules. \citet{nastase-etal-2010-wikinet} build a dataset based on Wikipedia article 
or category titles as well as the relations between categories and pages.

Wikipedia edit histories have been used for sentence compression \citep{yamangil-nelken-2008-mining,yatskar-etal-2010-sake}, writing assistants \citep{zesch-2012-measuring,cahill-etal-2013-robust,grundkiewicz2014wiked,boyd-2018-using}, paraphrasing \citep{max-wisniewski-2010-mining}, splitting and rephrasing \citep{botha-etal-2018-learning}, studying atomic edits \citep{faruqui-etal-2018-wikiatomicedits}, and modeling editors' behaviors \citep{jaidka-etal-2021-wikitalkedit}.

Additionally, by aligning sentences in Wikipedia to those in simple Wikipedia, Wikipedia has been used for text simplification \citep{zhu-etal-2010-monolingual} and learning sentence representations \citep{wieting-gimpel-2017-revisiting}. Through pairing Wikipedia with structured information (e.g., knowledge bases, such as Wikidata \citep{wikidata2014} and WordNet, or infoboxes on Wikipedia pages), researchers have created datasets for question answering \citep{hewlett-etal-2016-wikireading}, constructing knowledge graphs \citep{fabian2007yago,hoffart-etal-2013-yago2,safavi-koutra-2020-codex,wang-etal-2021-kepler}, table parsing \citep{herzig-etal-2020-tapas,yin-etal-2020-tabert}, and data-to-text generation \citep{lebret-etal-2016-neural,bao-etal-2018-aaai,jin-etal-2020-genwiki,agarwal-etal-2021-knowledge,wang-etal-2021-wikigraphs}.

\paragraph{Fandom.} Fandom has rich information for individual wiki items (e.g., episodes for television series). Similar to Wikipedia, wiki items on Fandom have consistent article structures and comprehensive information contributed by fans. However, unlike Wikipedia, it hosts wikis mainly on entertainment. Due to these characteristics, Fandom has been used for text summarization \citep{yu-etal-2016-unsupervised}, dialogue summarization \citep{rameshkumar-bailey-2020-storytelling}, paraphrase extraction \citep{regneri-wang-2012-using}, constructing sensorial lexicon \citep{tekiroglu-etal-2014-sensicon}, question answering \citep{maqsud-etal-2014-nerdle}, character description summarization \citep{shi-etal-2021-descgen}, entity linking \citep{logeswaran-etal-2019-zero}, and knowledge graph construction \citep{chu2021knowfi}.

\subsection{Additional Related Work}
\label{subsec:natural-structure-extra-related-work}

Researchers build text classification datasets from movie reviews and news articles \citep{maas-etal-2011-learning,NIPS2015_250cf8b5}. \citet{lan-etal-2017-continuously} create a paraphrase dataset by linking tweets through shared URLs. \citet{volske-etal-2017-tl} create a summarization dataset by taking advantage of the common practice of appending a "TL;DR" to long posts. \citet{fan-etal-2018-hierarchical} build a story generation dataset using the subreddit  \texttt{r/WritingPrompt}.
\citet{khodak-etal-2018-large} create a dataset for sarcasm detection by leveraging the fact that reddit users tend to add the marker ``/s'' to the end of sarcastic statements. \citet{yang-etal-2018-learning} learn sentence embeddings from Reddit using its conversational structures. \citet{joshi-etal-2017-triviaqa} construct question-answer pairs from 14 trivia and quiz-league websites. \citet{fan-etal-2019-eli5} build a question answering dataset with long-form answers using the subreddit \texttt{r/explainlikeimfive}. \citet{chakrabarty-etal-2019-imho} mine the acronyms IMO/IMHO (in my (humble) opinion) for claim detection.  \citet{zhang-etal-2020-dialogpt} train a GPT-like model on conversations from Reddit for dialogue generation. \citet{iyer-etal-2018-mapping} and \citet{agashe-etal-2019-juice} use GitHub to construct datasets for code generation. Other researchers have adapted Stack Overflow for question answering \citep{dhingra2017quasar}, question clarification \citep{rao-daume-iii-2018-learning}, semantic parsing \citep{ye-etal-2020-sketch}, and source code summarization \citep{iyer-etal-2016-summarizing} datasets. In addition to the material presented in this thesis, I have also contributed a dataset for ill-formed question rewriting based on Stack Exchange question edit histories \citep{Chu_Chen_Chen_Wang_Gimpel_Faruqui_Si_2020}.

\section{Sentence Variational Autoencoder}

Variational autoencoders are models that combine latent-variable modeling and neural networks. As a type of variational autoencoders, sentence variational autoencoders (SVAEs) seek to model a sentence through latent variables. The use of latent variables allows for design choices that reflect certain aspects of a sentence: e.g., sentiment and formality, leading to controllable and interpretable neural architectures. Below we briefly review the related work.

\subsection{Related Work}
SVAEs with a single latent variable often assumes that the latent variable follows Gaussian distributions, where researchers primarily focus on improving the SVAEs capabilities in language modeling (as measured by perplexities) and generating from interpolated samples from latent distributions   \citep{bowman-etal-2016-generating}. Relevant work studies the effect of different latent distributions, including von Mises-Fisher distributions \citep{xu-durrett-2018-spherical}, Gaussian mixtures \citep{ding-gimpel-2021-flowprior}, Gumbel-softmax distributions \citep{zhou-neubig-2017-multi}, discrete variables \citep{wiseman-etal-2018-learning,NEURIPS2018_b691334c,pmlr-v97-ziegler19a,pmlr-v119-stratos20a,jin-etal-2020-discrete}, piecewise constant distributions \citep{serban-etal-2017-piecewise}, sample-based \citep{fang-etal-2019-implicit} and flow-based distributions \citep{ma-etal-2019-flowseq,setiawan-etal-2020-variational,ding-gimpel-2021-flowprior}.

Apart from language modeling, researchers have used SVAEs for machine translations \citep{zhang-etal-2016-variational-neural,schulz-etal-2018-stochastic,eikema-aziz-2019-auto,ma-etal-2019-flowseq,calixto-etal-2019-latent,setiawan-etal-2020-variational}, relation extraction \citep{marcheggiani-titov-2016-discrete}, dependency parsing \citep{corro2018differentiable}, word representations \citep{rios-etal-2018-deep}, dialogue generation \citep{shen-etal-2017-conditional,bahuleyan-etal-2018-variational,Shen_Su_Niu_Demberg_2018}, text classification \citep{ding-gimpel-2019-latent,gururangan-etal-2019-variational,jin-etal-2020-discrete}, and question generation \citep{bahuleyan-etal-2018-variational}. Increasing attention has also been devoted to interpretable representation learning \citep{bastings-etal-2019-interpretable,zhou-etal-2020-interpretable,goyal-etal-2020-probabilistic,cheng-etal-2020-improving,mercatali-freitas-2021-disentangling-generative,vishnubhotla-etal-2021-evaluation} and controllable generation (e.g., controling sentiment \citep{hu17control,shen2017style} or formality \citep{ficler-goldberg-2017-controlling}).

\subsection{Formal Preliminaries}
\label{background-section:vae-formal-prelim}

Below we derive the training objectives for sentence variational autoencoders using Gaussian and vMF distributions. These materials will be used in \cref{section:vgvae-representation} and \cref{section:vgvae-generation}.

\paragraph{Variational Autoencoders.}
\begin{figure}
    \centering
    \includegraphics[scale=0.35]{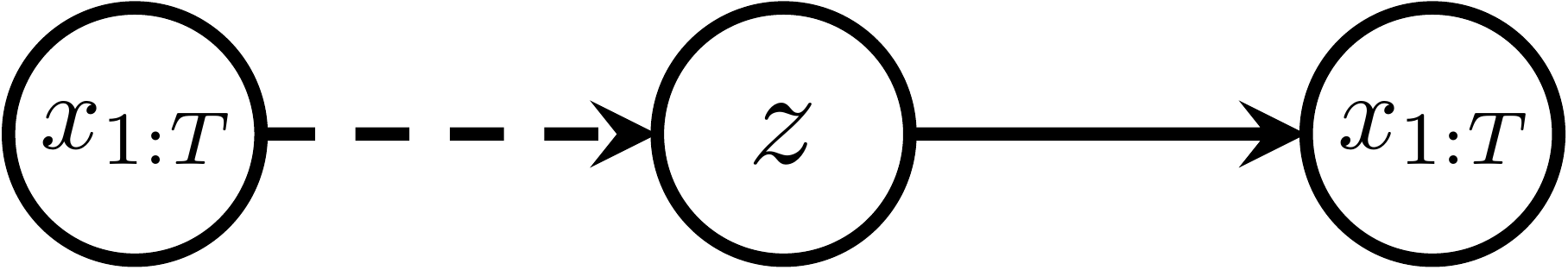}
    \caption{Graphical model of a sentence variational autoencoder. Dashed lines indicate inference model. Solid lines indicate generative model.}
    \label{background-fig:vae-graphical-model}
\end{figure}

We review variational autoencoders by describing an SVAE for an input sentence $x_{1:T}$. When using an SVAE, we assume a generative model that generates an input using a latent variable $z$, typically assumed to follow a multivariate Gaussian distribution (as shown in \cref{background-fig:vae-graphical-model}). We seek to maximize the marginal likelihood of inputs $x_{1:T}$ when marginalizing out the latent variable $z$. Since this is typically intractable, especially when using continuous latent variables, we instead maximize a lower bound on the marginal log-likelihood~\citep{kingma2013auto}:
\begin{equation}
\begin{aligned}
    \log~p_\theta(x_{1:T})&\geq
    \mathop\mathbb{E}_{z\sim q_\phi(\cdot\vert x_{1:T})}\!\!\left[\log p_\theta(x_{1:T}\vert z)-\log\frac{q_\phi(z\vert~x_{1:T})}{p_\theta(z)} \right]\\&=
    \underbrace{\mathop\mathbb{E}_{z\sim q_\phi(\cdot\vert x_{1:T})}\!\!\!\!\!\!\!\!\left[\log p_\theta(x_{1:T}\vert z)\right]}_\text{Reconstruction Loss}-\underbrace{\vphantom{\mathop\mathbb{E}_{z\sim~q_\phi(\cdot\vert x_{1:T})}\left[\log p_\theta(x_{1:T}\vert z)\right]}\kld(q_\phi(z\vert x_{1:T})\Vert p_\theta(z))}_\text{KL divergence}
\end{aligned}
\label{background_eq:vae_elbo}
\end{equation}

\noindent where KL divergence is a distance measure between two probability distributions \citep{kullback1951information} and we have introduced the variational posterior $q$ parametrized by new parameters $\phi$. $q$ is referred to as an ``inference model'' as it encodes an input into the latent space. We also have the generative model probabilities $p$ parametrized by $\theta$. The parameters are trained in a way that reflects a classical autoencoder framework: encode the input into a latent space, decode the latent space to reconstruct the input. These models are therefore referred to as ``variational autoencoders''.

The lower bound (also known as the evidence lower bound (ELBO)) consists of two terms: reconstruction loss and KL divergence.
The KL divergence term provides a regularizing effect during learning by ensuring that the learned posterior remains close to the prior over the latent variables.

\paragraph{von Mises-Fisher Distribution.} von Mises-Fisher (vMF) distribution can be regarded as a Gaussian distribution on a hypersphere with two parameters: $\mu$ and $\kappa$. $\mu\in\mathbb{R}^m$ is a normalized vector (i.e. $\Vert\mu\Vert_2=1$ ) defining the mean direction. $\kappa\in\mathbb{R}_{\geq 0}$ is often referred to as a concentration parameter analogous to the variance in a Gaussian distribution. vMF has been used for modeling similarity between two sentences~\citep{guu-etal-2018-generating}, which is particularly suited to our purposes, since we will evaluate our semantic representations in the context of modeling paraphrases (See \cref{section:vgvae-representation} and \cref{section:vgvae-generation} for more details).

When assuming $q_\phi(y\vert  x)$ follows $\vmf(\mu_\alpha(x),\kappa_\alpha(x))$ and the prior $p_\theta(y)$ follows the uniform distribution $\vmf(\cdot,0)$, the $\kld(q_\phi(y\vert  x)\Vert p_\theta(y))$ appearing in the ELBO can be computed in closed-form:
\begin{equation}
\begin{aligned}
    \kappa_\alpha\frac{\mathcal{I}_{m/2}(\kappa_\alpha)}{\mathcal{I}_{m/2-1}(\kappa_\alpha)} + (m/2 - 1)\log\kappa_\alpha - (m/2)\log(2\pi) -\\ \log\mathcal{I}_{m/2-1}(\kappa_\alpha)+\frac{m}{2}\log\pi+\log 2-\log\Gamma(\frac{m}{2})
\end{aligned}
\end{equation}

\noindent where $\mathcal{I}_v$ is the modified Bessel function of the first kind at order $v$ and $\Gamma(\cdot)$ is the Gamma function. In \cref{CHAPTER:DISENTANGLE}, we follow~\citet{davidson2018hyperspherical} and use an acceptance-rejection scheme to sample from vMF.

\paragraph{Gaussian Distribution.} We assume $q_\phi(z\vert  x)$ follows a Gaussian distribution $\mathcal{N}(\mu_\beta(x),\text{diag}(\sigma_\beta(x)))$ and that the prior $p_\theta(z)$ is $\mathcal{N}(0,I_{d})$, where $I_{d}$ is an $d\times d$ identity matrix. Since we only consider a diagonal covariance matrix, the KL divergence term $\kld(q_\phi(z\vert  x)\Vert p_\theta(z))$ in \cref{background_eq:vae_elbo} can be computed efficiently:
\begin{equation}
    \frac{1}{2}(-\sum_i\log\sigma_{\beta i} + \sum_i\sigma_{\beta i} + \sum_i{\mu_{\beta i} ^2} - d)
\end{equation}

\section{Summary}
In this chapter, we describe the background materials needed for the remainder of this thesis. In \cref{CHAPTER:SELFSUPERVISION}, we present our contributions to improving self-supervised training objectives for language model pretraining. The new training objectives help enhance the quality of general language representations and model performance on few-shot learning. \cref{CHAPTER:WIKIPEDIA} presents our contributions to exploit naturally-occurring data structures on \wikipedia for entity and sentence representations and textual entailment. \cref{CHAPTER:DISENTANGLE} presents our contributions on leveraging freely-available parallel corpora for disentangling semantic and syntactic representations. Then we apply the technique to controlling the syntax of generated sentences using a sentential exemplar. \cref{CHAPTER:EVALUATION} presents our contributed datasets for data-to-text generation, abstractive summarization, and story generation. They are tailored from naturally-occurring textual resources and have unique challenges in their respective task settings.

\chapter{Improving Self-Supervision for Language Pretraining}
\label{CHAPTER:SELFSUPERVISION}
This chapter describes our contributions to improving self-supervised training objectives for language model pretraining. Prior work has found that the next sentence prediction loss used for pretraining is ineffective in improving downstream task performance \citep{NEURIPS2019_dc6a7e65,liu2019roberta}. In \cref{sec:sentence-order-prediction}, we propose to replace it with the sentence ordering prediction loss and show the improved model leads to state-of-the-art performance.

Recent work has discovered that pretrained language models are capable of performing in-context few-shot learning \citep{brown-etal-gpt3-nips} and the performance can be improved by finetuning the models on human-annotated datasets \citep{mishra2021crosstask,ye-etal-2021-crossfit,wei2021finetuned}. \cref{sec:incontext-learning} shows that pretraining the models on self-supervised tasks can also lead to improved performance on downstream tasks.

The material in this chapter is adapted from \citet{Lan2020ALBERT:} and \citet{chen-etal-2022-improving}.

\section{Improving Language Representation Learning via Sentence Ordering Prediction}
\label{sec:sentence-order-prediction}

\subsection{Introduction}
Pretraining of large transformers has led to a series of breakthroughs in language representation learning \citep{radford2018improving, devlin-etal-2019-bert}. Many nontrivial NLP tasks, including those that have limited training data, have greatly benefited from these pretrained models. Given the importance of model sizes, we ask: Is having better NLP models as easy as having larger models?

\begin{table}[ht]
\small
\centering
\begin{tabular}{|lccc|}
\hline
Model  & Hidden Size & Parameters	 & Accuracy	 \\
\hline
    \bertsize{large}  &1024 &334M    &72.0\% \\
    \hline 
    \multicolumn{4}{|l|}{\textit{Our implementations}}\\
    \bertsize{large}	        & 1024                          &334M	 &73.9\%\\
    \bertsize{xlarge}      & 2048              	&1270M	 &54.3\%\\\hline
\end{tabular}
\caption{Model performance on RACE \citep{lai-etal-2017-race}, a reading comprehension dataset. Increasing the hidden size of \bertsize{large} leads to worse performance.}
\label{albert-table:large_vs_extra_large}
\end{table}

We observe that simply growing the hidden size of a model such as \bertsize{large} \citep{devlin-etal-2019-bert} can lead to worse performance. \cref{albert-table:large_vs_extra_large} shows an example where we increase the hidden size of \bertsize{large} to be 2 times larger but do not obtain better results.

In this paper, we address this problem by designing A Lite BERT (ALBERT) architecture that has significantly fewer parameters than a traditional BERT architecture. 

ALBERT incorporates two parameter reduction techniques: embedding matrix factorization \citep{pmlr-v70-grave17a, baevski2018adaptive} and cross-layer parameter sharing \citep{dehghani2018universal}. To further improve the performance of ALBERT, we also introduce a self-supervised loss for sentence ordering prediction (SOP). SOP primarily focuses on inter-sentence coherence and is designed to address the ineffectiveness \citep{NEURIPS2019_dc6a7e65,liu2019roberta} of the NSP loss proposed in the original \bert paper. 

As a result of these design choices, we are able to scale up to much larger ALBERT configurations that still have fewer parameters than BERT-large but achieve significantly better performance.
We also establish new state-of-the-art results on several popular NLP benchmarks.

\subsection{Related Work}

ALBERT uses a pretraining loss based on predicting the ordering of two consecutive segments of text. Several researchers have experimented with pretraining objectives that similarly relate to discourse coherence. 
Coherence and cohesion in discourse have been widely studied and many phenomena have been identified that connect neighboring text segments~\citep{DBLP:journals/cogsci/Hobbs79,halliday-76,grosz-etal-1995-centering}.
Most objectives found effective in practice are quite simple. 
Skip-thought~\citep{kiros-etal-skipthoughts-2015} and FastSent~\citep{hill-etal-2016-learning} sentence embeddings are learned by using an encoding of a sentence to predict words in neighboring sentences. 
Other objectives for sentence embedding learning include predicting future sentences rather than only neighbors~\citep{gan-etal-2017-learning} and predicting explicit discourse markers~\citep{jernite2017discourse,nie-etal-2019-dissent}.

Our loss is most similar to the sentence ordering objective of \citet{jernite2017discourse}, where sentence embeddings are learned in order to determine the ordering of two consecutive sentences. 
Unlike most of the above work, however, our loss is defined on textual segments rather than sentences. BERT uses a loss based on predicting whether the second segment in a pair has been swapped with a segment from another document.
We compare to this loss in our experiments and find that sentence ordering is a more challenging pretraining task and more useful for certain downstream tasks.

Concurrently to this work, \citet{Wang2020StructBERT:} also try to predict the order of two consecutive segments of text, but they combine it with the original next sentence prediction in a three-way classification task rather than empirically comparing the two. 

\subsection{Method}

The backbone of the ALBERT architecture is similar to BERT in that it uses a transformer encoder~\citep{attention_is_all_you_need} with GELU nonlinearities~\citep{hendrycks2016gelus}. 
We follow the BERT notation conventions and denote the vocabulary embedding size as $E$, the number of encoder layers as $L$, and the hidden size as $H$. Following \citet{devlin-etal-2019-bert}, 
we set the feed-forward/filter size to be $4H$ and the number of attention heads to be $H/64$.

\paragraph{Sentence Ordering Prediction.}
BERT is trained jointly by MLM and NSP. NSP is a binary classification task for predicting whether two segments appear consecutively in the original text. Positive examples are created by taking consecutive segments from the training corpus. Negative examples are created by pairing segments from different documents. Positive and negative examples are sampled with equal probability. Later studies~\citep{NEURIPS2019_dc6a7e65,liu2019roberta} show that NSP has little impact on improving downstream task performance.

We conjecture that the main reason behind NSP's ineffectiveness is its lack of difficulty as a task, as compared to MLM.
As formulated, NSP conflates topic prediction and coherence prediction. \footnote{Since a negative example is constructed using material from a different document, the negative-example segment is misaligned both from a topic and from a coherence perspective.}
However, topic prediction is easier to learn compared to coherence prediction.

We argue that inter-sentence modeling is an important aspect of language understanding. Therefore we propose a loss based primarily on coherence. 
That is, for ALBERT, we use a sentence ordering prediction (SOP) loss, which avoids topic prediction and instead focuses on modeling inter-sentence coherence.
The SOP constructs positive examples the same as BERT (two consecutive segments from the same document) and negative examples using the same two consecutive segments but with their order swapped.
This forces the model to learn finer-grained distinctions about discourse-level coherence properties.

\subsection{Experiments}
\paragraph{Pretraining Setup.}
To make our comparison to other models as meaningful as possible, we follow the setup used in pretraining BERT --- we use the \textsc{BookCorpus}~\citep{zhu2015aligning} and English Wikipedia for pretraining baseline models.
These two corpora consist of around 16GB of uncompressed text.
We format our inputs as \texttt{[CLS]} $x_1$ \texttt{[SEP]} $x_2$ \texttt{[SEP]} where $x_1$ and $x_2$ are two text segments.
We always limit the maximum input length to 512, and randomly generate input sequences shorter than 512 with a probability of 10\%.
Like \bert, we use a vocabulary size of 30,000, tokenized using SentencePiece~\citep{kudo-richardson-2018-sentencepiece} as in \xlnet~\citep{NEURIPS2019_dc6a7e65}.

We generate masked inputs for the MLM targets using $n$-gram masking~\citep{joshi-etal-2020-spanbert}, with the length of each $n$-gram mask selected randomly.
The probability for the length $n$ is given by
\begin{equation*}
    p(n) = \frac{1/n}{\sum_{k=1}^N{1/k}}
\end{equation*}

We set the maximum length of $n$-gram (i.e., $n$) to be 3 (i.e., the MLM target can consist of up to a 3-gram of complete words, such as ``White House correspondents'').

All the model updates use a batch size of 4096 and a \textsc{Lamb} optimizer with learning rate 0.00176~\citep{You2020Large}.
We train all models for 125,000 steps unless otherwise specified.
Training was done on Cloud TPU V3.
The number of TPUs used for training ranged from 64 to 1024, depending on model sizes.  Code and pretrained models are available at \url{https://github.com/google-research/albert}.

\paragraph{Evaluation Benchmarks.} Following \citet{NEURIPS2019_dc6a7e65} and \citet{liu2019roberta}, we evaluate our models on three popular benchmarks: GLUE benchmark~\citep{wang-etal-2018-glue}, two versions of the SQuAD Dataset \cite{rajpurkar-etal-2016-squad,rajpurkar-etal-2018-know}, and the RACE dataset \citep{lai-etal-2017-race}. 
For completeness, we provide description of these benchmarks in \cref{albert-appendix:downstream_detailed_description}.
As in \citet{liu2019roberta}, we perform early stopping on the development sets, on which we report all comparisons except for our final comparisons based on the task leaderboards, for which we also report test set results. For GLUE datasets, we report medians over five runs.

\paragraph{Comparing SOP to NSP.}
We compare head-to-head three experimental conditions for the additional inter-sentence loss: none (XLNet- and RoBERTa-style), NSP (BERT-style), and SOP (ALBERT-style), using an ALBERT-base configuration.
Results are shown in Table \cref{albert-table:sop-compare-result}, both over intrinsic and downstream tasks.

\begin{table}[!htbp] 
\small
\centering
\begin{tabular}{| l | c c c | c c c c  c | c |}
\hline
&\multicolumn{3}{|c|}{Intrinsic Tasks} & \multicolumn{6}{c|}{Downstream Tasks}  \\
	    	&MLM	&NSP & SOP &SQuAD1.1	&SQuAD2.0	&MNLI	&SST-2	&RACE & Avg. \\
\hline
None & \textbf{54.9} & 52.4 & 53.3 &88.6/81.5	&78.1/75.3	&81.5	&89.9	&61.7		& 79.0 \\
NSP	 &54.5 & \textbf{90.5} & 52.0 &88.4/81.5	&77.2/74.6	&81.6	&\textbf{91.1}	&62.3	& 79.2  \\
SOP	 &54.0 &78.9 &\textbf{86.5}  &\textbf{89.3/82.3}	&\textbf{80.0/77.1}	&\textbf{82.0}	&90.3	&\textbf{64.0}	 & \textbf{80.1}	\\\hline
\end{tabular}
\caption{The effect of using next sentence prediction (NSP) vs. sentence ordering prediction (SOP) on intrinsic and downstream tasks. We boldface the best results in each column. MLM=masked language modeling.}
\label{albert-table:sop-compare-result}
\end{table}

The results on the intrinsic tasks reveal that the NSP loss brings no discriminative power to the SOP task (52.0\% accuracy, similar to the random-guess performance for the ``None'' condition).
This allows us to conclude that NSP ends up modeling only topic shift.
In contrast, the SOP loss does solve the NSP task relatively well (78.9\% accuracy), and the SOP task even better (86.5\% accuracy).
Even more importantly, the SOP loss appears to consistently improve downstream task performance for multi-sentence encoding tasks (around +1\% for SQuAD1.1, +2\% for SQuAD2.0, +1.7\% for RACE), for an average score improvement of around +1\%.

\paragraph{Comparing ALBERT to State-of-the-Art Models.}

\begin{table}
\footnotesize
\centering
\begin{tabular}{|l   c c c c c c c c c c|}\hline
Models &MNLI	&QNLI	&QQP	&RTE	&SST	&MRPC	&CoLA	&STS	&WNLI	&Avg \\
\hline
\multicolumn{11}{|l|}{\textit{Single-task single models on dev}}\\
BERT-large	 &86.6	&92.3	&91.3	&70.4	&93.2	&88.0	&60.6	&90.0 &- &-	\\	
XLNet-large	 &89.8	&93.9	&91.8	&83.8	&95.6	&89.2	&63.6	&91.8 &- &-	\\	
RoBERTa-large&90.2	&94.7	&\textbf{92.2}	&86.6	&96.4	&\textbf{90.9}	&68.0	&92.4 &- &-	\\	
ALBERT (1M)   &90.4	&95.2	&92.0	&88.1	&96.8	&90.2	&68.7	&92.7 &- &-	\\
ALBERT (1.5M) &\textbf{90.8}	&\textbf{95.3}	&\textbf{92.2}	&\textbf{89.2}	&\textbf{96.9}	&\textbf{90.9}	&\textbf{71.4}	&\textbf{93.0} &- &-	\\	
\hline
\multicolumn{11}{|l|}{\textit{Ensembles on test (from leaderboard as of Sept. 16, 2019)}}\\
XLNet	    &90.2	&98.6	&90.3	&86.3	&96.8	&93.0	&67.8	&91.6	&90.4	&88.4 \\
RoBERTa	    &90.8	&98.9	&90.2	&88.2	&96.7	&92.3	&67.8	&92.2	&89.0	&88.5 \\
Adv-RoBERTa	&91.1	&98.8	&90.3	&88.7	&96.8	&93.1	&68.0	&92.4	&89.0	&88.8 \\
ALBERT 	    &\textbf{91.3}	&\textbf{99.2}	&90.5	&\textbf{89.2}	&\textbf{97.1}	&\textbf{93.4}	&69.1	&\textbf{92.5}	&\textbf{91.8}	&\textbf{89.4} \\\hline
\end{tabular}
\caption{GLUE benchmark results.
For single-task single-model results, we report ALBERT at 1M steps (comparable to RoBERTa) and 1.5M steps. The ALBERT ensemble use model checkpoints trained with 1M, 1.5M, and other numbers of steps.}
\label{albert-table:glue}
\end{table}

\begin{table}
\setlength{\tabcolsep}{4pt}
\small
\centering
\begin{tabular}{|lccc|}
\hline
Models		&SQuAD1.1 dev & SQuAD2.0 test &RACE test (Middle/High) \\
\hline
\multicolumn{4}{|l|}{\textit{Single model (from leaderboard as of Sept.~23, 2019)}}\\
BERT-large          & 90.9/84.1 & 89.1/86.3 & 72.0 (76.6/70.1) \\
XLNet               & 94.5/89.0 & 89.1/86.3 & 81.8 (85.5/80.2) \\
RoBERTa             & 94.6/88.9 & 89.8/86.8 & 83.2 (86.5/81.3) \\
UPM                 &  - & 89.9/87.2 & - \\
XLNet + SG-Net Verifier++ & - & 90.1/87.2 &  -\\
ALBERT (1M)          & 94.8/89.2 & - & 86.0 (88.2/85.1) \\
ALBERT (1.5M)        & \textbf{94.8/89.3} & \textbf{90.9/88.1} & \textbf{86.5 (89.0/85.5)} \\
\hline
\multicolumn{4}{|l|}{\textit{Ensembles (from leaderboard as of Sept.~23, 2019)}}\\
BERT-large          & 92.2/86.2   & - &  - \\
XLNet + SG-Net Verifier & -&	90.7/88.2	& -\\
UPM & - & 90.7/88.2 & \\ 
XLNet + DAAF + Verifier & - & 	90.9/88.6 & -\\
DCMN+               & - & - & 84.1 (88.5/82.3) \\
ALBERT     &    \textbf{95.5/90.1}  & \textbf{92.2/89.7}    &\textbf{89.4 (91.2/88.6)}  \\ \hline
\end{tabular}
\caption{Results for SQuAD and RACE.}
\label{albert-table:sota-squad-race}
\end{table}

The results we report in \cref{albert-table:glue} and \cref{albert-table:sota-squad-race} use the additional data described in \citet{liu2019roberta} and \citet{NEURIPS2019_dc6a7e65}. 
These two tables show that ALBERT achieves state-of-the-art results under two settings: single-model and ensembles.

When ensembling models, for GLUE benchmark and RACE, we average the model predictions for the ensemble models, where the candidates are fine-tuned from different training steps using the 12-layer and 24-layer architectures.
For SQuAD, we average the prediction scores for those spans that have multiple probabilities; we also average the scores of the ``unanswerable'' decision.

Both single-model and ensemble results indicate that ALBERT improves the state-of-the-art significantly for all three benchmarks. For RACE,
our single model achieves an accuracy of $86.5\%$, which is still $2.4\%$ better than the current state-of-the-art ensemble model.

\section{Improving In-Context Few-Shot Learning via Self-Supervised Training}
\label{sec:incontext-learning}

\subsection{Introduction}
In-context few-shot learning seeks to solve unseen tasks at inference time by conditioning on a few training examples. In particular, in this case we are interested in methods that forgo any weight updates \citep{brown-etal-gpt3-nips}. Prior work has been focused on improving inference time algorithms (e.g., rescoring generated outputs \citep{pmlr-v139-zhao21c}, selecting \citep{liu2021makes} and ordering \citep{lu2021fantastically} the given few-shot examples) and incorporating extra resources (e.g., finetuning models on human-annotated datasets \citep{mishra2021crosstask,ye-etal-2021-crossfit,wei2021finetuned}). 

We hypothesise that a different way to improve in-context few-shot learning is through designing self-supervised objectives that more closely resemble the format of tasks that the model will be asked to perform. To do so, we cast the self-supervised training as an intermediate training stage between language model pretraining and downstream few-shot evaluation. In particular, we construct training datasets based on the self-supervised objectives following similar formats used in the downstream tasks, finetune pretrained language model checkpoints on the training datasets, and then evaluate the models on benchmarks.

In experiments, we consider four self-supervised objectives, including masked word prediction and classification tasks related to next sentence prediction \citep{devlin-etal-2019-bert}. We evaluate models on two benchmarks (13 tasks in total): SuperGLUE \citep{alex-etal-2019-nips} and Natural-Instructions \citep{mishra2021crosstask}. SuperGLUE focuses on discriminative tasks, and Natural-Instructions is a set of generative tasks.

Empirically, we experiment with pretrained language models of two sizes: 125 million parameters and 1.3 billion parameters. We show that in our best setting, the 1.3 billion parameters model trained by the self-supervision performs better than the initial pretrained language models and two strong baselines on average.

Further analysis reveals that (1) the effectiveness of the self-supervision depends on the amount of training data, but the benefit of adding more data is diminishing; (2) the improvements brought by the self-supervision are in part due to the semantic similarity between the training and evaluation tasks; (3) adding more self-supervised objectives may not help model performance because adding them does not contribute to the diversity of the self-supervised tasks; (4) choosing similar task templates for both self-supervised and downstream tasks plays a vital role in improving model performance; (5) self-supervised tasks and human-annotated datasets are complementary; (6) generation examples show that compared to the initial pretrained language models, self-supervised-trained models are better at following the task instructions.
\subsection{Related Work}
\paragraph{In-Context Few-Shot Learning.} \citet{brown-etal-gpt3-nips} discover that large pretrained language models can solve unseen tasks at inference time. Recent work has improved the in-context few-shot performance by rescoring generated outputs \citep{pmlr-v139-zhao21c}, selecting \citep{liu2021makes} and ordering \citep{lu2021fantastically} the given few-shot examples.
Other work studies pretrained language models' cross-task generalization abilities for in-context few-shot or zero-shot learning using human-annotated datasets \citep{ye-etal-2021-crossfit,wei2021finetuned,sanh2021multitask,min2021metaicl,xu2022zeroprompt} via instructions \citep{weller-etal-2020-learning,efrat2020turking,mishra2021crosstask,ouyang2022training} and retrieved examples \citep{hu2022context,lin2022unsupervised}. Our work differs in that we focus on self-supervised training.

\paragraph{Finetuning for Few-Shot Learning.}
Pretrained language models for few-shot learning typically follows the ``pretrain then finetune'' paradigm \citep[\emph{inter alia}]{howard-ruder-2018-universal,radford2018improving,devlin-etal-2019-bert}, where recent work has focused on designing templates for few-shot finetuning \citep{reynolds2021prompt,schick-schutze-2021-exploiting,schick-schutze-2021-just,schick-schutze-2021-shot,le-scao-rush-2021-many,tam-etal-2021-improving,gao-etal-2021-making,sorensen2022informationtheoretic}, and optimizing soft prompts \citep{li-liang-2021-prefix,qin-eisner-2021-learning,lester2021power,gu2021ppt,zhang2022differentiable}.
Other work focuses on unifying task formats to maximize the benefits of human annotations, including question answering \citep{zhong-etal-2021-adapting-language}, textual entailment \citep{yin-etal-2019-benchmarking,yin-etal-2020-universal,wang2021entailment}, and many other tasks \citep{mccann2018natural,keskar2019unifying,2020t5,NEURIPS2021_8493eeac}.
In contrast, our focus is on in-context few-shot learning, without finetuning models on downstream task examples.

\paragraph{Pretraining for Few-Shot Learning.}
Several papers have adapted various resources for pretraining models to enhance their performances on few-shot learning, such as pretraining on hypertext \citep{aghajanyan2021htlm}, question-infused pretraining \citep{jia2021question}, and self-training \citep{du-etal-2021-self,vu-etal-2021-strata,wang2021list}. Pretraining approaches have targeted specific tasks, such as task-oriented dialog \citep{mi-etal-2021-self}, intent detection \citep{zhang-etal-2021-shot}, and data-to-text generation \citep{chen-etal-2020-kgpt}. Our work differs as we use plain text as opposed to (naturally-occurring) human-annotated resources. Relatedly, \citet{bansal-etal-2020-self} used self-supervised meta-learning for few-shot text classification rather than in-context few-shot learning.

\paragraph{Intermediate Finetuning.} Since our approach involves an extra training stage between pretraining and downstream evaluation, it is also related to prior work that uses multi-stage finetuning on human-annotated datasets for generic tasks \citep{phang2018sentence,pruksachatkun-etal-2020-intermediate,chang-lu-2021-rethinking-intermediate,aghajanyan-etal-2021-muppet,poth-etal-2021-pre} and text classification \citep{zhang-zhang-2021-qa}. Relevant work also studies intermediate finetuning using crosslingual supervision \citep{phang-etal-2020-english,moghe-etal-2021-cross}. \citet{rubino-sumita-2020-intermediate} use an intermediate self-supervised training stage for machine translation quality estimation.

\subsection{Method}
\label{incontext-sec:method}
We describe four self-supervised training objectives that will be used to train models before downstream evaluations.

\begin{figure}
    \centering
    \includegraphics[scale=0.4]{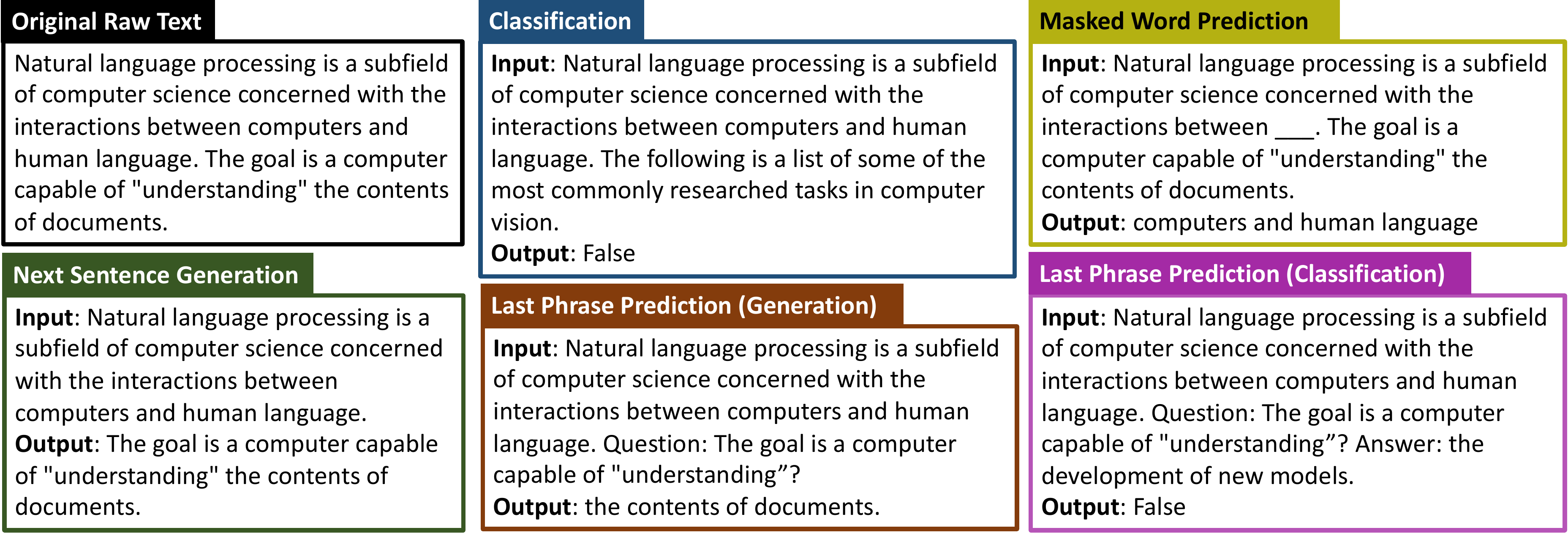}
    \caption{Examples of our self-supervised training tasks. Each example is an input-output pair constructed from the raw text.}
    \label{incontext-fig:selfsupervised_task_examples}
\end{figure}

\begin{figure}
    \centering
    \includegraphics[scale=0.37]{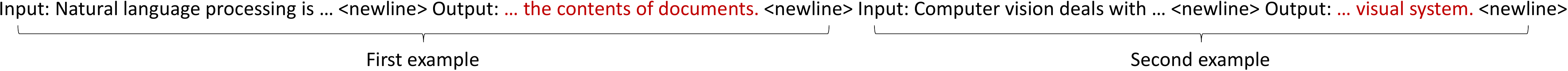}
    \caption{An example of a training instance. Each instance is formed by several training examples. During training, we use left-to-right language models and compute a cross-entropy loss on the output texts (indicated by the red color in the shown example). We note that when computing the loss on the second example, the first example can be seen as task demonstrations. For brevity, we show part of the input and output texts.}
    \label{incontext-fig:selfsupervised_instance_examples}
\end{figure}

We begin by defining the example and the instance used during our self-supervised training. An \textbf{example} is an input-output pair. To differentiate the input and the output, we append special tokens ``Input:'' and ``Output:'' to the beginning of input text and output text respectively where the two texts are also separated by the $\langle\text{newline}\rangle$ token (see \cref{incontext-fig:selfsupervised_task_examples} for examples).\footnote{We chose this special symbol because we always start the self-supervised training from a pretrained language model checkpoint.}

An \textbf{instance} is a linearized string formed by several examples from the same task (e.g., see \cref{incontext-fig:selfsupervised_instance_examples}). As we encode the text using causal attention, the examples closer to the beginning of input sequences can be seen as task demonstrations, resulting in efficient computation.

When constructing the training examples, we pick three or more consecutive sentences (depending on the minimum sequence length we enforce on the sentences) and then apply task-specific rules to automatically create training data. To form a training instance, we randomly select examples from the same task until reaching the maximum sequence length (i.e., 2048). During training, we compute a cross-entropy loss on tokens in the \textbf{output texts}. We describe details of the self-supervised tasks as follows.

\begin{figure}
    \centering
    \includegraphics[scale=0.33]{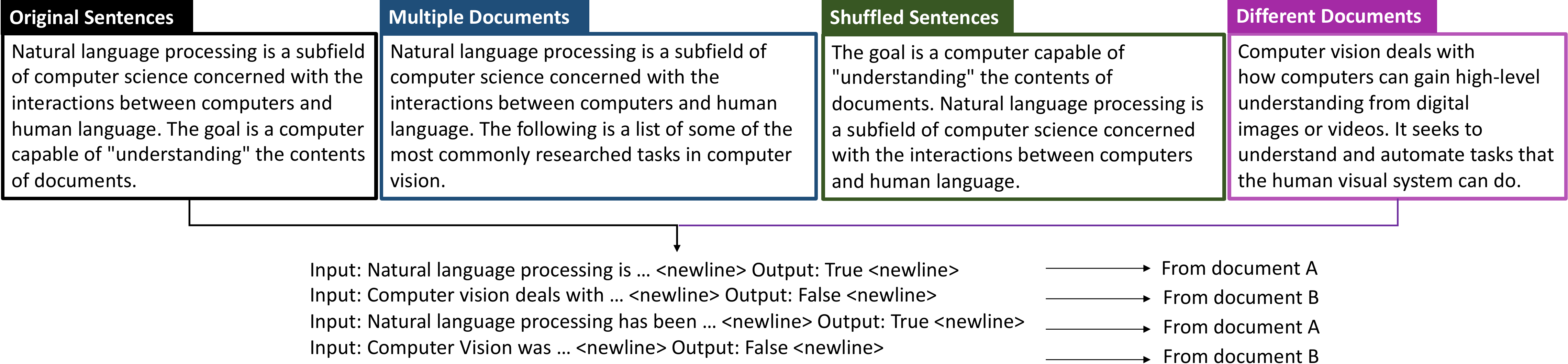}
    \caption{Example illustrating the construction of training instances for our classification task. There are four input types, and each training instance has two or three types. As the shown instance has the following two types: "original sentences" and "different documents", it comprises examples from two different documents. The instance resembles the next sentence prediction task, encouraging models to compare topical similarities between the two examples.}
    \label{incontext-fig:selfsupervised_classification_example}
\end{figure}

\paragraph{Next Sentence Generation.}

In light of the strong performance of language models on in-context few-shot learning \citep{brown-etal-gpt3-nips}, we incorporate the language modeling as one of our self-supervised tasks, which we call ``next sentence generation'' (NSG). NSG asks the model to generate the next sentence given previous sentences as context. When building data for this task, we use the last sentence as output and the rest of the sentences as input.

\paragraph{Masked Word Prediction.}
The second task we consider is based on masked word prediction (MWP) which is commonly used in pretraining generic text encoders \citep{devlin-etal-2019-bert,liu2019roberta}. The task asks the model to fill in the missing information based on the surrounding context. Specifically, MWP randomly replaces words in input sentences with a special symbol and requires models to recover the masked words in the input. For this task, we create input text by randomly replacing 1{\raise.17ex\hbox{$\scriptstyle\mathtt{\sim}$}}20 words in the input text with a special token\footnote{We randomly select the special token from the following list: \_\_\_, $\langle\langle\rangle\rangle$, @@@, (()), \$\$\$, \%\%\%, \#\#\#, ***, and +++. We use random symbols instead of a fixed symbol because we found that it gives better performance in our preliminary experiments.} and use the masked out words as the output text.

\paragraph{Last Phrase Prediction.}
Inspired by the \lambada dataset \citep{paperno-etal-2016-lambada}, a question answering dataset which asks models to predict the last word in a sentence given several sentences of context, we create a ``last phrase prediction'' (\lpp) task, which requires predicting the last phrase in a sentence. To solve this task, models need to draw relevant information from the context and the learned knowledge during pretraining. We cast \lpp as either a generation task or a classification task. The latter variant of \lpp is a binary classification task that labels if the given answer is the correct phrase. To facilitate a unified format of these two tasks, we append a special token ``Question:'' to the beginning of the last sentence and replace the last phrase with a question mark. For the classification \lpp, we separate the given answer and the previous context and sentences with a special token ``Answer:''. An example of this task is shown in \cref{incontext-fig:selfsupervised_task_examples}.

More specifically, we identify the last phrase of a sentence based on a set of function words (see \cref{incontext-appendix:lpp_function_words} for the list of function words). If there are multiple function words in a sentence, we pick the last one. Then we treat the text segment starting from the function word as the last phrase.\footnote{We ensure that the last sentence in raw text for this task always has at least one valid function word and the function word lies at the second half of the sentence.} When selecting negative answers, we randomly choose from the phrases extracted from the same function words (to make the negative answers more challenging).

\paragraph{Classification.}

Similar to the next sentence prediction task \citep{devlin-etal-2019-bert} and the sentence ordering prediction task \citep{jernite2017discourse,chen-etal-2019-evaluation} for pretraining language representations, we create a classification task (CL) for our self-supervised training. As shown in \cref{incontext-fig:selfsupervised_classification_example}, for this task, we consider four types of input: original sentences, shuffled sentences, sentences from a different document, and sentences from multiple documents. In particular, for original sentences, we directly use text from original human-written documents. For shuffled sentences, we randomly shuffle all the input sentences. For sentences from multiple documents, we randomly replace 50\% of the input sentences with sentences from another document. We also ensure that the selected sentences (from both the input and another document) are consecutive in their original documents. For sentences from different documents, we replace the input sentences with sentences from another document. See \cref{incontext-fig:selfsupervised_classification_example} for an example of each type of input.

When constructing a training instance, we randomly pick one or two additional input types and combine them with the original sentences to form a binary or three-way classification task. We also randomly assign label strings to input types in each instance to ensure that models follow the information given by earlier examples when making predictions.

The classification task is different from the other self-supervised tasks described in earlier subsections. It explicitly requires models to compare inputs across examples in a training instance to determine if the given input shares similar properties with the others. 

\subsection{Experimental Setup}

\paragraph{Training Setup.}
For the pretrained language model checkpoints, we use the 125 million parameters (125M) and the 1.3 billion parameters (1.3B) dense model from \citet{artetxe2021efficient}. These pretrained models have shown results comparable to GPT3 across various tasks.

For self-supervised training, we use a subset of documents from the RoBERTa training corpus \citep{liu2019roberta} that contains four domains: \textsc{BookCorpus} plus Wikipedia, \textsc{CC-News}, \textsc{OpenWebText}, and \textsc{Stories}. Specifically, we randomly sample 100k documents from each domain except \textsc{Stories} where we only sample 10k documents as the documents there are much longer than the others. The final training data contains approximately 1 million instances with 250k training instances per task.\footnote{The average numbers of example per instance for each data source are: 6.9 for \textsc{BookCorpus} plus Wikipedia, 5.3 for \textsc{CC-News}, 3.5 for \textsc{OpenWebText}, and 7.2 for \textsc{Stories}.} For the 125M model, we train for 10 epochs, which takes roughly 1 day on a V100 GPU. For the 1.3B model, we train for 5 epochs, which takes roughly 3 days on 2 V100 GPUs.

\paragraph{Evaluation Setup.} The instance and example during evaluation shares similar definition as those in \cref{incontext-sec:method} except that each evaluation instance has only one example from test splits and it is placed at the last position in the instance. The other examples in the instance (i.e., task demonstrations) come from either training splits or task-specific instructions depending on benchmarks.

We evaluate the models on two benchmarks: SuperGLUE and Natural-Instructions. SuperGLUE is a set of tasks focusing on natural language understanding. We use BoolQ (BQ; \citealp{clark-etal-2019-boolq}), CB \citep{demarneffe:cb}, COPA (CA; \citealp{roemmele2011choice}), MultiRC (MC; \citealp{khashabi-etal-2018-looking}), and RTE (RE; \citealp{giampiccolo-etal-2007-third,bentivogli2009fifth,dagan2006pascal,bar2006second}).\footnote{We exclude WSC \citep{levesque2011winograd} and ReCoRD \citep{zhang2018record} as pretrained models, including GPT3, require scoring algorithms at inference time to achieve competitive results. We exclude WiC \citep{pilehvar-camacho-collados-2019-wic} because GPT3-like models, including GPT3 and our models, do not give accuracies significantly better than random baselines.} We report results for the official development sets. The task demonstrations are examples randomly selected from the training sets. We report mean and standard deviations of five runs with different random seeds. Following GPT3, we use a ranking based approach when evaluating the models (i.e., pick the best label based on language modeling perplexities).

Natural-Instructions evaluates models' cross-task generalization abilities where all the tasks are generation tasks. It splits the tasks into two groups for training and evaluation. We use the same task split and evaluate models on the following task categories: question generation (QG), answer generation (AG), minimal modification (MM), and verification (VF).\footnote{We discard training tasks that share the same source datasets with evaluation tasks as we found that tasks with the same source dataset may contain leaked labels. We exclude the binary classification tasks because the class labels are severely imbalanced (i.e., more than 80\% of the class labels belong to one category).} Each task category has two tasks. Following the few-shot setting used in \citet{mishra2021crosstask}, we evaluate models using 100 examples per task, use greedy decoding, and report ROUGE-L \citep{lin-2004-rouge} scores per task category. For task demonstrations, we use the positive examples in the instructions in Natural-Instructions.

\begin{table}
    \centering\small
    \begin{tabular}{|l|p{0.8\textwidth}|}\hline
        \bf GPT3 & \$\{Context\}$\langle$newline$\rangle$\$\{Question\}$\langle$newline$\rangle$ - \bf\textcolor{red}{[\$\{Label\}] \$\{Answer\}} \\\hline
        \bf Ours & Input: \$\{Context\} Question: \$\{Question\} Answer: \$\{Answer\}$\langle$newline$\rangle$Output: \bf\textcolor{red}{\$\{Label\}} \\ \hline
    \end{tabular}
    \caption{Evaluation templates for MultiRC. \$\{$\cdot$\} represents values drawn from a particular data field. We alter the GPT3 template for this task to share a similar format with one of our self-supervised tasks (i.e., classification LLP in this case). The red, boldfaced texts are used to compute the language modeling perplexities for ranking the labels. We note that the shown template is for a single example and there could be multiple examples within an instance.}
    \label{incontext-tab:multirc_template}
\end{table}

As our self-supervised tasks are formatted as input-output pairs, we change the task-specific templates for SuperGLUE to make them more similar to our self-supervised tasks. For example, as shown in \cref{incontext-tab:multirc_template}, we make MultiRC similar to the classification LPP. More details of the template changes are in \cref{incontext-appendix:superglue_templates}.

For both benchmarks, we also report an averaged performance for each model. For SuperGLUE, the average performance is computed based on the means of task performances. When a task has two metrics, we take the average of the two as the task performance.

More details on the dataset statistics and metrics for each task for both benchmarks are in \cref{incontext-appendix:dataset_statistics}.

\paragraph{Baselines.} We consider four baselines: (1) directly evaluating pretrained language models on the benchmarks (LM) ; (2) performing additional language modeling training on the subset of the original data that is used for constructing the self-supervised tasks (ExtraLM). We use ExtraLM to approximately measure the contribution of additional computation; (3) fine-tuning on training sets for the tasks outside the evaluation sets (CrossTask). We use CrossTask to estimate the performances of cross-task supervision from human-annotated datasets; and (4) fine-tuning on training sets for the tasks in the evaluation sets (SameTask). SameTask serves as an oracle baseline estimating the approximated upperbound performances of cross-task supervision.

Since SuperGLUE does not have an official split for the CrossTask setting, we split the datasets into two groups according to the task category and report the CrossTask results based on ``CrossTask (QA$\rightarrow$NLI)'' and ``CrossTask (NLI$\rightarrow$QA)''.\footnote{``QA$\rightarrow$NLI'' suggests that we train models on the NLI tasks and evaluate on the QA tasks. Similarly, for ``NLI$\rightarrow$QA'', we train models on the QA tasks and evaluate on the NLI tasks.} As we alter the task templates, we report results for evaluating the pretrained language model checkpoints using the new templates (NewTemplate) to study the effect of new templates.

\subsection{Experimental Results}

\begin{table}[t]
    \centering\footnotesize
    \setlength{\tabcolsep}{3pt}
    \begin{tabular}{|l|c|c|c|c|c|c|c|}\hline
       \bf Model &\bf MS  &\bf BoolQ & \bf MultiRC &\bf COPA &\bf RTE &\bf CB &\bf Avg. \\\hline
       LM & 125M & 52.1(1.7) & 5.2(0.7)/49.5(1.1) & \bf 67.6(2.3) & 52.0(1.2) & 50.7(3.2)/34.8(2.5) & 48.4 \\
        ExtraLM & 125M & 51.5(1.7) & 5.1(0.8)/49.7(1.0) & 68.0(1.6) & 52.3(1.2) & 49.5(4.6)/35.5(5.6) & 48.3 \\
       NewTemplate & 125M & 52.2(1.8) & 5.2(0.6)/47.9(1.4) & 63.0(2.5) & 50.8(2.0) & 46.4(7.3)/30.1(6.4) & 46.2\\\hline
       CT(NLI$\rightarrow$QA) & 125M & 38.1(0.3) & 5.1(0.7)/43.5(2.5) & 65.4(2.1) & - & - & \multirow{2}{*}{42.2} \\
        CT (QA$\rightarrow$NLI) & 125M & - & - & - & \bf 53.6(0.5) & 39.6(1.5)/19.9(1.2) & \\\hline
        SameTask & 125M & 71.2 & 19.9/66.9 & 72.0 & 67.3 & 71.4/60.2 & 61.9 \\
        Self-Supervised & 125M & \bf 55.7(0.6) & \bf 7.0(1.0)/60.2(0.3) & \bf 67.6(2.1) & 53.0(1.5) & \bf 50.0(5.2)/39.8(3.0) & \bf 51.0 \\\hline\hline
        LM & 1.3B & 48.6(2.3) & 5.5(0.5)/53.7(0.7) & 83.4(1.7) & 51.9(1.2) & 53.6(5.2)/37.2(3.7) & 51.8 \\
        ExtraLM & 1.3B & 49.6(1.9) & 4.9(0.6)/54.8(0.6) & 82.6(1.5) & 52.9(1.9) & 51.4(7.5)/35.6(5.3) & 51.7  \\
        NewTemplate & 1.3B & 51.3(1.3) & 5.0(0.4)/52.8(1.2) & 81.2(2.4) & 50.8(2.3) & 49.3(4.7)/33.7(4.2) & 50.7 \\\hline
       CT (NLI$\rightarrow$QA) & 1.3B & 53.4(0.8) & 1.2(0.3)/57.2(0.3) & 76.2(2.9) & - & - & \multirow{2}{*}{49.6} \\
        CT (QA$\rightarrow$NLI) & 1.3B & - & - & - & \bf 54.3(1.2) & 44.6(3.6)/25.2(4.9) & \\\hline
        SameTask & 1.3B & 77.1 & 27.5/71.6 & 85.0 & 68.1 & 75.2/64.3 & 69.9 \\
        Self-Supervised & 1.3B & \bf 61.7(0.3) & \bf 5.2(0.1)/62.1(0.3) & \bf 84.0(2.7) & 53.1(0.7) & \bf 54.3(2.0)/37.0(1.9) & \bf 55.6 \\\hline
    \end{tabular}
    \caption{SuperGLUE results. We report mean and standard deviations (the numbers in parenthesis) of five runs. The best result (we take the average if there are two metrics) except SameTask in each column for each model size is boldfaced. MS=model size. CT=CrossTask.}
    \label{incontext-tab:main_result_superglue}
\end{table}

\begin{table}[t]
    \centering\small
    \begin{tabular}{|l|c|c|c|c|c|c|}\hline
        \bf Model & \bf MS & \bf QG & \bf AG & \bf MM & \bf VF & \bf Avg. \\\hline
        GPT3 & - & 43.0 & 50.0 & 70.0 & 32.0 & 48.8 \\\hline\hline
        LM & 125M & 33.7 & 12.9 & 53.0 & 14.7 & 28.6 \\ 
        ExtraLM & 125M &\bf 34.4 & 13.4 & 53.7 & 14.3 & 28.9 \\
        CrossTask & 125M & 22.0 & \bf 24.8 & 66.9 & 17.9 & \bf 32.9 \\
        SameTask & 125M & 54.8 & 42.3 & 77.3 & 78.3 & 63.2 \\
        SelfSup. & 125M & 16.9 & 14.6 & \bf 70.1 & \bf 18.9 & 30.0 \\\hline\hline
        LM & 1.3B & 40.9 & 32.5 & 74.0 & 27.8 & 43.8 \\ 
        ExtraLM & 1.3B & 41.1 & 32.7 & \bf 75.9 & 25.2 & 43.7 \\
        CrossTask & 1.3B & 38.1 & 41.6 & 69.2 & 23.0 & 42.9\\
        SameTask & 1.3B & 55.5 & 64.6 & 81.0 & 80.4 & 70.4\\
        SelfSup. & 1.3B & \bf 43.9 & \bf 37.5 & 72.3 & \bf 28.6 & \bf 45.5 \\\hline
    \end{tabular}
    \caption{Natural-Instructions results. The results for GPT3 are taken from \citet{mishra2021crosstask}. The best result except SameTask in each column for each model size is boldfaced. MS=model size.}
    \label{incontext-tab:main_result_naturalinstructions}
\end{table}

We report the results for SuperGLUE and Natural-Instructions in \cref{incontext-tab:main_result_superglue,incontext-tab:main_result_naturalinstructions}. Our findings are as follows:
\begin{enumeratesquish}
\item Our proposed self-supervised training achieves the best performance on average for both benchmarks.
\item ExtraLM and NewTemplate show similar performances as the pretrained language model checkpoints, suggesting that the improvements from our self-supervised training is unlikely to come from the additional training on the data and the task template changes.
\item Compared to the pretrained language model checkpoints, CrossTask shows worse performances on both benchmarks, which is likely due to the differences between training tasks and evaluation tasks.
\end{enumeratesquish}

\subsection{Analysis}

\paragraph{Effect of Amount of Data.}

\begin{figure}
    \centering\small
    \includegraphics[scale=0.5]{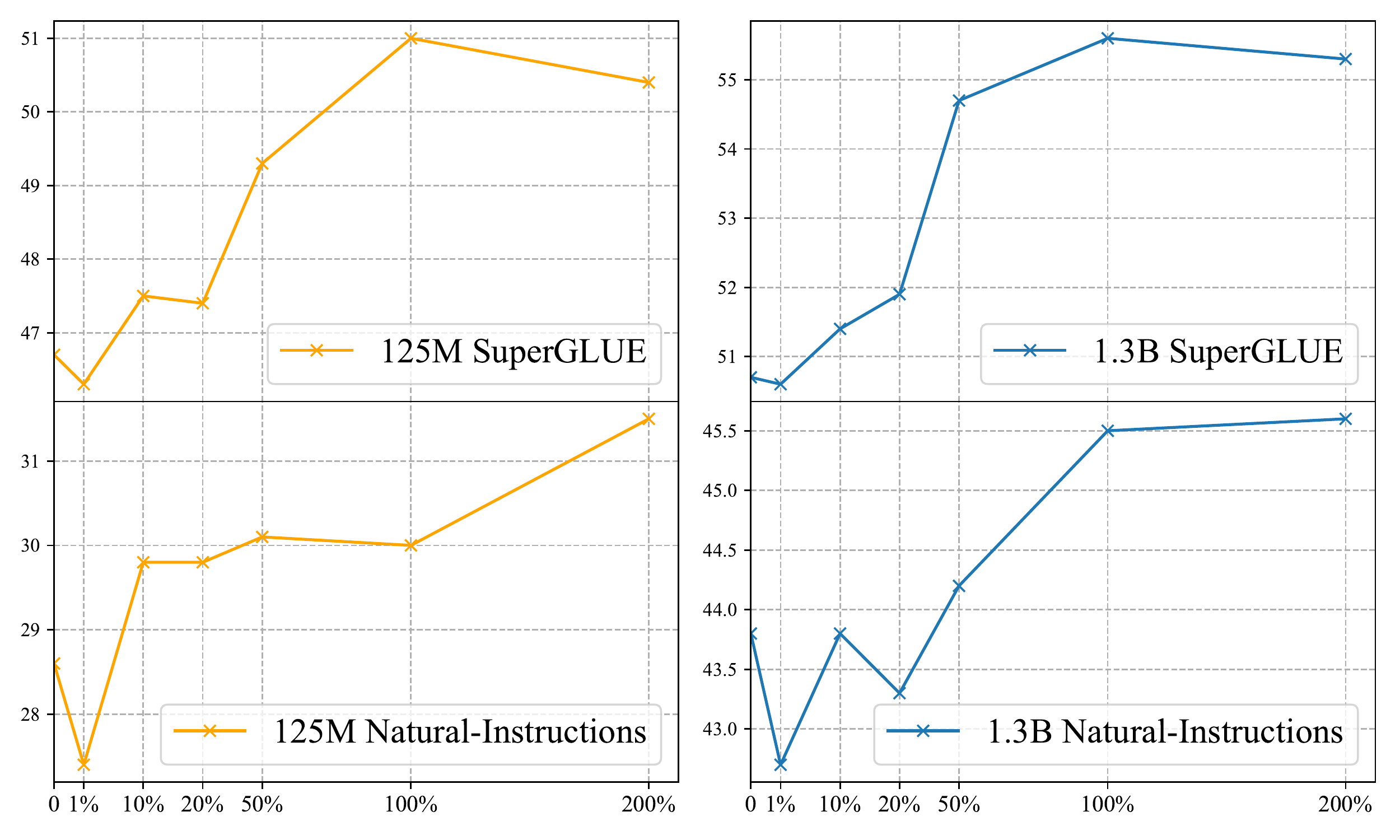}
    \caption{Average results for the 125M and 1.3B models on SuperGLUE and Natural-Instructions when varying the number of examples used for self-supervised training.}
    \label{incontext-fig:data_ratio}
\end{figure}

In \cref{incontext-fig:data_ratio}, we report model performances for the 125M and 1.3B models on SuperGLUE and Natural-Instructions with 1\%, 10\%, 20\%, 50\%, and 200\% of training examples.\footnote{We apply the same ratio to all the self-supervised tasks and use the same development sets for each task across these settings.} We train the models for ten epochs.\footnote{Upon manual inspection, we found that the development set loss values in these experiments have converged.} As shown in the figure, when the amount of training data for self-supervised tasks is similar to that for the CrossTask setting (i.e., 1\% data), the self-supervised tasks also lead to worse performances. The improvements become clearer when we increase the number of training data but it begins to plateau at around 100\% data. This suggests that one of the advantages of the self-supervised tasks compared to the tasks in the CrossTask setting is the amount of training data. We hypothesize that further increasing the amount of data not being helpful is because the data used for constructing the self-supervised tasks has already been used for language model pretraining. So, our models manage to learn to solve these tasks with a relatively limited amount of data. We have similar observations for the 125M model.\footnote{Our goal for this analysis is to show the rough trends of model performance when varying the amount of training data, rather than to provide an exact estimate of the training data required for the self-supervised training.}

\paragraph{Effect of Individual Self-Supervised Tasks.}

\begin{table}
    \centering\small
    \begin{tabular}{|l|c|c|c|}\hline
         & \bf MS & \bf GPT3 Template & \bf Our Template \\\hline
         LM & 125M & 48.4 & 46.2  \\
         SelfSup & 125M & 47.2 & 51.0 \\\hline
         LM & 1.3B & 51.8 & 50.7  \\
         SelfSup & 1.3B & 51.1 & 55.6 \\\hline 
    \end{tabular}
    \caption{Average results for SuperGLUE when using different task templates. MS=model size.}
    \label{incontext-tab:superglue_different_templates}
\end{table}

\begin{table}
    \centering\small
    \begin{tabular}{|l|c|c|c|c|c|c|}\hline
        \bf Model &\bf BQ &\bf MC &\bf CA &\bf RE &\bf CB &\bf Avg. \\\hline
         LM & 52.2 & 26.6 & 63.0 & 50.8 & 38.3 & 46.2 \\
         SelfSup. & 55.7 & 33.6 & 67.6 & 53.0 & 44.9 & 51.0\\\hline
         NSG & 52.1 & 25.9 & 64.0 & 51.0 & 41.2 & 46.9 \\
         CL & 52.5 & 26.8 & 61.4 & 50.9 & 48.1 & 47.9 \\
         MWP & 51.9 & 26.3 & 61.8 & 50.8 & 36.1 & 45.4 \\
         LPP & 53.5 & 29.5 & 61.6 & 52.0 & 40.3 & 47.4 \\\hline
    \end{tabular}
    \caption{SuperGLUE results when training the 125M model with only one of the self-supervised tasks.}
    \label{incontext-tab:effect_selfsupervised_superglue}
\end{table}

\begin{table}[t]
    \centering\small
    \begin{tabular}{|l|c|c|c|c|c|}\hline
        \bf Model & \bf QG & \bf AG & \bf MM & \bf VF & \bf Avg.  \\\hline
        LM & 33.7 & 12.9 & 53.0 & 14.7 & 28.6 \\
        SelfSup. & 16.9 & 14.6 & 70.1 & 18.9 & 30.0 \\\hline
        NSG & 32.3 & 12.5 & 54.0 & 13.8 & 28.2 \\
        CL & 8.3 & 0.3 & 1.0 & 2.7 & 3.1 \\
        MWP & 15.2 & 19.4 & 50.5 & 17.8 & 25.7 \\
        LPP & 11.3 & 16.6 & 49.5 & 19.9 & 24.3 \\\hline
    \end{tabular}
    \caption{Natural-Instructions results when training the 125M model with only one of the self-supervised tasks.}
    \label{incontext-tab:effect_selfsupervised_naturalinstructions}
\end{table}

We investigate the effect of individual self-supervised tasks by training models with only one task. We report the experiment results in \cref{incontext-tab:effect_selfsupervised_superglue,incontext-tab:effect_selfsupervised_naturalinstructions}. Our findings are:

\begin{enumeratesquish}
\item Combining all four self-supervised tasks results in the biggest improvements for most tasks, suggesting that the tasks are complementary.
\item Each self-supervised task improves a few downstream task performances (e.g., NSG helps COPA; CL helps MultiRC and CB). This is likely due to similarities between tasks.
\item It is worth noting that while CL hurts model performances on Natural-Instructions, it helps on the SuperGLUE. We hypothesis that this is because unlike Natural-Instructions, SuperGLUE is ranking based and, therefore, more favorable to classification-related training.
\item It is interesting to see that NSG and CL tasks are the two most beneficial to downstream performance among the four self-supervised tasks. This is likely due to (1) the generic task formulation of NSG, and (2) CL requires different inference abilities compared to the other self-supervised tasks. It is also interesting that training on only one of the self-supervised tasks can hurt the performance on Natural-Instruction.
\end{enumeratesquish}

\paragraph{Effect of More Self-Supervised Tasks.}
To investigate the effect of having more self-supervised tasks during training, we add two extra self-supervised tasks to the self-supervised training, following the same procedure as the other tasks. The additional tasks are: denoising autoencoding \citep{lewis-etal-2020-bart} and gap sentence generation \citep{pegasus-zhang20ae}. Denoising autoencoding is the task of reconstructing the original sentences from sentences corrupted by random noises, which has been shown effective for training generic language representations; gap sentence generation is to recover the missing sentence and has been found useful for abstractive summarization.

\begin{table}
    \centering\small
    \begin{tabular}{|l|c|c|}\hline
         & \bf ALL & \bf ALL+MoreTask \\\hline
         \multicolumn{3}{|c|}{SuperGLUE Results}
         \\\hline
         125M & 51.0 & 50.9 \\
         1.3B & 55.6 & 55.6 \\\hline 
         \multicolumn{3}{|c|}{Natural-Instructions Results} \\\hline
         125M & 30.0 & 31.7 \\
         1.3B & 45.5 & 45.4 \\\hline 
    \end{tabular}
    \caption{Average results when adding denoising autoencoding and gap sentence prediction to the self-supervised training. ALL: use all of the self-supervision described in \cref{incontext-sec:method}.}
    \label{incontext-tab:more_selfsupervised_tasks_results}
\end{table}

We report the results in \cref{incontext-tab:more_selfsupervised_tasks_results} where we do not find adding the two tasks improves downstream tasks. This is likely because the two tasks share similarities with our existing tasks (e.g., gap sentence generation shares a similar inference style as MWP). So, adding them does not promote diversity in the self-supervised tasks, leading to the fact that the models are not encouraged to learn different information.

\paragraph{Effect of Few-Shot Templates.}
\begin{table}
    \centering\small
    \begin{tabular}{|l|c|c|c|}\hline
         & \bf LM & \bf Correct Label & \bf Random Label \\\hline
         \multicolumn{4}{|c|}{SuperGLUE Results}
         \\\hline
         125M & 46.2 & 51.0 & 38.2 \\
         1.3B & 50.7 & 55.6 & 42.5 \\\hline 
         \multicolumn{4}{|c|}{Natural-Instructions Results} \\\hline
         125M & 28.6 & 30.0 & 19.1 \\
         1.3B & 43.8 & 45.5 & 31.5 \\\hline 
    \end{tabular}
    \caption{Average model performance comparing whether we assign random labels to the self-supervised tasks.}
    \label{incontext-tab:corrupt_results}
\end{table}
The self-supervised training brings two benefits: making models familiar with the few-shot templates and task semantics. To differentiate the effect of the two, we train models on the self-supervised tasks with random labels. For example, for NSG, we use random sentences as outputs rather than the true next sentences; for the binary classification tasks, we randomly select binary labels. As shown the results in \cref{incontext-tab:corrupt_results}, random labels hurt model performances, suggesting that what the models have learned is more than the few-shot format.

We also investigate the effect of task templates for SuperGLUE by evaluating models using different templates. We report results in \cref{incontext-tab:superglue_different_templates} where we find that having the templates for downstream tasks similar to the ones used for pretraining gives the models significantly better performance.

\paragraph{Zero-Shot vs. One-Shot vs. Few-Shot.}
\begin{table}
    \centering\small
    \begin{tabular}{|l|c|c|c|c|c|c|}\hline
         & \multicolumn{2}{|c|}{\bf Zero-Shot} & \multicolumn{2}{|c|}{ \bf One-Shot} & \multicolumn{2}{|c|}{\bf Few-Shot} \\
             & LM & SS & LM & SS & LM & SS \\\hline
        125M & 46.7 & 44.3  & 42.6 & 46.1  & 46.2 & 51.0 \\\cline{2-7}
        \multicolumn{1}{|c|}{$\Delta$} &\multicolumn{2}{c|}{(-2.4)}  & \multicolumn{2}{c|}{(+3.5)} & \multicolumn{2}{c|}{\bf (+4.8)}  \\\hline
        1.3B & 49.5 & 49.9 & 46.5 & 50.8  & 50.7 & 55.6 \\\cline{2-7}
        \multicolumn{1}{|c|}{$\Delta$} & \multicolumn{2}{c|}{(+0.4)} & \multicolumn{2}{c|}{(+4.3)} & \multicolumn{2}{c|}{\bf (+4.9)} \\
        \hline
    \end{tabular}
    \caption{Average results for SuperGLUE showing the zero-shot, one-shot, and few-shot model performances for the LM and the self-supervised model (SS). The numbers in parenthesis are the performance differences between the LM and the SS with the positive numbers indicating improvements. We boldface the largest improvement for each model.}
    \label{incontext-tab:zeroshot_vs_oneshot_vs_fewshot}
\end{table}

\begin{table}
    \centering\small\setlength{\tabcolsep}{5pt}
    \begin{tabular}{|l|c|c|c|c|c|c|}\hline
        \bf Model & \bf MS & \bf QG & \bf AG & \bf MM & \bf VF & \bf Avg.  \\\hline
        LM & 125M & \bf 33.7 & 12.9 & 53.0 & 14.7 & 28.6 \\
        CrossTask & 125M & 22.0 & 24.8 & 66.9 & 17.9 & 32.9 \\
        SelfSup. & 125M & 16.9 & 14.6 & 70.1 & \bf 18.9 & 30.0 \\
        Combined & 125M & 23.5 & \bf 25.2 & \bf 70.3 & 18.5 & \bf 34.4 \\\hline\hline
        LM & 1.3B & 40.9 & 32.5 & 74.0 & 27.8 & 43.8 \\
        CrossTask & 1.3B & 38.1 & 41.6 & 69.2 & 23.0 & 42.9\\
        SelfSup. & 1.3B & \bf 43.9 & 37.5 & 72.3 & 28.6 & 45.5 \\
        Combined & 1.3B & 42.1 & \bf 42.5 & \bf 74.1 & \bf 28.7 & \bf 46.9 \\\hline
    \end{tabular}
    \caption{Natural-Instructions results when combining the self-supervised tasks and the tasks in the CrossTask setting. The best performance in each column for each model size is boldfaced. MS=model size.}
    \label{incontext-tab:crosstask_selfsupervised_combine_naturalinstructions}
\end{table}

\paragraph{Generation Examples.}
\begin{table*}
    \centering
    \footnotesize
    \begin{tabular}{|p{0.12\textwidth}|p{0.3\textwidth}|p{0.13\textwidth}|p{0.13\textwidth}|p{0.13\textwidth}|}\hline
        Task Prompt & Task Input & Reference & LM & Self-Supervised \\\hline
        Construct a question from the given fact by a simple rearrangement of words. & Fact: Pollen seeds come from male gametes of plants.
         & what seeds come from male gametes of plants? &  What might cause harm to plants? & What would you use to measure the number of male gametes of plants? \\\hline
        Ask a question on ``event duration'' based on the provided sentence. & Sentence: At the sight of the great man, Spear flushed crimson, and then his look of despair slowly disappeared; and into his eyes there came incredulously hope and gratitude. & How long did Spear see the great man? & How long did he stay in the Embassy? & How long did it take for Spear to look at the great man? \\\hline
        Answer the given question. Your answer must be a single span in the passage. & Passage: ... The following year he won a scholarship to the Royal Academy of Music, ... The principal of the Academy, Sir Alexander Mackenzie, had forbidden ... Question: What was the full name of the school Sir Alexander Mackenzie was principal of? & Royal Academy of Music. & Oliver. & the Royal Academy of Music.\\\hline
        Answer the given question. Your answer must be a single span in the passage. & Passage: ... Epitaph Records, founded by Brett Gurewitz of Bad Religion, was the base for many future pop punk bands ... The mainstream pop punk of latter-day bands such as Blink-182 is criticized by many punk rock devotees; in critic Christine Di Bella's words ... Question: What is the full name of the person that is very critical of modern mainstream pop punk bands? & Christine Di Bella. & the ``Bad Religion''. & many punk rock devotees. \\ \hline
    \end{tabular}
    \caption{Generation examples by the 1.3B model. The examples are taken from Natural-Instructions. The first two examples are from QG, and the other two are from AG. We only show part of the passages relevant to the outputs for QA for brevity.}
    \label{incontext-tab:gen_examples}
\end{table*}

We show zero-shot, one-shot, and few-shot performances for the LM and the self-supervised model in \cref{incontext-tab:zeroshot_vs_oneshot_vs_fewshot}. We find that among the three settings, the self-supervised training is the most helpful in the few-shot setting and does not help in the zero-shot setting, suggesting that the self-supervised training improves the models' in-context learning capabilities.

\paragraph{Combine Self-Supervision with Cross-Task Human-Supervision.}

We investigate the relations between the self-supervised tasks and the human-annotated tasks. We combine the tasks from the self-supervision and those from the CrossTask and report the results in Table \cref{incontext-tab:crosstask_selfsupervised_combine_naturalinstructions}. Interestingly, combining the two kinds of tasks results in better performances on average, showing that they are complementary.

We show generation examples in \cref{incontext-tab:gen_examples}. In general, we find that compared to the vanilla pretrained language models, the self-supervised models are better at using information from task input following task requirements. Specifically, for the first two examples in \cref{incontext-tab:gen_examples}, the LM suffers from more severe semantic drift than the self-supervised model (e.g., ``male gametes of plants'' is more specific and relevant to the task input than ``plants''). We have similar observations for the third example, where ``Oliver'' is a name from the task demonstration rather than the passage. Interestingly, for the last example, the answer generated by the LM is from the passage but is actually ``the base for many future pop punk bands'' instead of what the question looks for (i.e., ``very critical of modern mainstream pop punk bands''). While the answer generated by the self-supervised model does not exactly match the reference, it is partially correct as the mainstream pop punk ``is criticized by many punk rock devotees''.

\section{Summary}
In this chapter, we described the SOP loss for pretraining language models and a pretraining technique for improving model performance on in-context few-shot learning. In \cref{sec:sentence-order-prediction}, we compared SOP to NSP on both intrinsic and extrinsic tasks, finding that the two losses encode different information. We showed that the models pretrained with SOP achieve state-of-the-art performance across various benchmarks. In \cref{sec:incontext-learning}, we benchmarked 4 self-supervised tasks on 13 tasks, finding that similar to human-annotated datasets, self-supervision can also lead to improved downstream task performance. Our analysis uncovered several factors that contributed the improvements, including dataset sizes for the self-supervised tasks, the few-shot templates, and the semantic similarities between the training and evaluation tasks. In addition, we also experimented with concatenating forward and backward language modeling losses to achieve bidirectional training for language models, finding that our proposed approaches showed better performance than the unidirectional language modeling loss but worse than masked language modeling on downstream tasks  (see \cref{appendix-sec:bidir} for more details).
\chapter{Learning Semantic Knowledge from Wikipedia}
\label{CHAPTER:WIKIPEDIA}
In this chapter, we describe our contributions to exploiting rich, naturally-occurring structures on Wikipedia for various NLP tasks. In \cref{sec:wikipedia-entity-representations}, we use hyperlinks to learn entity representations. The resultant models use contextualized representations rather than a fixed set of vectors for representing entities (unlike most prior work). In \cref{sec:wikipedia-discourse-sentence-representations}, we use article structures (e.g., paragraph positions and section titles) to make sentence representations aware of the broader context in which they situate, leading to improvements across various discourse-related tasks. In \cref{sec:wikipedia-concept-hierarchies}, we use article category hierarchies to learn concept hierarchies that improve model performance on textual entailment tasks.

The material in this chapter is adapted from \citet{chen-etal-2019-enteval}, \citet{chen-etal-2019-evaluation}, and \citet{chen-etal-2020-mining}.

\section{Learning Entity Representations from Hyperlinks}
\label{sec:wikipedia-entity-representations}
\subsection{Introduction}

Entity representations play a key role in numerous important problems including language modeling~\citep{ji-etal-2017-dynamic}, 
dialogue generation~\citep{he-etal-2017-learning}, entity linking~\citep{gupta-etal-2017-entity}, and story generation~\citep{clark-etal-2018-neural}. 
One successful line of work on learning entity representations has been learning \emph{static} embeddings: 
that is, assign a unique vector to each entity in the training data
\citep{yamada-etal-2016-joint,gupta-etal-2017-entity,yamada-etal-2017-learning}. 
While these embeddings are useful in many applications, they have the obvious drawback of not accommodating unknown entities. 

Motivated by the recent success of contextualized word representations (henceforth: CWRs) from pretrained models~\citep{mccann-etal-2017-cove,peters-etal-2018-deep,devlin-etal-2019-bert,NEURIPS2019_dc6a7e65,liu2019roberta}, we propose to encode the mention context or 
the description to dynamically represent an entity. 
In addition, we perform an in-depth comparison of ELMo and BERT-based embeddings 
and find that they show different characteristics on different tasks. 
We analyze each layer of the CWRs and make the following observations: 
\begin{itemizesquish}
\item The dynamically encoded entity representations show a strong improvement on the entity disambiguation task compared to prior work using static entity embeddings.
\item BERT-based entity representations require further supervised training to perform well on downstream tasks, while ELMo-based representations are more capable of performing zero-shot tasks. 
\item In general, higher layers of ELMo and BERT-based CWRs are more transferable to entity-related tasks. 
\end{itemizesquish}

To further improve contextualized and descriptive entity representations (CER/DER), we leverage natural hyperlink annotations in Wikipedia. 
We identify effective objectives for incorporating the contextual information in hyperlinks 
and improve ELMo-based CWRs on a variety of entity related tasks.

\subsection{Related Work}

The training objectives considered in this work are built on previous works that involve reasoning over entities. We give a brief overview of relevant works. 

Entity linking is a fundamental task in information extraction 
with a wealth of literature~\citep{he-etal-2013-learning,Guo:2014:REL:2661829.2661887,ling-etal-2015-design, huang2015leveraging,francis-landau-etal-2016-capturing,le-titov-2018-improving,martins-etal-2019-joint}.
The goal of this task is to map a mention in context to the 
corresponding entity in a database. A natural approach is to learn 
entity representations that enable this mapping. 
Recent works focused on learning a fixed embedding for each entity 
using Wikipedia hyperlinks~\citep{yamada-etal-2016-joint,ganea-hofmann-2017-deep,le-titov-2019-boosting}. 
\citet{gupta-etal-2017-entity} additionally train context and description embeddings jointly, but this mainly aims to improve 
the quality of the fixed entity embeddings rather than 
using the context and description embeddings directly; 
we find that their context and description encoders 
perform poorly on \enteval tasks.

A closely related concurrent work by \citep{logeswaran-etal-2019-zero} jointly encodes 
a mention in context and an entity description 
from Wikipedia to perform zero-shot entity linking. 
In contrast, here we seek to pretrain a general purpose entity representations that can function well either 
given or not given entity descriptions or mention contexts.

Other entity-related tasks involve entity typing~\citep{yaghoobzadeh-schutze-2015-corpus,murtyfiner,del-corro-etal-2015-finet,rabinovich-klein-2017-fine,choi-etal-2018-ultra,onoe-durrett-2019-learning,obeidat-etal-2019-description} and coreference resolution~\citep{durrett-klein-2013-easy,wiseman-etal-2016-learning,lee-etal-2017-end,webster-etal-2018-mind,kantor-globerson-2019-coreference}. 

\subsection{Method}

We are interested in two approaches: \contextrep (henceforth: CER) and \descrep (henceforth: DER), both encoding fixed-length vector representations for entities. 

The \contextrep encodes an entity based on the context it appears regardless of whether the entity is seen before. The motivation behind \contextrep is that we want an entity encoder that does not depend on entries in a knowledge base, but 
is capable of inferring knowledge about an entity from the context it appears. 

As opposed to \contextrep, \descrep do rely on entries in Wikipedia. We use a model-specific function $f$ to obtain a fixed-length vector representation 
from the entity's textual description.

\paragraph{Encoders for Contextualized Entity Representations.}
\label{enteval-sec:ctx-ent-rep}
For defining these encoders, we assume we have a sentence $s = (w_1, \dots, w_T)$ where span $(w_i, \dots, w_j)$ refers to an entity mention. When using \elmo, we first encode the sentence: $(c_1, \dots, c_T) = \mathrm{ELMo}(w_1, \cdots, w_T)$, and we use the average of contextualized hidden states corresponding to the entity span as the contextualized entity representation. That is, $f_\text{\elmo}(w_{1:T},i,j)=\frac{\sum_{k=i}^j c_k}{j-i+1}$. 

With \bert, following \citet{onoe-durrett-2019-learning}, we concatenate the full sentence with the entity mention, starting with $\texttt{[CLS]}$ and separating the two by $\texttt{[SEP]}$, i.e.,  $\texttt{[CLS]}, w_1, \dots, w_T, \texttt{[SEP]}, w_i, \dots, w_j, \texttt{[SEP]}$. We encode the full sequence using BERT and use the output from the $\texttt{[CLS]}$ token as the entity mention  representation. 

\paragraph{Encoders for Descriptive Entity Representations.}
\label{enteval-sec:desc-ent-rep}
We encode an entity description by treating the entity description as a sentence, and use the average of the hidden states from \elmo as the entity description representation. 
With \bert, we use the output from the $\texttt{[CLS]}$ token as the description representation. 

\paragraph{Hyperlink-Based Training.}
\label{enteval-sec:hyperlink}

An entity mentioned in a Wikipedia article is often linked to its Wikipedia page, which provides a useful description of the mentioned entity. 
\begin{figure}
    \centering
    \includegraphics[scale=0.6]{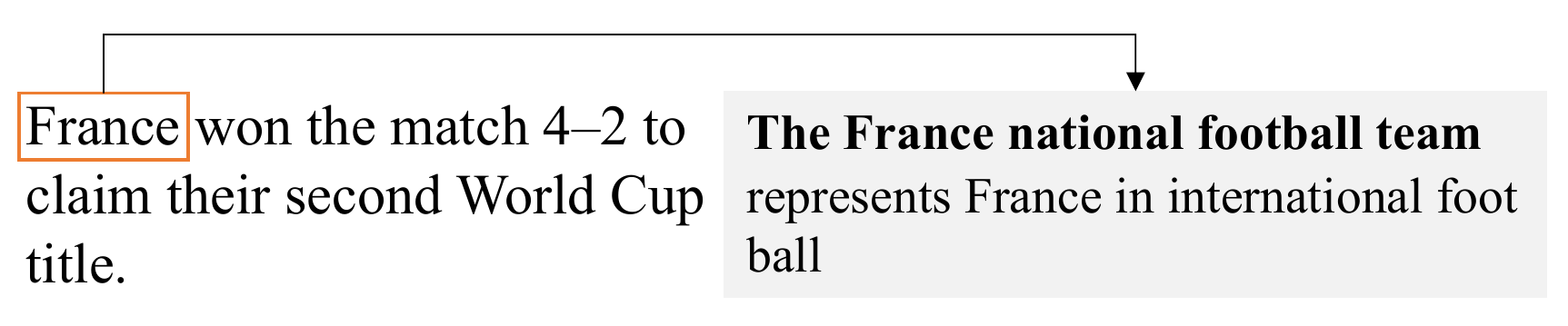}
    \caption{An example of hyperlinks in Wikipedia. ``France'' is linked to the Wikipedia page of ``France national football team'' instead of the country France.}
    \label{enteval-fig:wiki-link}
\end{figure}
The same Wikipedia page may correspond to many different entity mentions. Likewise, the same entity mention may refer to different Wikipedia pages depending on its context. 
For instance, as shown in \cref{enteval-fig:wiki-link}, based on the context, ``France'' is linked to the Wikipedia page of ``France national football team'' instead of the country. 
The specific entity in the knowledge base can be inferred from the context information. 
In such cases, we believe Wikipedia provides valuable complementary information to the current pretrained CWRs such as \bert and \elmo.

To incorporate such information during training, we automatically construct a hyperlink-enriched dataset from Wikipedia. Prior work has used similar resources~\citep{singh12:wiki-links,gupta-etal-2017-entity}. The dataset consists of sentences with contextualized entity mentions and their corresponding descriptions obtained via hyperlinked Wikipedia pages. 
When processing descriptions, we only keep the first 100 word tokens at most as the description of a Wikipedia page; similar truncation has been done in prior work~\citep{gupta-etal-2017-entity}. 
For context sentences, we remove those without hyperlinks from the training data and duplicate those with multiple hyperlinks. We also remove context sentences for which we cannot find matched Wikipedia descriptions. These processing steps result in a training set of approximately 85 million instances and over 3 million unique entities. 

We define a hyperlink-based training objective and add it to \elmo. In particular, we use \contextrep to decode the hyperlinked Wikipedia description, and also use the \descrep to decode the linked context. We use bag-of-words decoders in both decoding processes. More specifically, given a context sentence $x_{1:T_x}$ with mention span $(i,j)$ and a description sentence $y_{1:T_y}$, we use the same bidirectional language modeling loss $l_\text{lang}(x_{1:T_x})+l_\text{lang}(y_{1:T_y})$ in \elmo where $l_\text{lang}$ is defined in \cref{background-eq:elmo-loss}. 
\noindent In addition, we define the two bag-of-words reconstruction losses:
\begin{align*}
    l_\text{ctx}  &= -\sum_t \log q(x_t\vert f_\text{\elmo}(\texttt{[BOD]} y_{1:T_y}, 1, T_y)) \\
    l_\text{desc} &= -\sum_t \log q(y_t\vert f_\text{\elmo}(\texttt{[BOC]} x_{1:T_x}, i, j))
\end{align*}
\noindent where $\texttt{[BOD]}$ and $\texttt{[BOC]}$ are special symbols prepended to sentences to distinguish descriptions from contexts.  
The distribution $q$ is parameterized by a linear layer that transforms 
the conditioning embedding into weights over the vocabulary. 
The final training loss is
\begin{equation}
    l_\text{lang}(x_{1:T_x})+l_\text{lang}(y_{1:T_y})+l_\text{ctx}+l_\text{desc}\nonumber \label{enteval-eq:loss}
\end{equation}
Same as the original \elmo, each log loss is approximated with negative sampling~\citep{jean-etal-2015-using}. 
We write \entelmo to denote the model trained by \cref{enteval-eq:loss}. When using \entelmo for \contextrep  and \descrep, we use it analogously to \elmo.

To evaluate CERs and DERs, we propose a \enteval comprised of a wide range of entity related tasks. Specifically, \enteval consists of the tasks below:
\begin{itemizesquish}
\item The task of entity typing (ET) is to assign types to an entity given only the context of the entity mention. ET is context-sensitive, making it an effective approach to probe the knowledge of context encoded in pretrained representations.
\item Given two entities and the associated context, coreference arc prediction (CAP) seeks to determine whether they refer to the same entity. Solving this task may require the knowledge of entities.
\item The entity factuality prediction (EFP) task involves determining the correctness of statements regarding entities. 
\item The task of contexualized entity relationship prediction (CERP) modeling determines the connection between two entities appeared in the same context.
\item Given two entities with their descriptions from Wikipedia, entity similarity and Relatedness (ESR) is to determine their similarity or relatedness.
\item As another popular resource for common knowledge, we propose a entity relationship typing (ERT) task, which uses Freebase~\citep{bollacker2008freebase} for probing the encoded knowledge by classifying the types of relations between pair of entities.
\item Named entity disambiguation (NED) is the task of linking a named-entity mention to its corresponding instance in a knowledge base such as Wikipedia.
\end{itemizesquish}

For more details about \enteval, see \citet{chen-etal-2019-enteval}. Since our purpose is for examining the learned entity representations, we only use a linear classifier and freeze the entity representations when performing the following tasks. Unless otherwise noted, when the task involves a pair of entities, the input to the classifier are the entity representations $x_1$ and $x_2$, concatenated with their element-wise product and absolute difference: $[x_1, x_2, x_1 \odot x_2, |x_1 - x_2|]$. This input format has been used in SentEval~\citep{conneau-kiela-2018-senteval}.
\subsection{Experiments}

\begin{table*}[t]
\small
    \centering\begin{tabular}{|l|ccccccc|c|}
    \hline
& CAP & CERP & EFP & ET & ESR & ERT & NED & Average \\
\hline
GloVe & 71.9 & 52.6 & 67.0 & 10.3 & 50.9 & 40.8 & 41.2 & 47.8 \\
\bertsize{base} & \bf 80.6 & 65.6 & 74.8 & 32.0 & 28.8 & 42.2 & 50.6 & 53.5 \\
\bertsize{large} & 79.1 & \bf 66.9 & \bf 76.7 & 32.3 & 32.6 & \bf 48.8 & \bf 54.3 & 55.8 \\
ELMo & 80.2 & 61.2 & 75.8 & \bf 35.6 & 60.3 & 46.8 & 51.6 & \bf 58.8 \\
\hline
EntELMo baseline & 78.0 & 59.6 & 71.5 & 31.3 & \bf 61.6 & 46.5 & 48.5 & 56.7 \\
\entelmo & 76.9 & 59.9 & 72.4 & 32.2 & 59.7 & 45.7 & 49.0 & 56.5 \\
\entelmo w/o $l_\text{ctx}$ & 73.5 & 59.4 & 71.1 & 33.2 & 53.3 & 44.6 & 48.9 & 54.9 \\
\entelmo w/ $l_\text{etn}$ & 76.2 & 60.4 & 70.9 & 33.6 & 49.0 & 42.9 & 49.3 & 54.6 \\
\hline
\end{tabular}
    \caption{Performance of entity representations on \enteval tasks. Best performing model in each task is boldfaced. CAP: coreference arc prediction, CERP: contexualized entity relationship prediction, EFP: entity factuality prediction, ET: entity typing, ESR: entity similarity and relatedness, ERT: entity relationship typing, NED: named entity disambiguation. 
    \entelmo baseline is trained on the same dataset as \entelmo but not using the hyperlink-based training. \entelmo w/ $l_\text{etn}$ is trained with a modified version of $l_\text{ctx}$, where we only decode entity mentions instead of the whole context.
    }
    \label{enteval-tab:results}
\end{table*}

\paragraph{Setup.}

As a baseline for hyperlink-based training, we train \entelmo on our constructed dataset with only a bidirectional language model loss. Due to the limitation of computational resources, both variants of \entelmo are trained for one epoch (3 weeks time) with smaller dimensions than \elmo. We set the hidden dimension of each directional LSTM layer to be 600, and project it to 300 dimensions.
The resulting vectors from each layer are thus 600 dimensional. We use 1024 as the negative sampling size for each positive word token. For bag-of-words reconstruction, we randomly sample at most 50 word tokens as positive samples from the the target word tokens. Other hyperparameters are the same as \elmo. \entelmo is implemented based on the official \elmo implementation.\footnote{Our implementation is available at \url{https://github.com/mingdachen/bilm-tf}}

As a baseline for contextualized and descriptive entity representations, we use GloVe word averaging of the entity mention as the ``contextualized'' entity representation, and use word averaging of the truncated entity description text as its description representation. We also experiment two variants of \entelmo, namely \entelmo w/o $l_\text{ctx}$ and \entelmo with $l_\text{etn}$. For second variant, we replace $l_\text{ctx}$ with $l_\text{etn}$, where we only decode entity mentions instead of the whole context from descriptions.
We lowercased all training data as well as the evaluation benchmarks.

We evaluate the transferrability of \elmo, \entelmo, and \bert by using trainable mixing weights for each layer. For \elmo and \entelmo, we follow the recommendation from \citet{peters-etal-2018-deep} to first pass mixing weights through a softmax layer and then multiply the weighted-summed representations by a scalar. For \bert, we find it better to just use unnormalized mixing weights. In addition, we investigate per-layer performance for both models in \cref{enteval-sec:analysis}. Code and data are available at \url{https://github.com/ZeweiChu/EntEval}.

\paragraph{Results.}

\cref{enteval-tab:results} shows the performance of our models on the \enteval tasks. Our findings are detailed below:
\begin{itemizesquish}
\item Pretrained CWRs (ELMo, BERT) perform the best on \enteval overall, indicating that they capture knowledge about entities in contextual mentions or as entity descriptions. 
\item BERT performs poorly on entity similarity and relatedness tasks. 
Since this task is zero-shot, it validates the recommended setting of finetuning BERT on downstream tasks, while the embedding of the $\texttt{[CLS]}$ token does not necessarily capture the semantics of the entity.
\item \bertsize{large} is better than \bertsize{base} on average, showing large improvements in ERT and NED. To perform well at ERT, a model must either glean particular relationships from pairs of lengthy entity descriptions or else leverage knowledge from pretraining about the entities considered.  Relatedly, performance on NED is expected to increase with both the ability to extract knowledge from descriptions and by starting with increased knowledge from pretraining. The Large model appears to be handling these capabilities better than the Base model.

\item \entelmo improves over the \entelmo baseline (trained without the hyperlinking loss) on some tasks but suffers on others. The hyperlink-based training helps on CERP, EFP, ET, and NED. Since the hyperlink loss is closely-associated to the NED problem, it is unsurprising that NED performance is improved. 
Overall, we believe that hyperlink-based training benefits contextualized entity representations but does not benefit descriptive entity representations (see, for example, the drop of nearly 2 points on ESR, which is based solely on descriptive representations). This pattern may be due to the difficulty of using descriptive entity representations to reconstruct their appearing context.

\end{itemizesquish}
\subsection{Analysis}
\label{enteval-sec:analysis}

\paragraph{Is descriptive entity representation necessary?}

\begin{table}
\centering
\setlength{\tabcolsep}{4pt}
\small
\begin{tabular}{|c|cc|cc|cc|}
\hline
& \multicolumn{2}{|c|}{ Rare } & \multicolumn{2}{c|}{CoNLL} & \multicolumn{2}{c|}{ERT}\\
& Des. & Name & Des. & Name & Des. & Name \\
\hline
\elmo & 38.1 & 36.7 & 63.4 & 71.2 & 46.8 & 31.5\\
\bertsize{base} & 42.2 & 36.6 & 64.7 & 74.3 & 42.2 & 34.3\\
\bertsize{large} & 48.8 & 44.0 & 64.6 & 74.8 & 48.8 & 32.6 \\ 
\hline
\end{tabular}
\caption{Accuracies (\%) in comparing the use of description encoder (Des.) to entity name (Name).}
    \label{enteval-tab:name_vs_desc}
\end{table}
A natural question to ask is whether the entity description is needed, as for humans, the entity names carry sufficient amount of information for a lot of tasks. 
To answer this question, we experiment with encoding entity names by the descriptive entity encoder for ERT (entity relationship typing) and NED (named entity disambiguation) tasks. The results in \cref{enteval-tab:name_vs_desc} show that encoding the entity names by themselves already captures a great deal of knowledge regarding entities, especially for \yago. However, in tasks like ERT, the entity descriptions are crucial as the names do not reveal enough information to categorize their relationships.

\begin{table}
\centering
\setlength{\tabcolsep}{5pt}
\small
\begin{tabular}{|c|c|}
\hline
& CoNLL\\
\hline
\elmo & 71.2 \\
\citet{gupta-etal-2017-entity} & 65.1 \\
Deep ED & 66.7 \\
\hline
\end{tabular}
\caption{Accuracies (\%) on \yago with static or non-static entity representations.}
    \label{enteval-tab:conll_yago}
\end{table}

\cref{enteval-tab:conll_yago} reports the performance of different descriptive entity representations on the \yago task. 
The three models all use ELMo as the context encoder. 
``ELMo'' encodes the entity name with ELMo as descriptive encoder, while both \citet{gupta-etal-2017-entity} and Deep ED~\citep{ganea-hofmann-2017-deep} use their trained static entity embeddings. \footnote{We note that the numbers reported here are not strictly comparable to the ones in their original paper since we keep all the top 30 candidates from Crosswiki while prior work employs different pruning heuristics. }
As \citet{gupta-etal-2017-entity} and Deep ED have different embedding sizes from ELMo, %
we add an extra linear layer after them to map to the same dimension. These two models are designed for entity linking, which gives them potential advantages. Even so, ELMo outperforms them both by a wide margin.

\paragraph{Per-Layer Analysis.}
We evaluate each \elmo and \entelmo layer, i.e., the character CNN layer and two bidirectional LSTM layers, as well as each \bert layer on the \enteval tasks.
\begin{figure}
    \centering
    \includegraphics[scale=0.4]{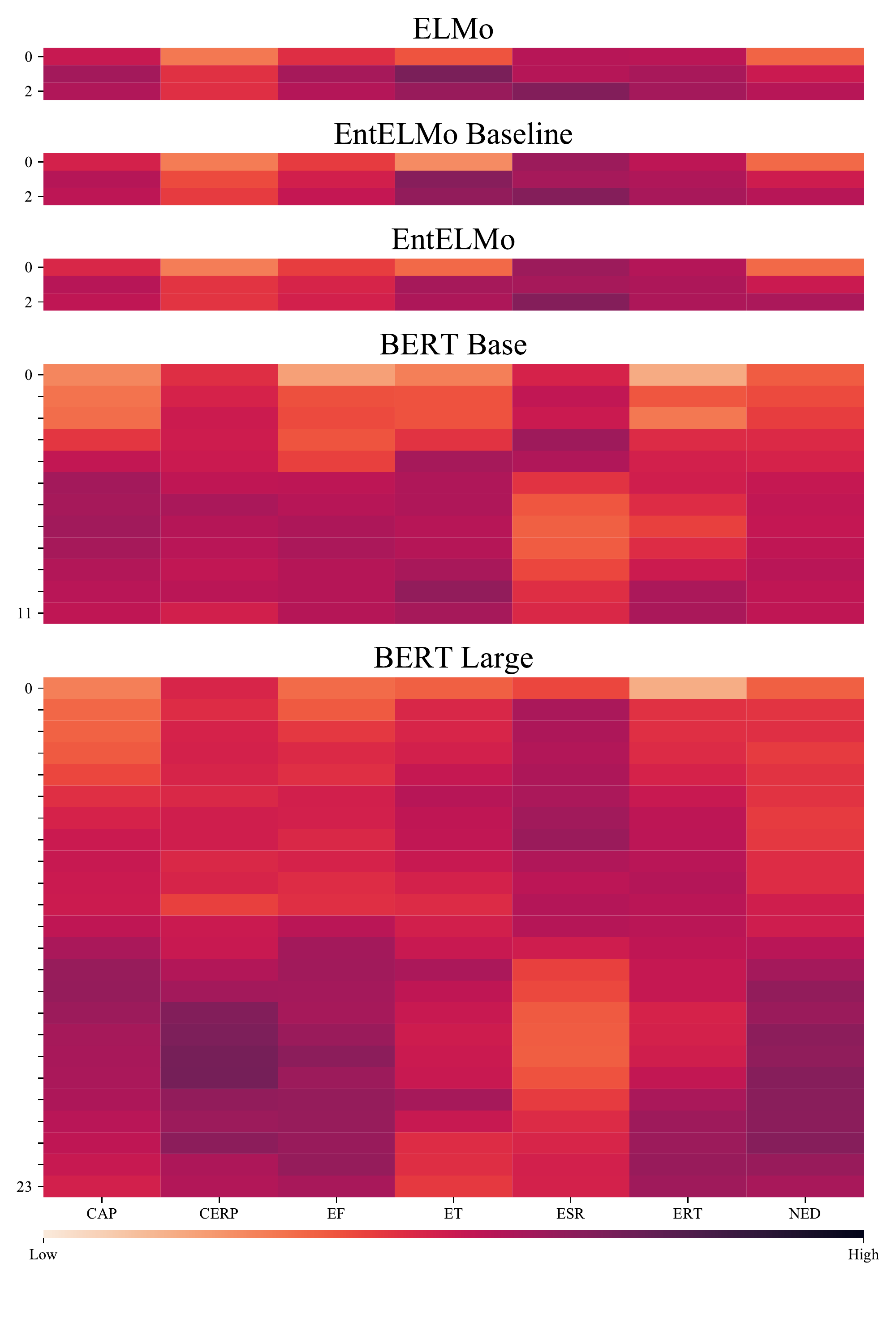}
    \caption{Heatmap showing per-layer performances for \elmo, \entelmo baseline, \entelmo, \bertsize{base}, and \bertsize{large}. }
    \label{enteval-fig:per-layer-analysis}
\end{figure}
\cref{enteval-fig:per-layer-analysis} reveals that for ELMo models, the first and second LSTM layers capture most of the entity knowledge from context and descriptions. 
The BERT layers show more diversity. Lower layers perform better on ESR (entity similarity and relatedness), while for other tasks higher layers are more effective.

\section{Learning Discourse-Aware Sentence Representations from Document Structures}
\label{sec:wikipedia-discourse-sentence-representations}
\subsection{Introduction}

Pretrained sentence representations have been found useful in various downstream tasks such as visual question answering~\citep{tapaswi2016movieqa}, script inference~\citep{pichotta-mooney-2016-using}, and information retrieval~\citep{le-etal-distributed-14, Palangi:2016:DSE:2992449.2992457}. However, the focus of pretrained sentence representations has been primarily on a stand-alone sentence rather than the broader context in which it is situated.

\begin{figure}
\small
\centering
\includegraphics[scale=0.5]{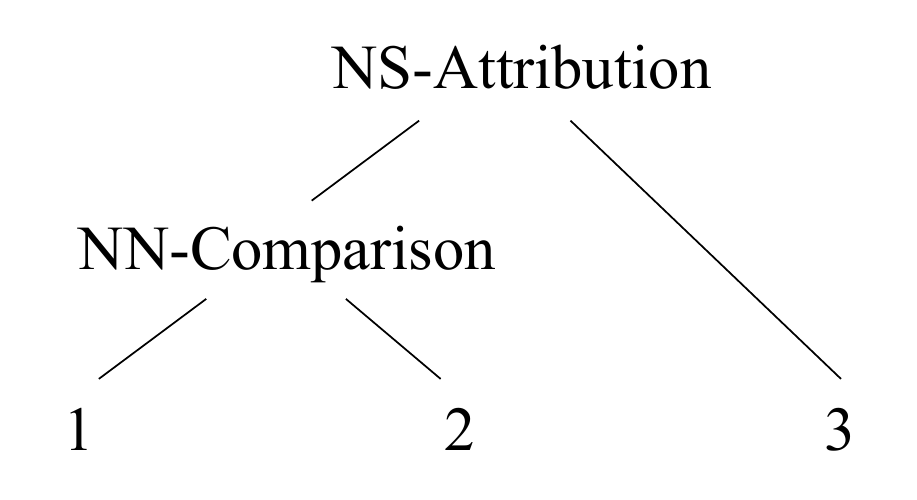}
\captionsetup{font=footnotesize}
\caption*{[The European Community's consumer price index rose a provisional 0.6\% in September from August]\textsubscript{1}
[and was up 5.3\% from September 1988,]\textsubscript{2}
[according to Eurostat, the EC's statistical agency.]\textsubscript{3}}
\captionsetup{font=normal}
\caption{An RST discourse tree from the RST Discourse Treebank. ``N'' represents ``nucleus'', containing basic information for the relation. ``S'' represents ``satellite'', containing additional information about the nucleus.}\label{discoeval:rst-example}
\end{figure}

In this work, we seek to incorporate discourse knowledge in general purpose sentence representations. 
A discourse is a coherent, structured group of sentences that acts as a fundamental type of structure in natural language~\citep{jurafsky-martin-spl}. A discourse structure is often characterized %
by the arrangement of semantic elements across multiple sentences, such as entities and pronouns. The simplest such arrangement (i.e., linearly-structured) can be understood as sentence ordering, where the structure is manifested in the timing of introducing entities. Deeper discourse structures use more complex relations among sentences (e.g., tree-structured; see \cref{discoeval:rst-example}).

Theoretically, discourse structures have been approached through Centering Theory~\citep{grosz1995centering} for studying distributions of entities across text and Rhetorical Structure Theory (RST; \citealp{mann1988rhetorical}) for modelling the logical structure of natural language via discourse trees. Researchers have found modelling discourse useful in a range of tasks~\citep{guzman-etal-2014-using,narasimhan-barzilay-2015-machine,liu-lapata-2018-learning,pan-etal-2018-discourse}, including summarization~\citep{gerani-etal-2014-abstractive}, text classification~\citep{ji-smith-2017-neural}, and text generation~\citep{bosselut-etal-2018-discourse}.

We propose a set of novel multi-task learning objectives building upon standard pretrained sentence encoders, which depend only on the natural structure in structured document collections like Wikipedia. Empirically, we benchmark our models and several popular sentence encoders on our proposed benchmark datasets \disceval and \senteval~\citep{conneau-kiela-2018-senteval}. Specifically, the \disceval has
the following tasks:
\begin{itemizesquish}
\item As the most direct way to probe discourse knowledge, we consider the task of predicting annotated discourse relations among sentences. We use two human-annotated datasets: the RST Discourse Treebank~(RST-DT; \citealp{carlson-etal-2001-building}) and the Penn Discourse Treebank~(PDTB; \citealp{Prasad08pdtb}).
\item \spos (SP), which can be seen as way to probe the knowledge of linearly-structured discourse, where the ordering corresponds to the timings of events.
\item \bso (BSO) is a binary classification task to determine the order of two sentences. The fact that BSO only has a pair of sentences as input makes it different from \spos, where there is more context, and we hope that BSO can evaluate the ability of capturing local discourse coherence in the given sentence representations.
\item Inspired by prior work on chat disentanglement~\citep{elsner-charniak-2008-talking,elsner-charniak-2010-disentangling} and sentence clustering~\citep{wang-etal-2018-picking}, we propose a sentence disentanglement task which seeks to determine whether a sequence of six sentences forms a coherent paragraph.
\item The \ssp (SSP) task is defined as determining the section of a given sentence. The motivation behind this task is that sentences within certain sections typically exhibit similar patterns because of the way people write coherent text.
\end{itemizesquish}

\noindent See \citet{chen-etal-2019-enteval} for more details about the benchmark. We find that our proposed training objectives help the models capture different characteristics in the sentence representations.
Additionally, we find that \elmo shows strong performance on \senteval, whereas \bert performs the best among the pretrained embeddings on \disceval. Both \bert and \skipthought vectors, which have training losses explicitly related to surrounding sentences, perform much stronger compared to their respective prior work, demonstrating the effectiveness of incorporating losses that make use of broader context.
\subsection{Related Work}

Discourse modelling and discourse parsing have a rich history~\cite[\emph{inter alia}]{marcu2000discourse,barzilay-lapata-2008-modeling,zhou-etal-2010-predicting,kalchbrenner-blunsom-2013-recurrent-convolutional,ji-eisenstein-2015-one,li-jurafsky-2017-neural,wang-etal-2018-toward,liu-etal-2018-discourse,lin-etal-2019-unified}, much of it based on recovering linguistic annotations of discourse structure.

There is work on incorporating discourse related objectives into the training of sentence representations. \citet{jernite2017discourse} propose binary sentence ordering, conjunction prediction (requiring manually-defined conjunction groups), and next sentence prediction. Similarly, \citet{sileo-etal-2019-mining} and \citet{nie-etal-2019-dissent} create training datasets automatically based on discourse relations provided in the Penn Discourse Treebank~(PDTB; \citealp{lin-etal-2009-recognizing}). 

Our work differs from prior work in that we propose novel training signals based on document structure, including sentence position and section titles, without requiring additional human annotation. 
\subsection{Method}\label{discoeval-sec:method}
All models in our experiments are composed of a single encoder and multiple decoders.
The encoder, parameterized by a bidirectional GRU, encodes the sentence, either in training or in evaluation of the downstream tasks, to a fixed-length vector representation (i.e., the average of the hidden states across positions).

The decoders take the aforementioned encoded sentence representation, and predict the targets we define in the sections below. We first introduce Neighboring Sentence Prediction, the loss for our baseline model. We then propose additional training losses to encourage our sentence embeddings to capture other context information. 

\paragraph{Neighboring Sentence Prediction (NSP).}

Similar to prior work on sentence embeddings~\citep{kiros-etal-skipthoughts-2015,hill-etal-2016-learning}, 
we use an encoded sentence representation to predict its surrounding sentences. 
In particular, we predict the immediately preceding and succeeding sentences.
All of our sentence embedding models use this loss. Formally, the loss is defined as

\begin{equation}
    \text{NSP}=-\log p_\theta(s_{t-1}\vert s_t)-\log p_\phi(s_{t+1}\vert s_t)\nonumber 
\end{equation}
\noindent where we parameterize $p_\theta$ and $p_\phi$ as separate feedforward neural networks and compute the log-probability of a target sentence using its bag-of-words representation.

\paragraph{Nesting Level (NL).}

A table of contents serves as a high level description of an article, outlining its organizational structure. Wikipedia articles, for example, contain rich tables of contents with many levels of hierarchical structure. The ``nesting level'' of a sentence (i.e., how many levels deep it resides) provides information about its role in the overall discourse. 
To encode this information into our sentence representations, we introduce a discriminative loss to predict a sentence's nesting level in the table of contents:
\begin{equation}
    \text{NL}=-\log p_\theta(l_t\vert s_t)\nonumber
\end{equation}
\noindent where $l_t$ represents the nesting level of the sentence $s_t$ and $p_\theta$ is parameterized by a feedforward neural network. Note that sentences within the same paragraph share the same nesting level. 
In Wikipedia, there are up to 7 nesting levels.

\paragraph{Sentence and Paragraph Position (SPP).}
Similar to nesting level, we add a loss based on using the sentence representation to predict its position in the paragraph and in the article. 
The position of the sentence can be a strong indication of 
the relations between the topics of the current sentence and the topics in the entire article. 
For example, the first several sentences often cover the general topics to be discussed more thoroughly in the following sentences. To encourage our sentence embeddings to capture such information, 
we define a position prediction loss
\begin{equation}
    \text{SPP}=-\log p_\theta(sp_t\vert s_t) - \log p_\phi(pp_t\vert s_t)\nonumber
\end{equation}
\noindent where $sp_t$ is the sentence position of $s_t$ within the current paragraph and $pp_t$ is the position of the current paragraph in the whole document.

\paragraph{Section and Document Title (SDT).}
Unlike the previous position-based losses, this loss makes use of section and document titles, which gives the model more direct access to the topical information at different positions in the document. The loss is defined as
\begin{equation}
    \text{SDT}=-\log p_\theta(st_t\vert s_t)-\log p_\phi(dt_t\vert s_t)\nonumber
\end{equation}
\noindent where $st_t$ is the section title of sentence $s_t$, $dt_t$ is the document title of sentence $s_t$, and $p_\theta$ and $p_\phi$ are two different bag-of-words decoders.

\subsection{Experiments}
\paragraph{Setup.}
We train our models on Wikipedia as it is a knowledge rich textual resource and has consistent structures over all documents. Details on hyperparameters are in the supplementary material. When evaluating on \disceval,
we encode sentences with pretrained sentence encoders. Following \senteval, we freeze the sentence encoders and only learn the parameters of the downstream classifier. The ``Baseline'' row in \cref{discoeval-table:results} are embeddings trained with only the NSP loss. The subsequent rows are trained with extra losses defined in \cref{discoeval-sec:method} in addition to the NSP loss. Code and data are available at \url{https://github.com/ZeweiChu/DiscoEval}.

\begin{table}
    \footnotesize
    \centering\setlength{\tabcolsep}{3pt}
\begin{tabular}{|ccccc|cccccccc|}
\hline
& \multicolumn{4}{c|}{ SentEval } & \multicolumn{8}{c|}{ \disceval } \\
& USS & SSS & SC & Prob & SP & BSO & DC & SSP & PDTB-E & PDTB-I & RST-DT & avg. \\
\hline
\skipthought & 41.7 & 81.2 & 78.4 & 70.1 & 47.5 & 64.6 & 55.2 & 77.5 & 39.3 & 40.2 & \textbf{59.7} & 54.8 \\
\infersent & \textbf{63.4} & \textbf{83.3} & 79.7 & 71.8 & 45.8 & 62.9 & 56.3 & 62.2 & 37.3 & 38.8 & 52.3 & 50.8 \\
\dissent & 50.0 & 79.2 & 80.5 & 74.0 & 47.7 & 64.9 & 54.8 & 62.2 & 42.2 & 40.7 & 57.8 & 52.9 \\
\elmo & 60.9 & 77.6 & 80.8 & 74.7 & 47.8 & 65.6 & \textbf{60.7} & 79.0 & 41.3 & 41.8 & 57.5 & 56.2 \\
\bertsize{base} & 30.1 & 66.3 & 81.4 & 73.9 & 53.1 & 68.5 & 58.9 & 80.3 & 41.9 & 42.4 & 58.8 & 57.7 \\
\bertsize{large} & 43.6 & 70.7 & \textbf{83.4} & \textbf{75.0} & \textbf{53.8} & \textbf{69.3} & 59.6 & \textbf{80.4} & \textbf{44.3} & \textbf{43.6} & 59.1 & \textbf{58.6} \\
\hline
Baseline (NSP) & 57.8 & 77.1 & 77.0 & 70.6 & 47.3 & 63.8 & \underline{61.0} & 77.8 & 36.5 & 39.1 & \underline{56.7} & 54.6 \\
+ SDT & \underline{59.0} & 77.3 & 76.8 & 69.7 & 45.8 & 62.9 & 60.3 & 78.0 & 36.6 & 39.1 & 55.7 & 54.1 \\
+ SPP & 56.0 & 77.5 & \underline{77.4} & \underline{70.7} & 48.4 & \underline{65.3} & 60.2 & 78.4 & \underline{38.1} & 39.9 & 56.4 & 55.2 \\
+ NL & 56.7 & \underline{78.2} & 77.2 & 70.6 & 46.9 & 64.0 & \underline{61.0} & \underline{78.9} & 37.6 & 39.9 & 56.5 & 55.0 \\
+ SPP + NL & 55.4 & 76.7 & 77.0 & 70.4 & \underline{48.5} & 64.7 & 59.9 & \underline{78.9} & 37.8 & \underline{40.5} & \underline{56.7} & \underline{55.3} \\
+ SDT + NL & 58.5 & 76.9 & 77.2 & 70.2 & 46.1 & 63.0 & 60.8 & 78.1 & 36.7 & 38.1 & 56.2 & 54.1 \\
+ SDT +SPP & 58.4 & 77.4 & 76.6 & 70.2 & 46.5 & 63.9 & 60.4 & 77.6 & 35.2 & 38.6 & 56.3 & 54.1 \\
ALL & 58.8 & 76.3 & 77.0 & 70.2 & 46.1 & 63.7 & 60.0 & 78.6 & 36.3 & 37.6 & 55.3 & 53.9 \\
\hline
\end{tabular}
\caption{Results for \senteval and \disceval. The highest number in each column is boldfaced. The highest number for our models in each column is underlined. ``All'' uses all four losses. ``avg.'' is the averaged accuracy for all tasks in \disceval. 
}
  \label{discoeval-table:results}
\end{table}

Additionally, we benchmark several popular pretrained sentence encoders on \disceval, including~\skipthought,\footnote{\href{https://github.com/ryankiros/skip-thoughts}{\nolinkurl{github.com/ryankiros/skip-thoughts}}}~\infersent~\citep{conneau-etal-2017-supervised},\footnote{\href{https://github.com/facebookresearch/InferSent}{\nolinkurl{github.com/facebookresearch/InferSent}}}~\dissent~\citep{nie-etal-2019-dissent},\footnote{\href{https://github.com/windweller/DisExtract}{\nolinkurl{github.com/windweller/DisExtract}}}~\elmo,\footnote{\href{https://github.com/allenai/allennlp}{\nolinkurl{github.com/allenai/allennlp}}} and~\bert.\footnote{\href{https://github.com/huggingface/pytorch-pretrained-BERT}{\nolinkurl{github.com/huggingface/pytorch-pretrained-BERT}}} For \elmo, we use the averaged vector of all three layers and time steps as the sentence representations. For \bert, we use the averaged vector at the position of the \texttt{[CLS]} token across all layers. We also evaluate per-layer performance for both models in \cref{discoeval-sec:analysis}.

When reporting results for \senteval, we compute the averaged Pearson correlations for Semantic Textual Similarity tasks from 2012 to 2016~\citep{agirre-etal-2012-semeval,agirre-etal-2013-sem,agirre-etal-2014-semeval,agirre-etal-2015-semeval,agirre-etal-2016-semeval}. We refer to the average as unsupervised semantic similarity (USS) since those tasks do not require training data. 
We compute the averaged results for the STS Benchmark \citep{cer-etal-2017-semeval}, textual entailment, and semantic relatedness \citep{marelli-etal-2014-semeval} and refer to the average as supervised semantic similarity (SSS). 
We compute the average accuracy for movie review~\citep{pang-lee-2005-seeing}; customer review~\citep{hu2004mining}; opinion polarity~\citep{wiebe2005annotating}; subjectivity classification~\citep{pang-lee-2004-sentimental}; Stanford sentiment treebank~\citep{socher-etal-2013-recursive}; question classification~\citep{li-roth-2002-learning}; and paraphrase detection~\citep{dolan-etal-2004-unsupervised}, and refer to it as sentence classification (SC). For the rest of the linguistic probing tasks~\citep{conneau-etal-2018-cram}, we report the average accuracy and report it as ``Prob''.

\paragraph{Results.}

\cref{discoeval-table:results} shows the experiment results over all \senteval and \disceval tasks. Different models and training signals have complex effects when performing various downstream tasks. We summarize our findings below: 

\begin{itemizesquish}
    \item On \disceval, \skipthought performs best on RST-DT. \dissent performs strongly for PDTB tasks but it requires discourse markers from PDTB for generating training data. \bert has the highest average by a large margin, but \elmo has competitive performance on multiple tasks.
    
    \item The NL or SPP loss alone has complex effects across tasks in \disceval, but when they are combined, the model achieves the best performance, outperforming our baseline by 0.7\% on average. In particular, it yields 40.5\% accuracy on PDTB-I, outperforming Skip-thought by 0.3\%. This is presumably caused by the differing, yet complementary, effects of these two losses (NL and SPP).
    
    \item The SDT loss generally hurts performance on \disceval, especially on the position-related tasks (SP, BSO). This can be explained by the notion that consecutive sentences in the same section are encouraged to have the same sentence representations when using the SDT loss. However, the SP and BSO tasks involve differentiating neighboring sentences in terms of their position and ordering information. 

    \item On \senteval, SDT is most helpful for the USS tasks, presumably because it provides the most direct information about the topic of each sentence, which is a component of semantic similarity. SDT helps slightly on the SSS tasks. NL gives the biggest improvement in SSS.
    \item In comparing \bert to \elmo and \skipthought to \infersent on \disceval, we can see the benefit of adding information about neighboring sentences. Our proposed training objectives show complementary improvements over NSP, which suggests that they can potentially benefit these pretrained representations.

\end{itemizesquish}

\subsection{Analysis}\label{discoeval-sec:analysis}

\paragraph{Per-Layer analysis.}
\begin{figure}
    \centering
    \includegraphics[scale=0.55]{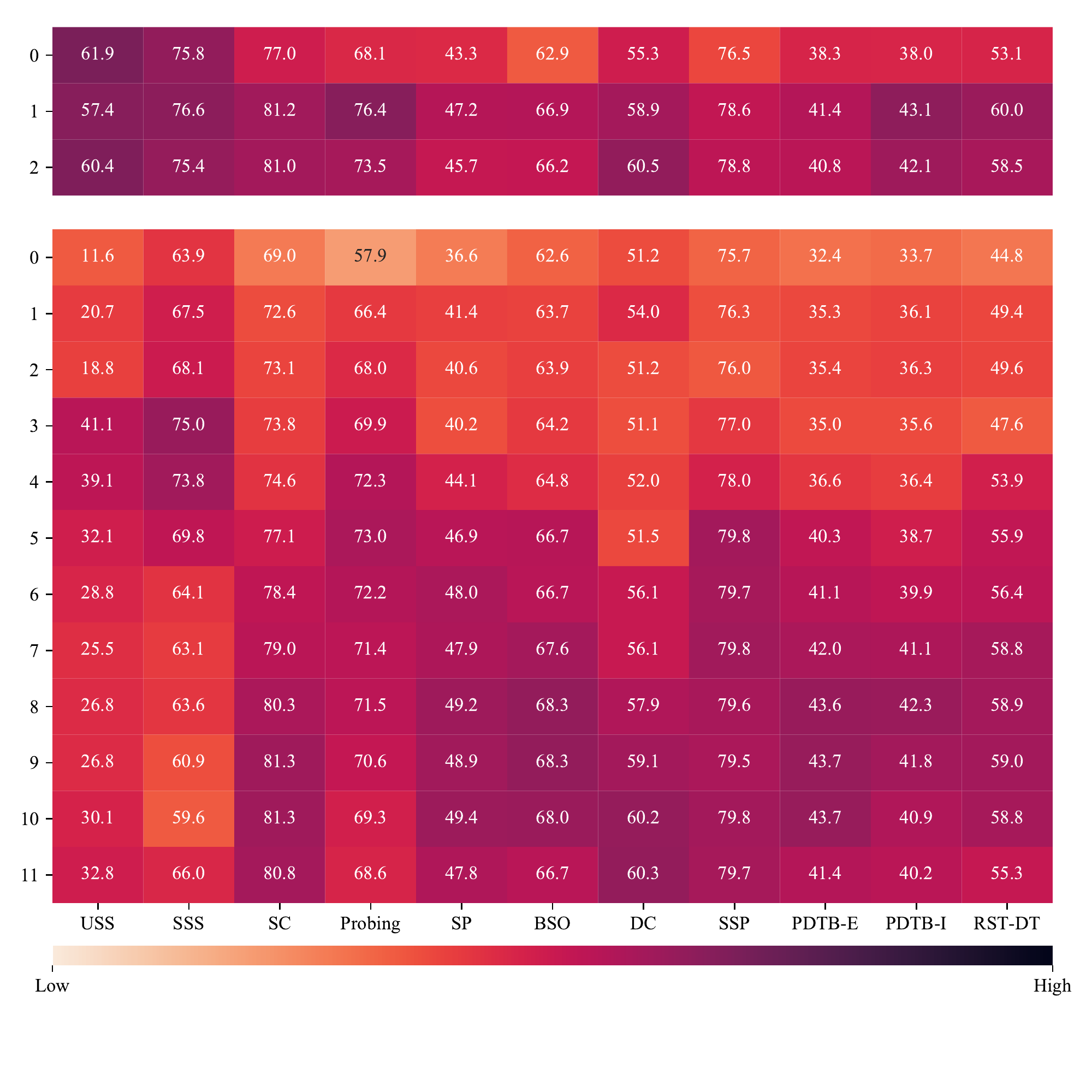}
    \caption{Heatmap for individual hidden layers of \bertsize{base} (lower part) and \elmo (upper part).
    }
    \label{discoeval-fig:bert_elmo_per_layer_perf}
\end{figure}

\begin{table}
    \small
    \centering
    \begin{tabular}{|ccc|}
        \hline
        & \elmo & \bertbase \\
        \hline
        \senteval & 0.8 & 5.0 \\
        \disceval & 1.3 & 8.9 \\
        \hline
    \end{tabular}
    \caption{Average of the layer number for the best layers in \senteval and \disceval. }
    \label{discoeval-table:bert_elmo_best_layer}
\end{table}

To investigate the performance of individual hidden layers, we evaluate \elmo and \bert on both \senteval and \disceval using each hidden layer. For \elmo, we use the averaged vector from the targeted layer. For \bertbase, we use the vector from the position of the \texttt{[CLS]} token. \cref{discoeval-fig:bert_elmo_per_layer_perf} shows the heatmap of performance for individual hidden layers. We note that for better visualization, colors in each column are standardized. On \senteval, \bertbase performs better with shallow layers on USS, SSS, and Probing (though not on SC), but on \disceval, the results using \bertbase gradually increase with deeper layers. To evaluate this phenomenon quantitatively, we compute the average of the layer number for the best layers for both \elmo and \bertbase and show it in \cref{discoeval-table:bert_elmo_best_layer}. 
From the table, we can see that \disceval requires deeper layers to achieve better performance. We assume this is because deeper layers can capture higher-level structure, which aligns with the information needed to solve the discourse tasks.

\paragraph{\disceval architectures.}

\begin{table}
    \small
    \centering
    \begin{tabular}{|cc|}
        \hline
        Baseline w/o hidden layer & 52.0 \\
        Baseline w/ hidden layer & 61.0 \\
        \hline
    \end{tabular}
    \caption{Accuracies with baseline encoder on \dc task, with or without a hidden layer in the classifier.}
    \label{discoeval-table:hidlayer}
\end{table}
In all \disceval tasks except \dcshort, we use no hidden layer in the neural architectures, following the example of SentEval. However, some tasks are unsolvable with this simple architecture. In particular, the \dcshort tasks have low accuracies with all models unless a hidden layer is used. As shown in \cref{discoeval-table:hidlayer}, when adding a hidden layer of 2000 to this task, the performance on \dcshort improves dramatically. This shows that \dcshort requires more complex comparison and inference among input sentences. Our human evaluation below on \dcshort also shows that human accuracies exceed those of the classifier based on sentence embeddings by a large margin.

\paragraph{Human Evaluation.}
\begin{table*}
    \small
    \centering
    \begin{tabular}{|l|ccc|ccc|cc|}
        \hline
        &\multicolumn{3}{c|}{\spos} & \multicolumn{3}{c|}{\bso} & \multicolumn{2}{c|}{\dc} \\
        \hline
        Human & \multicolumn{3}{c|}{77.3} & \multicolumn{3}{c|}{84.7} & \multicolumn{2}{c|}{87.0} \\
        \bertlarge & \multicolumn{3}{c|}{53.8} & \multicolumn{3}{c|}{69.3} & \multicolumn{2}{c|}{59.6} \\
        \hline
                   & Wiki & arXiv & ROC & Wiki & arXiv & ROC & Wiki & Ubuntu \\
        Human      & 84.0 & 76.0  & 94.0 & 64.0 & 72.0  & 96.0  & 98.0  & 74.0 \\
        \bertsize{large} & 50.7 & 47.3  & 63.4 & 70.4 & 66.8  & 70.8  & 65.1  & 54.2 \\
        \hline
    \end{tabular}
    \caption{Accuracies (\%) for a human annotator and \bertsize{large} on \spos, \bso, and \dc tasks.}
    \label{discoeval-table:dc-human}
\end{table*}

We conduct a human evaluation on the \spos, \bso, and \dc datasets.
A native English speaker was provided with 50 examples per domain for these tasks. While the results in \cref{discoeval-table:dc-human} show that the overall human accuracies exceed those of the classifier based on \bertsize{large} by a large margin, we observe that within some specific domains, for example Wiki in BSO, \bertsize{large} demonstrates very strong performance.

\paragraph{Does context matter in \spos?}

\begin{table}
    \small
    \centering
    \begin{tabular}{|cl|}
        \hline
        Random & 20 \\
        Baseline w/o context & 43.2 \\
        Baseline w/ context & 47.3 \\
        \hline
    \end{tabular}
    \caption{Accuracies (\%) for baseline encoder on \spos task when using downstream classifier with or without context.}
    \label{discoeval-table:context}
\end{table}

In the \sposshort task, the inputs are the target sentence together with 4 surrounding sentences. We study the effect of removing the surrounding 4 sentences, i.e., only using the target sentence to predict its position from the start of the paragraph.  

\cref{discoeval-table:context} shows the comparison of the baseline model performance on \spos with or without the surrounding sentences and a random baseline. Since our baseline model is already trained with NSP, it is expected to see improvements over a random baseline. The further improvement from using surrounding sentences demonstrates that the context information is helpful in determining the sentence position.

\section{Learning Concept Hierarchies from Document Categories}
\label{sec:wikipedia-concept-hierarchies}
\subsection{Introduction}

Natural language inference (NLI) is the task of classifying the relationship, such as entailment or contradiction, between sentences.
It has been found useful in downstream tasks, such as summarization \citep{mehdad-etal-2013-abstractive} and long-form text generation \citep{holtzman-etal-2018-learning}.
NLI involves rich natural language understanding capabilities, many of which relate to world knowledge. To acquire such  knowledge, researchers have found benefit from external knowledge bases like \wordnet \citep{wordnet1998}, FrameNet \citep{baker-2014-framenet}, \wikidata \citep{wikidata2014}, and large-scale human-annotated datasets \citep{bowman-etal-2015-large,williams-etal-2018-broad,nie-etal-2020-adversarial}. Creating these resources generally requires expensive human annotation. In this work, we are interested in automatically generating a large-scale dataset from Wikipedia categories that can improve performance on both NLI and lexical entailment (LE) tasks.

One key component of NLI tasks is recognizing lexical and phrasal hypernym relationships. For example, vehicle is a hypernym of car. In this paper, we take advantage of the naturally-annotated Wikipedia category graph, where we observe that most of the parent-child category pairs are entailment relationships, i.e., a child category entails a parent category. Compared to \wordnet and \wikidata, the \wikipedia category graph has more fine-grained connections, which could be helpful for training models.
Inspired by this observation, we construct \wikinli, a dataset for training NLI models constructed automatically from the \wikipedia category graph, by automatic filtering from the \wikipedia category graph. The dataset has 428,899 pairs of phrases and contains three categories that correspond to the entailment and neutral relationships in NLI datasets. 

To empirically demonstrate the usefulness of \wikinli, we pretrain \bert and \roberta on \wikinli, \wordnet, and \wikidata, before finetuning on various LE and NLI tasks. Our experimental results show that \wikinli gives the best performance averaging over 8 tasks for both \bert and \roberta. 

We perform an in-depth analysis of approaches to handling the Wikipedia category graph and the effects of pretraining with \wikinli and other data sources under different configurations. We find that 
\wikinli brings consistent improvements in a low resource NLI setting where there are limited amounts of training data, and the improvements plateau as the number of training instances increases; more \wikinli instances for pretraining are beneficial for downstream finetuning tasks with pretraining on a fourway variant of \wikinli showing more significant gains for the task requiring higher-level conceptual knowledge; \wikinli also introduces additional knowledge related to lexical relations benefiting finer-grained LE and NLI tasks.

We also construct \wikinli in other languages and benchmark several resources on XNLI \citep{conneau-etal-2018-xnli}, showing that \wikinli benefits performance on NLI tasks in the corresponding languages.
\subsection{Related Work}
\label{wikinli-sec:related}

We build on a rich body of literature on leveraging specialized resources 
(such as knowledge bases) to enhance model performance.
These works either (1) pretrain the model on datasets extracted from such resources, 
or (2) use the resources directly by changing the model itself. 

The first approach aims to improve performance at test time by designing 
useful signals for pretraining, for instance using hyperlinks \citep{logeswaran-etal-2019-zero,chen-etal-2019-enteval}
or document structure in Wikipedia \citep{chen-etal-2019-evaluation}, 
knowledge bases \citep{logan-etal-2019-baracks}, and discourse markers \citep{nie-etal-2019-dissent}. 
Here, we focus on using category hierarchies in Wikipedia. 
There are some previous works that also use category relations derived from knowledge bases
\citep{shwartz-etal-2016-improving,riedel-etal-2013-relation}, but they are used in a particular form of distant supervision in which they are matched 
with an additional corpus to create noisy labels. 
In contrast, we use the category relations directly without requiring such additional steps. \citet{onoe2020fine} use the direct parent categories of hyperlinks for training entity linking systems.

Within this first approach, there have been many efforts aimed at harvesting inference rules from raw text~\citep{Lin:2001:DIR:973890.973894,szpektor-etal-2004-scaling,bhagat-etal-2007-ledir,szpektor-dagan-2008-learning,Yates:2009:UMD:1622716.1622724,bansal-etal-2014-structured,berant-etal-2015-efficient,hosseini-etal-2018-learning}. 
Since \wikinli uses category pairs in which one is a hyponym of the other, it is more closely related to work in extracting hyponym-hypernym pairs from text~\citep{hearst-1992-automatic,NIPS2004_2659,snow-etal-2006-semantic,pasca-07,mcnamee-etal-2008-learning,le-etal-2019-inferring}. \citet{pavlick-etal-2015-adding} automatically generate a large-scale phrase pair dataset with several relationships by training classifiers on a relatively small amount of human-annotated data. However, most of this prior work uses raw text or raw text combined with either annotated data or curated resources like WordNet. \wikinli, on the other hand, seeks a middle road, striving to find large-scale, naturally-annotated data that can improve performance on NLI tasks. 

The second approach aims to enable the model to leverage knowledge resources during prediction,
for instance by computing attention weights over lexical relations in WordNet \citep{chen-etal-2018-neural} 
or linking to reference entities in knowledge bases within the transformer block \citep{peters-etal-2019-knowledge}. 
While effective, this approach requires nontrivial and domain-specific modifications of the model itself.  
In contrast, we develop a simple pretraining method to leverage knowledge bases that 
can likewise improve the performance of already strong baselines such as BERT
without requiring such complex model modifications.
\subsection{Experimental Setup}
To demonstrate the effectiveness of \wikinli, we pretrain \bert and \roberta on \wikinli and other resources, and then finetune them on several NLI and LE tasks. We assume that if a pretraining resource is better aligned with downstream tasks, it will lead to better downstream performance of the models pretrained on it.

\paragraph{Training.} Following \citet{devlin-etal-2019-bert} and \citet{liu2019roberta}, we use the concatenation of two texts as the input to \bert and \roberta. Specifically, for a pair of input texts $x_1$, $x_2$, the input would be $\texttt{[CLS]} x_1 \texttt{[SEP]} x_2 \texttt{[SEP]}$. We use the encoded representations at the position of \texttt{[CLS]} as the input to a two-layer classifier, and finetune the entire model.

We start with a pretrained \bertlarge or \robertalarge model and further pretrain it on different pretraining resources. After that, we finetune the model on the training sets for the downstream tasks, as we will elaborate on below.

\paragraph{Evaluation.}
We use several NLI and LE datasets. Statistics for these datasets are shown in \cref{wikinli-tab:dataset_stats} and details are provided below. 

\begin{table}[t]
    \small
    \centering
    \begin{tabular}{|l|c|c|c|c|}
    \hline
        \textbf{dataset} & \textbf{\#train} & \textbf{\#dev} & \textbf{\#test} & \textbf{\#train per cat.} \\\hline
        \multicolumn{5}{|c|}{Natural Language Inference} \\\hline
        MNLI & 3,000 & 9,815 & 9,796 & 1,000 \\\hline
        SciTail & 3,000 & 1,304 & 2,126 & 1,500 \\\hline
        RTE & 2,490 & 277 & 3,000 & 1,245 \\\hline
        PPDB & 13,904 & 4,633 & 4,641 & 1,545 \\\hline
        Break & - & - & 8,193 & - \\\hline
        \multicolumn{5}{|c|}{Lexical Entailment} \\\hline
        K2010 & 739 & 82 & 621 & 370 \\\hline
        B2012 & 791 & 87 & 536 & 396 \\\hline
        T2014 & 539 & 59 & 507 & 270 \\\hline
    \end{tabular}
    \caption{Dataset statistics.}
    \label{wikinli-tab:dataset_stats}
\end{table}

\paragraph{MNLI.} The Multi-Genre Natural Language Inference (MNLI;~\citealp{williams-etal-2018-broad}) dataset is a human-annotated multi-domain NLI dataset. MNLI has three categories: entailment, contradiction, and neutral. Since the training split for this dataset has a large number of instances, models trained on it are capable of picking up information needed regardless of the quality of the pretraining resources we compare, which makes the effects of pretraining resources negligible. To better compare pretraining resources, we simulate a low-resource scenario by randomly sampling 3,000 instances\footnote{The number of training instances is chosen based on the number of instances per category, as shown in the last column of \cref{wikinli-tab:dataset_stats}, where we want the number to be close to 1-1.5K.} from the original training split as our new training set, but use the standard ``matched'' development and testing splits.

\paragraph{SciTail.} SciTail is created from science questions and the corresponding answer candidates, and premises from relevant web sentences retrieved from a large corpus \citep{scitail}. SciTail has two categories: entailment and neutral. Similar to MNLI, we randomly sample 3,000 instances from the training split as our training set.

\paragraph{RTE.}
We evaluate models on the GLUE \citep{wang-etal-2018-glue} version of the recognizing textual entailment (RTE) dataset \citep{dagan2006pascal,bar2006second,giampiccolo-etal-2007-third,bentivogli2009fifth}. RTE is a binary task, focusing on identifying if a pair of input sentences has the entailment relation.

\paragraph{PPDB.} We use the human-annotated phrase pair dataset from \citet{pavlick-etal-2015-adding}, which has 9 text pair relationship labels. The labels are: hyponym, hypernym, synonym, antonym, alternation, other-related, NA, independent, and none. We directly use phrases in PPDB to form input data pairs. We include this dataset for more fine-grained evaluation. 
Since there is no standard development or testing set for this dataset, we randomly sample 60\%/20\%/20\% as our train/dev/test sets.

\paragraph{Break.} \citet{glockner-etal-2018-breaking} constructed a challenging NLI dataset called ``Break'' using external knowledge bases such as \wordnet. Since sentence pairs in the dataset only differ by one or two words, similar to a pair of adversarial examples, it has broken many NLI systems.

Due to the fact that Break does not have a training split, we use the aforementioned subsampled MNLI training set as a training set for this dataset. 
We select the best performing model on the development set of MNLI and evaluate it on Break. 

\paragraph{Lexical Entailment.} We use the lexical splits for 3 datasets from \citet{levy-etal-2015-supervised}, including K2010 \citep{kotlerman-etal-2009-directional}, B2012 \citep{baroni-etal-2012-entailment}, and T2014 \citep{Turney2015ExperimentsWT}. These datasets all similarly formulate lexical entailment as a binary task, and they were constructed from diverse sources, including human annotations, \wordnet, and \wikidata.

\paragraph{Baselines.}
We consider three baselines for both \bert and \roberta, namely the original model, the model pretrained on \wordnet, and the model pretrained on \wikidata.

\wordnet is a widely-used lexical knowledge base, where words or phrases are connected by several lexical relations. We consider direct hyponym-hypernym relations available from \wordnet, resulting in 74,645 pairs. 

\wikidata is a database that stores items and relations between these items. Unlike \wordnet, \wikidata consists of items beyond word types and commonly seen phrases, offering more diverse domains similar to \wikinli. The available conceptual relations in \wikidata are: ``subclass of'' and ``instance of''. In this work, we consider the ``subclass of'' relation in \wikidata because (1) it is the most similar relation to category hierarchies from Wikipedia; (2) the relation ``instance of'' typically involves more detailed information, which is found less useful empirically (see the supplementary material for details). The filtered data has 2,871,194 pairs.

We create training sets from both \wordnet and \wikidata following the same procedures used to create \wikinli. All three datasets are constructed from their corresponding parent-child relationship pairs. Neutral pairs are first randomly sampled from non-ancestor-descendant relationships and top ranked pairs according to cosine similarities of ELMo embeddings are kept. We also ensure these datasets are balanced among the three classes. Code and data are available at \url{https://github.com/ZeweiChu/WikiNLI}.

\subsection{Experimental Results}

\begin{table*}
    \small
    \centering
    \begin{tabular}{l|c|c|c|c|c|c|c|c|c|}
        \cline{2-9}
        \multirow{2}*{} & \multicolumn{5}{|c|}{Natural Language Inference} & \multicolumn{3}{c|}{Lexical Entailment}\\ \cline{2-10}
        & MNLI & RTE & PPDB & Break & SciTail & K2010  & B2012 & T2014 & avg. \\
        \hline
        \multicolumn{1}{|l|}{\bert} & 75.0 & 69.9 & 66.7 & 80.2 &\underline{92.3} & \underline{85.2} &  79.4 & 63.3  & 76.5 \\
        \multicolumn{1}{|l|}{$+$\wordnet} & 75.8 & \underline{71.3} & 71.1 & 83.5 & 90.8 & 83.5 & 94.3 & \underline{71.2} & 80.2\\
        \multicolumn{1}{|l|}{$+$\wikidata} & 75.7 & \underline{71.3} & \bf\underline{75.0} & 81.3 & 91.5 & 82.3 & 95.3 & 70.5 & 80.4 \\
        \multicolumn{1}{|l|}{$+$\wikinli} & \underline{76.4} & 70.9 & 70.7 & \bf\underline{85.7}& 91.8  & 84.9 & \bf\underline{96.1}  & \underline{71.2} & \underline{81.0} \\
        \hline
        \multicolumn{1}{|l|}{\roberta} & 82.5 & 78.8 & 65.9 & 81.3 & 93.6 & 85.3 & 65.9 & 66.8 &  77.5 \\
        \multicolumn{1}{|l|}{$+$\wordnet} & 83.8 & 82.2 & 72.0 & 82.3 & \bf\underline{93.9}   & 82.5& 88.6  & 70.7 & 82.0 \\
        \multicolumn{1}{|l|}{$+$\wikidata} &  84.0 & 82.3 & \underline{72.5} & 83.2 & 92.9  & 82.4 & 94.8 & 71.0 & 82.9 \\
        \multicolumn{1}{|l|}{$+$\wikinli} & \bf\underline{84.4} &\bf \underline{83.1} & 71.7 & \underline{83.8}& 93.0 & \bf\underline{85.4}  & \underline{95.7} & \bf\underline{72.9} &\bf\underline{83.8} \\\hline
    \end{tabular}
    \caption{Test set performance for baselines and models pretrained on various resources. We report accuracy (\%) for NLI tasks and $F_1$ score (\%) for LE tasks. The highest results for each model (\bert or \roberta) are underlined. The highest numbers in each column are boldfaced.
    } 
    \label{wikinli-tab:results}
\end{table*}

The results are summarized in \cref{wikinli-tab:results}. 
In general, pretraining on \wikinli, \wikidata, or \wordnet improves the performances on downstream tasks, and pretraining on \wikinli achieves the best performance on average. Especially for Break and MNLI, \wikinli can lead to much more substantial gains than the other two resources. Although \bertlarge + \wikinli is not better than the baseline \bertlarge on RTE, \roberta + \wikinli shows much better performance. Only on PPDB, \wikidata is consistently better than \wikinli. We note that \bertlarge + \wikinli still shows a sizeable improvement over the \bertlarge baseline.
More importantly, the improvements to both \bert and \roberta brought by \wikinli show that the benefit of the \wikinli dataset can generalize to different models. We also note that pretraining on these resources has little benefit for SciTail.
\subsection{Analysis}

We perform several kinds of analysis, including using \bert to compare the effects of different settings. Due to the submission constraints of the GLUE leaderboard, we will report dev set results (medians of 5 runs) for the tables in this section, except for Break which is only a test set.

\paragraph{Lexical Analysis.}

\begin{table}[t]
    \centering
    \small
\begin{tabular}{|c|c|c|}\hline
\wikinli & \wikidata & \wordnet \\\hline
albums & protein & genus \\
songs & gene & dicot \\
players & putative & family \\
male & protein-coding & unit \\
people & conserved & fish \\
American & hypothetical & tree \\
British & languages & bird \\
writers & disease & person \\
(band) & RNA & fern \\
female & language & plant \\
templates & function & mammal \\
music & process & monetary \\
articles & group & animal \\
women & unknown & process \\
films & syndrome & arthropod \\
French & activity & order \\
artists & cell & acid \\
German & binding & disease \\
rock & food & vein \\
musicians & mineral & system \\
culture & transport & herb \\
\hline
\end{tabular}
    \caption{Top 20 most frequent words in \wikinli, \wikidata, and \wordnet.}
    \label{wikinli-tab:lexical_analysis}
\end{table}

To qualitatively investigate the differences between \wikinli, \wikidata, and \wordnet, we list the top 20 most frequent words in these three resources in \cref{wikinli-tab:lexical_analysis}. Interestingly, \wordnet contains mostly abstract words, such as ``unit'', ``family'', and ``person'', while \wikidata contains many domain-specific words, such as ``protein'' and ``gene''. 
In contrast, \wikinli strikes a middle ground, covering topics broader than those in \wikidata but less generic than those in \wordnet.

\paragraph{Fourway vs. Threeway vs. Binary Pretraining.}

We investigate the effects of the number of categories for \wikinli by empirically comparing three settings: fourway, threeway, and binary classification.
For fourway classification, we add an extra relation ``sibling'' in addition to child, parent, and neutral relationships. A sibling pair consists of two categories that share the same parent. We also ensure that neutral pairs are non-siblings, meaning that we separate a category that was considered as part of the neutral relations to provide a more fine-grained pretraining signal.

We construct two versions of \wikinli with binary class labels. One classifies the child against the rest, including parent, neutral, and sibling (``child vs. rest''). The other classifies child or parent against neutral or sibling (``child/parent vs. rest''). The purpose of these two datasets is to find if a more coarse training signal would reduce the gains from pretraining. %

These dataset variations are each balanced among their classes and contain 100,000 training instances and 5,000 development instances. 

\begin{table}
    \small\setlength{\tabcolsep}{4pt}
    \centering
    \begin{tabular}{l|c|c|c|c|c|}
        \cline{2-6}
        & MNLI & RTE & PPDB & Break  & avg. \\\hline
        \multicolumn{1}{|l|}{Threeway} & \bf 75.6 & \bf 74.4 & \bf 71.2 & 85.7 & \bf 76.7 \\
        \multicolumn{1}{|l|}{Fourway}  & \bf 75.6 & 74.0 & 69.8  & \bf 86.9 & 76.6 \\
        \multicolumn{1}{|l|}{Binary (C vs. R)}  & 75.1 & 72.6 & 70.5 & 81.7 & 75.0 \\
        \multicolumn{1}{|l|}{Binary (C/P vs. R)}  & 74.3 & 72.2 & 69.8 & 80.5 & 74.3  \\\hline
    \end{tabular}
    \caption{Comparing binary, threeway, and fourway classification for pretraining. C = child; P = parent; R = rest. The highest numbers in each column are boldfaced.}
    \label{wikinli-tab:results_fourway}
\end{table}

\cref{wikinli-tab:results_fourway} shows results
of MNLI, RTE, and PPDB.
Overall, fourway and threeway classifications are comparable, although they excel at different tasks. Interestingly, we find that pretraining with  child/parent vs.~rest is worse than pretraining with child vs.~rest. We suspect this is because the child/parent vs.~rest task resembles topic classification. The model does not need to determine direction of entailment, but only whether the two phrases are topically related, as neutral pairs are generally either highly unrelated or only vaguely related. 
The child vs.~rest task still requires reasoning about entailment as the models still need to differentiate between child and parent. 

In addition, we explore pruning levels in Wikipedia category graphs, and incorporating sentential context, finding that relatively higher levels of knowledge from \wikinli have more potential of enhancing the performance of NLI systems and sentential context shows promising results on the Break dataset (see supplementary material for more details).

\paragraph{Larger Training Sets.}
\label{wikinli-sec:large_train_set}
\begin{table}[t]
    \small\setlength{\tabcolsep}{4pt}
    \centering
    \begin{tabular}{l|c|c|c|c|c|}
        \cline{2-6}
        & MNLI & RTE & PPDB & Break  & avg. \\\hline
        \multicolumn{1}{|l|}{Threeway 100k}  & 75.6 & 74.4 & \bf 71.2 & 85.7 & 76.7 \\
        \multicolumn{1}{|l|}{Fourway 100k}  & 75.6 & 74.0  & 69.8  & 86.9 & 76.6  \\
        \multicolumn{1}{|l|}{Threeway 400k} & \bf 75.7 & \bf 75.5 & 70.9 & 83.0
        & 76.3 \\
        \multicolumn{1}{|l|}{Fourway 400k} & 75.6 & 75.1 & 70.8 & \bf 89.5
        & \bf 77.8 \\
        \hline
    \end{tabular}
    \caption{The effect of the number of \wikinli pretraining instances. The highest numbers in each column are boldfaced.}
    \label{wikinli-tab:results_full_dataset}
\end{table}
We train on larger numbers of \wikinli instances, approximately 400,000, for both threeway and fourway classification. We note that we only pretrain models on \wikinli for one epoch as it leads to better performance on downstream tasks. The results are in \cref{wikinli-tab:results_full_dataset}. We observe that except for PPDB, adding more data generally improves performance. For Break, we observe significant improvements when using fourway \wikinli for pretraining, whereas threeway \wikinli seems to hurt the performance.

\paragraph{\wikipedia Pages, Mentions, and Layer Pruning.}\label{wikinli-sec:mention-and-pages}
\begin{table}[t]
    \small\setlength{\tabcolsep}{4pt}
    \centering
    \begin{tabular}{l|c|c|c|c|c|}
        \cline{2-6}
        & MNLI & RTE & PPDB & Break  & avg. \\\hline
        \multicolumn{1}{|l|}{\bert} & 74.4 & 71.8 & 68.6 & 80.2 & 73.3  \\
        \multicolumn{1}{|l|}{\bert \& \wikinli}  & \bf 75.6 & \bf 74.4 & \bf 71.2 & 85.7 & \bf 76.7 \\
        \multicolumn{1}{|l|}{$-$ one cat. layer} & 74.6 & 73.3 & 70.9 & \bf 87.0 & 76.5 \\
        \multicolumn{1}{|l|}{$-$ two cat. layers}   & 75.4  &  72.9  & \bf 71.2  & 82.5 & 75.5 \\
        \multicolumn{1}{|l|}{$+$ page titles} & 74.2 & 73.6 & 70.6 & 80.7 & 74.8 \\
        \hline
    \end{tabular}
    \caption{Comparing pruning levels for hierarchies available in \wikipedia. The highest numbers in each column are boldfaced. }
    \label{wikinli-tab:results_pruning}
\end{table}
The variants of \wikinli we considered so far have used categories as the lowest level of hierarchies. We are interested in whether adding \wikipedia page titles would bring in additional knowledge for inference tasks.

We experiment with including \wikipedia page titles that belong to \wikipedia categories to \wikinli. We treat these page titles as the leaf nodes of the \wikinli dataset. Their parents are the categories that the pages belong to.

Although \wikipedia page titles are additional source of information, they are more specific compared to \wikipedia categories. A majority of \wikipedia page titles are person names, locations, or historical events. They are not general summaries of concepts. To explore the effect of more general concepts, we try pruning leaf nodes from the \wikinli category hierarchies. 
As higher-level nodes are more general and abstract concepts compared to lower-level nodes, we hypothesize that pruning leaf nodes would make the model learn higher-level concepts. 
We experiment with pruning one layer and two layers of leaf nodes in \wikinli category hierarchies. 

\cref{wikinli-tab:results_pruning} compares the results of adding page titles and pruning different numbers of layers. Adding page titles mostly gives relatively small improvements to the model performance on downstream tasks, which shows that the page title is not a useful addition to \wikinli. Pruning layers also slightly hurts the model performance. One exception is Break, which shows that solving it requires knowledge of higher-level concepts. 

\paragraph{\wikisent.}
\begin{table*}
\setlength{\tabcolsep}{5pt}
\footnotesize
\begin{center}
\begin{tabular}{|p{0.45\textwidth}|p{0.45\textwidth}|c|}
\hline
Sentence 1 & Sentence 2 & Rel. \\
\hline
He then moved to \textbf{Scottish society} as an actuary & He then moved to \textbf{Edinburgh} as an actuary & \multirow{3}{*}{P}\\
for Standard Life Assurance Company. However, & for Standard Life Assurance Company. However, & \\
he transferred back to London with the company. & he transferred back to London with the company.  & \\\hline
Dobroselo () is a village in \textbf{Croatia} . It is & Dobroselo () is a village in \textbf{Southern European} & \multirow{4}{*}{C} \\
connected by the D218 highway. According to & \textbf{countries} . It is connected by the D218 highway.  & \\
the 2011 census, Dobroselo had 117 inhabitants. & Accordingto the 2011 census, Dobroselo had 117 & \\
& inhabitants.  & \\\hline
His oldest brother Charuhasan, like Kamal, is a & His oldest brother Charuhasan, like Kamal, is a & \multirow{3}{*}{N} \\
National Film Award-winning actor who appeared &  National Film Award-winning actor who appeared & \\
in the \textbf{ladino-language} film "Tabarana Kathe". & in the \textbf{Kannada} film "Tabarana Kathe". & \\\hline
\end{tabular}
\end{center}
\caption{\label{wikinli-tab:example_wikisent}Examples from \wikisent. C = child; P = parent; N = neutral. Boldfaced are page titles, \wikinli categories, or their mentions in the context. }
\end{table*}

\begin{figure}
    \centering
    \includegraphics[scale=0.4]{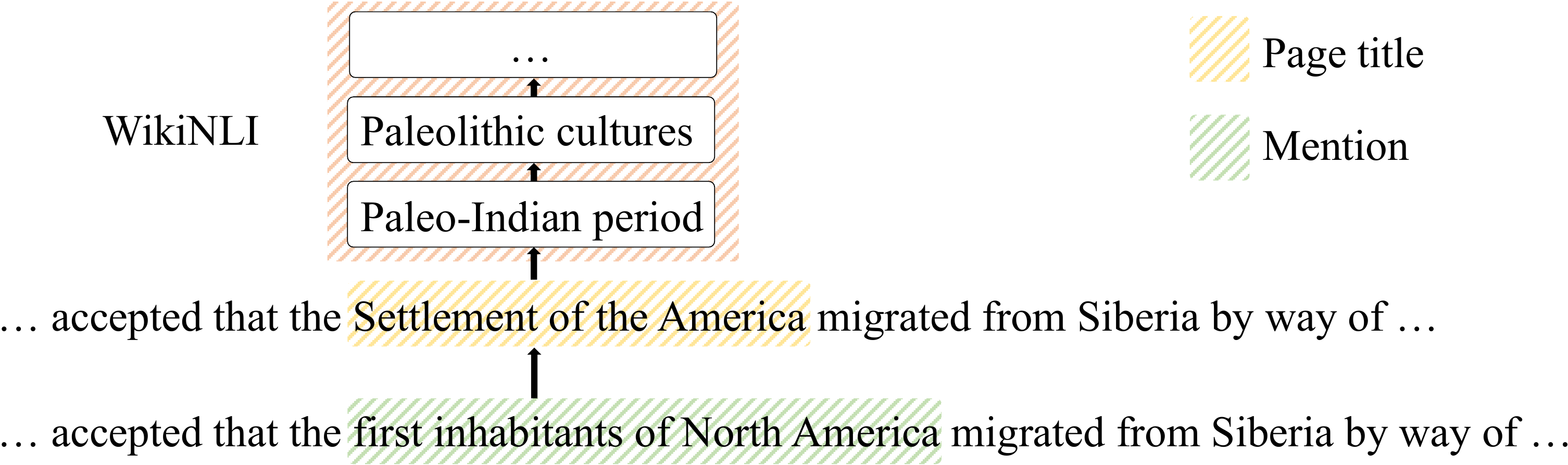}
    \caption{An example of \wikisent and higher-level categories that are used to construct \wikinli.}
    \label{wikinli-fig:wikisent_example}
\end{figure}

To investigate the effect of sentential context, we construct another dataset, which we call \wikisent, that is made up of full sentences. The general idea is to create sentence pairs that only differ by several words by using the hyperlinks in the \wikipedia sentences. More specifically, for a sentence with a hyperlink (if there are multiple hyperlinks, we will consider them as different instances), we form new sentences by replacing the text mention (marked by the hyperlink) with the page title as well as the categories describing that page. 
We consider these two sentences forming candidate child-parent relationship pairs. An example is shown in \cref{wikinli-fig:wikisent_example}.
As some page titles or category names do not fit into the context of the sentence, we score them by \bertsize{large}, averaging over the loss spanning that page title or category name. We pick the candidate with the lowest loss. 
To generate neutral pairs, we randomly sample 20 categories for a particular page mention in the text and pick the candidate with the lowest loss by \bertsize{large}. 
\wikisent is also balanced among three relations (child, parent and neutral), and we experiment with 100k training instances and 5k development instances. 
\cref{wikinli-tab:example_wikisent} are some examples from \wikisent. 

\begin{table}
    \small\setlength{\tabcolsep}{4pt}
    \centering
    \begin{tabular}{l|c|c|c|c|c|}
        \cline{2-6}
        & MNLI & RTE & PPDB & Break  & avg. \\
        \hline
        \multicolumn{1}{|l|}{\bert} & 74.4 & 71.8 & 68.6 & 80.2 & 73.3  \\
        \multicolumn{1}{|l|}{\wikinli} &  \bf 76.4 & \bf 74.4 & \bf 71.2 & \bf 85.7 & \bf 76.7 \\
        \multicolumn{1}{|l|}{$+$ page \& mention}  & 72.2 & 69.0 & 70.8 & 58.5 & 67.6  \\
        \multicolumn{1}{|l|}{\wikisent} & 67.0 & 62.8 & 69.1 & 56.9 & 64.0 \\
        \multicolumn{1}{|l|}{\wikisent cat.} & 71.8 & 67.1 & 70.6 & 84.0 & 73.4\\
        \hline
    \end{tabular}
    \caption{Comparison using \wikisent. The highest numbers in each column are boldfaced.}
    \label{wikinli-tab:results_wikisent}
\end{table}

\cref{wikinli-tab:results_wikisent} shows the results. In comparing \wikinli to \wikisent, we observe that adding extra context to \wikinli does not help on the downstream tasks. It is worth noting that the differences between \wikinli and \wikisent are more than sentential context. The categories we considered in \wikisent are always immediately after \wikipedia pages, limiting the exposure of higher-level categories.

To look into the importance of those categories, we construct another version of \wikisent by treating the mentions and page title layer as the same level (``\wikisent cat.''). This effectively gives models pretrained on this version of \wikisent access to higher-level categories. Practically, when creating child sentences, we randomly choose between keeping the original sentences or replacing the text mention with its linked page title. When creating parent sentences, we replace the text mention with the parent categories of the linked page. Then, we perform the same steps as described in the previous paragraph. Pretraining on  \wikisent cat.~gives a sizable improvement compared to pretraining on \wikisent.

Additionally, we try to add mentions to \wikinli, which seems to impair the model performance greatly. This also validates our claim that specific knowledge tends to be noisy and less likely to be helpful for downstream tasks. More interestingly, these variants seem to affect Break the most, which is in line with our previous finding that Break favors higher-level knowledge. While most of our findings with sentential context are negative, the \wikisent cat.~variant shows promising improvements over \bert in some of the downstream tasks, demonstrating that a more appropriate way of incorporating higher-level categories can be essential to benefit from \wikisent in practice.

\paragraph{Combining Multiple Data Sources.}

\begin{table}[t]
    \small\setlength{\tabcolsep}{4pt}
    \centering
    \begin{tabular}{l|c|c|c|c|c|}
        \cline{2-6}
        & MNLI & RTE & PPDB & Break  & avg. \\\hline
        \multicolumn{1}{|l|}{\TC{1} 100k}      & 75.6 & 74.4 & 71.2 & 85.7 & 76.7 \\
        \multicolumn{1}{|l|}{\TC{1} 50k}      & 74.9 & 74.7  & 70.8 & 76.9  & 74.3  \\
        \multicolumn{1}{|l|}{\TC{1} 50k + \TC{2} 50k} & 75.0 & 71.5 &  70.9 & 80.2 & 74.4 \\
        \multicolumn{1}{|l|}{\TC{1} 50k + \TC{3} 50k} & 75.0  &  73.6  & 70.7 & 81.5 &  75.3 \\
        \hline
    \end{tabular}
    \caption{Combining \wikinli with other datasets for pretraining. \TC{1}=\wikinli; \TC{2}=\wordnet; \TC{3}=\wikidata.}
    \label{wikinli-tab:results_combine}
\end{table}
We combine multiple data sources for pretraining. In one setting we combine 50k instances of \wikinli with 50k instances of \wordnet, while in the other setting we combine 50k instances of \wikinli with 50k instances of \wikidata. \cref{wikinli-tab:results_combine} compares these two settings for pretraining. \wikinli works the best when pretrained alone.

\begin{table*}
    \footnotesize
    \centering
    \begin{tabular}{|l|l|c|c|c|c|c|}
    \hline
    phrase 1 & phrase 2 & gold & \bert & \wikinli & \wordnet & \wikidata \\ \hline
        car & the trunk & hypernym & other-related & hypernym & hypernym & hypernym \\
        return & return home & hypernym & synonym & hypernym & hypernym & hypernym \\
        boys are & the children are & hyponym & synonym & hyponym & hyponym & hyponym \\
        foreign affairs & foreign minister & other-related & hypernym & other-related & hypernym & hypernym \\
        company & debt & other-related & independent & independent & other-related & other-related \\ 
        europe & japan & alternation & hypernym & alternation & independent & alternation\\ 
        family & woman & independent & independent & hypernym & independent & other-related\\\hline
    \end{tabular}
    \caption{Examples from PPDB development set showing the effect of pretraining resources.}
    \label{wikinli-tab:effect_pretrain}
\end{table*}
\paragraph{Effect of Pretraining Resources.}
\begin{table}
    \small
    \centering
    \begin{tabular}{c|c|c|c|c|}\cline{2-5}
                 & antonym & alternation  & hyponym & hypernym \\ \hline
        \multicolumn{1}{|l|}{w/ } & 34& 51 & 276 & 346  \\
        \multicolumn{1}{|l|}{w/o}& 1 & 35 & 231 & 248 \\\hline
    \end{tabular}
    \caption{Per category numbers of correctly predicted instances by \bert with or without pretraining on \wikinli.}
    \label{wikinli-tab:effect_category}
\end{table}
We show several examples of predictions from PPDB in \cref{wikinli-tab:effect_pretrain}. In general, we observe that without pretraining, \bert tends to predict symmetric categories, such as synonym, or other-related, instead of predicting entailment-related categories. For example, the phrase pair ``car'' and ``the trunk'', ``return'' and ``return home'', and ``boys are'' and ``the children are''. These are either ``hypernym'' or ``hyponym'' relationship, but \bert tends to conflate them with symmetric relationships, such as other-related. To quantify this hypothesis, we compute the numbers of correctly predicted antonym, alternation, hyponym and hypernym and show them in \cref{wikinli-tab:effect_category}.\footnote{We observed similar trends when pretraining on the other resources.} It can be seen that with pretraining those numbers increase dramatically, showing the benefit of pretraining on these resources.

We also observe that the model performance can be affected by the coverage of pretraining resources. In particular, for phrase pair ``foreign affairs'' and ``foreign minister'',
\wikinli has a closely related term ``foreign affair ministries'' and ``foreign minister'' under the category ``international relations'', whereas \wordnet does not have these two, and \wikidata only has ``foreign minister''.

As another example, consider the phrase pair ``company'' and ``debt''. In \wikinli, ``company'' is under the ``business'' category and debt is under the ``finance'' category. They are not directly related. In \wordnet, due to the polysemy of ``company'', ``company'' and ``debt'' are both hyponyms of %
``state'', and in \wikidata, they are both a subclass of ``legal concept''.

For the phrase pair ``family''/``woman'',
in \wikinli, ``family'' is a parent category of ``wives'', and in \wikidata, they are related in that ``family'' is a subclass of ``group of humans''. In contrast, \wordnet does not have such knowledge.

\paragraph{Finetuning with Different Amounts of Data.}

\begin{table}[t]
    \small
    \centering
    \begin{tabular}{l|c|c|c|c|c|}
        \cline{2-6}
        & 2k & 3k & 5k & 10k & 20k  \\
        \hline
        \multicolumn{6}{|c|}{MNLI}\\\hline
        \multicolumn{1}{|l|}{\bert} & 72.2  & 74.4 & 76.6 & 78.8 & 80.4  \\
        \multicolumn{1}{|l|}{\wikinli} & 74.5 & 75.6 & 77.3  & 79.1 & 80.6 \\\hline
        \multicolumn{1}{|c|}{$\Delta$} & +2.3 & +1.2 & +0.7  & +0.3 & +0.2 \\\hline
        \multicolumn{6}{|c|}{PPDB}\\ \hline
        \multicolumn{1}{|l|}{\bert} & 55.5 & 59.2 & 59.9  & 68.1 & 68.6 \\
        \multicolumn{1}{|l|}{\wikinli} & 65.0 & 66.4 & 67.9  & 70.2 & 71.2 \\\hline
        \multicolumn{1}{|c|}{$\Delta$} & +9.5 & +7.2 & +8.0 & +2.1 & +2.6 \\\hline
    \end{tabular}
    \caption{Results for varying numbers of MNLI or PPDB training instances. The rows ``$\Delta$'' show improvements from \wikinli. We use all the training instances for PPDB in the ``20k'' setting.
    }
    \label{wikinli-tab:results_mnli}
\end{table}
We now look into the relationship between the benefit of \wikinli and the number of training instances from downstream tasks (\cref{wikinli-tab:results_mnli}). We compare \bertlarge to \bertlarge pretrained on \wikinli when finetuning on 2k, 3k, 5k, 10k, and 20k MNLI or PPDB training instances accordingly. In general, the results show that \wikinli has more significant improvement with less training data, and the gap between \bertlarge and \wikinli narrows as the training data size increases. We hypothesize that the performance gap does not reduce as expected between 3k and 5k or 10k and 20k due in part to the imbalanced number of instances available for the categories. For example, even when using 20k training instances, some of the PPDB categories are still quite rare.

\paragraph{Evaluating on Adversarial NLI.}

\begin{table}
    \centering
    \small
    \begin{tabular}{|c|c|c|c|}\hline
& R1 & R2 & R3 \\\hline
BERT & 39.8 & 37.0 & 41.3 \\
$+$ \wordnet & 41.1 & 38.2 & 39.9 \\
$+$ \wikidata & 43.2 & 39.0 & 41.8 \\
$+$ \wikinli & 39.6 & 38.2 & 39.3 \\\hline
RoBERTa & 46.1 & 39.3 & 39.4 \\
$+$ \wordnet & 53.7 & 38.7 & 37.9 \\
$+$ \wikidata & 51.5 & 39.6 & 39.8 \\
$+$ \wikinli & 51.2 & 38.1 & 39.4 \\\hline
\end{tabular}
    \caption{Test results for ANLI.}
    \label{wikinli-tab:anli}
\end{table}

Adversarial NLI (ANLI; \citealp{nie-etal-2020-adversarial}) is collected via an iterative human-and-model-in-the-loop procedure. ANLI has three rounds that progressively increase the difficulty. When finetuning the models for each round, we use the sampled 3k instances from the corresponding training set, perform early stopping on the original development sets, and report results on the original test sets. As shown in \cref{wikinli-tab:anli}, our pretraining approach has diminishing effect as the round number increases. This may due to the fact that humans deem the NLI instances that require world knowledge as the hard ones, and therefore when the round number increases, the training set is likely to have more such instances, which makes pretraining on similar resources less helpful.
\cref{wikinli-tab:anli} also shows that \wordnet and \wikidata show stronger performance than \wikinli. We hypothesize that this is because ANLI has a context length almost 3 times longer than MNLI on average, in which case our phrase-based resources or pretraining approach are not optimal choices. Future research may focus on finding better ways to incorporate sentential context into \wikinli. For example, we experiment with such a variant of \wikinli (i.e., \textsc{WikiSentNLI}) in the supplementary material.

We have similar observations that our phrase-based pretraining has complicated effect (e.g., only part of the implicature results shows improvements) when evaluating these resources on IMPPRES \citep{jeretic-etal-2020-natural}, which focuses on the information implied in the sentential context (please refer to the supplementary materials for more details).

\subsection{Multilingual \wikinli}

\begin{table}
    \small\setlength{\tabcolsep}{6pt}
    \centering
    \begin{tabular}{l|c|c|c|c|c|}
        \cline{2-6}
        & fr & ar & ur & zh & avg. \\\hline
        \multicolumn{1}{|l|}{mBERT}      & 61.5  & 57.3  & 49.3 & 57.9 & 56.5 \\
        \multicolumn{1}{|l|}{m\wikinli}  & 62.5 & 56.8 & 51.5 & 59.9 & 57.7 \\
        \multicolumn{1}{|l|}{tr\wikinli}  & 63.0 & \bf 57.7  & 51.3  &  59.9  & 58.0 \\
        \multicolumn{1}{|l|}{\wikinli}  & \bf 63.3 & 57.1 & \bf 51.8  &\bf 60.0 & \bf 58.1 \\
        \multicolumn{1}{|l|}{\wikidata} & 63.2 & 56.9 &  49.5 & 59.8 & 57.4 \\
        \multicolumn{1}{|l|}{\wordnet}  & 63.1 & 56.0 & 50.5 & 58.6 & 57.1 \\\hline
    \end{tabular}
    \caption{Test set results for XNLI. m\wikinli is constructed from Wikipedia in other languages. tr\wikinli is translated from the English \wikinli. The highest numbers in each column are boldfaced.}
    \label{wikinli-tab:results_other_langs}
\end{table}

Wikipedia has different languages, which naturally motivates us to extend \wikinli to other languages.
We mostly follow the same procedures as English \wikinli to construct a multilingual version of \wikinli from Wikipedia in other languages, except that (1) we filter out instances that contain English words for Arabic, Urdu, and Chinese; and (2) we translate the keywords into Chinese when filtering the Chinese \wikinli. 
We will refer to this version of \wikinli as ``m\wikinli''. As a baseline, we also consider ``tr\wikinli'', where we translate the English \wikinli into other languages using Google Translate. We benchmark these resources on XNLI in four languages: French (fr), Arabic (ar), Urdu (ur), and Chinese (zh).
When reporting these results, we pretrain multilingual BERT (mBERT; \citealp{devlin-etal-2019-bert}) on the corresponding resources, finetune it on 3000 instances of the training set, perform early stopping on the development set, and test it on the test set. We always use XNLI from the corresponding language. In addition, we pretrain mBERT on English \wikinli, \wikidata, and \wordnet, finetune and evaluate them on other languages using the same language-specific 3000 NLI pairs mentioned earlier. 
We note that when pretraining on m\wikinli or tr\wikinli, we use the versions of these datasets with the same languages as the test sets. 

\cref{wikinli-tab:results_other_langs} summarizes the test results on XNLI. In general, pretraining on \wikinli gives the best results. \citet{phang-etal-2020-english} also observed that training on English intermediate tasks helps in cross-lingual tasks but in a zero-shot setting. While m\wikinli is not the best resource, it still gives better results on average than \wikidata, \wordnet, and no pretraining at all. The  exception is Arabic, where only tr\wikinli performs better than the mBERT baseline. In comparing among different versions of \wikinli, we find that tr\wikinli  performs almost as good as \wikinli, but for Urdu, tr\wikinli is the worst resource among the three. 
The variance of tr\wikinli may arise from the variable quality of machine translation across languages.

\begin{CJK*}{UTF8}{gbsn}
\begin{table}
    \centering
    \small
\begin{tabular}{|c|c|}\hline
\wikinli & Chinese m\wikinli \\\hline
albums & 中国 (China) \\
songs & 中华人民共和国 (P. R. C.) \\
players & 行政区划 (administrative division) \\
male & 人 (man) \\
people & 政治 (politics) \\
American & 人物 (people) \\
British & 各国 (countries) \\
writers & 组织 (organization) \\
(band) & 各省 (provinces) \\
female & 建筑物 (building) \\
templates & 美国 (American) \\
music & 历史 (history) \\
articles & 作品 (work) \\
women & 文化 (culture) \\
films & 属 (category) \\
French & 校友 (alumnus) \\
artists & 官员 (officer) \\
German & 地理 (geography) \\
rock & 公司 (company) \\
musicians & 城市 (city) \\
culture & 单位 (unit) \\
\hline
\end{tabular}

    \caption{Top 20 most frequent words in \wikinli, and m\wikinli in Chinese. Each Chinese word is followed by a translation in parenthesis.
    }
    \label{wikinli-tab:en_vs_zh_word_list}
\end{table}
\end{CJK*}

The accuracy differences between  m\wikinli and \wikinli could be partly attributed to domain differences across languages. To measure the  differences, we compile a list of the top 20 most frequent words in the Chinese m\wikinli, shown in Table \cref{wikinli-tab:en_vs_zh_word_list}. The most frequent words for m\wikinli in Chinese are mostly related to political concepts, whereas \wikinli offers a broader range of topics. 

Future research will be required to obtain a richer understanding of how training on \wikinli benefits non-English languages more than training on the language-specific m\wikinli resources. One possibility is the presence of emergent cross-lingual structure in mBERT~\citep{conneau-etal-2020-emerging}. Nonetheless, we believe m\wikinli and our training setup offer a useful framework for further research into multilingual learning with pretrained models.

\section{Summary}
In this chapter, we described approaches to exploiting various naturally-occurring structures on Wikipedia. In \cref{sec:wikipedia-entity-representations}, we used hyperlinks as natural supervision for two kinds of entity representations: CER and DER. For CER, we asked models to predict entity descriptions given context sentences in which the entities appear. For DER, we asked models to predict mention texts in the context sentences. Our proposed approaches were evaluated on a benchmark for entity representations and showed promising results.

In \cref{sec:wikipedia-discourse-sentence-representations}, we used article structures to train sentence encoders. The article structures are cast as multi-task learning objectives, encouraging the sentence-level models to encode information with respect to the broader context in which it situates. We evaluated the models on a discourse-related benchmark, finding that using the losses beyond sentences helped model performance on the discourse tasks.

In \cref{sec:wikipedia-concept-hierarchies}, we used Wikipedia category graphs to induce knowledge related to textual entailment. We treated the parent-child relations in Wikipedia category graphs as the entailment relations and built a training dataset for textual entailment. We found that training on our proposed datasets improves model performance on low-resource textual entailment tasks and we obtained similar improvements when extending our approaches to multilingual settings.
\chapter{Disentangling Latent Representations for Interpretability and Controllability}
\label{CHAPTER:DISENTANGLE}
In this chapter, we describe our contributions to disentangling latent representations using naturally-occurring structures of paired data. In \cref{section:vgvae-representation}, we presented a multi-task, latent-variable model that disentangles semantics and syntax in sentence representations. The model leverages the fact that the semantics of a paraphrase pair is shared but syntax varies. In \cref{section:vgvae-generation}, we extend this framework for controlling the syntax of generated text. In this controlled generation setting, we propose to use a sentential exemplar to control the syntax.

The material in this chapter is adapted from \citet{chen-etal-2019-multi} and \citet{chen-etal-2019-controllable}.

\section{Disentangling Semantics and Syntax in Sentence Representations}
\label{section:vgvae-representation}

\subsection{Introduction}
As generative latent variable models, especially of the continuous variety~\citep{kingma2013auto,goodfellow2014generative}, have become increasingly important in natural language processing~\citep{bowman-etal-2016-generating,Gulrajani2017wgan}, there has been increased interest in learning models where the latent representations are disentangled~\citep{hu17control}. Much of the recent NLP work on learning disentangled representations of text has focused on disentangling the representation of attributes such as sentiment from the representation of content, typically in an effort to better control text generation~\citep{shen2017style,zhao-etal-2017-learning,fu2018style}.

In this work, we instead focus on learning sentence representations that disentangle the syntax and the semantics of a sentence. We are moreover interested in disentangling these representations not for the purpose of controlling generation, but for the purpose of calculating semantic or syntactic similarity between sentences (but not both). To this end, we propose a generative model of a sentence which makes use of both semantic and syntactic latent variables, and we evaluate the induced representations on both standard semantic similarity tasks and on several novel syntactic similarity tasks. 

We use a deep generative model consisting of von Mises Fisher (vMF) and Gaussian priors on the semantic and syntactic latent variables (respectively) and a deep bag-of-words decoder that conditions on these latent variables. Following much recent work, we learn this model by optimizing the ELBO with a VAE-like~\citep{kingma2013auto,Rezende2014} approach.

Our learned semantic representations are evaluated on the SemEval semantic textual similarity (STS) tasks~\citep{agirre-etal-2012-semeval,cer-etal-2017-semeval}. Because there has been less work on evaluating syntactic representations of sentences, we propose several new syntactic evaluation tasks, which involve predicting the syntactic analysis of an unseen sentence to be the syntactic analysis of its nearest neighbor (as determined by the latent syntactic representation) in a large set of annotated sentences.

In order to improve the quality and disentanglement of the learned representations, we incorporate simple additional losses in our training, which are designed to force the latent representations to capture different information. In particular, our semantic multi-task losses make use of aligned paraphrase data, whereas our syntactic multi-task loss makes use of word-order information. Additionally, we explore different encoder and decoder architectures for learning better syntactic representations.

Experimentally, we find that by training in this way we are able to force the learned representations to capture different information (as measured by the performance gap between the latent representations on each task). Moreover, we find that we achieve the best performance on all tasks when the learned representations are most disentangled.

\subsection{Related Work}
There is a growing amount of work on learning interpretable or disentangled latent representations in various NLP applications, including sentence sentiment and style transfer~\cite[\emph{inter alia}]{hu17control,shen2017style,fu2018style,pmlr-v80-zhao18b}, morphological reinflection~\citep{zhou-neubig-2017-multi}, semantic parsing~\citep{yin-etal-2018-structvae}, text generation~\citep{wiseman-etal-2018-learning}, and sequence labeling~\citep{chen-etal-2018-variational}. Another related thread of work is text-based variational autoencoders~\citep{10.5555/3045390.3045573,bowman-etal-2016-generating}.

In terms of syntax and semantics in particular, there is a rich history of work in analyzing their interplay in sentences~\citep{jurafsky-1988-issues,van2005exploring}. We do not intend to claim that the two can be entirely disentangled in distinct representations. Rather, our goal is to propose modica of knowledge via particular multi-task losses and measure the extent to which this knowledge leads learned representations to favor syntactic or semantic information from a sentence. 

There has been prior work with similar goals for representations of words~\citep{mitchell-steedman-2015-orthogonality} and bilexical dependencies~\citep{mitchell-2016-decomposing}, finding that decomposing syntactic and semantic information can lead to improved performance on semantic tasks. We find similar trends in our results, but at the level of sentence representations. A similar idea has been explored for text generation~\citep{iyyer-etal-2018-adversarial}, where adversarial examples are generated by controlling syntax.

Some of our losses use sentential paraphrases, relating them to work in paraphrase modeling~\citep{wieting-16-full,wieting-gimpel-2018-paranmt}. 
\citet{deudon2018learning} proposed a variational framework for modeling paraphrastic sentences and \citet{chen-gimpel-2020-learning} learned probabilistic representations from paraphrases, but our focus here is on learning disentangled representations.

As part of our evaluation, we develop novel syntactic similarity tasks for sentence representations learned without any syntactic supervision. These evaluations relate to the broad range of work in unsupervised parsing~\citep{klein-manning-2004-corpus} and part-of-speech tagging~\citep{christodoulopoulos-etal-2010-two}. However, our evaluations differ from previous evaluations in that we employ $k$-nearest-neighbor syntactic analyzers using our syntactic representations to choose nearest neighbors. 
There is a great deal of work on applying multi-task learning to various NLP tasks~\cite[\emph{inter alia}]{plank-etal-2016-multilingual,rei-2017-semi,augenstein-sogaard-2017-multi,bollmann-etal-2018-multi} and, recently, as a way of improving the quality or disentanglement of learned representations~\citep{zhao-etal-2017-learning,NIPS2017_900c563b,du-etal-2018-variational,john-etal-2019-disentangled}.

\subsection{Model and Parameterization}

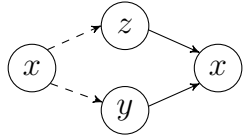
\begin{figure}
    \centering
    \begin{tikzpicture}
  [node distance=1.5em,
  start chain=going right,
  scale=0.5,
  every node/.style={scale=0.5}]
	 \node[bignode, gen_circle] (x1) {$x$};
	 \node[right=2.0em of x1] (in) {};
	 \node[bignode, gen_circle, above=0.4em of in] (z) {$z$};
	 \node[bignode, gen_circle, below=0.4em of in] (y) {$y$};
	 \node[bignode, gen_circle, right=2.0em of in] (x2) {$x$};
	 
	 \draw[->,thin] (y.10) -> (x2.220);
	 \draw[->,thin] (z.350) -> (x2.140);

	 \draw[->,thin, dashed] (x1.50) -> (z.180);
	 \draw[->,thin, dashed] (x1.320) -> (y.160);
\end{tikzpicture}
    \caption{Graphical model of \vgvae. Dashed lines indicate inference model. Solid lines indicate generative model.}
    \label{vgvae-representation-fig:graph}
\end{figure}

Our goal is to extract the disentangled semantic and syntactic information from sentence representations. To achieve this, we introduce the vMF-Gaussian Variational Autoencoder (\vgvae). As shown in \cref{vgvae-representation-fig:graph}, \vgvae assumes a sentence is generated by conditioning on two independent variables: semantic variable $y$ and syntactic variable $z$. In particular, our model gives rise to the following joint likelihood
\begin{align*}
    p_{\theta}(x, y, z) &= p_{\theta}(y) p_{\theta}(z) p_{\theta}(x\vert y, z) \\
    &= p_{\theta}(y) p_{\theta}(z) \prod_{t=1}^T p(x_t\vert y, z)
\end{align*}
where $x_t$ is the $t$th word of $x$, $T$ is the sentence length, and $p(x_t \vert y, z)$ is given by a softmax over a vocabulary of size $V$. Further details on the parameterization are given below.

To perform inference, we assume a factored posterior $q_\phi(y,z\vert  x)=q_\phi(y\vert x)q_\phi(z\vert x)$, as has been used in prior work~\citep{zhou-neubig-2017-multi,chen-etal-2018-variational}. Learning of \vgvae maximizes a lower bound on marginal log-likelihood:
\begin{equation*}
\begin{aligned}
    \log p_\theta(x)&\geq\mathop\mathbb{E}_{\substack{y\sim q_\phi(y\vert  x)\\z\sim q_\phi(z\vert  x)}}[\log p_\theta(x\vert z,y)-\log\frac{q_\phi(z\vert  x)}{p_\theta(z)}
    -\log\frac{q_\phi(y\vert  x)}{p_\theta(y)}]\\
    &=\mathop\mathbb{E}_{\substack{y\sim q_\phi(y\vert  x)\\z\sim q_\phi(z\vert  x)}}[\log p_\theta(x\vert  z,y)]-\kld(q_\phi(z\vert  x)\Vert p_\theta(z))
    -\kld(q_\phi(y\vert  x)\Vert p_\theta(y))
\end{aligned}
\label{vgvae-representation-eq:elbo}
\end{equation*}

\vgvae uses two distribution families in defining the posterior over latent variables, namely, the von Mises-Fisher (vMF) distribution and the Gaussian distribution (See \cref{background-section:vae-formal-prelim} for background materials on these two distributions).

\paragraph{Inference and Generative Models.}

The inference models $q_\phi(y\vert x)$ and $q_\phi(z\vert x)$ are two independent word averaging encoders with additional linear feedforward neural networks for producing $\mu(x)$ and $\sigma(x)$ (or $\kappa(x)$). The generative model $p_\theta(x\vert  y,z)$ is a feedforward neural network $g_\theta$ with the output being a bag of words. In particular, the expected output log-probability (the first term in \cref{vgvae-representation-eq:elbo}) is computed as follows:
\begin{equation}
\begin{aligned}
    \mathop\mathbb{E}_{\substack{y\sim q_\phi(y\vert  x)\\z\sim q_\phi(z\vert  x)}}[\log p_\theta(x\vert  y,z)]=
    \mathop\mathbb{E}_{\substack{y\sim q_\phi(y\vert  x)\\z\sim q_\phi(z\vert  x)}}\left[\sum_{t=1}^T \log \frac{\exp{g_\theta([y;z])}_{x_t}}{\sum_{j=1}^{V}\exp{g_\theta([y;z])}_j}\right]\nonumber
\end{aligned}
\end{equation}
\noindent where $V$ is the vocabulary size, $[;]$ indicates concatenation, $T$ is the sentence length and $x_t$ is the index of the $t$'th word's word type.

\paragraph{Recurrent Neural Networks.}

To facilitate better learning of syntax, we also consider replacing both the generative and inference models with RNN-based sequence models, rather than bag-of-words models. In this setting, the generative model $p_\theta(x\vert y,z)$ is a unidirectional LSTM and a linear feedforward neural network for predicting the word tokens. The expected output log-probability is computed as follows:
\begin{equation}
    \mathop\mathbb{E}_{\substack{y\sim q_\phi(y\vert  x)\\z\sim q_\phi(z\vert  x)}}[\log p_\theta(x\vert  y,z)]=
    \mathop\mathbb{E}_{\substack{y\sim q_\phi(y\vert  x)\\z\sim q_\phi(z\vert  x)}}\left[\sum_{t=1}^T \log p_\theta(x_t\vert y,z,x_{1:t-1})\right]\nonumber
\end{equation}
\noindent where $V$ is the vocabulary size, $T$ is the sentence length and $x_t$ is the index of the $t$'th word's word type.

The inference model $q_\phi(y\vert x)$ is still a word averaging encoder, but $q_\phi(z \vert x)$ is parameterized by a bidirectional LSTM, where we concatenate the forward and backward hidden states and then take the average. The output of the LSTM is then used as input to a feedforward network with one hidden layer for producing $\mu(x)$ and $\sigma(x)$ (or $\kappa(x)$).

In the following sections, we will introduce several losses that will be added into the training of our base model, which empirically shows the ability of further disentangling the functionality between the semantic variable $y$ and the syntactic variable $z$.

\subsection{Multi-Task Training}
\label{vgvae-representation-sec:multitask}

\begin{figure}
    \centering
    \begin{tikzpicture}
  [node distance=1.5em,
  start chain=going right,
  scale=0.5,
  every node/.style={scale=0.5}]
 \node[regularnode, punktchain] (e1) {$z$ encoder};
 \node[bignode, gen_circle, right=2.0em of e1] (z) {$z_1$};
 \node[bignode, gen_circle, left=1.0em of e1] (x1) {$x_1$};
 \node[regularnode, punktchain, below=2.0em of e1] (e2) {$y$ encoder};
 \node[bignode, gen_circle, right=2.0em of e2] (y) {$y_1$};
 \node[bignode, gen_circle, left=1.0em of e2] (x2) {$x_1$};
 
 \node[regularnode, punktchain, below=2.0em of e2] (e3) {$y$ encoder};
 \node[bignode, gen_circle, right=2.0em of e3] (z2) {$y_2$};
 \node[bignode, gen_circle, left=1.0em of e3] (x4) {$x_2$};
 \node[regularnode, punktchain, below=2.0em of e3] (e4) {$z$ encoder};
 \node[bignode, gen_circle, right=2.0em of e4] (y2) {$z_2$};
 \node[bignode, gen_circle, left=1.0em of e4] (x5) {$x_2$};

 \node[below=1.0em of e1] (middle) {};
 \node[bignode, gen_circle, right=6.5em of middle] (x3) {$x_1$};
 
 \draw[->,thin, dashed] (e1.0) -> (z.180);
 \draw[->,thin, dashed] (x1.0) -> (e1.180);
 
 \draw[->,thin, dashed] (e2.0) -> (y.180);
 \draw[->,thin, dashed] (x2.0) -> (e2.180);
 
 \%draw[->,thin] (z.350) -> (x3.135);
 
 \node[below=1.0em of e3] (middle) {};
 \node[bignode, gen_circle, right=6.5em of middle] (x6) {$x_2$};
 
 \draw[->,thin, dashed] (e3.0) -> (z2.180);
 \draw[->,thin, dashed] (x4.0) -> (e3.180);
 
 \draw[->,thin, dashed] (e4.0) -> (y2.180);
 \draw[->,thin, dashed] (x5.0) -> (e4.180);

 \node[below=0.8em of e2] (dpl_pos) {};
 \node[bignode, below=0.4em of e2] (dpl_des) {DPL};
 \node[rectangle, rounded corners,
 	   draw=black, thick, minimum width=5em,
 	   minimum height=13em, right=3.4em of dpl_pos, dashed] (dpl) {};

 \draw[->,thin, dotted] (dpl_des.east) -> (dpl.178);

 \draw[->,thin, dash dot] (y2.10) -> (x6.220);
 \draw[->,thin, dash dot] (z2.30) -> (x3.245);
 
 \draw[->,thin, dash dot] (y.330) -> (x6.120);
 \draw[->,thin, dash dot] (z.350) -> (x3.135);
\end{tikzpicture}
    \caption{Diagram showing the training process when using the discriminative paraphrase loss (\spl; dotted lines) and paraphrase reconstruction loss (\prl; dash-dotted lines). The pair $(x_1, x_2)$ is a sentential paraphrase pair, the $y$'s are the semantic variables corresponding to each $x$, and the $z$'s are syntactic variables.
    }
    \label{vgvae-representation-fig:losses}
\end{figure}
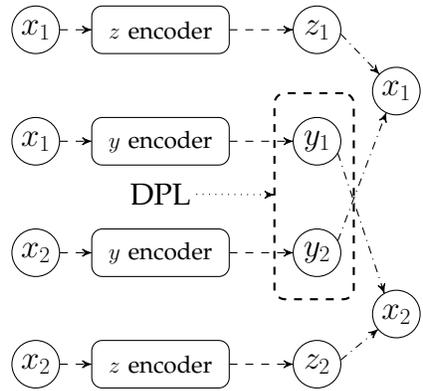

We attempt to improve the quality and disentanglement of our semantic and syntactic representations by introducing additional losses, which encourage $y$ to capture semantic information and $z$ to capture syntactic information. We elaborate on these losses below.

\paragraph{Paraphrase Reconstruction Loss}
\label{vgvae-representation:para-recon-loss}
Our first loss is a paraphrase reconstruction loss (\prl). 
The key assumption underlying the PRL is that for a paraphrase pair $x_1$, $x_2$, the semantic information is equivalent between the two sentences and only the syntactic information varies. To impose such constraints, \prl is defined as
\begin{equation*}
    \mathop\mathbb{E}_{\substack{y_2\sim q_\phi(y\vert x_2)\\z_1\sim q_\phi(z\vert x_1)}}[-\log p_\theta(x_1\vert y_2,z_1)] + \mathop\mathbb{E}_{\substack{y_1\sim q_\phi(y\vert x_1)\\z_2\sim q_\phi(z\vert x_2)}}[-\log p_\theta(x_2\vert y_1,z_2)]
\label{vgvae-representation-eq:prl}
\end{equation*}
That is, we swap the semantic variables, keep the syntactic variables, and attempt to reconstruct the sentences (shown in \cref{vgvae-representation-fig:losses}). 
While instead of using a multi-task objective we could directly model paraphrases $x_1$ and $x_2$ as being generated by the same $y$ (which naturally suggests a product-of-experts style posterior, as in \citet{NEURIPS2018_1102a326}), we found that for the purposes of our downstream tasks training with the multi-task loss gave superior results. 

\paragraph{Discriminative Paraphrase Loss}
\label{vgvae-representation:disc-para-loss}

Our second loss is a discriminative paraphrase loss (\spl). The \spl explicitly encourages the similarity of paraphrases $x_1$, $x_2$ to be scored higher than the dissimilar sentences $n_1$, $n_2$ (i.e., negative samples; see \cref{vgvae-representation:neg-sample} for more details) by a given margin $\delta$. As shown in \cref{vgvae-representation-fig:losses}, the similarity function in this loss only uses the semantic variables in the sentences. The loss is defined as
\begin{equation*}
    \max(0,\delta-d(x_1,x_2)+d(x_1,n_1))+
    \max(0,\delta-d(x_1,x_2)+d(x_2,n_2))
\end{equation*}
\noindent The similarity function we choose is the cosine similarity between the mean directions of the semantic variables from the two sentences: 
\begin{equation*}
    d(x_1, x_2)=\mathrm{cosine}(\mu_\alpha(x_1), \mu_\alpha(x_2))
\end{equation*}
\paragraph{Word Position Loss}
It has been observed in previous work that word order typically contributes little to the modelling of semantic similarity~\citep{wieting-16-full}. We interpret this as evidence that word position information is more relevant to syntax than semantics, at least as evaluated by STS tasks. To guide the syntactic variable to represent word order, we introduce a word position loss (\wpl). Although our word averaging encoders only have access to the bag of words of the input, using this loss can be viewed as a denoising autoencoder where we have maximal input noise (i.e., an orderless representation of the input) and the encoders need to learn to reconstruct the ordering. 

For both word averaging encoders and LSTM encoders, \wpl is parameterized by a three-layer feedforward neural network $f(\cdot)$ with input from the concatenation of the samples of the syntactic variable $z$ and the embedding vector $e_i$ at input position $i$; we then attempt to predict a one-hot vector representing the position $i$. More specifically, we define
\begin{equation*}
    \text{WPL}\stackrel{\text{def}}{=\joinrel=}\mathop\mathbb{E}_{z\sim q_\phi(z\vert x)}\left[-\sum_{i}\log\textrm{softmax}(f([e_i;z]))_i\right]
\end{equation*}
\noindent where $\textrm{softmax}(\cdot)_i$ indicates the probability at position $i$. 

\subsection{Experimental Setup}
\paragraph{KL Weight.} Following previous work on VAEs \citep{higgins2016beta,alemi2016deep}, we attach a weight to the KL divergence and tune it based on development set performance.

\paragraph{Negative Samples.}
\label{vgvae-representation:neg-sample}
When applying \spl, we select negative samples based on maximizing cosine similarity to sentences from a subset of the data. In particular, we accumulate $k$ mini-batches during training, yielding a ``mega-batch'' $\mathcal{S}$ \citep{wieting-gimpel-2018-paranmt}. Then the negative samples are selected based on the following criterion:
\begin{equation}
    n_1=\argmax_{n\in\mathcal{S}\wedge n\neq x_2} \mathrm{cosine}(\mu_\alpha(x_1),\mu_\alpha(n))\nonumber
\end{equation}
\noindent where $x_1$, $x_2$ forms the paraphrase pair and the mega-batch size is fixed to $k=20$ for all of our experiments. Since all of our models are trained from scratch, we observed some instabilities with \spl during the initial stages of training. We suspect that this is because the negative samples at these initial stages are of low quality. To overcome this issue, \spl is included starting at the second epoch of training so that the models can have a warm start. 

\paragraph{Training Setup.}
We subsampled half a million paraphrase pairs from ParaNMT-50M~\citep{wieting-gimpel-2018-paranmt} as our training set. We use SemEval semantic textual similarity (STS) task 2017~\citep{cer-etal-2017-semeval} as a development set. For semantic similarity evaluation, we use the STS tasks from 2012 to 2016~\citep{agirre-etal-2012-semeval,agirre-etal-2013-sem,agirre-etal-2014-semeval,agirre-etal-2015-semeval,agirre-etal-2016-semeval} and the STS benchmark test set~\citep{cer-etal-2017-semeval}. For evaluating syntactic similarity, we propose several evaluations. One uses the gold parse trees from the Penn Treebank~\citep{marcus-etal-1993-building}, and the others are based on automatically tagging and parsing five million paraphrases from ParaNMT-50M; we describe these tasks in detail below.

For hyperparameters, the dimensions of the latent variables are 50. The dimensions of word embeddings are 50. We use cosine similarity as similarity metric for all of our experiments. We tune the weights for \prl and reconstruction loss from 0.1 to 1 in increments of 0.1 based on the development set performance. We use one sample from each latent variable during training. When evaluating \vgvae based models on STS tasks, we use the mean direction of the semantic variable $y$, while for syntactic similarity tasks, we use the mean vector of the syntactic variable $z$.

\paragraph{Baselines} Our baselines are a simple word averaging (\wordavg) model and bidirectional LSTM averaging (\lstmavg) model, both of which have been shown to be very competitive for modeling semantic similarity when trained on paraphrases \citep{wieting-gimpel-2018-paranmt}. Specifically, \wordavg takes the average over the word embeddings in the input sequence to obtain the sentence representation. \lstmavg uses the averaged hidden states of a bidirectional LSTM as the sentence representation, where forward and backward hidden states are concatenated. These models use 50 dimensional word embeddings and 50 dimensional LSTM hidden vectors per direction. These baselines are trained with \spl only. Additionally, we scramble the input sentence for \lstmavg since it has been reported beneficial for its performance in semantic similarity tasks \citep{wieting-gimpel-2017-revisiting}. 

We also benchmark several pretrained embeddings on both semantic similarity and syntactic similarity datasets, including GloVe,\footnote{We use 300 dimensional Common Crawl  embeddings available at \href{https://nlp.stanford.edu/projects/glove}{\nolinkurl{nlp.stanford.edu/projects/glove}}} \skipthought,\footnote{\href{https://github.com/ryankiros/skip-thoughts}{\nolinkurl{github.com/ryankiros/skip-thoughts}}} InferSent,\footnote{We use model V1 available at \href{https://github.com/facebookresearch/InferSent}{\nolinkurl{github.com/facebookresearch/InferSent}}} ELMo,\footnote{We use the original model available at \href{https://allennlp.org/elmo}{\nolinkurl{allennlp.org/elmo}}} and BERT.\footnote{We use bert-large-uncased available at \href{https://github.com/huggingface/pytorch-pretrained-BERT}{\nolinkurl{github.com/huggingface/pytorch-pretrained-BERT}}} For GloVe, we average word embeddings to form sentence embeddings. For ELMo, we average the hidden states from three layers and then average the hidden states across time steps. For BERT, we use the averaged hidden states from the last attention block. Code is available at \url{https://github.com/mingdachen/disentangle-semantics-syntax}.
\subsection{Experimental Results}

\paragraph{Semantic Similarity.}

\begin{table}[t]
\setlength{\tabcolsep}{4pt}
\centering
\small
\begin{tabular}{|l|c|c|c|c|}\hline
        & \multicolumn{2}{c|}{semantic var.}& \multicolumn{2}{c|}{syntactic var.}\\
        & bm & avg &  bm & avg \\
\hline
GloVe & 39.0 & 48.7 & - & - \\
SkipThought  & 42.1 & 42.0 & - & - \\
InferSent & 67.8 & 61.0 & - & - \\
ELMo & 57.7 & 60.3 & - & - \\
BERT & 4.5 & 15.0 & - & - \\
\hline
\wordavg & 71.9 & 64.8 &  -   & - \\
\lstmavg & 71.4 & 64.4 &  -   & - \\
\hline
\vgvae             & 45.5 & 42.7 & 40.8 & 43.2 \\
\vgvae+ \wpl          & 51.5 & 49.3 & 28.1 & 31.0 \\
\vgvae+ \spl         & 68.4 & 58.2 & 37.8 & 40.5 \\
\vgvae+ \prl          & 67.9 & 57.8 & 29.6 & 32.7 \\
\vgvae+ \prl+ \wpl       & 69.8 & 61.3 & 23.2 & 27.9 \\
\vgvae+ \prl+ \spl       & 71.2 & 64.2 & 31.7 & 33.9 \\
\vgvae+ \spl+ \wpl       & 71.0 & 63.5 & 24.1 & 29.0 \\
ALL    & 72.3 & 65.1 & 20.1 & 24.2 \\
ALL + LSTM enc. & 72.5 & 65.1 & 16.3 & 24.5 \\
ALL + LSTM enc. \& dec. & \textbf{72.9} &  \textbf{65.5} & \textbf{11.3} & \textbf{19.3} \\\hline

\end{tabular}
\caption{Pearson correlation (\%) for STS test sets. bm: STS benchmark test set. avg: the average of Pearson correlation for each domain in the STS test sets from 2012 to 2016. Results are in bold if they are highest in the ``semantic variable'' columns or lowest in the ``syntactic variable'' columns. ``ALL'' indicates all of the multi-task losses are used.}
\label{vgvae-representation:sts-res}
\end{table}

As shown in \cref{vgvae-representation:sts-res}, the semantic and syntactic variables of our base \vgvae model show similar performance on the STS test sets. As we begin adding multi-task losses, however, the performance of these two variables gradually diverges, indicating that different information is being captured in the two variables. More interestingly, note that when \textit{any} of the three losses is added to the base \vgvae model (even the WPL loss which makes no use of paraphrases), the performance of the semantic variable increases and the performance of the syntactic variable decreases; this suggests that each loss is useful in encouraging the latent variables to learn complementary information. 

Indeed, the trend of additional losses both increasing semantic performance and decreasing syntactic performance holds even as we use more than two losses, except for the single case of \vgvae + PRL + DPL, where the syntactic performance increases slightly. Finally, we see that when the bag-of-words \vgvae model is used with all of the multi-task losses (``ALL''), we observe a large gap between the performance of the semantic and syntactic latent variables, as well as strong performance on the STS tasks that outperforms all baselines. 

Using LSTM modules further strengthens the disentanglement between the two variables and leads to even better semantic performance. While using an LSTM encoder and a bag-of-words decoder is difficult to justify from a generative modeling perspective, we include results with this configuration to separate out the contributions of the LSTM encoder and decoder. 

\paragraph{Syntactic Similarity.}
So far, we have only confirmed empirically that the syntactic variable has learned to \textit{not} capture semantic information. To investigate what the syntactic variable has captured, we propose several syntactic similarity tasks.

\begin{table}
\setlength{\tabcolsep}{2pt}
\centering
\footnotesize
\begin{tabular}{|l|c|c|c|c|c|c|}\hline
        & \multicolumn{2}{c|}{Constituent Parsing (TED, $\downarrow$)} & \multicolumn{2}{c|}{Constituent Parsing ($F_1$, $\uparrow$)} & \multicolumn{2}{c|}{POS Tagging (\%Acc., $\uparrow$)}
        \\ 
\hline
GloVe & \multicolumn{2}{c|}{120.8} & \multicolumn{2}{c|}{27.3} & \multicolumn{2}{c|}{23.9}  \\
\skipthought &  \multicolumn{2}{c|}{99.5} & \multicolumn{2}{c|}{30.9} & \multicolumn{2}{c|}{29.6}  \\
InferSent &  \multicolumn{2}{c|}{138.9} & \multicolumn{2}{c|}{28.0} & \multicolumn{2}{c|}{25.1}  \\
ELMo  &  \multicolumn{2}{c|}{103.8} & \multicolumn{2}{c|}{30.4} & \multicolumn{2}{c|}{27.8}  \\
BERT  & \multicolumn{2}{c|}{101.7} & \multicolumn{2}{c|}{28.6} & \multicolumn{2}{c|}{25.4}  \\
\hline
Random baseline & \multicolumn{2}{c|}{121.4} & \multicolumn{2}{c|}{19.2} & \multicolumn{2}{c|}{12.9} \\
Upper bound perf. & \multicolumn{2}{c|}{51.6} & \multicolumn{2}{c|}{71.1} & \multicolumn{2}{c|}{62.3} \\
\hline
\wordavg   & \multicolumn{2}{c|}{107.0} & \multicolumn{2}{c|}{25.5} & \multicolumn{2}{c|}{21.4} \\
\lstmavg   & \multicolumn{2}{c|}{106.8} & \multicolumn{2}{c|}{25.7} & \multicolumn{2}{c|}{21.6} \\
\hline
        & semantic var.& syntactic var.& semantic var.& syntactic var.& semantic var.& syntactic var. \\
\hline
\vgvae                   & 109.3 & 111.4 & 25.2 & 25.0 & 21.1 & 21.0 \\
\vgvae+ \wpl             & 112.3 & 105.9 & \textbf{24.1} & 28.2 & \textbf{20.3} & 24.2 \\
\vgvae+ \spl             & 108.1 & 110.6 & 25.1 & 26.1 & 21.3 & 21.8 \\
\vgvae+ \prl             & 111.9 & 110.9 & 24.7 & 26.9 & 21.0 & 22.2 \\
\vgvae+ \spl+ \wpl       & 111.2 & 105.0 & 25.1 & 28.8 & 21.5 & 24.6 \\
\vgvae+ \prl+ \spl       & 108.0 & 110.4 & 25.0 & 26.2 & 21.1 & 22.1 \\
\vgvae+ \prl+ \wpl       & 109.4 & 105.1 & 24.4 & 28.1 & 20.6 & 23.6 \\
ALL                   & 110.0 & 104.7 & 25.4 & 29.3 & 21.4 & 25.5  \\
ALL + LSTM enc.       & 112.0 & 101.0 & 25.7 & 37.3 & 22.1  & 34.0 \\
ALL + LSTM enc. \& dec. & \textbf{114.6} & \textbf{100.5} & 25.3 & \textbf{38.8} & 21.4 & \textbf{35.7}\\\hline
\end{tabular}
\caption{Syntactic similarity evaluations, showing tree edit distance (TED) and labeled $F_1$ score for constituent parsing, and accuracy (\%) for part-of-speech tagging. Numbers are boldfaced if they are worst in the ``semantic variable'' column or best in the ``syntactic variable'' column. ``ALL'' indicates all the multi-task losses are used.} \label{vgvae-representation:tot-syntax-res}
\end{table}

In particular, we consider using the syntactic latent variable in calculating nearest neighbors for a 1-nearest-neighbor syntactic parser or part-of-speech tagger. We use our latent variables to define the similarity function in these settings and evaluate the quality of the output parses and tag sequences using several metrics. 

Our first evaluation involves constituency parsing, and we use the standard training and test splits from the Penn Treebank. We predict a parse tree for each sentence in the test set by finding its nearest neighbor in the training set based on the cosine similarity of the mean vectors for the syntactic variables. The parse tree of the nearest neighbor will then be treated as our prediction for the test sentence. Since the train and test sentences may differ in length, standard parse evaluation metrics are not applicable, so we use tree edit distance~\cite{zhang1989simple}\footnote{\href{https://github.com/timtadh/zhang-shasha}{\nolinkurl{github.com/timtadh/zhang-shasha}}} 
to compute the distance between two parse tree without considering word tokens.

To better understand the difficulty of this task, we introduce two baselines. The first randomly selects a training sentence. We calculate its performance by running it ten times and then reporting the average. We also report the upper bound performance given the training set. Since computing tree edit distance is time consuming, we subsample 100 test instances and compute the minimum tree edit distance for each sampled instance. Thus, this number can be seen as the approximated upper bound performance for this task given the training set.

To use a more standard metric for these syntactic similarity tasks, we must be able to retrieve training examples with the same number of words as the sentence we are trying to parse. We accordingly parse and tag the five million paraphrase subset of the ParaNMT training data using  Stanford CoreNLP~\citep{manning-etal-2014-stanford}. To form a test set, we group sentences in terms of sentence length and subsample 300 sentences for each sentence length. After removing the paraphrases of the sentences in the test set, we use the rest of the training set as candidate sentences for nearest neighbor search, and we restrict nearest neighbors to have the same sentence length as the sentence we are attempting to parse or tag, which allows us to use standard metrics like labeled $F_1$ score and tagging accuracy for evaluation.

As shown in \cref{vgvae-representation:tot-syntax-res}, the syntactic variables and semantic variables demonstrate similar trends across these three syntactic tasks. Interestingly, both \spl and \prl help to improve the performance of the syntactic variables, even though these two losses are only imposed on the semantic variables. We saw an analogous pattern in \cref{vgvae-representation:sts-res}, which again suggests that by pushing the semantic variables to learn information shared by paraphrastic sentences, we also encourage the syntactic variables to capture complementary syntactic information. 
We also find that adding \wpl brings the largest improvement to the syntactic variable, and keeps the syntactic information carried by the semantic variables at a relatively low level. Finally, when adding all three losses, the syntactic variable shows the strongest performance across the three tasks.

In addition, we observe that the use of the LSTM encoder improves syntactic performance by a large margin and the LSTM decoder improves further, which suggests that the use of the LSTM decoder contributes to the amount of syntactic information represented in the syntactic variable. 

Among pretrained representations, \skipthought shows the strongest performance overall and ELMo has the second best performance in the last two columns. While InferSent performs worst in the first column, it gives reasonable performance for the other two. BERT performs relatively well in the first column but worse in the other two.

\subsection{Analysis}

\begin{table}
\small
\setlength{\tabcolsep}{6pt}
\centering
\begin{tabular}{|l|l|}%
\hline
\multirow{2}{*}{starting} & \textit{syntactic:} getting heading sitting chasing taking \\
& \textit{semantic:} begin start stopping forward rising \\
\hline
\multirow{2}{*}{area} & \textit{syntactic:} engines certificate guests bottle pieces  \\
 & \textit{semantic:} sector location zone fields rooms field \\
\hline
\multirow{2}{*}{considered} & \textit{syntactic:} stable limited odd scary classified awful \\
& \textit{semantic:} thought assumed regard reasons wished \\
\hline
\multirow{2}{*}{jokes} & \textit{syntactic:} gentlemen photos finding baby missile \\
& \textit{semantic:} funny humor prize stars cookie ideal\\
\hline
\multirow{2}{*}{times} & \textit{syntactic:} princess officer wounds plan gang user\\
& \textit{semantic:} twice later thousand often seven time\\
\hline

\end{tabular}
\caption{Examples of the most similar words to particular query words using syntactic variable (first row) or semantic variable (second row). 
}
\label{vgvae-representation:lex-examples}
\end{table}

\begin{table}
\small
\setlength{\tabcolsep}{2pt}
\centering
\begin{tabular}{|p{0.34\textwidth}|p{0.31\textwidth}|p{0.32\textwidth}|}
\hline
\multicolumn{1}{|c|}{Query Sentence} & \multicolumn{1}{c|}{Semantically Similar} & \multicolumn{1}{c|}{Syntactically Similar} \\
\hline
i have much more colours at home . & even if there was food , would n't it be at least 300 years old ? & you have a beautiful view from here . \\\hline
victor had never known darkness like it . & he had never experienced such darkness as this . & you seem like a really nice kid . \\\hline
this is , uh , too serious . & but this is too serious . & it is , however , illegal discrimination .\\\hline
you 're gon na save her life . & you will save her . & you 're gon na give a speech . \\\hline
we 've got to get a move on . & come on , we got ta move . & you 'll have to get in there . \\\hline
and that was usually the highlight of my day . & i really enjoyed it when i did it . & and yet that was not the strangest aspect of the painting . \\\hline
we do need to collect our taxes somehow . & we have to earn the money we need . & now i have to do my job .\\
\hline
this is just such a surprise . & oh . this is a surprise . & this is just a little gain . \\\hline
okay . aw , that 's so romantic . & it 's so romantic ! & oh . well , that 's not good . \\\hline
we 're gon na have to do something about this . & we 'll have to do something about that . & we 're gon na have to do something about yours .\\\hline
\end{tabular}
\caption{Examples of most similar sentences to particular query sentences in terms of the semantic variable or the syntactic variable. }
\label{vgvae-representation:sentence-examples}
\end{table}

\paragraph{Qualitative Analysis.}
To qualitatively evaluate our latent variables, we find (via cosine similarity) nearest neighbor sentences to test set examples in terms of both the semantic and syntactic representations. We also find nearest neighbors of words (which we view as single-word sentences). We discuss the results of this analysis below.

\paragraph{Lexical Analysis.}
\cref{vgvae-representation:lex-examples} shows word nearest neighbors for both syntactic and semantic representations. We see that the most similar words found by the syntactic variable share the same part-of-speech tags with the query words. For example, ``starting'' is close to ``getting'' and ``taking,''  even though these words are not semantically similar. Words retrieved according to the semantic variable, however, are more similar semantically, e.g., ``begin'' and ``starts''. As another example, ``times'' is similar to words that are either related to descriptions of frequency (e.g., ``twice'' and ``often'') or related to numbers (e.g., ``thousand'', ``seven'').

\paragraph{Sentential Analysis.}
As shown in \cref{vgvae-representation:sentence-examples}, sentences that are similar in terms of their semantic variables tend to have similar semantics. However, sentences that are similar in terms of their syntactic variables are mostly semantically unrelated but have similar surface forms. For example, ``you 're gon na save her life .'' has the same meaning as ``you will save her .'' while having a similar syntactic structure to ``you 're gon na give a speech .'' (despite having very different meanings). As another example, although the semantic variable does not find a good match for ``i have much more colours at home .'', which can be attributed to the limited size of candidate sentences, the nearest syntactic neighbor (``you have a beautiful view from here .'') has a very similar syntactic structure to the query sentence.

\section{Controllable Paraphrase Generation with a Syntactic Exemplar}
\label{section:vgvae-generation}
\subsection{Introduction}
Controllable text generation has recently become an area of intense focus in the NLP community. Recent work has focused both on generating text satisfying certain stylistic requirements such as being formal or exhibiting a particular sentiment~\citep{hu17control,shen2017style,ficler-goldberg-2017-controlling}, as well as on generating text meeting structural requirements, such as conforming to a particular template~\citep{iyyer-etal-2018-adversarial,wiseman-etal-2018-learning}. 

These systems can be used in various application areas, such as text summarization~\citep{fan-etal-2018-controllable}, adversarial example generation~\citep{iyyer-etal-2018-adversarial}, dialogue~\citep{niu-bansal-2018-polite}, and data-to-document generation~\citep{wiseman-etal-2018-learning}. However, prior work on controlled generation has typically assumed a known, finite set of values that the controlled attribute can take on. In this work, we are interested instead in the novel setting where the generation is controlled through an exemplar sentence (where any syntactically valid sentence is a valid exemplar). We will focus in particular on using a sentential exemplar to control the syntactic realization of a generated sentence. This task can benefit  natural language interfaces to information systems by suggesting alternative invocation phrases for particular types of queries~\citep{kumar2017just}. It can also bear on dialogue systems that seek to generate utterances that fit particular functional categories~\citep{ke-etal-2018-generating}. 

To address this task, we propose a deep generative model with two latent variables, which are designed to capture semantics and syntax. To achieve better disentanglement between these two variables, we design multi-task learning objectives that make use of paraphrases and word order information. To further facilitate the learning of syntax, we additionally propose to train the syntactic component of our model with word noising and latent word-cluster codes. Word noising randomly replaces word tokens in the syntactic inputs based on a part-of-speech tagger used only at training time. Latent codes create a bottleneck layer in the syntactic encoder, forcing it to learn a more compact notion of syntax. The latter approach also learns interpretable word clusters. Empirically, these learning criteria and neural architectures lead to better generation quality and generally better disentangled representations.

\begin{figure}
    \small
    \centering
    \begin{subfigure}[t]{\textwidth}
        \centering
        \begin{Frame}
        $X$: {\color{red}his teammates' eyes got an ugly, hostile expression.}\\
        $Y$: {\color{blue}the smell of flowers was thick and sweet.}\\
        $Z$: the eyes of his teammates had turned ugly and hostile.
        \end{Frame}
    \end{subfigure}
    \begin{subfigure}[t]{\textwidth}
        \centering
        \begin{Frame}
        $X$: {\color{red}we need to further strengthen the agency's capacities.}\\
        $Y$: {\color{blue}the damage in this area seems to be quite minimal.}\\
        $Z$: the capacity of this office needs to be reinforced even further.
        \end{Frame}
    \end{subfigure}
    \caption{Examples from our annotated evaluation dataset of paraphrase generation using semantic input $X$ (red), syntactic exemplar $Y$ (blue), and the reference output $Z$ (black).}
    \label{vgvae-generation:example}
\end{figure}

To evaluate this task quantitatively, we manually create an evaluation dataset containing triples of a semantic exemplar sentence, a syntactic exemplar sentence, and a reference sentence incorporating the semantics of the semantic exemplar and the syntax of the syntactic exemplar. This dataset is created by first automatically finding syntactic exemplars and then heavily editing them by ensuring (1) semantic variation between the syntactic inputs and the references, (2) syntactic similarity between the syntactic inputs and the references, and (3) syntactic variation between the semantic input and references. Examples are shown in \cref{vgvae-generation:example}. This dataset allows us to evaluate different approaches quantitatively using standard metrics, including BLEU~\citep{papineni-etal-2002-bleu} and ROUGE~\citep{lin-2004-rouge}. As the success of controllability of generated sentences also largely depends on the syntactic similarity between the syntactic exemplar and the reference, we propose a ``syntactic similarity'' metric based on evaluating tree edit distance between constituency parse trees of these two sentences after removing word tokens. 

Empirically, we benchmark the syntactically-controlled paraphrase network (SCPN) of \citet{iyyer-etal-2018-adversarial} on this novel dataset, which shows strong performance with the help of a supervised parser at test-time but also can be sensitive to the quality of the parse predictor. We show that using our word position loss effectively characterizes syntactic knowledge, bringing consistent and sizeable improvements over syntactic-related evaluation. The latent code module learns interpretable latent representations. Additionally, all of our models can achieve improvements over baselines. Qualitatively, we show that our models do suffer from the lack of an abstract syntactic representation, though we also show that SCPN and our models  exhibit similar artifacts. 
\subsection{Related Work}
We focus primarily on the task of paraphrase generation, which has received significant recent attention~\citep[\emph{inter alia}]{quirk-etal-2004-monolingual,prakash-etal-2016-neural,mallinson-etal-2017-paraphrasing,dong-etal-2017-learning,ma-etal-2018-query,li-etal-2018-paraphrase}.

In seeking to control generation with exemplars, our approach relates to recent work in controllable text generation. Whereas much work on controllable text generation seeks to control distinct attributes of generated text (e.g., its sentiment or formality)~\citep[\emph{inter alia}]{hu17control,shen2017style,ficler-goldberg-2017-controlling,fu2018style,pmlr-v80-zhao18b}, there is also recent work which attempts to control structural aspects of the generation, such as its latent~\citep{wiseman-etal-2018-learning} or syntactic~\citep{iyyer-etal-2018-adversarial} template. 
Our work is closely related to this latter category, and to the syntactically-controlled paraphrase generation of~\citet{iyyer-etal-2018-adversarial} in particular, but our proposed model is different in that it simply uses a single \textit{sentence} as a syntactic exemplar rather than requiring a supervised parser. This makes our setting closer to style transfer in computer vision, in which an image is generated that combines the content from one image and the style from another~\citep{gatys2016image}. In particular, in our setting, we seek to generate a sentence that combines the semantics from one sentence with the syntax from another, and so we only require a pair of (unparsed) sentences. We also note concurrent work that attempts to use sentences as exemplars in controlling generation~\citep{lin-etal-2020-data} in the context of data-to-document generation~\citep{wiseman-etal-2017-challenges}.

Another related line of work builds generation upon sentential exemplars~\citep{guu-etal-2018-generating,weston-etal-2018-retrieve,pandey-etal-2018-exemplar,cao-etal-2018-retrieve,peng-etal-2019-text} in order to improve the quality of the generation itself, rather than to allow for control over syntactic structures.

\subsection{Model and Training}

Given two sentences $X$ and $Y$, our goal is to generate a sentence $Z$ that follows the syntax of $Y$ and the semantics of $X$. We refer to $X$ and $Y$ as the semantic template and syntactic template, respectively.

To solve this problem, we adapt the \vgvae model described in \cref{section:vgvae-representation}. 
In particular, we assume a generative model that has two latent variables: $y$ for semantics and $z$ for syntax (as depicted in \cref{vgvae-representation-fig:graph}).

\paragraph{Decoders.}
\begin{figure}
    \centering
    \includegraphics[scale=0.7]{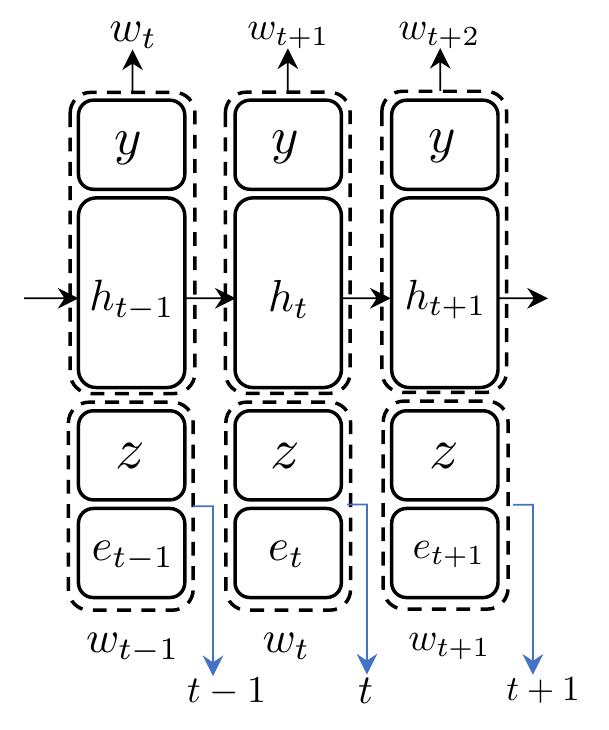}
    \caption{Diagram showing training of the decoder. Blue lines indicate the word position loss (\wpl).}
    \label{vgvae-generation-fig:decoder}
\end{figure}
As shown in \cref{vgvae-generation-fig:decoder}, at each time step, we concatenate the syntactic variable $z$ with the previous word's embedding as the input to the decoder and concatenate the semantic variable $y$ with the hidden vector output by the decoder for predicting the word at the next time step. Note that the initial hidden state of the decoder is always set to zero.

\paragraph{Latent Codes for Syntactic Encoder}
Since what we want from the syntactic encoder is only the syntactic structure of a sentence, using standard word embeddings tends to mislead the syntactic encoder to believe the syntax is manifested by the exact word tokens. An example is that the generated sentence often preserves the exact pronouns or function words in the syntactic input instead of making necessary changes based on the semantics. To alleviate this, we follow \citet{chen-gimpel-2018-smaller} to represent each word with a latent code (\lc) for word clusters within the word embedding layer. Our goal is for this to create a bottleneck layer in the word embeddings, thereby forcing the syntactic encoder to learn a more abstract representation of the  syntax. However, since our purpose is not to reduce model size (unlike \citealp{chen-gimpel-2018-smaller}), we marginalize out the latent code to get the embeddings during both training and testing. That is,
\begin{equation}
    e_w=\sum_{c_w}p(c_w)v_{c_w}\nonumber
\end{equation}
\noindent where $c_w$ is the latent code for word $w$, $v_{c_w}$ is the vector for latent code $c_w$, and $e_w$ is the resulting word embedding for word $w$. In our models, we use 10 binary codes produced by 10 feedforward neural networks based on a shared word embedding, and then we concatenate these 10 individual cluster vectors to get the final word embeddings.

\paragraph{Multi-Task Training.} Aside from ELBO, we use multi-task training losses: \prl and \wpl, as described in \cref{vgvae-representation-sec:multitask}.

\paragraph{Word Noising via Part-of-Speech Tags.}
\begin{figure}
    \centering
    \includegraphics[scale=0.6]{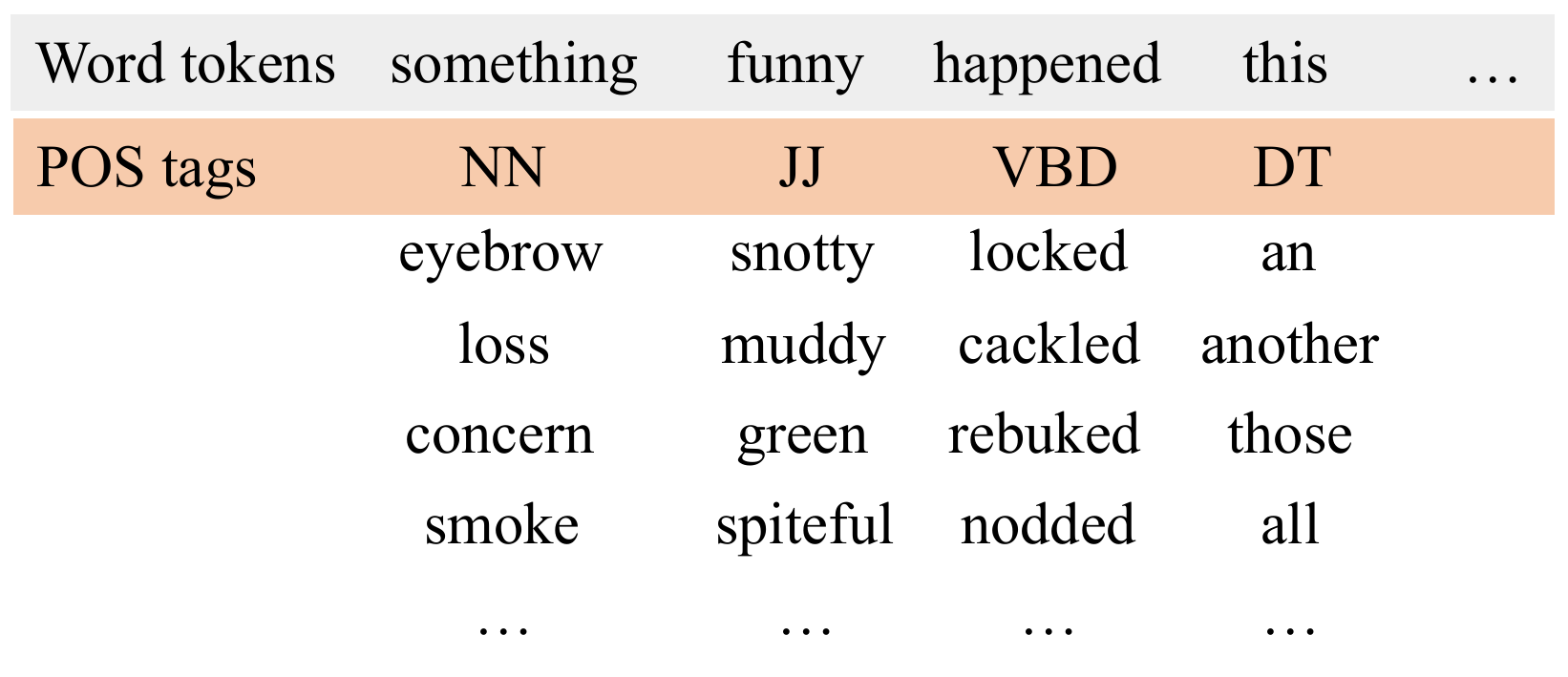}
    \caption{An example of word noising. For each word token in the training sentences, we randomly replace it with other words that share the same POS tags.
    }
    \label{vgvae-generation-fig:word-noise}
\end{figure}
In practice, we often observe that the syntactic encoder tends to remember word types instead of learning syntactic structures. To provide a more flexible notion of syntax, we add word noising (\wn) based on part-of-speech (POS) tags. More specifically, we tag the training set using the Stanford POS tagger~\citep{toutanova-etal-2003-feature}. Then we group the word types based on the top two most frequent tags for each word type. During training, as shown in \cref{vgvae-generation-fig:word-noise}, we noise the syntactic inputs by randomly replacing word tokens based on the groups and tags we obtained. This provides our framework many examples of word interchangeability based on POS tags, and discourages the syntactic encoder from memorizing the word types in the syntactic input. When using \wn, the probability of noising a word is tuned based on development set performance.
\subsection{Experimental Setup}

\paragraph{Training Setup.} For training with the PRL, we require a training set of sentential paraphrase pairs. We use ParaNMT, a dataset of approximately 50 million paraphrase pairs. 
To ensure there is enough variation between paraphrases, we filter out paraphrases with high BLEU score between the two sentences in each pair, which leaves us with around half a million paraphrases as our training set. All hyperparameter tuning is based on the BLEU score on the development set (see appendix for more details). Code and data are available at \url{https://github.com/mingdachen/syntactic-template-generation}.

\paragraph{Evaluation Dataset and Metrics.}
To evaluate models quantitatively, we manually annotate 1300 instances based on paraphrase pairs from ParaNMT independent from our training set. Each instance in the annotated data has three sentences: semantic input, syntactic input, and reference, where the semantic input and the reference can be seen as human generated paraphrases and the syntactic input shares its syntax with the reference but is very different from the semantic input in terms of semantics. The differences among these three sentences ensure the difficulty of this task. \cref{vgvae-generation:example} shows examples.

The annotation process involves two steps. We begin with a paraphrase pair $\langle u, v\rangle$. First, we use an automatic procedure to find, for each sentence $u$, a syntactically-similar but semantically-different other sentence $t$. We do this by seeking sentences $t$ with high edit distance of predicted POS tag sequences and low BLEU score with $u$. Then we manually edit all three sentences to ensure (1) strong semantic match and large syntactic variation between the semantic input $u$ and reference $v$, (2) strong semantic match between the syntactic input $t$ and its post-edited version, and (3) strong syntactic match between the syntactic input $t$ and the reference $v$. We randomly pick 500 instances as our development set and use the remaining 800 instances as our test set. We perform additional manual filtering and editing of the test set to ensure quality.

For evaluation, we consider two categories of automatic evaluation metrics, designed to capture different components of the task. To measure roughly the amount of semantic content that matches between the predicted output and the reference, we report BLEU score (BL), METEOR score (MET; \citealp{banerjee-lavie-2005-meteor}) and three ROUGE scores, including ROUGE-1 (R-1), ROUGE-2 (R-2) and ROUGE-L (R-L). Even though these metrics are not purely based on semantic matching, we refer to them in this paper as ``semantic metrics'' to differentiate them from our second metric category, which we refer to as a ``syntactic metric''. For the latter, to measure the syntactic similarity between generated sentences and the reference, we report the syntactic tree edit distance (\sted). To compute \sted, we first parse the sentences using Stanford CoreNLP, and then compute the tree edit distance~\citep{zhang1989simple} between constituency parse trees after removing word tokens.

\paragraph{Baselines.} We report results for three baselines. The first two baselines directly output the corresponding syntactic or semantic input for each instance. For the last baseline, we consider SCPN~\citep{iyyer-etal-2018-adversarial}. As SCPN requires parse trees for both the syntactic and semantic inputs, we follow the process in their paper and use the Stanford shift-reduce constituency parser~\citep{manning-etal-2014-stanford} to parse both, then use the parsed sentences as inputs to SCPN. We report results for SCPN when using only the top two levels of the parse as input (template) and using the full parse as input (full parse).
\subsection{Experimental Results}

\begin{table}[t]
\centering
\setlength{\tabcolsep}{3pt}
\footnotesize
\begin{tabular}{|l|cccccc|}\hline
        & BLEU ($\uparrow$) & ROUGE-1 ($\uparrow$) & ROUGE-2 ($\uparrow$) & ROUGE-L ($\uparrow$) & METEOR ($\uparrow$) & ST ($\downarrow$) \\
\hline
\multicolumn{7}{c}{Return-input baselines}\\
\hline
Semantic input & 18.5 & 50.6 & 23.2 & 47.7 & 28.8 & 12.0 \\
Syntactic input & 3.3 & 24.4 & 7.5 & 29.1 & 12.1 & 5.9 \\
\hline
\multicolumn{7}{c}{Our work}\\
\hline
\vgvae              & 3.5  & 24.8 & 7.3  & 29.7 & 12.6 & 10.6 \\
\vgvae + \wpl       & 4.5  & 26.5 & 8.2  & 31.5 & 13.3 & 10.0 \\
\vgvae + \lc        & 3.3  & 24.0 & 7.2  & 29.4 & 12.5 & 9.1 \\
\vgvae + \lc + \wpl & 5.9  & 29.1 & 10.2 & 33.0 & 14.5 & 9.0 \\
\vgvae + \wn        & 13.0 & 43.2 & 20.2 & 47.0 & 23.8 & 6.8 \\
\vgvae + \wn + \wpl & 13.2 & 43.4 & 20.3 & 47.0 & 23.9 & 6.7 \\
ALL & 13.6 & 44.7 & 21.0 & 48.3 & 24.8 & 6.7 \\
\hline
\multicolumn{7}{c}{Prior work using supervised parsers}\\
\hline
SCPN + template & 17.8 & 47.9 & 22.8 & 48.5 & 27.3 & 9.9 \\
SCPN + full parse & 19.2 & 50.4 & 26.1 & 53.5 & 28.4 & 5.9 \\\hline
\end{tabular}
\caption{Test results. The final metric (ST) measures the syntactic match between the output and the reference. ALL: \vgvae + \lc + \wn + \wpl.
}
\label{vgvae-generation:test-set-res}
\end{table}

As shown in \cref{vgvae-generation:test-set-res}, simply outputting the semantic input shows strong performance across the BLEU, ROUGE, and METEOR scores, which are more relevant to semantic similarity, but shows much worse performance in terms of \sted. On the other hand, simply returning the syntactic input leads to lower BLEU, ROUGE, and METEOR scores but also a very strong \sted score. These trends provide validation of the evaluation dataset, as they show that the reference and the semantic input match more strongly in terms of their semantics than in terms of their syntax, and also that the reference and the syntactic input match more strongly in terms of their syntax than in terms of their semantics. The goal in developing systems for this task is then to produce outputs with higher semantic metric scores than the syntactic input baseline and simultaneously higher syntactic scores than the semantic input baseline.

Among our models, adding \wpl leads to gains across both the semantic and syntactic metric scores. The gains are much larger without \wn, but even with \wn, adding \wpl improves nearly all scores. Adding \lc typically helps the semantic metrics (at least when combined with WPL) without harming the syntactic metric (\sted).  We see the largest improvements, however, by adding \wn, which uses an automatic part-of-speech tagger at training time only. Both the semantic and syntactic metrics increase consistently with \wn, as the syntactic variable is shown many examples of word interchangeability based on POS tags.

While the SCPN yields very strong metric scores, there are several differences that make the SCPN results difficult to compare to those of our models. In particular, the SCPN uses a supervised parser both during training and at test time, while our strongest results merely require a POS tagger and only use it at training time. Furthermore, since \sted is computed based on parse trees from a parser, systems that explicitly use constituency parsers at test time, such as SCPN, are likely to be favored by such a metric. This is likely the reason why SCPN can match the syntactic input baseline in \sted. Also, SCPN trains on a much larger portion of ParaNMT. We find large differences in metric scores when SCPN only uses a parse template (i.e., the top two levels of the parse tree of the syntactic input). In this case, the results degrade, especially in \sted, showing that the performance of SCPN depends on the quality of the input parses. 
Nonetheless, the SCPN results show the potential benefit of explicitly using a supervised constituency parser at both training and test time. 
Future work can explore ways to combine syntactic parsers with our models for more informative training and more robust performance. 

\subsection{Analysis}

\paragraph{Effect of Paraphrase Reconstruction Loss.}\label{vgvae-generation-sec:effect-prl}
\begin{table}
\setlength{\tabcolsep}{5pt}
\centering
\small
\begin{tabular}{|l|c|c|c|c|c|c|}\hline
        & BL & R-1 & R-2 & R-L & MET & ST \\
\hline
\vgvae w/o \prl   & 2.0  & 23.4 & 4.3  & 26.4 & 11.3 & 11.8 \\
\vgvae w/ \prl    & 3.5  & 24.8 & 7.3  & 29.7 & 12.6 & 10.6 \\\hline
\end{tabular}
\caption{Test results when including \prl.}
\label{vgvae-generation:prl-res}
\end{table}
We investigate the effect of \prl by removing \prl from training, which effectively makes \vgvae a variational autoencoder. As shown in \cref{vgvae-generation:prl-res}, making use of pairing information can improve performance both in the semantic-related metrics and syntactic tree edit distance.

\paragraph{Effect of Position of Word Position Loss.}
\begin{table}
\setlength{\tabcolsep}{5pt}
\centering
\small
\begin{tabular}{|l|c|c|c|c|c|c|}\hline
        & BL & R-1 & R-2 & R-L & MET & ST \\
\hline
\vgvae w/o \wpl       & 3.5  & 24.8 & 7.3  & 29.7 & 12.6 & 10.6 \\
Dec. hidden state     & 3.6  & 24.9 & 7.3  & 29.7 & 12.6 & 10.5 \\
Enc. emb.             & 3.9  & 26.1 & 7.8  & 31.0 & 12.9 & 10.2 \\
Dec. emb.             & 4.1  & 26.3 & 8.1  & 31.3 & 13.1 & 10.1  \\
Enc. \&\ Dec. emb.      & 4.5  & 26.5 & 8.2  & 31.5 & 13.3 & 10.0 \\\hline
\end{tabular}
\caption{Test results with \wpl at different positions. }
\label{vgvae-generation:wpl-res}
\end{table}

We also study the effect of the position of \wpl by (1) using the decoder hidden state, (2) using the concatenation of word embeddings in the syntactic encoder and the syntactic variable, (3) using the concatenation of word embeddings in the decoder and the syntactic variable,  or (4) adding it on both the encoder embeddings and decoder word embeddings. \cref{vgvae-generation:wpl-res} shows that adding \wpl on hidden states can help improve performance slightly but not as good as adding it on word embeddings. In practice, we also observe that the value of \wpl tends to vanish when using \wpl on hidden states, which is presumably caused by the fact that LSTMs have sequence information, making the optimization of \wpl trivial. We also observe that adding \wpl to both the encoder and decoder brings the largest improvement.

\paragraph{Semantic Similarity.}

\begin{table}
\setlength{\tabcolsep}{5pt}
\centering
\small
\begin{tabular}{|l|c|c|}\hline
        & Semantic var. & Syntactic var. \\
\hline
\vgvae                    &   64.8 &   14.5 \\
\vgvae + \wpl             &   65.2 &   10.5 \\
\vgvae + \lc              &   67.2 &   29.0 \\
\vgvae + \lc + \wpl       &   67.9 &   8.5  \\
\vgvae + \wn              &   71.1 &   10.2  \\
\vgvae + \wn + \wpl       &   72.9 & 9.8 \\
\vgvae + \lc + \wn + \wpl &   74.3 & 7.4 \\\hline
\end{tabular}
\caption{Pearson correlation (\%) for STS Benchmark test set.}
\label{vgvae-generation:encoder-sts}
\end{table}

We use cosine similarity between two variables encoded by the inference networks as the predictions and then compute Pearson correlations on the STS Benchmark test set~\citep{cer-etal-2017-semeval}. As shown in \cref{vgvae-generation:encoder-sts}, the semantic variable $y$ always outperforms the  syntactic variable $z$ by a large margin, suggesting that different variables have captured different information. Every time when we add \wpl the differences in performance between the two variables increases. Moreover, the differences between these two variables are correlated with the performance of models in \cref{vgvae-generation:test-set-res}, showing that a better generation system has a more disentangled latent representation.

\paragraph{Syntactic Similarity.}

\begin{table}
\setlength{\tabcolsep}{5pt}
\centering
\small
\begin{tabular}{|l|c|c|c|c|}\hline
&\multicolumn{2}{c|}{Semantic var.} & \multicolumn{2}{c|}{Syntactic var.} \\
   & $F_1$ & Acc. & $F_1$ & Acc.  \\
\hline
Random & 19.2 & 12.9 & - & - \\
Best   & 71.1 & 62.3 & - & - \\
\hline
\vgvae                &20.7 & 24.9 &25.9 & 28.8 \\
\vgvae + \wpl         &21.2 & 25.3 &31.1 & 33.3 \\
\vgvae + \lc          &21.6 & 25.5 &29.0 & 32.4  \\
\vgvae + \lc + \wpl   &18.9 & 23.5 &31.2 & 33.5 \\
\vgvae + \wn          &20.6 & 18.1 &28.4 & 30.4 \\
\vgvae + \wn + \wpl   &20.0 & 24.6 &43.7 & 40.8 \\
\vgvae + \lc +\wn + \wpl & 20.3 & 24.8 & 43.7 & 40.9\\\hline
\end{tabular}
\caption{Labeled $F_1$ score (\%) and accuracy (\%) on syntactic similarity tasks from \cref{section:vgvae-representation}.} 
\label{vgvae-generation:encoder-syntax}
\end{table}

We use the syntactic evaluation tasks from \cref{section:vgvae-representation} to evaluate the syntactic knowledge encoded in the encoder. The tasks are based on a 1-nearest-neighbor constituency parser or POS tagger. To understand the difficulty of these two tasks, \cref{vgvae-generation:encoder-syntax} shows results for two baselines. ``Random'' means randomly pick candidates as predictions. The second baseline (``Best'') is to compute the pairwise scores between the test instances and the sentences in the candidate pool and then take the maximum values. It can be seen as the upper bound performance for these tasks.

As shown in \cref{vgvae-generation:encoder-syntax}, similar trends are observed as in  \cref{vgvae-generation:test-set-res,vgvae-generation:encoder-sts}. When adding \wpl or \wn, there is a boost in the syntactic similarity for the syntactic variable. Adding \lc also helps the performance of the syntactic variable slightly.

\paragraph{Latent Code Analysis.}
\begin{table}[t]
\setlength{\tabcolsep}{4pt}
\centering
\small
\begin{tabular}{|l|c|}
\hline
12 & does must could shall do wo 's did ai 'd 'll should \\\hline
451 & watching wearing carrying thrown refuse drew  \\ \hline
11 & ? : * $\gg$ ! ; ) . '' , '  \\ \hline
18 & maybe they because if where but we when how \\ \hline
41279 & elvish festive freeway anteroom jennifer terrors  \\ \hline
10 & well $\langle$unk$\rangle$ anyone okay now everybody someone \\ \hline
165 & supposedly basically essentially rarely officially  \\ \hline
59 & using on by into as the with within under quite \\ \hline
\end{tabular}
\caption{Examples of learned word clusters. Each row is a different clusters. Numbers in the first column indicate the number of words in that cluster.}
\label{vgvae-generation:lc-analysis}
\end{table}

We look into the learned word clusters by taking the argmax of latent codes and treating it as the cluster membership of each word. Although these are not the exact word clusters we would use during test time (because we marginalize over the latent codes), it provides us intuition on what individual cluster vectors have contributed to the final word embeddings. As shown in \cref{vgvae-generation:lc-analysis}, the words in the first and last rows are mostly function words. The second row has verbs. The third row has special symbols. The fourth row also has function words but somewhat different from the first row. The fifth row is a large cluster populated by content words, mostly nouns and adjectives. The sixth row has words that are not very important semantically and the seventh row has mostly adverbs. We also observe that the size of clusters often correlates with how strongly it relates to topics. In \cref{vgvae-generation:lc-analysis}, clusters that have size under 20 are often function words while the largest cluster (5th row) has words with the most concrete meanings.

\begin{table}[t]
\setlength{\tabcolsep}{5pt}
\centering
\small
\begin{tabular}{|l|c|c|c|c|c|c|}\hline
        & BL & R-1 & R-2 & R-L & MET & ST \\
\hline
\lc        & 13.6 & 44.7 & 21.0 & 48.3 & 24.8 & 6.7 \\
Single \lc & 12.9 & 44.2 & 20.3 & 47.4 & 24.1 & 6.9 \\
\hline
\end{tabular}
\caption{Test results when using a single code.}
\label{lc-varaints-res}
\end{table}

We also compare the performance of \lc by using a single latent code that has 50 classes. The results in Table~\ref{lc-varaints-res} show that it is better to use smaller number of classes for each cluster instead of using a cluster with a large number of classes. 

\paragraph{Effect of Decoder Structure.}
\begin{figure}
    \centering
    \begin{subfigure}[t]{0.1\textwidth}
    \includegraphics[scale=0.7]{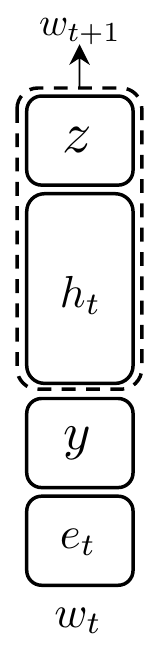}
    \end{subfigure}\hspace{5em}%
    \begin{subfigure}[t]{0.1\textwidth}
    \includegraphics[scale=0.7]{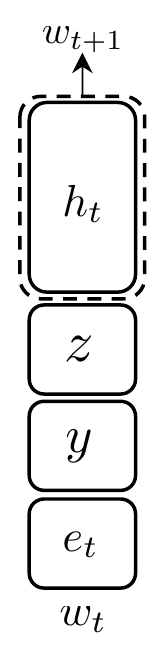}
    \end{subfigure}\hspace{5em}%
    \begin{subfigure}[t]{0.1\textwidth}
    \includegraphics[scale=0.7]{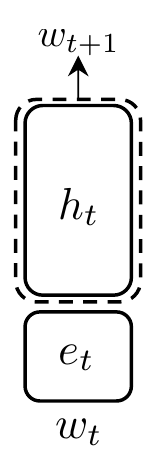}
    \end{subfigure}
    \caption{Variants of decoder. Left (\decswap): we swap the position of variable $y$ and $z$. Middle (\decconcat): we concatenate word embedding with $y$ and $z$ as input to decoder. Right (\decinit): we use word embeddings as input to the decoder and use the concatenation of $y$ and $z$ to compute the initial hidden state of the decoder.}
    \label{vgvae-generation-fig:decoder-variants}
\end{figure}

\begin{table}
\setlength{\tabcolsep}{5pt}
\centering
\small
\begin{tabular}{|l|c|c|c|c|c|c|}\hline
        & BL & R-1 & R-2 & R-L & MET & ST \\
\hline
\vgvae     & 4.5  & 26.5 & 8.2 & 31.5 & 13.3 & 10.0 \\
\decinit   & 3.5  & 22.7 & 6.0 & 24.9 & 9.8  & 11.5 \\
\decconcat & 4.0  & 23.9 & 6.6 & 27.9 & 11.2 & 10.9 \\
\decswap   & 4.3  & 25.6 & 7.5 & 30.4 & 12.5 & 10.5 \\\hline
\end{tabular}
\caption{Test results with decoder variants. }
\label{vgvae-generation:dec-variants-res}
\end{table}

As shown in \cref{vgvae-generation-fig:decoder-variants}, we evaluate three variants of the decoder, namely \decinit, \decconcat, and \decswap. For \decinit, we use the concatenation of semantic variable $y$ and syntactic variable $z$ for computing the initial hidden state of decoder and then use the word embedding as input and hidden state to predict the next word. For \decconcat, we move both $y$ and $z$ to the input of the decoder and use the concatenation of these two variables as input to the decoder and use the hidden state for predicting the next word. For \decswap, we swap the position of $y$ and $z$ to use the concatenation of $y$ and word embeddings as input to the decoder and the concatenation of $z$ and hidden states as output for predicting the next word. Results for these three settings are shown in \cref{vgvae-generation:dec-variants-res}. \decinit performs the worst across the three settings. Both \decconcat and \decswap have variables in each time step in the decoder, which improves performance. \decswap arranges variables in different positions in the decoder and further improves over \decconcat in all metrics.

\paragraph{Generated Sentences.}

\begin{table}
\setlength{\tabcolsep}{4pt}
\centering
\small
\begin{tabular}{|p{0.18\textwidth}|p{0.18\textwidth}|p{0.18\textwidth}|p{0.18\textwidth}|p{0.18\textwidth}|}
\hline
\multicolumn{1}{|c|}{Semantic input} & \multicolumn{1}{c|}{Syntactic input} & \multicolumn{1}{c|}{Reference} & \multicolumn{1}{c|}{SCPN + full parse} & \multicolumn{1}{c|}{Our best model} \\
\hline
don't you think that's a quite aggressive message? & that's worth something, ain't it? & that's a pretty aggressive message, don't you think? & that's such news, don't you? & that's impossible message, aren't you? \\\hline

if i was there, i would kick that bastard in the ass.& they would've delivered a verdict in your favor. & i would've kicked that bastard out on his ass. & you'd have kicked the bastard in my ass. & she would've kicked the bastard on my ass. \\\hline

with luck, it may turn out you're right. & of course, i could've done better. & if lucky, you will be proved correct. & with luck, i might have gotten better. & of course, i'll be getting lucky. \\\hline

they can't help, compassion is unbearable. & love is straightforward and it is lasting. & their help is impossible and compassion is insufferable. & compassion is unbearable but it is excruciating. & compassion is unacceptable and it is intolerable. \\\hline

her yelling sounds sad.  &  she looks beautiful. shining like a star. & she sounds sad. yelling like that. & she's sad. screaming in the air. & she sounds sad. screaming like a scream. \\\hline

me, scare him? & how dare you do such thing? & how can i scare him? & why do you have such fear? & why do you scare that scare?\\
\hline

\end{tabular}
\caption{Examples of generated sentences.}
\label{vgvae-generation:gen-sents}
\end{table}

We show several generated sentences in \cref{vgvae-generation:gen-sents}. We observe that both SCPN and our model suffer from the same problems. When comparing syntactic input and results from both our models and SCPN, we find that they are always the same length. This can often lead to problems like the first example in \cref{vgvae-generation:gen-sents}. The length of the syntactic input is not sufficient for expressing the semantics in the semantic input, which causes the generated sentences from both models to end at ``you?'' and omit the verb ``think''. Another problem is in the consistency of pronouns between the generated sentences and the semantic inputs. An example is the second row in \cref{vgvae-generation:gen-sents}. Both models alter ``i'' to be either ``you'' or ``she'' while the ``kick that bastard in the ass'' becomes ``kicked the bastard in my ass''. 

We found that our models sometimes can generate nonsensical sentences, for example the last row in \cref{vgvae-generation:gen-sents}.
while SCPN, which is trained on a much larger corpus, does not have this problem. Also, our models can sometimes be distracted by the word tokens in the syntactic input as shown in the 3rd row in \cref{vgvae-generation:gen-sents}, where our model directly copies ``of course'' from the syntactic input while since SCPN uses a parse tree, it outputs ``with luck''. In some rare cases where the function words in both syntactic inputs and the references are the exactly the same, our models can perform better than SCPN, e.g., the last two rows in \cref{vgvae-generation:gen-sents}. Generated sentences from our model make use of the word tokens ``and'' and ``like'' while SCPN does not have access to this information and generates inferior sentences.

\section{Summary}
In this chapter, we demonstrated the utility of our proposed latent-variable framework in the context of representation learning (\cref{section:vgvae-representation}) and controllable generation (\cref{section:vgvae-generation}). In both cases, we leveraged the structures of paired data to disentangle semantics and syntax in sentence representations. We found that the syntactic and semantic latent variables showed desirable characteristics. For controlled generation, we provided human-annotated evaluation sets to promote future research in this direction. In addition, in a follow-up work, we showed that the multi-task, latent-variable framework can generalize to bilingual text corpora \citep{chen2020exemplar}.
\chapter{Tailoring Textual Resources for Evaluation Tasks}
\label{CHAPTER:EVALUATION}

This chapter describes our contributions to building evaluation tasks from naturally-occurring textual resources. In \cref{sec:wikitablet}, we cast generating arbitrary Wikipedia sections as a data-to-text generation problem. We leverage different data sources to create tabular data for a given section text. In \cref{sec:summscreen} and \cref{sec:tvstorygen}, we use fan-contributed websites to create summarization and story generation datasets. Due to the rich information provided on these websites, the resulting datasets offer unique challenges in their respective task settings.

The material in this chapter is adapted from \citet{chen-etal-2022-summscreen}, \citet{chen-etal-2021-wikitablet}, and \citet{chen2021tvrecap}.

\section{Long-Form Data-to-Text Generation}
\label{sec:wikitablet}
\subsection{Introduction}

Data-to-text generation \citep{kukich-1983-design,mckeown1992text} is the task of generating text based on structured data. Most existing data-to-text datasets focus on single-sentence generation, such as \wikibio \citep{lebret-etal-2016-neural}, LogicNLG \citep{chen-etal-2020-logical}, and ToTTo \citep{parikh-etal-2020-totto}. Other datasets are relatively small-scale and focus on long-form text generation, such as \textsc{RotoWire} \citep{wiseman-etal-2017-challenges} and MLB \citep{puduppully-etal-2019-data}. In this work, we cast generating Wikipedia sections as a data-to-text generation task and build a large-scale dataset targeting multi-sentence data-to-text generation with a variety of domains and data sources.

To this end, we create a dataset that we call \wikitablet (``Wikipedia Tables to Text'') that pairs Wikipedia sections with their corresponding tabular data and various metadata. The data resources we consider are relevant either to entire Wikipedia articles, such as Wikipedia infoboxes and Wikidata tables, or to particular sections. Data from the latter category is built automatically from either naturally-occurring hyperlinks or from named entity recognizers. This data construction approach allows us to collect large quantities of instances while still ensuring the coverage of the information in the table. We also perform various types of filtering to ensure dataset quality.

\wikitablet contains millions of instances  covering a broad range of topics and a variety of flavors of generation with different levels of flexibility.
\cref{wikitablet-fig:dataset-example} shows two examples from \wikitablet. The first instance has more flexibility as it involves generating a fictional character biography in a comic book, whereas the second is more similar to standard data-to-text generation tasks, where the input tables contain all of the necessary information for generating the text. While the open-ended instances in \wikitablet are to some extent similar to story generation \citep{propp1968morphology,mcintyre-lapata-2009-learning,fan-etal-2018-hierarchical}, the fact that these instances are still constrained by the input tables enables different evaluation approaches and brings new challenges (i.e., being coherent and faithful to the input tables at the same time).

Because of the range of knowledge-backed generation instances in \wikitablet, models trained on our dataset can be used in assistive writing technologies for a broad range of topics and types of knowledge. For example, technologies can aid students in essay writing by drawing from multiple kinds of factual sources. Moreover, \wikitablet can be used as a pretraining dataset for other relatively small-scale data-to-text datasets (e.g., \textsc{RotoWire}). A similar idea that uses data-to-text generation to create corpora for pretraining language models has shown promising results \citep{agarwal-etal-2021-knowledge}. In experiments, we train several baseline models on \wikitablet and empirically compare training and decoding strategies. We find that the best training strategies still rely on enforcing hard constraints to avoid overly repetitive texts. Human evaluations reveal that (1) humans are unable to differentiate the human written texts from the generations from our neural models; (2) while the annotations show that grammatical errors in the reference texts and the generations may prevent humans from fully understanding the texts, the best decoding strategy (i.e., beam search with $n$-gram blocking \citep{paulus2018a}) does not have such a problem and shows the best performance on several aspects; (3) the degree of topical similarity between the generations and the reference texts depends on the open-endedness of the instances.

Our analysis shows that the generations are fluent and generally have high quality, but the models sometimes struggle to generate coherent texts for all the involved entities, suggesting future research directions. For example, when the instance has a high degree of flexibility, we find the models making mistakes about what a particular entity type is capable of. We also find errors in terms of the factuality of the generated text, both in terms of contradictions relative to the tables and commonsense violations. 

\subsection{Related Work}
There have been efforts in creating data-to-text datasets from various resources, including sports summaries \citep{wiseman-etal-2017-challenges,puduppully-etal-2019-data}, weather forecasts \citep{liang-etal-2009-learning}, and commentaries \citep{david2008sport}. Most of the recent datasets focus on generating single sentences given tables, such as \wikibio, ToTTo, LogicNLG, and WikiTableText \citep{bao2018table}, or other types of data formats, such as data triples \citep{vougiouklis2018neural,gardent-etal-2017-webnlg,nan-etal-2021-dart}, abstract meaning representations \citep{flanigan-etal-2016-generation}, minimal recursion semantics \citep{hajdik-etal-2019-neural}, or a set of concepts \citep{lin-etal-2020-commongen}. Other than single sentences, there have been efforts in generating groups of sentences describing humans and animals \citep{wang-etal-2018-describing}, and generating a post-modifier phrase for a target sentence given a sentence context \citep{kang-etal-2019-pomo}. In this work, our focus is long-form text generation and we are interested in automatically creating a large-scale dataset containing multiple types of data-to-text instances. 
As shown in \cref{wikitablet-tab:data-stats}, \wikitablet differs from these datasets in that it is larger in scale and contains multi-sentence texts. More details are in the next section.

Wikipedia has also been used to construct datasets for other text generation tasks, such as generating Wikipedia movie plots \citep{orbach-goldberg-2020-facts2story,rashkin-etal-2020-plotmachines} and short Wikipedia event summaries \citep{gholipour-ghalandari-etal-2020-large}, and summarizing Wikipedia documents  \citep{zopf-2018-auto,j.2018generating} or summaries of aspects of interests \citep{hayashi-etal-2021-wikiasp} from relevant documents.

As part of this work involves finding aligned tables and text, it is related to prior work on aligning Wikipedia texts to knowledge bases \citep{elsahar-etal-2018-rex,logan-etal-2019-baracks}.

\subsection{\wikitablet}
\begin{figure}
    \centering
    \begin{subfigure}[t]{1.0\textwidth}
    \centering
    \includegraphics[scale=0.46]{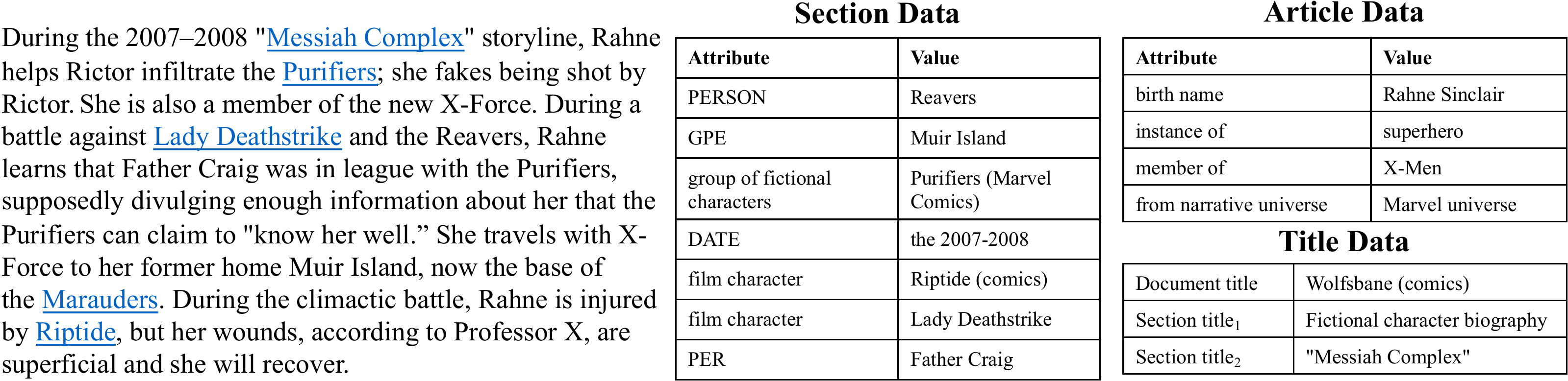}
    \end{subfigure}
    \begin{subfigure}[t]{1.0\textwidth}
    \vspace{0.1em}
    \centering
    \includegraphics[scale=0.46]{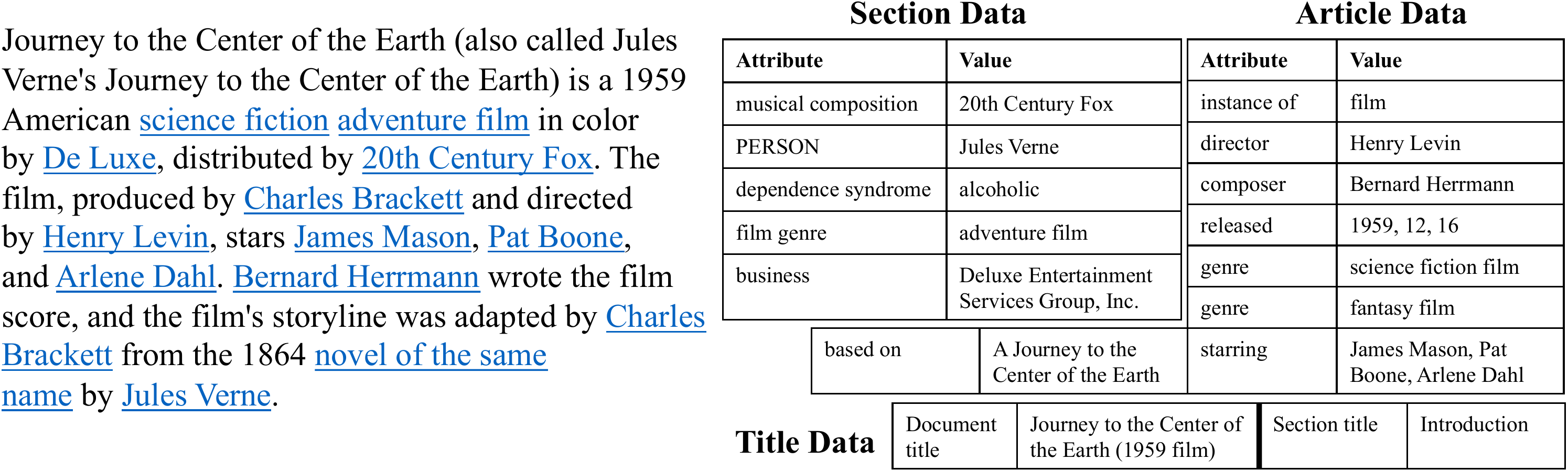}
    \end{subfigure}
    \caption{Two examples from \wikitablet. Only parts of the tables are shown due to space constraints. Underlined texts are hyperlinks. Records with the attributes ``DATE'', ``PER'', ``PERSON'', 
    or ``GPE'' are from NER. The subscripts for section titles indicate the ordering of nesting, where smaller numbers are for higher level sections.
    }
    \label{wikitablet-fig:dataset-example}
\end{figure}

The \wikitablet dataset pairs Wikipedia sections\footnote{We define a Wikipedia section to be all text starting after a (sub)section heading and proceeding until the next (sub)section heading. We include Wikipedia sections at various nesting levels. For example, a top level section may start with a few paragraphs describing general information followed by two subsections with more specific information, in which case the example will be converted into three instances in our dataset.} with their corresponding tabular data and various metadata; some of this data is relevant to entire Wikipedia articles (``article data'') or article structure (``title data''), while some is section-specific (``section data''). Each data table consists of a set of \textbf{records}, each of which is a tuple containing an \textbf{attribute} and a \textbf{value}.  

The instances in \wikitablet cover a range of flavors of language generation. Some have more flexibility, requiring models to generate coherent stories based on the entities and knowledge given in the tables. The first instance in \cref{wikitablet-fig:dataset-example} is such an example. The text is from the Wikipedia article entitled ``Wolfsbane (comics)'' and resides within two nested sections: the higher-level section ``Fictional character biography'' and the lower-level section ``Messiah Complex''. The task is challenging as models need to generate a coherent passage that can connect all the entities in the section data, and the story also needs to fit the background knowledge provided in the article data.

Other instances are more similar to standard data-to-text generation tasks, where the input tables contain all the necessary information for generating the text. The second instance in \cref{wikitablet-fig:dataset-example} is an example of this sort of task. However, these tasks are still challenging due to the wide variety of topics contained in \wikitablet.

\paragraph{Dataset Construction.}

We begin by describing the steps we take to construct \wikitablet. More details are in \cref{wikitablet-appendix-sec:dataset-construction}. In general, the steps can be split into two parts: collecting data tables and filtering out texts.
When collecting data, we consider five resources: \wikidata tables, infoboxes in Wikipedia pages,\footnote{Wikidata is a consistently-structured knowledge base (e.g., has a fixed set of attributes), whereas infoboxes are not consistently-structured and this flexibility sometimes allows the infobox to contain extra information. Therefore, we consider using infoboxes as extra resources.} hyperlinks in the passage, named entities in the passage obtained from named entity recognition (NER), and Wikipedia article structure.
For a given Wikipedia article, we use the same infobox and \wikidata table for all sections. These tables can serve as background knowledge for the article.
For each section in the article, we create a second table corresponding to section-specific data, i.e., section data. 
The section data contains records constructed from hyperlinks and entities identified by a named entity recognizer.\footnote{We use the NER tagger from spaCy \citep{spacy2} and a BERT model  \citep{devlin-etal-2019-bert} finetuned on CoNLL03 data \citep{tjong-kim-sang-de-meulder-2003-introduction}.}

We form records for named entities by using the type of the entity as the attribute and the identified entity as the value. We form records for hyperlinks as follows. For the attribute, for a hyperlink with surface text $t$ and hyperlinked article $\ell$, we use the value of the ``instance of'' or ``subclass of'' tuple in the \wikidata table for $\ell$. For example, the first instance in Figure \ref{wikitablet-fig:dataset-example} will be turned into a record with attribute ``superhero'' and value ``Wolfsbane (comics)''. If $\ell$ does not have a \wikidata table or no appropriate tuple, we consider the parent categories of $\ell$. For the value of the tuple, we use the document title of $\ell$ rather than the actual surface text $t$ to avoid giving away too much information in the reference text.

Complementary to the article data, we create a title table that provides information about the position in which the section is situated, which includes the article title and the section titles for the target section. As the initial sections in Wikipedia articles do not have section titles, we use the section title ``Introduction'' for these.

We also perform various filtering to ensure the quality of the data records, the coverage of the input data, and the length of the reference text. The final dataset contains approximately 1.5 million instances. We randomly sample 4533 instances as the development set and 4351 as the test set. We also ensure that there are no overlapping Wikipedia articles among splits.

\begin{table}
    \centering
    \footnotesize
\begin{tabular}{|l|ccccccc|}
\hline
& Vocab. & Tokens & Examples & AL & RT & AR & Domain \\
\hline
WikiTableText & - & 185.0k & 13.3k & 13.9 & 3.0k & 4.1 & Wikipedia \\
\textsc{WikiBio} & 400.0k & 19.0M & 728.0k & 26.1 & 1.7k & 19.7 & Biography \\
\textsc{RotoWire} & 11.3k & 1.6M & 4.9k & 337.1 & 39.0 & 628.0 & Sports \\
MLB & 38.9k & 14.3M & 26.3k & 542.1 & 53.0 & 565.0 & Sports \\
LogicNLG & 122.0k & 52.7k & 37.0k & 14.2 & 11.7k & 13.5 & Wikipedia \\
ToTTo & 136.0k & 1.3M & 136.0k & 17.4 & 41.8k & 32.7 & Wikipedia \\
DART & 33.2k & 717.1k  & 82.2k & 21.6 & - & - & Wikipedia+Restaurant\\
\wikitablet & 1.9M & 169.0M & 1.5M$^\ast$
& 115.9 & 147.4k$^\dag$
& 51.9 & Wikipedia \\
\hline
\end{tabular}
\caption{
Statistics for several data-to-text datasets. \wikitablet combines a large number of examples, moderate generation length (typically more than one sentence), and a large variety of record types.
We omit record types and avg.~records for DART as its input units are triple sets instead of table records. $^\ast$887.7k unique Wikipedia articles. $^\dag$Number of record types for each resource: 31.8k (Infobox), 1.7k (\wikidata), 115.6k (Hyperlinks), 17 (NER). AL=Average Length. RT=Record Types. AR=Average Records.
}\label{wikitablet-tab:data-stats}
\end{table}

\paragraph{Dataset Characteristics.}
\cref{wikitablet-tab:data-stats} shows statistics for \wikitablet and related datasets. While the average length of a \wikitablet instance is not longer than some of the existing datasets, \wikitablet offers more diverse topics than the sports-related datasets \textsc{RotoWire} and MLB, or the biography-related dataset \wikibio. Compared to the prior work that also uses Wikipedia for constructing datasets, \wikibio, LogicNLG, ToTTo, and DART \citep{nan-etal-2021-dart} all focus on sentence generation, whereas \wikitablet requires generating Wikipedia article sections, which are typically multiple sentences and therefore more challenging. \wikitablet is also much larger than all existing datasets.

\begin{table}[t]
    \centering
    \small
\begin{tabular}{|l|c|}\hline
Category & Fraction (\%) \\\hline
human & 45.62 \\
film & \phantom{1}4.61 \\
single (music) & \phantom{1}1.74 \\
human settlement & \phantom{1}1.53 \\
album & \phantom{1}1.41 \\
sports season & \phantom{1}1.26 \\
television series & \phantom{1}1.17 \\
village & \phantom{1}1.12 \\
taxon & \phantom{1}0.89 \\
\hline
\end{tabular}

    \caption{Top 10 most frequent article categories and their corresponding proportions in \wikitablet.}
    \label{wikitablet-tab:data-doc-cat-freq}
\end{table}

To demonstrate the diversity of topics covered in \wikitablet, we use either the ``instance of'' or ``subclass of'' relation from \wikidata as the category of the article.\footnote{When there are multiple values in these two relations, we pick the one that has the smallest number of words, as it often is the most generic phrase, suitable for representing the topic.} 
We show the top 10 most frequent document categories in \cref{wikitablet-tab:data-doc-cat-freq}. Due to the criteria we use for filtering, only 1.05\% of articles in \wikitablet do not have these relations or \wikidata entries, and we omit these articles in the table. As the table demonstrates, more than 50\% of the articles in \wikitablet are not about people (i.e., the topic of \wikibio), within which the most frequent category covers only 4.61\%.

We highlight two challenges of \wikitablet:
\begin{enumeratesquish}
    \item In contrast to work on evaluating commonsense knowledge in generation where reference texts are single sentences describing everyday scenes \citep{lin-etal-2020-commongen}, \wikitablet can serve as a testbed for evaluating models’ abilities to use world knowledge for generating coherent long-form text.
    \item Compared to other long-form data-to-text datasets such as \textsc{RotoWire} where the input tables are box scores, the input tables in \wikitablet are more diverse, including both numbers (e.g., economy and population data of an area throughout years), and short phrases. This makes \wikitablet more challenging and applicable to various scenarios.
\end{enumeratesquish}

\subsection{Method}

In this section, we describe details of models that we will benchmark on \wikitablet.

Our base model is based on the transformer \citep{attention_is_all_you_need}. To encode tables, we linearize the tables by using special tokens to separate cells and using feature embeddings to represent records in tables. For the title table in the first instance in \cref{wikitablet-fig:dataset-example} 
the linearized table will be
\begin{equation}
\begin{aligned}
    &\langle\text{boc}\rangle_1 \text{Doc.}_1\text{ title}_1 \langle\text{bov}\rangle_1 \text{ Wolfsbane}_1\text{ (comics)}_1\\ &\langle\text{boc}\rangle_2 \text{Sec.}_2\text{ title}_2 \langle\text{bov}\rangle_2 \text{ Fictional}_2\text{ character}_2\\ &\text{ biography}_2\langle\text{boc}\rangle_3 \cdots \langle\text{eoc}\rangle
\end{aligned}
\label{wikitablet-eq:linear-table}
\end{equation}

As shown in \cref{wikitablet-eq:linear-table}, we employ several techniques when encoding tables: (1) we use special tokens $\langle\text{boc}\rangle$ and $\langle\text{bov}\rangle$ to separate attributes and values, and  $\langle\text{eoc}\rangle$ to indicate the end of a sequence; (2) we use subscript indices to indicate unique ID embeddings that are added to the embeddings for each record, which helps models align attributes with values; and (3) we restart the positional embeddings at each $\langle\text{boc}\rangle$, such that models will not use the ordering of the input records. In addition, we add a special embedding to each record to indicate if it is from the section table or the article/title table. In \wikidata, there could be multiple qualifiers attached to a record, in which case we replicate the record for each qualifier separately.

Similar linearization approaches have been used in prior work \citep{dhingra-etal-2019-handling,hwang2019comprehensive,herzig-etal-2020-tapas,yin-etal-2020-tabert}. With linearized tables, training and inference become similar to other sequence-to-sequence settings.
We train our models with teacher-forcing and standard cross entropy loss unless otherwise specified.

We experiment with three types of modifications to standard sequence-to-sequence training and a few decoding strategies:

\paragraph{\entmax.} \entmax~\citep{peters-etal-2019-sparse} is a mapping from scores to a distribution that permits varying the level of sparsity in the distribution. This mapping function has been used in machine translation~\citep{peters-etal-2019-sparse} and text generation~\citep{martins-etal-2020-sparse}. When using \entmax in the decoder, we also replace the cross entropy loss with the \entmax loss~\citep{peters-etal-2019-sparse}. Both \entmax and the \entmax loss have a hyperparameter $\alpha$. We follow \citet{martins-etal-2020-sparse} and use $\alpha=1.2$ as they found it to be the best value for reducing repetition in generation. 

\paragraph{Copy Mechanism.} Similar to prior work on data-to-text generation \citep{wiseman-etal-2017-challenges,puduppully-etal-2019-data}, we use pointer-generator network style copy attention \citep{see-etal-2017-get} in the decoder.

\paragraph{Cyclic Loss.}
Cyclic losses have been shown to be effective in textual style transfer \citep{shetty2018a4nt,pang-gimpel-2019-unsupervised} and neural machine translation~\citep{cheng-etal-2016-semi,dual_nips2016,tu2017neural}. \citet{wiseman-etal-2017-challenges} also used this for data-to-text and found it helpful for generating long sequences. In this work, we experiment with adding the cyclic loss to our transformer models, where the backward model can be seen as an information extraction system. We expect that adding the cyclic loss should enable a data-to-text model to generate sentences that are more faithful to the conditioned tables. In particular,
we denote the linearized table where the values are replaced with a special $\langle\text{mask}\rangle$ token by $u_1,\cdots,u_n$, and denote the reference text by $x_1,\cdots,x_m$. Formally, the training loss is
\begin{equation}
    \sum_{w\in S}-\log p(w\vert u_1,\cdots,u_n,\bs{v}_1,\cdots,\bs{v}_m)\nonumber
\end{equation}
\noindent where $S$ represents the set of masked tokens, and $\bs{v}_1,\cdots,\bs{v}_m$ is the sequence of token-level probabilities predicted by the forward model (in our experiments, these could either come from the softmax function, or the \entmax function). Specifically, we multiply the backward transformer's input embedding matrix by the $\bs{v}$ probability vectors to obtain the input representations to the first encoder layer. We find that it is helpful to add a ``reference loss'' while training with the cyclic loss, defined as
\begin{equation}
    \sum_{w\in S}-\log p(w\vert u_1,\cdots,u_n,x_1,\cdots,x_m)
\end{equation}
\noindent this loss does not contain the generation model in it explicitly, but it does lead to an improved backward model by training it with clean inputs. Improving the backward model then increases the benefits of the cyclic loss.\footnote{We experimented with initializing the backward model with pretrained checkpoints, but did not find it helpful.} The cyclic loss is used during training only and does not affect the models during inference.

\paragraph{Decoding Strategies.}

\citet{massarelli-etal-2020-decoding} showed that the choice of decoding strategy can affect the faithfulness or repetitiveness of text generated by language models. We are also interested in these effects in the context of data-to-text generation, and therefore benchmark several decoding strategies on \wikitablet. 
Our models use byte-pair encoding (BPE; \citealp{sennrich-etal-2016-neural}) and for all of the following strategies, we always set the minimum number of decoding steps to 100 as it improves most of the evaluation metrics, and the maximum number of decoding steps to 300. The details of the decoding algorithms we benchmark are as follows.

\paragraph{Greedy Decoding.} In this setting, we feed the previous predicted tokens to the encoder, and then pick the word with the highest predicted probability as the prediction. This can be seen as a baseline strategy.

\paragraph{Nucleus Sampling.} Generating long sequences usually suffers from repetitions. Nucleus sampling \cite{Holtzman2020The} aims to reduce the repetitions in generations by sampling from truncated probability distributions. The truncation is based on whether the cumulative probability is above a threshold. We set the threshold to be 0.9 as suggested in \citet{Holtzman2020The}.

\paragraph{Beam Search.} Beam search is a widely used deterministic decoding algorithm. It maintains a fixed size set of $k$ candidate sequences. At each time step, it adds words from the vocabulary to these candidates, scores the new sequences, and then picks the ones with the top-$k$ probabilities. We set the beam size $k$ to be 5 by default.

\paragraph{Beam Search with $n$-gram Blocking.} \citet{paulus2018a} found it effective to reduce the repetitions during beam search by ``blocking'' $n$-grams that have been generated in previous decoding steps. We follow their approach by using trigram blocking and setting the probability of repeated trigrams to be 0 during beam search. 

Specifically, we benchmark (1) greedy decoding; (2) nucleus sampling \citep{Holtzman2020The} with threshold 0.9 as suggested by \citet{Holtzman2020The}; (3) beam search; and (4) beam search with $n$-gram blocking \citep{paulus2018a} where we set the probabilities of repeated trigrams to be 0 during beam search. We set the beam size to be 5 by default.

\subsection{Experiments}

\paragraph{Setup.}

\begin{table}[t]
    \centering
    \footnotesize
    \begin{tabular}{|l|c|c|c|c|c|c|c|}\hline
& REP & BLEU & RL & MET & PAR-P & PAR-R & PAR-F1 \\\hline
References & 1.2 & 100.0 & 100.0 & 100.0 & 100.0 & 59.2 & 72.9 \\
Linearized article tables & 8.0 & 2.2 & 14.7 & 9.3 & 100.0 & 16.3 & 25.6 \\
Linearized section tables & 1.0 & 1.9 & 27.9 & 15.5 & 100.0 & 20.9 & 33.4 \\
Linearized tables & 7.9 & 6.4 & 22.0 & 18.3 & 100.0 & 48.3 & 63.0 \\
Linearized tables + references & 7.6 & 36.5 & 61.3 & 56.5 & 99.9 & 100.0 & 100.0 \\\hline
\multicolumn{8}{c}{ Base models trained on the 500k training set (beam search) } \\\hline
Base & 33.0 & 15.6 & 36.9 & 20.3 & 66.3 & 28.8 & 37.7 \\
Base + entmax & 25.9 & 15.4 & 36.2 & 20.3 & 64.6 & 29.0 & 37.7 \\
Base + copy & 30.1 & 15.9 & 37.5 & 20.7 & 67.1 & 29.4 & 38.5 \\
Base + copy + cyclic loss & 28.0 & 15.7 & 37.5 & 20.8 & 67.5 & 29.7 & 38.9 \\\hline
\multicolumn{8}{c}{ Large models trained on the full training set (different decoding strategies) } \\\hline
Large + greedy & 26.8 & 18.9 & 38.5 & 23.5 & 60.4 & 33.1 & 40.4 \\
Large + nucleus sampling & 2.3 & 18.3 & 36.1 & 23.7 & 54.2 & 32.5 & 38.7 \\
Large + beam search & 18.8 & 19.5 & 39.9 & 23.9 & 65.8 & 34.3 & 42.8 \\
Large + beam search + $n$-gram blocking & 1.9 & 19.3 & 39.3 & 24.4 & 62.2 & 35.3 & 43.0 \\\hline
\end{tabular}

    \caption{Test set results for our models. When training the large models, we use the ``copy + cyclic loss'' setting as it gives the best performance for the base models for most of the metrics.}
    \label{wikitablet-tab:main-result}
\end{table}

We experiment with two sizes of transformer models. One is ``Base'', where we use a 1-layer encoder and a 6-layer decoder, each of which has 512 hidden size and 4 attention heads. The other one is ``Large'', where we use a 1-layer encoder and a 12-layer decoder, each of which has 1024 hidden size and 8 attention heads. Models similar to the base configuration have shown strong performance on \textsc{RotoWire}~\citep{gong-etal-2019-enhanced}.\footnote{When training the base model with entmax on \wikibio, it achieves BLEU-4 45.75 and ROUGE-4 39.39 on the test set using greedy decoding, which are comparable to the current state-of-the-art results of~\citet{liu2017aaai}.} Due to limited computational power, we parameterize our backward model as a transformer model with a 2-layer encoder and a 2-layer decoder.\footnote{We did not experiment with pretrained models because they typically use the entirety of Wikipedia, which would presumably overlap with our test set.}

We use BPE with 30k merging operations. We randomly sample 500k instances from the training set and train base models on them when exploring different training strategies. We train a large model with the best setting (using the copy mechanism and cyclic loss)  on the full training set. We train both models for 5 epochs. During training we perform early stopping on the development set using greedy decoding.

We report BLEU, ROUGE-L (RL), METEOR (MET), and PARENT \citep{dhingra-etal-2019-handling}, including precision (PAR-P), recall (PAR-R), and F1 (PAR-F1) scores. The first three metrics consider the similarities between generated texts and references, whereas PARENT also considers the similarity between the generation and the table. When using PARENT, we use all three tables, i.e., the section, article, and title tables.

As we are also interested in the repetitiveness of generated texts, we define a metric based on $n$-gram repetitions which we call ``REP''. REP computes the ratio of the number of repeated $n$-grams to the total number of $n$-grams within a text, so when REP has higher value, it indicates that the text has more repetitions. Here we consider $n$-grams that appear 3 or more times as repetitions and the $n$-grams we consider are from bigrams to 4-grams. When reporting REP scores for a dataset, we average the REP scores for each instance in the dataset. Similar metrics have been used in prior work \citep{Holtzman2020The,Welleck2020Neural}. Code and the dataset are available at \url{https://github.com/mingdachen/WikiTableT}.

\paragraph{Results.}

In \cref{wikitablet-tab:main-result}, we report the test results for both our base models and large models. We also report a set of baselines that are based on simply returning the linearized tables and their concatenations with the references. The linearized table baselines show how much information is already contained in the table, while the reference baselines show the upper bound performance for each metric.

In comparing training strategies, we find that using \entmax improves REP significantly but not other metrics. Adding the cyclic loss or the copy mechanism helps improve performance for the PAR scores and REP, and combining both further improves these metrics.

When comparing decoding strategies, we find that both nucleus sampling and $n$-gram blocking are effective in reducing repetition. Nucleus sampling harms the PAR scores, especially PAR-P, but has less impact on the other metrics, indicating that it makes the model more likely to generate texts that are less relevant to the tables. Using beam search improves all metrics significantly when compared to greedy decoding, especially the PAR-P and REP scores. Adding $n$-gram blocking further reduces the REP score, pushing it to be even lower than that from nucleus sampling, but still retains the improvements in PAR scores from beam search. The best overall decoding strategy appears to be beam search with $n$-gram blocking. 

\subsection{Analysis}

We now describe a manual evaluation and analyze some generated examples. All results in this section use the development set.

\paragraph{Effect of \entmax.}

\begin{table}
    \centering
    \small\setlength{\tabcolsep}{5pt}
\begin{tabular}{|l|c|c|c|c|c|}\hline
 & REP & BLEU & PAR-P & PAR-R & PAR-F1 \\\hline
 \multicolumn{6}{c}{Greedy decoding}
\\\hline
base & 38.1 & 14.7 & 61.6 & 27.7 & 35.8 \\
+ ent. + ent. loss & 36.0 & 16.2 & 62.2 & 28.9 & 37.0 \\
+ ent. & 44.5 & 13.9 & 63.5 & 25.5 & 33.9 \\
+ ent. + copy & 43.7 & 14.8 & 64.2 & 26.6 & 35.2 \\
+ copy & 37.8 & 15.8 & 61.3 & 28.3 & 36.3 \\
\hline
 \multicolumn{6}{c}{Beam search (beam size 5)} \\
\hline
base & 33.0 & 15.6 & 66.3 & 28.8 & 37.7 \\
+ ent. + ent. loss & 25.9 & 15.4 & 64.6 & 29.0 & 37.7 \\
+ ent. & 34.7 & 13.8 & 67.2 & 26.6 & 35.8 \\
+ ent. + copy & 34.1 & 15.0 & 69.4 & 28.1 & 37.6 \\
+ copy & 30.1 & 15.9 & 67.1 & 29.4 & 38.5 \\
\hline
\end{tabular}
    \caption{Effect of using \entmax and \entmax loss. When not using the \entmax loss, we use standard cross entropy loss.}
    \label{wikitablet-tab:effect_entmax}
\end{table}

In this section, we disentangle the effect of \entmax and that of \entmax loss. We note that (1) when not using the \entmax loss, we use standard cross entropy loss (e.g., in the case of ``base+ent.'' we maximize the log probabilities generated by \entmax); (2) when combining \entmax and copy mechanism, we aggregate the probabilities generated by \entmax and those from softmax. This is because we use the first attention head in the transformer decoder as the copy attention, following the implementation in OpenNMT \citep{klein-etal-2017-opennmt}. While it is feasible to combine the \entmax and \entmax loss with the copy mechanism if we use the sparse transformer \citep{correia-etal-2019-adaptively}, we leave this for future study. We report the results in \cref{wikitablet-tab:effect_entmax}. It is interesting to see that when using greedy decoding, ``ent. + ent. loss'' outperforms the baseline model by a significant margin on all the metrics, however the improvement disappears (except for repetition) after we switch to use beam search as the decoding strategy. This is likely because \entmax promotes sparsity in the generated probabilities, making beam search decoding unnecessary. Removing the \entmax loss hurts the performance, but its gains become larger in switching to beam search decoding. Adding copy mechanism improves the performance, leading to comparable performance to the baseline model. Although ``base+ent.+copy'' still underperforms ``base+copy'' when using beam search, we believe that combining \entmax and \entmax loss with the copy mechanism is promising as (1) \entmax is not used in our large models and the initial results have shown that \entmax and the copy mechanism are complementary, so it may further improve our current best performance; (2) \entmax already shows the best performance when using greedy decoding, which has speed and optimization advantages compared to the beam search based decoding strategies especially considering the long-form characteristic of \wikitablet.

\paragraph{Human Evaluation.}

We conduct a human evaluation using generations from the large model on the development set. We choose texts shorter than 100 tokens and that cover particular topics as we found during pilot studies that annotators struggled with texts that were very long or about unfamiliar topics.\footnote{We did not find the filtering to change the observed trends for the automatic metrics and provide the list of selected topics in \cref{wikitablet-appendix-sec:human-eval}.} 

We design two sets of questions. The first focuses on the text itself (i.e., grammaticality and coherence) and its faithfulness to the input article table. Since this set does not involve the reference, we can ask these questions about both generated texts and the reference texts themselves. 

The second set of questions evaluates the differences between the generations and the reference texts (i.e., relevance and support), allowing us to see if the generated text matches the human written section text. Specifically, relevance evaluates topical similarity between generations and references, and support evaluates whether the facts expressed in the generations are supported by or contradictory to those in the references. %
The full questions and numerical answer descriptions %
are in \cref{wikitablet-appendix-sec:human-eval}.  %

\begin{table}[t]
    \centering
    \small\setlength{\tabcolsep}{4pt}
\begin{tabular}{|l|c|c|c|}\hline
 & Grammar & Coherence & Faithfulness \\\hline
Reference & 4.0 (1.0) & 4.1 (0.9) & 3.8 (0.8) \\\hline
Beam search & 4.0 (1.0) & 4.0 (1.0) & 3.9 (1.0) \\
Nucleus sampling & 4.0 (0.8) & 4.1 (0.9) & 3.9 (0.8) \\
$n$-gram blocking & 4.2 (0.9) & 4.2 (0.9) & 3.9 (1.0) \\
\hline
\end{tabular}
    \caption{Average human ratings (standard deviations in parentheses) for grammaticality, coherence, and faithfulness to the input article table. }
    \label{wikitablet-tab:human_eval_table}
\end{table}

\begin{table}[t]
    \centering
    \small\setlength{\tabcolsep}{5pt}
\begin{tabular}{|l|c|c|}\hline
 & Relevance & Support \\\hline
Beam search & 3.8 (1.1) & 3.6 (1.2) \\
Nucleus sampling & 3.7 (1.2) & 3.8 (1.1) \\
$n$-gram blocking & 3.9 (1.0) & 3.8 (1.0) \\
\hline
\end{tabular}
    \caption{Average human ratings (standard deviations in parentheses) of relevance and support when comparing to the reference text. }
    \vspace{-1.5em}
    \label{wikitablet-tab:human_eval_reference}
\end{table}

We report results in \cref{wikitablet-tab:human_eval_table,wikitablet-tab:human_eval_reference}. The scores are on a 1-5 scale with 5 being the best. For the first set, we collect 480 annotations from 38 annotators. For the second set, we collect 360 annotations from 28 annotators. We also ensure that each system has the same number of annotations.\footnote{We used Amazon Mechanical Turk. To ensure annotation quality, we only recruited annotators with master qualification. We collected one annotation for each instance (so that we can cover more instances) and paid 30 cents per annotation. The amount of wage per annotation is decided by (1) the amount of time each annotator spent on the task during our pilot study and (2) a target hourly wage of approximately \$11.}

\begin{table}
    \centering\small\setlength{\tabcolsep}{4pt}
\begin{tabular}{|p{0.08\textwidth}|p{0.815\textwidth}|c|c|}\hline
\multicolumn{1}{|c|}{Method} &\multicolumn{1}{c|}{Text} & G & C \\\hline

Reference & He contested the parliamentary seat of Meriden at the 1987 general election, where he was defeated by the sitting Conservative MP Iain Mills by a margin of 16,820. He was then selected to fight the Conservative-held marginal seat of Birmingham Northfield ... & 3 & 3 \\\hline

Reference & Boscawen married on 23 April 1700 in Henry VII's Chapel, Westminster Abbey, Charlotte Godfrey elder daughter and coheir of Colonel Charles Godfrey, master of the jewel office and his wife Arabella Churchill ... & 3 & 4 \\\hline

Sampling & 7th Marquess of Exeter married, firstly, Edith Csanady de Telegd (born \textbf{1 September 1935} in England; died 16 June 1956 in London), on \textbf{17 January 1934} ... & 4 & 5 \\\hline

Blocking & ... He averaged 10.9 rebounds and 3.0 assists per game \textbf{as a senior in 1987-88}. He was selected to the Sweet 16 of the NCAA Tournament \textbf{as a junior in 1988-89} ... & 5 & 5 \\\hline
\end{tabular}

    \caption{Human annotation examples for grammaticality (G) and coherence (C). Due to space constraints, only parts of the texts are shown. We highlight texts that are incoherent.}
    \label{wikitablet-tab:human_eval_examples}
\end{table}

It is interesting to note from \cref{wikitablet-tab:human_eval_table} that human annotators are unable to differentiate the human written texts from the generations from our neural models. Since the Wikipedia section texts are parts of Wikipedia articles, showing the section texts in isolation can make them difficult to understand, potentially resulting in noisy annotations. As shown by the first instance in \cref{wikitablet-tab:human_eval_examples}, the text uses the pronoun ``he'' without clarifying what the pronoun refers to. The paragraph is rated 3 for coherence, presumably due to this ambiguity. Also, Wikipedia texts are sometimes grammatically complex and annotators can mistake them for being ungrammatical, e.g., the second instance in \cref{wikitablet-tab:human_eval_examples}. 

On the other hand, the coherence errors in the generated texts are not always easy to spot. See, for example, the last two instances in \cref{wikitablet-tab:human_eval_examples}, where the incoherence lies in the facts that (1) it is impossible to marry a person before the person is born, and (2) senior year takes place after junior year. These details are embedded in long contexts, which may be overlooked by annotators and lead to results favorable to these neural models.

To study the relationship between coherence and grammaticality, we compute Spearman's correlations between the human annotations for coherence and grammaticality after removing the ones with perfect scores for coherence. \cref{wikitablet-tab:human_eval_spearman} shows the results. The correlations are much higher for references, beam search, and nucleus sampling than for $n$-gram blocking. This trend suggests that the imperfect coherence scores for the reference texts are likely because annotators find the texts to contain grammatical errors (or to possess grammatical complexity) which may prevent them from fully understanding the texts. However, $n$-gram blocking does not have this problem and thus achieves the best results for both coherence and grammaticality. We hypothesize that $n$-gram blocking is able to avoid the types of grammatical errors that prevent understanding because (1) unlike nucleus sampling, $n$-gram blocking does not rely on randomness to avoid repetition; (2) $n$-gram blocking does not suffer from repetitions like beam search.

\begin{table}
    \centering
    \small\setlength{\tabcolsep}{5pt}
\begin{tabular}{|l|c|c|c|c|}\hline
 & Ref. & Beam & Samp. & Block. \\\hline
Spearman corr. & 39.6 & 39.7 & 40.8 & 16.4 \\
\# annotations & 67 & 80 & 76 & 67  \\
\hline
\end{tabular}
    \caption{Spearman correlations between the human evaluation results for grammaticality and coherence. We omit annotations with perfect scores for coherence.}
    \label{wikitablet-tab:human_eval_spearman}
\end{table}

\begin{table}
    \centering
    \small\setlength{\tabcolsep}{5pt}
\begin{tabular}{|l|c|c|c|c|c|}\hline
 & 1 & 2 & 3 & 4 & 5 \\\hline
Relevance & 24.2 & 19.2 & 13.6  & 12.0 & 8.9 \\
\# annotations & 10 & 48 & 65 & 124 & 113 \\\hline
Support & 17.0 & 11.0 & 17.5 & 12.5  & 9.4  \\
\# annotations & 13 & 47 & 68 & 135 & 97 \\
\hline
\end{tabular}
    \caption{Averaged perplexities and the corresponding numbers of annotations for each option for the relevance and support questions (5 is the best option). We aggregate annotations for different decoding algorithms. We note that the perplexities are computed based on the reference texts using the large model.}
    \label{wikitablet-tab:human_eval_avg_ppl}
\end{table}

We report results for the second set of questions in Table \ref{wikitablet-tab:human_eval_reference}. The three evaluated systems show similar performance. To investigate the relationship between the degree of open-endedness of a \wikitablet instance and its corresponding evaluation scores, we compute the averaged perplexities (based on our large models) for each option in \cref{wikitablet-tab:human_eval_avg_ppl}. The most relevant generations are typically from more closed-ended or constrained instances.\footnote{\citet{li2015nlp} use entropy as a proxy to quantify complexity of tasks. In this work, we use perplexity to measure how open-ended the instances are.} Similarly for the support scores, more open-ended instances are distributed at score 3, which means that there is no fact supported by or contradictory to the shown tables. While the open-endedness of an instance usually depends on its topics (e.g., movie plots are open-ended), there are many cases where the models can benefit from better entity modeling, such as understanding what a particular entity type is capable of (e.g., see the last example in \cref{wikitablet-sec:generation_example}).

Recent work has also found conducting human evaluation for long-form generation to be challenging, for example in the context of 
question answering \citep{krishna-etal-2021-hurdles} and story generation \citep{akoury-etal-2020-storium}.  Our observations for data-to-text generation complement theirs and we hope that our dataset can inspire future research on human evaluation for long-form text generation.

\begin{table}
    \centering\small
    \begin{tabular}{|l|r|r|}\hline
        percentile & train perp. & dev perp. \\\hline
        10 &  2.3 & 2.5 \\
        20 &  3.1 & 3.6 \\
        30 &  4.0 & 4.7 \\
        40 &  4.9 & 6.1 \\
        50 &  6.1 & 7.6 \\
        60 &  7.6 & 9.8 \\
        70 & 9.8 & 12.9 \\
        80 & 13.3 & 18.2 \\
        90 & 20.6 & 29.3 \\
        100 & 278.8 & 129.4 \\\hline
    \end{tabular}
    \caption{Percentiles of perplexities for training and development splits of our dataset. We use the large model.}
    \label{wikitablet-tab:dataset_ppl}
\end{table}

\begin{table}
    \centering\small\setlength{\tabcolsep}{5pt}
\begin{tabular}{|p{0.08\textwidth}|p{0.87\textwidth}|}\hline
\multicolumn{1}{|c|}{Method} &\multicolumn{1}{c|}{Generated Text} \\\hline
Nucleus sampling (threshold 0.9) & In a two-issue mini-series titled "Messiah Complex", portrayed by Maisie Williams, Father Craig is ordered to rescue the world from Reavers and run out of the X-Force causing the X-Force to permanently withdraw. The original humans were replaced by slots and Muir Island being pulled away. They also possessed the entire X-Force after issue \#9 in which the X-Force appeared. Claremont and Sinclair later reprised their parts in "The Donderful World of Wolverine" with Wolfsbane as one of the new X-Men and her group taken back to their reservation for unknown reasons. \\\hline
$n$-gram blocking & In the "Messiah Complex" storyline, Wolfsbane is a member of the X-Men, a group of mutants recruited by Father Craig to fight the Reavers in Muir Island. Father Craig is killed in the battle with Reavers, but Father Craig manages to rescue the mutant from the Rain of Gesham and retrieve the mutant. She is later revealed to have been killed by Rain in a battle with Father Craig. In the aftermath, Father Craig tells Father Craig that she is the only mutant to be killed by a mutant. \\\hline\hline
Nucleus sampling (threshold 0.9) & In May 2013, McCoy allegedly assault a Spoiler business official. The incident motivated him to take a business trip to Harrisburg, Pennsylvania and sign a letter from the Buffalo Bills paying him over \$50,000 as well as verbally prohibiting legal kickoff work. In September 2014, McCoy shared the issue with some former players of the first two film franchises. McCoy played a leading role in the 2012 Marvel Cinematic Universe film "Erab of the Press", and was named one of the film's biggest stars on Twitter. On September 9, 2014, McCoy filed a lawsuit against PYT to change its Twitter name to LeSean McCoy Productions. \\\hline
$n$-gram blocking & On September 9, 2014, McCoy was arrested in Harrisburg, Pennsylvania on suspicion of assault. He was charged with assault and battery. In May 2013, he was fined over \$50,000 by the Buffalo Bills. In September 2014, he was suspended for two games by the PYT for violating the Marvel Cinematic Universe. He was released by the Bills in October of the same year. He was cleared of all charges on Twitter, and was banned from playing in the 2014 Pro Bowl due to his Twitter account. \\\hline

\end{tabular}

    \caption{Generation examples from the large model. The first example corresponds to the first instance in \cref{wikitablet-fig:dataset-example}.}
    \label{wikitablet-tab:generation_examples}
\end{table}

\paragraph{Distribution of Perplexity.}
To determine the fraction of \wikitablet that can be seen as constrained, we report the percentiles of perplexities for training and development splits in \cref{wikitablet-tab:dataset_ppl}. From \cref{wikitablet-tab:human_eval_avg_ppl}, it can be observed that instances with perplexities around 9.0 generally lead to model generations that are closely relevant to the reference texts and mostly supported by the input tables, and therefore are likely to be the constrained instances. From \cref{wikitablet-tab:dataset_ppl}, we see that at least half of our dataset has perplexities lower than 9.0, so we conjecture that half of our dataset consists of constrained instances.

\paragraph{Generation Examples.}
\label{wikitablet-sec:generation_example}

\cref{wikitablet-tab:generation_examples} shows generation examples for nucleus sampling and beam search with $n$-gram blocking. We observe very different trends between the two instances in \cref{wikitablet-fig:dataset-example}. For the first instance about the X-Men, although both generations look fluent, their stories differ dramatically. The generated text for nucleus sampling describes a story that starts by saying Father Craig rescues the world from Reavers and ends with Wolfsbane joining as one of the new X-Men. On the other hand, $n$-gram blocking generates a story where Wolfsbane already is a member of X-Men, and the story says Father Craig fought and was killed by the Reavers, but manages to rescue the mutant. For the less open-ended instances (e.g., the second instance in \cref{wikitablet-fig:dataset-example}), different decoding strategies mostly generate similar details.

Despite having different details, these generations appear to try to fit in as many entities from the tables as possible, in contrast to beam search which mostly degenerates into repetition for more open-ended instances. This explains our previous observation that $n$-gram blocking helps with the PAR-R score.

Even though the generations are of good quality for most instances, their implausibility becomes more apparent when readers have enough background knowledge to understand the involved entities. For example, the second instance in \cref{wikitablet-tab:generation_examples} comes from the Wikipedia page ``LeSean McCoy'' (a football player) under the sections ``Personal life'' and ``Controversies''. The generation from nucleus sampling is implausible/nonsensical in some places (``assault a Spoiler business official'') and factually incorrect elsewhere (McCoy did not play a leading role in any film, and ``Erab of the Press'' is not an actual film). The fourth generation is implausible because a player is unlikely to be suspended for ``violating the Marvel Cinematic Universe'', and it is unlikely for a person to be cleared of all charges on Twitter. Our models have limited access to knowledge about entities, e.g., the capabilities of a social media company like Twitter. Future research may incorporate extra resources, make use of pretrained models, or incorporate factuality modules to solve these problems.

\section{Long-Form Text Summarization}
\label{sec:summscreen}
\subsection{Introduction}
\begin{figure}
    \centering
    \includegraphics[scale=0.8]{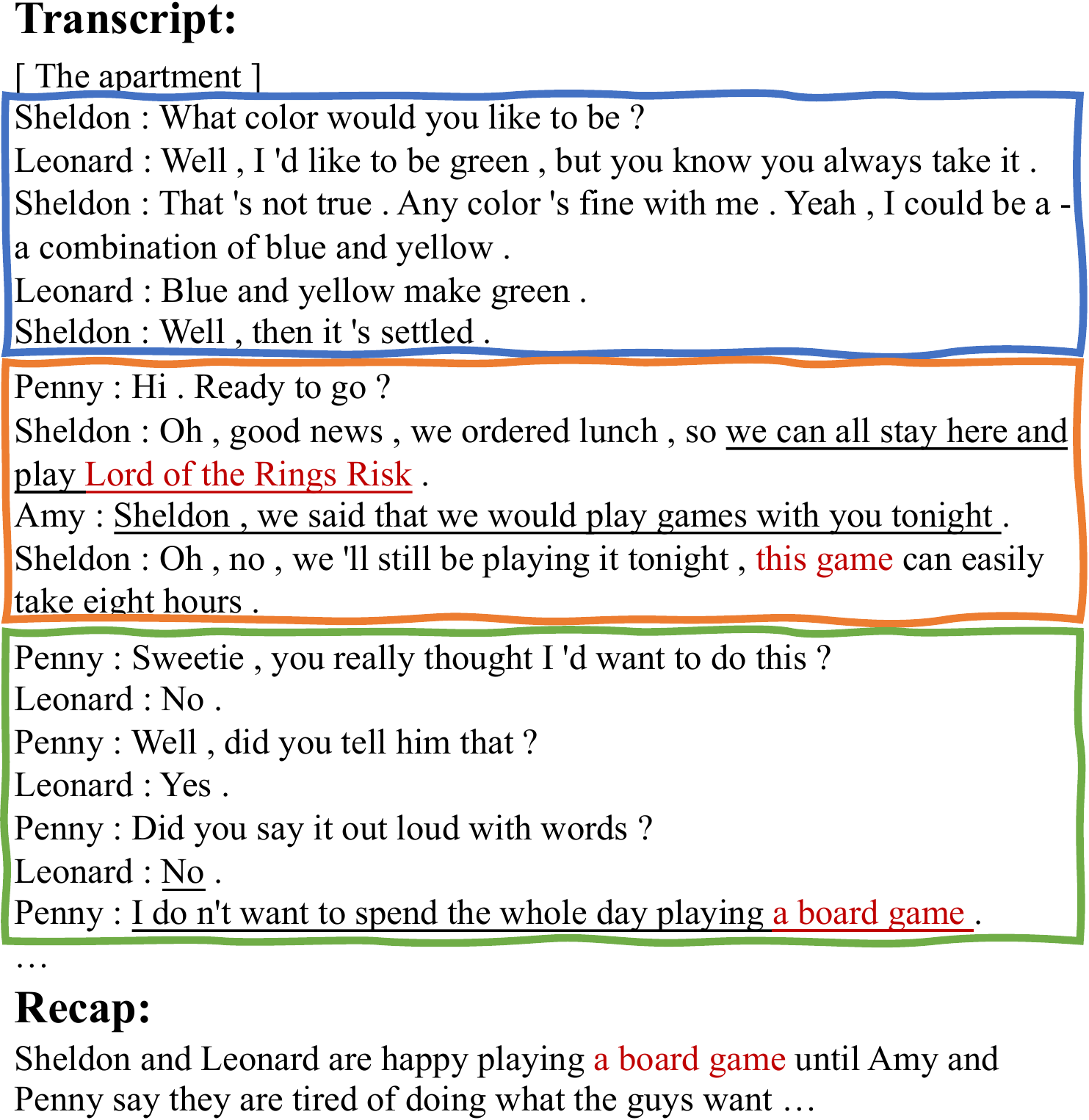}
    \caption{Excerpts from an example from \summscreen. The transcript and recap are from the TV show ``The Big Bang Theory''.
    Generating this sentence in the recap requires discerning the characters' feelings (clues in the transcript are underlined) about playing the board game (references are shown in red). Colored boxes indicate utterances belonging to the same conversations. }
    \label{summscreen-fig:intro_example}
\end{figure}

Abstractive summarization aims to produce a summary that concisely expresses key points of the input document rather than merely extracting pieces of it. Existing datasets are constructed from various domains, such as news \citep{linguistic2008evan,teaching2015hermann,rush-etal-2015-neural,narayan-etal-2018-dont,grusky-etal-2018-newsroom}, online forums \citep{volske-etal-2017-tl}, meeting dialogues \citep{icsi2003,carletta2005ami}, and webpages \citep{10.1145/3366423.3380206}. However, few datasets exist for abstractive summarization of narrative text, which focuses on entities and dialogue among entities, with plot details often communicated indirectly via dialogue. In this work, we build \summscreen, an abstractive summarization dataset combining TV series transcripts and episode recaps. \cref{summscreen-fig:intro_example} shows an example from \summscreen. 

Several aspects of \summscreen make it a challenging testbed for abstractive summarization. First, the relationship between character dialogue and plot details is not straightforward. Plot events are often expressed indirectly in dialogue, and dialogue contains other information that is not directly relevant to the plot, such as character development and humor. Also, a typical episode has multiple subplots that proceed in parallel, with consecutive scenes often describing different subplots. Solving \summscreen requires drawing information from utterances across a wide range of the input and integrating the information to form concise plot descriptions. Moreover, since actual TV episodes ground their scripts with audio-visual accompaniment, many details may be omitted from the transcript itself. This omission of details and the other challenging aspects mentioned above have inspired research into other NLP tasks on TV show transcripts, such as entity tracking \citep{chen-choi-2016-character,choi-chen-2018-semeval} and coreference resolution \citep{chen-etal-2017-robust,zhou-choi-2018-exist}. 

Another prominent characteristic of TV series transcripts is their focus on characters. To reflect this aspect, we propose two entity-centric metrics to evaluate the quality of generated plot summaries. One is based on bags of characters, which measures the overlap of the characters that appear in both the generated and reference recaps. The other metric measures character relations: the overlap of cooccurrences of character pairs in generations and recaps.

We empirically evaluate several types of methods on \summscreen. We consider nearest neighbor models, which look up similar transcripts or recaps, neural abstractive summarization models, and hybrid models, which use the nearest neighbor models as content selectors followed by abstractive summarization. Oracle extractive approaches outperform all models on all the automatic metrics.
These results suggest that the benchmarked methods are unable to fully exploit the input transcripts and that improving content selection may be a promising research direction.

Human evaluations show that our non-oracle hybrid models are competitive with their oracle counterparts in terms of generating faithful plot events. Hybrid models may be promising approaches for future research. Qualitative analysis shows that neural models tend to generate generic summaries, hybrid models can benefit from better content selection, and hybrid models sometimes generate unfaithful details. 

\subsection{Related Work}

There has been prior work on \textbf{extractive} screenplay summarization \citep{gorinski-lapata-2015-movie,papalampidi-etal-2020-screenplay}, and analyzing crime drama \citep{frermann-etal-2018-whodunnit}. The majority of TV show transcripts are dialogues, relating our work to prior work on dialogue and meeting summarization. 
Relevant datasets have been studied for medical dialogues \citep{joshi-etal-2020-dr,krishna-etal-2021-generating}, chitchat (SAMSum; \citealp{gliwa-etal-2019-samsum}), podcasts \citep{clifton-etal-2020-100000}, meetings (AMI; \citealp{carletta2005ami}; ICSI; \citealp{icsi2003}; QMSum; \citealp{zhong-etal-2021-qmsum}), livestreams (StreamHover; \citealp{cho-etal-2021-streamhover}), online forums (ForumSum; \citealp{khalman-etal-2021-forumsum-multi}) and news interviews (MediaSum; \citealp{zhu-etal-2021-mediasum}).

There have been attempts in summarizing long-form text (other than screenplays), such as books \citep{mihalcea-ceylan-2007-explorations}, scientific articles (PubMed and arXiv; \citealp{cohan-etal-2018-discourse}), multiple news articles (Multi-News; \citealp{fabbri-etal-2019-multi}),  opinionated text (RottenTomatoes; \citealp{wang-ling-2016-neural}), patents \citep{sharma-etal-2019-bigpatent}, TV show stories (TVRecap; \citealp{chen2021tvrecap}) and (extractive summarization of) chapters of novels \citep{ladhak-etal-2020-exploring}. More detailed discussion on the differences between these datasets and \summscreen is in the next section.

Recently there have been efforts on adapting resources for TV shows for different tasks, including question answering \citep{ma-etal-2018-challenging,yang-choi-2019-friendsqa}, speaker identification \citep{ma-etal-2017-text}, sarcasm detection \citep{joshi-etal-2016-harnessing}, emotion detection \citep{zahiri2017emotion,hsu-ku-2018-socialnlp}, character relation extraction \citep{yu-etal-2020-dialogue}, and story generation \citep{chen2021tvrecap}.

\subsection{\summscreen}

An instance in \summscreen contains a transcript from TV series and its corresponding recap. The transcripts consist of dialogue utterances with speaker names, and descriptions of scenes or character actions. The recaps are human-written summaries of the corresponding transcripts. \cref{summscreen-fig:intro_example} shows an example in \summscreen from the TV show ``The Big Bang Theory''. The transcript documents a dialogue involving four characters (Sheldon, Leonard, Penny, and Amy) about playing a board game, and the recap summarizes the dialogue into sentences. 

\paragraph{Dataset Construction.}
\label{summscreen-sec:dataset_construct}

\begin{table}[t]
    \centering\small
    \setlength{\tabcolsep}{5pt}
    \begin{tabular}{|l|r|r|r|r|r|r|}\hline
         & uni. & bi. & tri. & four. & src. & tgt. \\\hline
        \multicolumn{7}{|c|}{\summscreen}\\\hline
        \foreverdreamshort & 81.6 & 29.9 & 5.6 & 1.3 & 7.6k & 113.7 \\
        \tvmegasiteshort & 86.5  & 34.1 & 6.9 & 2.1 & 6.4k & 380.6 \\\hline
        \multicolumn{7}{|c|}{Other summarization datasets}\\\hline
        XSum\textsuperscript{\dag} & 64.2 & 16.6 & 4.5 & 1.5 & 431.1 & 23.3 \\
        CNNDM\textsuperscript{\S} & 80.5 & 43.1 & 25.6 & 17.2 & 810.6 & 56.2\\
        MNews\textsuperscript{\S} & 82.2 & 42.9 & 24.3 & 17.7 & 2.1k & 264.7 \\\hline
    \end{tabular}
    \caption{Fraction (\%) of n-grams in the \textbf{output summaries} that also appear in the inputs, and the average numbers of tokens for the inputs and outputs. Datasets with smaller fractions of overlapping n-grams tend to favor abstractive summarization approaches. Results marked by \dag\xspace and \S\xspace are from \citet{narayan-etal-2018-dont} and \citet{fabbri-etal-2019-multi} respectively.}
    \label{summscreen-tab:overlap_ratio_summarization_dataset_compare}
\end{table}

We use two sources to construct \summscreen: 
The TV MegaSite, Inc. %
(\tvmegasiteshort)\footnote{\url{http://tvmegasite.net/}} 
and \foreverdream (\foreverdreamshort),\footnote{\url{transcripts.foreverdreaming.org}} both 
of which provide
community-contributed transcripts. As \foreverdreamshort does not provide recaps, we obtain recaps 
of \foreverdreamshort shows
from Wikipedia and TVMaze.\footnote{\url{www.tvmaze.com}, an online TV database curated by TV fans. } 
To ensure dataset quality of \summscreen, we filter out instances based on two criteria. First, the overlap ratio of TV show characters appearing in the recap and its transcript should be higher than 85\%. We use this criterion to ensure that the alignments between recaps and transcripts are correct. 
Second,
the number of transcript lines that have speaker information (``character utterances'') should be larger than 100. We use this criterion to eliminate transcripts that are essentially subtitles, i.e., utterances without speaker information. In practice, for each transcript line, if a colon symbol appears in the first 8 tokens and there exists at least one character name in front of the colon symbol, we will count it as a character utterance. We note that \foreverdreamshort and \tvmegasiteshort do not have overlapping TV series.

In \cref{summscreen-tab:overlap_ratio_summarization_dataset_compare}, we compute n-gram overlap ratios between recaps and transcripts for measuring the abstractiveness of \summscreen. From the results, We find that despite \summscreen has longer summaries, its fraction of overlapping four-gram is comparable to XSum \citep{narayan-etal-2018-dont} which is known for abstractiveness, suggesting that \summscreen favors abstractive approaches.

\begin{table}
    \centering\small
\begin{tabular}{|l|r|r|}\hline
& \foreverdreamshort & \tvmegasiteshort \\\hline
number of shows & 88 & 10 \\
number of episodes & 4348 & 22503 \\
min. \# episodes per show & 1 & 168 \\
max. \# episodes per show & 568 & 3784 \\
median \# episodes per show & 9.0 & 1973.5 \\
avg. \# episodes per show & 49.4 & 2250.0 \\
\hline
avg. \# tokens in recaps & 113.7 & 380.6 \\
avg. \# tokens in transcripts & 7605.4 & 6420.7 \\
avg. \# lines in transcripts & 447.6 & 360.8 \\
avg. \# char. utterances in transcripts & 330.7 & 327.0 \\
avg. \# uniq. char. in recaps & 5.8 & 14.3 \\
avg. \# uniq. char. in transcripts & 20.6 & 29.8 \\\hline
\end{tabular}
    \caption{Detailed dataset statistics for \summscreen.}
    \label{summscreen-tab:dataset_stats}
\end{table}

\begin{figure}
    \centering
    \begin{subfigure}[t]{0.4\textwidth}
    \centering
    \small
    \begin{tabular}{|l|r|}%
    \hline
      Genre   & Count \\\hline
    Drama & 65\\
    Romance & 24\\
      Comedy & 23\\
     Crime & 18\\
      Action & 15 \\
    Science-Fiction & 12\\
      Adventure & 9 \\
    Supernatural & 9\\
    Mystery & 8\\
    Thriller & 5\\
    Family & 5\\
    Medical & 5\\
    Fantasy & 4\\
    Horror & 4\\
    History & 3\\
    Sports & 3\\
    Western & 3\\
      Children & 2\\
    Legal & 2\\
    Espionage & 1\\
    Music & 1\\
    \hline
    \end{tabular}
    \end{subfigure}%
    \begin{subfigure}[t]{0.4\textwidth}
    \centering
    \small
    \begin{tabular}{|l|r|}\hline
       Genre  & Count \\\hline
        Drama & 10 \\
       Romance & 6 \\
       Family & 4 \\
       Medical & 1 \\
        \hline
    \end{tabular}
    \end{subfigure}
    \caption{Left: TV show genres from \foreverdream. Right: TV show genres from \tvmegasite.
    }
    \label{summscreen-fig:dataset_genres}
\end{figure}

\cref{summscreen-tab:dataset_stats} shows data statistics and \cref{summscreen-fig:dataset_genres} shows the genres of the TV shows from the two sources.\footnote{The genre information is from TVMaze where a TV show may correspond to multiple genres.}
When computing the number of unique characters in TV shows,
we first collect the character names from TVMaze and the named entities\footnote{We use the named entity recognizer from spaCy.} preceding the colon symbols in transcripts. We then 
perform string matching to obtain numbers of TV show characters in recaps and transcripts. From these two tables, we observe
that \foreverdreamshort and \tvmegasiteshort are different in many aspects. First, \foreverdreamshort covers more diverse genres than \tvmegasiteshort. This is partly due to the fact that TV shows 
from \tvmegasiteshort are soap operas. 
Second, transcripts from \foreverdreamshort are longer, which is caused by the fact that the transcripts from \foreverdreamshort tend to have more descriptions about environments or character actions, whereas the ones from \tvmegasiteshort are mostly made up of dialogue (see \cref{summscreen-tab:dataset_stats}). Third, recaps from \foreverdreamshort are shorter whereas recaps from \tvmegasiteshort seek to cover more details. Fourth, writing styles are more diverse in \foreverdreamshort than those in \tvmegasiteshort. In light of these differences, we treat \foreverdreamshort and \tvmegasiteshort as different datasets in the following experiments.

\begin{table}
    \centering\small
\begin{tabular}{|l|ccc|}
\hline
\foreverdream & Train & Dev & Test \\ \hline
\# shows & 66 & 78 & 81 \\ %
\# episodes & 3673 & 338 & 337 \\
\hline
\tvmegasite & Train & Dev & Test \\ \hline
\# shows & 10 & 10 & 10 \\ %
\# episodes & 18915 & 1795 & 1793 \\ %
\hline
\end{tabular}
    \caption{Statistics of train/dev/test splits for \foreverdream and \tvmegasite. }
    \label{summscreen-tab:split_stat}
\end{table}

We create train/dev/test splits for \foreverdreamshort and \tvmegasiteshort by ensuring the ratio to be roughly 10:1:1, and filter out instances in the dev/test splits if the reference texts are shorter than 30 word tokens. The statistics of the splits are shown in \cref{summscreen-tab:split_stat}.

\begin{table*}
    \centering\footnotesize
    \setlength{\tabcolsep}{3pt}
    \begin{tabular}{|l|r|r|r|r|c|}\hline
         & \# instances & \# tokens (input) & \# tokens (summary) & \# speakers & Domain \\\hline
        Multi-News & 56.2k & 2103.5 & 264.7 & - & News \\
        RottenTomatoes & 3.7k & 2124.7 & 22.2 & - & Reviews \\
        arXiv  & 215k & 4938.0 & 220.0 & - & Science \\
        PubMed & 113k & 3016.0 & 203.0 & - & Science \\
        GovReport & 19.5k & 9409.4 & 553.4 & - & Government Reports \\
        TVRecap & 29.0k & 1868.7 & 221.6 & - & Television Series \\
        \hline
        SAMSum & 16.4k & 83.9 & 20.3 & 2.2 & Chitchat \\
        ForumSum & 4.1k & 303.5 & 36.0 & 6.7 & Forum Messages \\
        MediaSum & 463.6k & 1553.7 & 14.4 & 6.5 & News Interviews \\
        AMI  & 137 & 4757.0 & 322.0 & 4.0 &  Meetings \\
        ICSI  & 59  & 10189.0 & 534.0 & 6.2 &  Meetings \\
        QMSum & 1.8k & 9069.8 & 69.6 & 9.2 & Meetings \\
        \summscreen & 26.9k & 6612.5 & 337.4 & 28.3 & Television Series         \\\hline
    \end{tabular}
    \caption{Statistics for datasets focusing on abstractive summarization for long-form text or dialogue. The numbers are averaged over instances. We omit number of speakers for datasets that do not contain dialogue. \summscreen combines long source inputs, large numbers of speakers, and a moderate number of instances.
    }
    \label{summscreen-tab:compare_dataset}
\end{table*}

\paragraph{Dataset Comparison.}

We compare \summscreen to other abstractive dialogue summarization datasets in \cref{summscreen-tab:compare_dataset}. \summscreen differs from other datasets in several ways:
\begin{enumeratesquish}
\item %
Compared to recently proposed large-scale dialogue summarization datasets (i.e., SAMsum and MediaSUM), \summscreen has longer source inputs. 
\item %
Compared to other dialogue summarization datasets, \summscreen has larger numbers of speakers per instance. The TV series genre focuses on narrative, which is typically entity-centric and can include multiple parallel subplots in a single episode. 
\item %
Compared to AMI, ICSI and QMSum, which are long-input meeting summarization datasets, \summscreen has far more instances. 
\item
Unlike most of the other datasets, \summscreen contains many episodes of a single show (e.g., more than 3k episodes for \tvmegasiteshort). This episodic structure could be used to model character arcs, the evolution of character personality traits and character relationships over episodes, among others.
\end{enumeratesquish}
Properties (1) and (2) above make extracting information from transcripts more challenging than other datasets. The third property means that \summscreen is large enough to train and evaluate neural methods. 

The Spotify Podcast Dataset \citep{clifton-etal-2020-100000} and StreamHover  \citep{cho-etal-2021-streamhover} are similar to \summscreen in that they contain transcribed speech and summaries. However, the transcriptions are obtained automatically and therefore contain errors.\footnote{For this reason, we do not include their statistics in \cref{summscreen-tab:compare_dataset}.} The datasets therefore involve speech processing (or at least handling speech recognition errors) compared to \summscreen, which has human-written transcripts. 

Since MediaSum is constructed from news transcripts, it is the most similar dataset in \cref{summscreen-tab:compare_dataset} to \summscreen. However, the summaries in MediaSum are twenty times shorter than those in \summscreen, and the average number of speakers per instance is only a quarter of that in \summscreen. Furthermore, our results in \cref{summscreen-sec:main_results} indicate that our dataset is much harder than MediaSum as the pretrained models perform worse on our dataset than on MediaSum according to automatic metrics.

\paragraph{Dataset Challenges.}
We qualitatively analyze the challenging aspects of \summscreen. Since the transcripts focus on dialogue among characters, along with limited descriptions of scenes and actions, it leads to the challenge that plot information is not stated explicitly but rather only implied in the dialogue. For example, the transcript in \cref{summscreen-fig:intro_example} does not explicitly describe what Sheldon and Leonard are playing. However, it is implied by Sheldon when he mentions playing ``Lord of the Rings Risk,'' and later by Penny when she says that she does not ``want to spend the whole day playing a board game.''

\begin{figure*}
    \centering
    \begin{subfigure}{0.7\textwidth}
    \centering
    \includegraphics[scale=0.7]{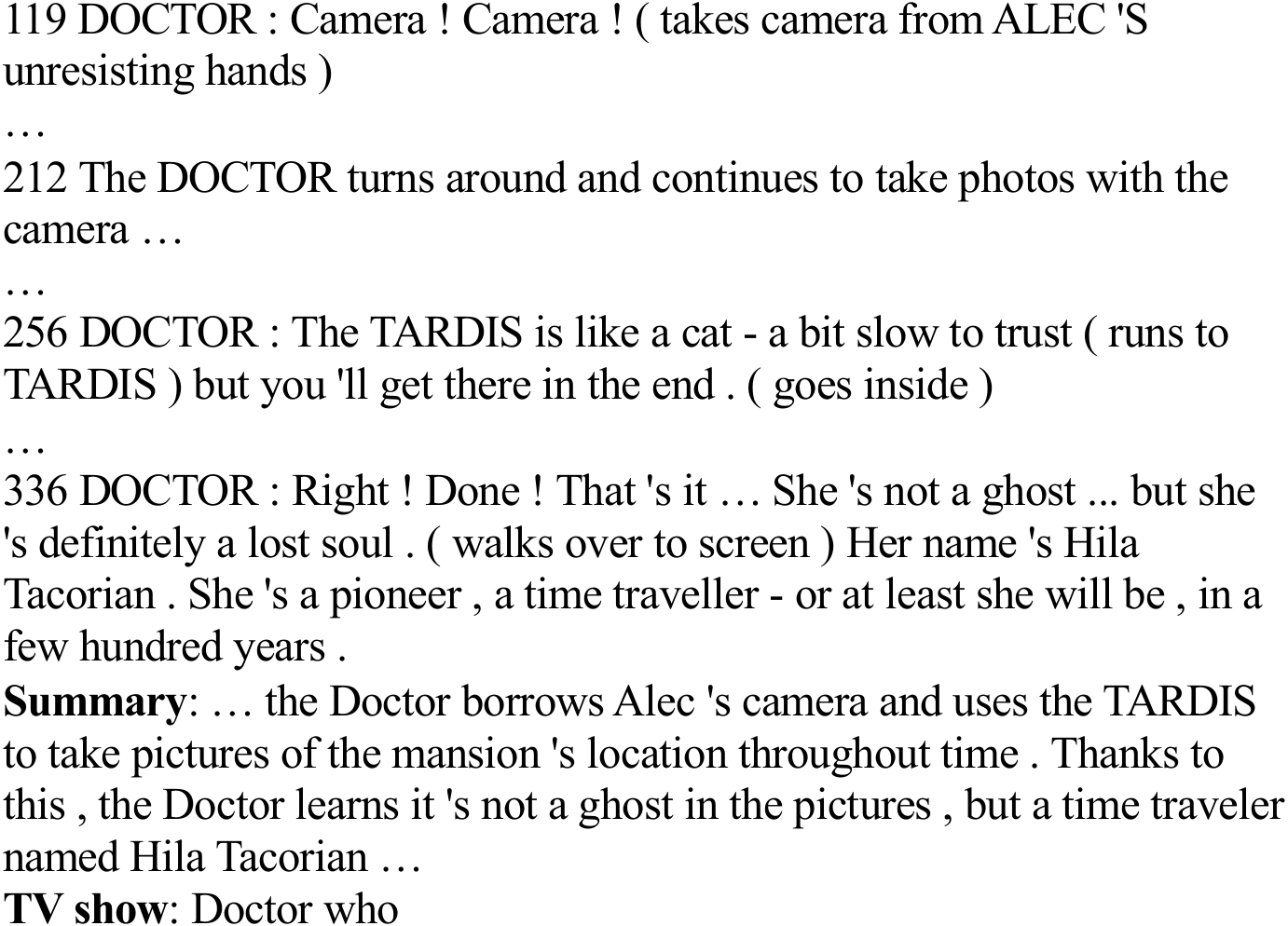}
    \subcaption{An excerpt from the TV show ``Doctor Who''.}
    \end{subfigure}\vspace{1em}
    \begin{subfigure}{0.7\textwidth}
    \centering
    \includegraphics[scale=0.7]{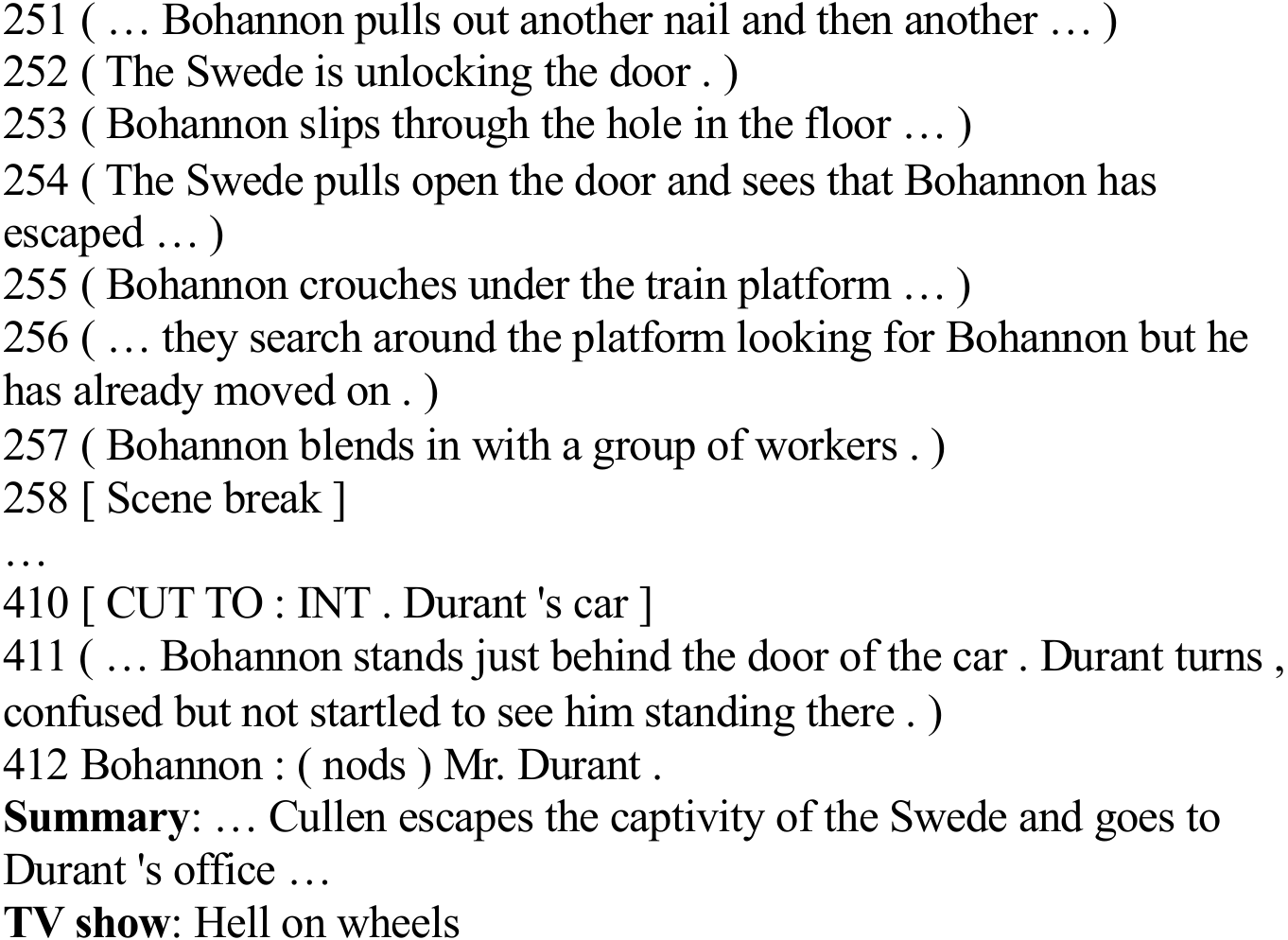}
    \subcaption{An excerpt from the TV show ``Hell on wheels''.}
    \end{subfigure}
    \caption{Two excerpts from \summscreen showing that generating summaries from TV show transcripts requires drawing information from a wide range of the input transcripts. We only show lines in the transcripts that are closely related to the shown parts of summaries. The number at the beginning of each line is the line number in the original transcript. For the first instance, we omit a few lines containing clues about the doctor taking pictures of the mansion at different times due to space constraints.}
    \label{summscreen-fig:long_range_example}
\end{figure*}

A related challenge is the need to understand the context in which characters' utterances are situated. In the example, the recap describes four characters taking sides regarding playing a board game. The transcript expresses the characters' sentiments through their interactions with one another. The conflict does not occur until Sheldon proposes to ``stay here and play Lord of the Rings Risk'', and it becomes more apparent when Penny mentions she does not want to play the board game. Given the context, Leonard's series of yes and no responses to Penny's questions is largely due to the awkward situation, and it actually shows that he is happy playing the game as he and Sheldon are doing so at the beginning of the scene. Similarly, Amy mentions their previous agreement with Sheldon as a way of politely declining Sheldon's plan. The sentiments of characters are not necessarily easily discernible from their utterances but rather must be inferred using context and knowledge about the characters. 

Another challenge in \summscreen is the need to draw information from a wide range of the input transcripts, which arises for two primary reasons. First, there are many utterances that serve a purpose other than driving the plot forward. They may help to develop characters or character relationships, or to add humor or suspense. These lines enrich the narrative but their information content is  often omitted from the summaries. For example, in the first instance in \cref{summscreen-fig:long_range_example}, we show key lines from the transcript that pertain to the excerpt of the summary. There are many other lines between the lines shown, which are conversations between the doctor and other characters. 
This property necessitates the models' ability to correctly attend to major events across the transcript when generating summaries. 
The pattern can also be observed in \cref{summscreen-tab:dataset_stats} through the differences between the number of unique characters in recaps and  transcripts. More than half of the characters in the transcripts are not contained in the recaps. 

The second reason why information needs to be combined across wide ranges of the input relates to scene breaks and multiple plots. As a TV show often narrates a few plots in parallel, scene breaks are used to separate the stories. The discontinuity sometimes requires models to connect subplots hundreds of lines apart. For example, for the second instance in \cref{summscreen-fig:long_range_example}, the show uses scene breaks to express what is happening when Cullen Bohannon escapes from the Swede, which is why there are almost two hundred lines between Cullen Bohannon's escape and his appearance at Durant's office.

\subsection{Method}

In this section, we describe modeling approaches that we benchmark on \summscreen. We note that since the meaning of sentences in transcripts is highly context-dependent, extractive summarization approaches are not expected to be useful for this dataset. We report the results from nearest neighbor-based extractive summarizers mostly for characterizing the dataset.

\paragraph{Neural Models.}

We use transformer based sequence-to-sequence architectures \citep{attention_is_all_you_need}. 
Because transcripts are quite long, we limit the number of encoder hidden vectors that are used in the decoder's attention mechanism. To do so, when encoding transcripts, we first append a special token \texttt{[EOS]} to each line of the transcript, and then linearize the transcript. We then only feed the vectors representing these special tokens to the decoder. We use the Longformer \citep{beltagy2020longformer} as our encoder architecture, and set the \texttt{[EOS]} tokens to use global attention.
For our decoders, we use the standard transformer architecture.

\paragraph{Nearest Neighbor Models (NNMs).}

We consider two metrics when finding nearest neighbors: BM25 (\citealp{robertson1995okapi}; a popular metric for information retrieval),
and ROUGE scores. We use ROUGE scores as they are used for evaluation, and we use BM25 because it is designed for retrieving long documents whereas ROUGE scores are not. When using ROUGE scores, we use the average of ROUGE-1, ROUGE-2, and ROUGE-L. We consider three types of nearest neighbor search: transcript-to-transcript, recap-to-transcript, and recap-to-recap.

\paragraph{Recap-to-transcript (NNM-r2t).} We use each sentence in the recap as queries and the lines in the corresponding transcript as candidates. The generation is formed by the nearest neighbors for each sentence. We use BM25 or ROUGE scores as the metric. This method can serve as an oracle result for an extractive summarization system, showing roughly how much information can be extracted at the utterance level from the source transcript.

\paragraph{Transcript-to-transcript (NNM-t2t).} We use the transcripts in the test sets as queries, the transcripts in the training sets as candidates, and then find the nearest neighbors using BM25. The generations are the corresponding recaps. his baseline measures the similarity of instances between training and test splits.

\paragraph{Recap-to-recap (NNM-r2r).} This setting is similar to the ``transcript-to-transcript'' setting, but we use recaps for both queries and candidates, and we use ROUGE and our proposed entity-centric scores (see \cref{summscreen-sec:entity_metrics} for more details) as the metric. When using the entity metrics, we use the average of the 4 metric scores. This is an oracle baseline of the ``transcript-to-transcript'' setting and also measures the similarity of the splits.

\paragraph{Hybrid Models.}

As content selection has been shown to be helpful in prior work \citep{gehrmann-etal-2018-bottom,j.2018generating}, we use the ``recap-to-transcript'' nearest neighbor model and BM25 as the metric to select the most salient content from transcripts, and then apply neural models to the selected content when performing generation. As these methods combine nearest neighbor models and neural models, we refer to them as hybrid models.

In particular, for each sentence in the recap, we find the top three most similar lines in the transcript, include two extra lines that come before or after the selected lines as context, and also include a line that is retrieved by using the whole recap.
As the selected content is significantly shorter than the original transcript, it allows us to use pretrained models.\footnote{After the selection steps, the average number of tokens of the transcripts for \foreverdreamshort and \tvmegasiteshort reduces to 1138.9 and 3252.7 respectively.} Therefore, in this setting, we fine-tune a pretrained BART-large model \citep{lewis-etal-2020-bart}. We note that as the nearest neighbor models rely on the gold standard recaps, this hybrid model demonstrates an approximate upper bound on performance when using powerful content selectors.\footnote{We use the maximum sequence length of 1024 (i.e., we truncate the input sequences if they are longer than 1024) for BART-large due to computational constraints. }

To establish a non-oracle baseline, we train neural models to predict the selected lines, and then fine-tune BART-large models on the predicted lines. Details of the architecture for this component, which we call our ``neural content selector'', are as follows. We use a 3-layer longformer encoder followed by a 2-layer feedforward network with GELU activations \cite{hendrycks2016gelus}. We perform early stopping based on F1 scores on the development sets, where the threshold is chosen by averaging over the oracle thresholds for each instance. When selecting content, we use the threshold chosen based on the development set and ensure that no less than 10\% of lines for each transcript are selected. The model achieves test performance (F1 scores) of 19.0 on \foreverdreamshort, 19.2 on anonymized \foreverdreamshort, 41.5 on \tvmegasiteshort, and 40.1 on anonymized \tvmegasiteshort.

\subsection{Experiments}

\begin{table*}
    \centering
    \footnotesize\setlength{\tabcolsep}{4pt}
\begin{tabular}{|l|ccccc|ccccc|}
\hline%
& \multicolumn{5}{c|}{ Generic Metrics} & \multicolumn{5}{c|}{ Entity Metrics} \\
& BLEU & R1 & R2 & RL & avg. & BoC-p & BoC-r & BoR-p & BoR-r & avg. \\
\hline
\multicolumn{11}{c}{ \foreverdream } \\\hline
NNM-r2t (oracle, BM25) & 3.4 & 34.3 & 6.6 & 29.6 & 18.5 & 70.5 & 61.9 & 36.4 & 16.1 & 46.2 \\
NNM-r2t (oracle, RG) & 3.9 & 34.8 & 8.5 & 31.5 & 19.7 & \bf 76.7 & 63.3 & \bf 46.5 & 21.3 & 52.0 \\
NNM-r2r (oracle, RG) & \bf 9.9 & \bf 38.8 & \bf 11.5 & \bf 33.9 & \bf 23.5 & 50.6 & 51.4 & 24.6 & 26.8 & 38.4 \\
NNM-r2r (oracle, Entity Metrics) & 5.5 & 31.1 & 6.8 & 27.1 & 17.6 & 58.6 & \bf 79.6 & 26.4 & \bf 43.7 & \bf 52.1 \\
NNM-t2t & 7.9 & 31.3 & 7.8 & 27.4 & 18.6 & 56.5 & 59.2 & 28.2 & 29.4 & 43.3 \\\hline
Neural model & 2.6 & 25.9 & 4.2 & 23.8 & 14.1 & 54.7 & 38.5 & 22.8 & 15.1 & 32.8 \\
Hybrid model & 2.4 & 25.3 & 3.9 & 23.1 & 13.7 & 61.2 & 51.4 & 29.8 & 23.6 & 41.5 \\
Hybrid model (oracle) & 3.0 & 26.4 & 5.0 & 23.3 & 14.4 & 70.0 & 57.8 & 36.9 & 29.1 & 48.5 \\
\hline
\multicolumn{11}{c}{ \tvmegasite } \\\hline
NNM-r2t (oracle, BM25) & 6.7 & 45.0 & 10.2 & 43.0 & 26.2 & 82.5 & \bf 80.4 & 57.7 & 18.1 & 59.7 \\
NNM-r2t (oracle, RG) & \bf 8.5 & 44.1 & 11.7 & 42.4 & 26.7 & 85.2 & 76.8 & \bf 61.2 & 16.9 & 60.0 \\
NNM-r2r (oracle, RG) & 7.9 & \bf 49.0 & 11.6 & \bf 46.9 & \bf 28.9 & 59.2 & 59.0 & 29.5 & 29.9 & 44.4 \\
NNM-r2r (oracle, Entity Metrics) & 4.9 & 42.8 & 8.8 & 40.4 & 24.2 & 60.8 & 81.7 & 26.0 & \bf 37.5 & 51.5 \\
NNM-t2t & 6.2 & 43.2 & 8.6 & 41.4 & 24.9 & 63.2 & 69.3 & 31.8 & 35.3 & 49.9 \\\hline
Neural model & 7.9 & 42.9 & 11.9 & 41.6 & 26.1 & \bf 86.1 & 48.7 & 48.9 & 22.3 & 51.5 \\
Hybrid model & 5.5 & 38.8 & 10.2 & 36.9 & 22.8 & 84.5 & 57.2 & 51.0 & 29.3 & 55.5 \\
Hybrid model (oracle) & 8.9 & 42.1 & \bf 11.9 & 40.9 & 25.9 & 84.0 & 69.5 & 56.4 & 36.8 & \bf 61.7 \\
\hline
\end{tabular}
    \caption{
    Results on the \summscreen test sets. 
    BLEU, R1, R2, and RL are BLEU and ROUGE scores between model generated and reference recaps. 
    Bo\{C,R\}-\{p,r\} are precision and recall for bag of characters and bag of character relations, respectively. The highest numbers for each dataset in each column are in bold.
    }
    \label{summscreen-tab:result}
\end{table*}

\paragraph{Experimental Setup.}
\label{summscreen-sec:setup}

\label{summscreen-sec:entity_metrics}
We report BLEU, ROUGE-1 (R1), ROUGE-2 (R2), and ROUGE-L (RL). We report the average of these four metrics as it generally shows the semantic similarities between generations and references. We will refer to these metrics as generic metrics as they treat each word equally. 

As characters are fundamental to TV show plots, we believe the quality of plot summaries also depends on including the right characters. To take this factor into account, we compute several bag of character (BoC) metrics based on the fraction of the overlapping characters between generated and gold standard recaps. Formally, we define the BoC precision to be
\begin{equation}
    \frac{\vert f(\text{generation})\&f(r)\vert}{\vert f(\text{generation})\vert}\nonumber
\end{equation}
\noindent where $f$ is a function that extracts the bag of characters from some text, where we perform string matching based on the character names that are automatically extracted during dataset construction (see \cref{summscreen-sec:dataset_construct}), $\&$ computes the intersection of two bags, $\vert\cdot\vert$ returns the size of its inputs, and $r$ is the gold standard recap. Similarly, we define the BoC recall to be
\begin{equation}
    \frac{\vert f(\text{generation})\&f(r)\vert}{\vert f(r)\vert}\nonumber
\end{equation}
Since BoC does not consider relations between characters, we also report bag of character relations (BoR) metrics based on the cooccurrence of character pairs. We assume two characters are related when they appear in the same sentence. After obtaining the character relations from the gold standard recaps and the generations, we compute recall and precision against the recaps following the same approach as BoC. We note that the extracted relations are non-directional, and BoR does not consider frequency of the cooccurrences. We also report the averages of both precisions and recalls from both the BoC and BoR metrics. Code and the dataset are available at \url{https://github.com/mingdachen/SummScreen}.

\paragraph{Hyperparameters.} We set the maximum sequence length to be 14336 for the encoder and 1024 for the decoder. We use byte-pair encoding \cite{sennrich-etal-2016-neural} with approximately 10k vocabulary size. 
We use a 1-layer encoder and a 12-layer decoder with 1024 hidden units unless otherwise specified. We use an effective batch size of 200, and train the models for 50 epochs. During training, we perform early stopping on the development sets based on perplexities. During testing, we use beam search with trigram blocking \cite{paulus2018a} and a beam size of 5.

\paragraph{Experimental Results.}
\label{summscreen-sec:main_results}

We report test results for \foreverdreamshort and \tvmegasiteshort in \cref{summscreen-tab:result}.
Our findings for the nearest neighbor models are as follows:

\begin{enumeratesquish}
\item
We find that the nearest neighbor models give strong performance on our dataset. In particular, NNM-r2t shows the best performance in general. 
This demonstrates that there is still room for improving the ability of our neural models to extract the most useful information from transcripts, suggesting that improved transcript modeling may be a fruitful research direction for these datasets.
\item
We observe that NNM-r2r exhibits different strengths when based on different metrics, for example, using ROUGE scores will lead to results favorable to generic metrics.
\end{enumeratesquish}

As for the results involving neural models, our findings are as follows:

\begin{enumeratesquish}
\item
The neural model shows strong performance in generic semantic matching but it is relatively weak in entity metrics compared to the non-oracle baselines. 
\item
The hybrid model is better than the neural model in terms of generating character mentions and relations. With the help of the oracle content selector, the hybrid model improves significantly in both semantic matching and entity-related metrics, showing that future research may find improvement by designing better content selectors.
\end{enumeratesquish}

\subsection{Analysis}

\paragraph{Anonymized \summscreen.}

As plots for TV shows are typically about a limited number of characters, models trained on \summscreen may focus on those characters and their typical behaviors rather than the actual 
actions taking place in the input transcripts. To eliminate this effect, we create an anonymized version of \summscreen by replacing character names with random character IDs. We ensure that the IDs of particular characters in different episodes are randomly assigned (i.e., IDs are not consistent across episodes). 

\begin{figure}
    \centering
    \includegraphics[scale=0.7]{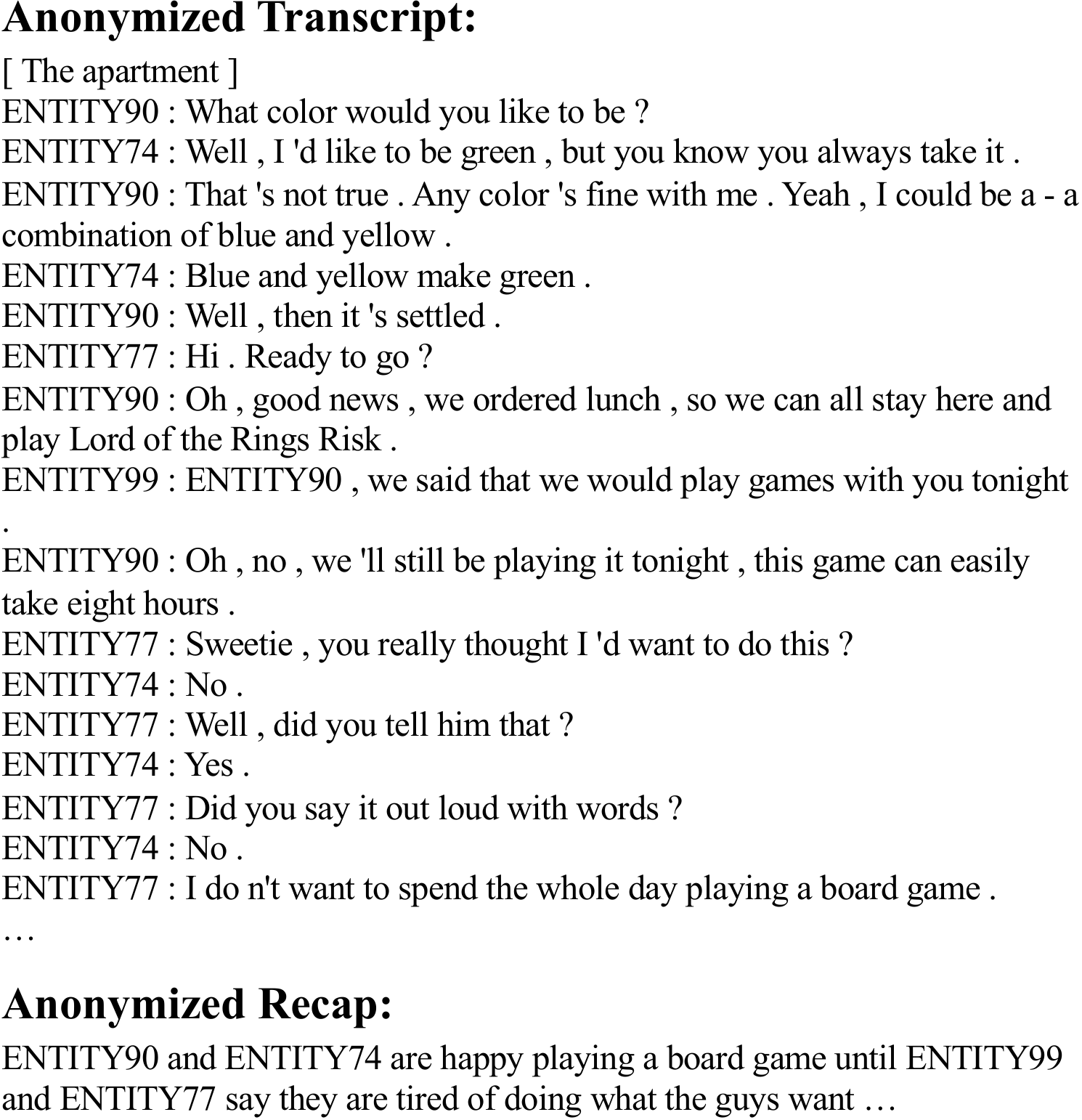}
    \caption{An excerpt from anonymized \summscreen that corresponds to the instance in \cref{summscreen-fig:intro_example}. Character names are replaced with IDs that are permuted across episodes.}
    \label{summscreen-fig:example}
\end{figure}

\cref{summscreen-fig:example} shows an example from anonymized \summscreen. Anonymized question answering datasets have also been created out of  similar concerns to those just described \citep{teaching2015hermann}.

\paragraph{Results for Anonymized \summscreen.}

\begin{table*}
    \centering
    \small\setlength{\tabcolsep}{4pt}
\begin{tabular}{|l|ccccc|ccccc|}
\hline%
& \multicolumn{5}{c|}{ Generic Metrics} & \multicolumn{5}{c|}{ Entity Metrics} \\
& BLEU & R1 & R2 & RL & avg. & BoC-p & BoC-r & BoR-p & BoR-r & avg. \\
\hline
\multicolumn{11}{c}{ Anonymized \foreverdream } \\\hline
NNM-r2t (oracle, BM25) & 3.5 & 34.5 & 6.8 & 30.0 & 18.7 & 70.4 & 60.4 & 37.5 & 16.7 & 46.2 \\
NNM-r2t (oracle, RG) & 4.0 & \bf 34.7 & 8.5 & \bf 31.4 & 19.7 & \bf 76.8 & \bf 63.4 & \bf 49.1 & 22.6 & \bf 53.0 \\
NNM-r2r (oracle, RG) & \bf 7.9 & 34.3 & \bf 9.1 & 30.1 & \bf 20.4 & 5.4 & 6.3 & 0.2 & 0.1 & 3.0 \\
NNM-t2t & 6.0 & 26.2 & 6.0 & 23.0 & 15.3 & 21.5 & 6.6 & 5.0 & 0.2 & 8.3 \\\hline
Neural model & 2.6 & 28.6 & 4.6 & 25.1 & 15.2 & 65.0 & 57.7 & 27.9 & \bf 30.6 & 45.3 \\
Hybrid model & 2.3 & 23.1 & 3.9 & 20.6 & 12.5 & 12.2 & 2.3 & 0.3 & 0.0 & 3.7 \\
Hybrid model (oracle) & 2.9 & 26.0 & 5.0 & 22.2 & 14.0 & 33.9 & 8.8 & 3.6 & 0.6 & 11.7 \\
\hline
\multicolumn{11}{c}{ Anonymized \tvmegasite } \\\hline
NNM-r2t (oracle, BM25) & 6.9 & \bf 45.0 & 10.2 & \bf 42.9 & 26.2 & 82.6 &\bf 80.5 & 58.9 & 20.7 & 60.7 \\
NNM-r2t (oracle, RG) & \bf 8.7 & 44.1 &  \bf 11.7 & 42.3 & \bf 26.7 & 85.3 & 76.7 & \bf 61.8 & 19.3 & 60.8 \\
NNM-r2r (oracle, RG) & 6.0 & 42.8 & 9.3 & 41.1 & 24.8 & 46.3 & 14.7 & 3.8 & 0.6 & 16.3 \\
NNM-t2t & 4.4 & 26.2 & 6.0 & 23.0 & 14.9 & 47.7 & 15.2 & 3.8 & 0.5 & 16.8 \\\hline
Neural model & 7.1 & 41.6 & 11.6 & 40.4 & 25.2 & \bf 86.8 & 53.6 & 32.0 & 15.2 & 46.9 \\
Hybrid model & 6.2 & 37.7 & 9.3 & 36.4 & 22.4 & 82.5 & 62.3 & 47.4 & 30.2 & 55.6 \\
Hybrid model (oracle) & 6.1 & 38.9 & 10.1 & 37.6 & 23.2 & 84.3 & 68.1 & 55.6 & \bf 38.8 & \bf 61.7 \\
\hline
\end{tabular}
    \caption{
    Results on the anonymized \summscreen test sets.
    BLEU, R1, R2, and RL are BLEU and ROUGE scores between model generated and reference recaps. 
    Bo\{C,R\}-\{p,r\} are precision and recall for \emph{Bag of Characters} and \emph{Bag of Character Relations}, respectively. The highest numbers for each dataset in each column are in bold. 
    }
    \label{summscreen-tab:anon_result}
\end{table*}

In \cref{summscreen-tab:anon_result}, it is interesting to observe the performance differences of the nearest neighbor models between the anonymized and non-anonymized datasets. The gaps show that the anonymization does not lead to much difference regarding the similarities between recaps and transcripts, but it makes correlations among recaps and transcripts much weaker especially for those entities.

\paragraph{Effect of Anonymization.}
\label{summscreen-sec:effect_anonym}
\begin{table}
    \centering\small
    \begin{tabular}{|l|c|c|}\hline
        Fraction & \tvmegasiteshort & Anonymized \tvmegasiteshort \\\hline
        All &  61.7 & 61.7 \\
        80\% & 19.1 & 25.5 \\
        60\% & 11.0 & 17.0 \\
        \hline
    \end{tabular}
    \caption{Average scores of entity metrics computed on various subsets of entities, dropping the most common entities when forming subsets. For example, the ``80\%'' row corresponds to omitting the most frequent 20\% of entities in each TV show. Results are based on the oracle hybrid model.}
    \label{summscreen-tab:effect_anonym}
\end{table}

We study the effect of anonymization by investigating performance on rare entities. To do so, we first compute entity frequencies for each TV show 
from the training set, rank the entities by their frequencies, pick the rare entities according to the rank, and evaluate performance for the selected entities. We summarize the results in \cref{summscreen-tab:effect_anonym}. We find that models trained on the anonymized \tvmegasiteshort dataset give better performance on rare entities, suggesting that anonymization helps in modeling rare entities. The fact that the two models have the same performance in the ``all'' setting shows that anonymization also makes the learning of common entities harder, matching our expectations.

\paragraph{Effect of Copy Mechanism.}
\label{summscreen-sec:effect_copy}

\begin{table}
    \centering\small
    \begin{tabular}{|l|c|c|}\hline
         & Generic & Entity \\\hline
        \multicolumn{3}{c}{\foreverdream} \\\hline
        w/o copy mechanism & 12.4 & 29.3 \\
        w/ copy mechanism & 12.6 & 27.1 \\\hline
        \multicolumn{3}{c}{Anonymized \foreverdream} \\\hline
        w/o copy mechanism & 11.2 & 11.9 \\
        w/ copy mechanism & 12.1 & 31.2 \\\hline
    \end{tabular}
    \caption{Comparing models with and without the copy mechanism on \foreverdream.}
    \label{summscreen-tab:copy_mechanism}
\end{table}

We report results on \foreverdream in \cref{summscreen-tab:copy_mechanism} comparing models with and without the copy mechanism. We note that models used in this table use 6-layer decoders with 512 hidden units, so the results are not directly comparable to other results. From the results in \cref{summscreen-tab:copy_mechanism}, we find that the copy mechanism helps tremendously on the anonymized dataset, but gives mixed results on the non-anonymized dataset. This is likely due to the fact that for the anonymized dataset, there is not enough training data for the character ID embeddings, and the copy mechanism helps to reduce the required supervision. While there may be better ways of handling the character IDs that may avoid this issue (e.g., sampling IDs from exponential-like distributions rather than uniform distribution), we leave this for future research.  

However, this benefit does not hold for the non-anonymized dataset as the models are able to exploit more information when learning character name embeddings by having access to the character names. 
\paragraph{Effect of Combining \foreverdreamshort and \tvmegasiteshort.}

\begin{table}
    \centering\small
    \begin{tabular}{|l|c|c|}\hline %
        & Generic & Entity \\
        \hline %
        \multicolumn{3}{c}{\foreverdream} \\
        \hline %
        \foreverdreamshort Only & 16.5 & 47.3  \\
        \tvmegasiteshort + \foreverdreamshort & 16.9 & 50.1 \\
        \hline %
        \multicolumn{3}{c}{Anonymized \foreverdream} \\
        \hline %
        Anonymized \foreverdreamshort Only & 13.7 & 11.3 \\
        Anonymized (\tvmegasiteshort + \foreverdreamshort) & 17.1 & 52.9 \\
        \hline %
        \multicolumn{3}{c}{\tvmegasite} \\\hline %
        \tvmegasiteshort Only & 25.9 & 61.7 \\
        \tvmegasiteshort + \foreverdreamshort & 23.2 & 58.0 \\
        \hline %
        \multicolumn{3}{c}{Anonymized \tvmegasite} \\\hline %
        Anonymized \tvmegasiteshort Only & 23.2 & 61.7 \\
        Anonymized (\tvmegasiteshort + \foreverdreamshort) & 22.7 & 59.8 \\
        \hline %
    \end{tabular}
    \caption{Results of 
    the oracle hybrid model comparing training on both datasets (\tvmegasiteshort + \foreverdreamshort) to training on the in-domain dataset only. The metrics are averaged scores of the generic and entity metrics. Training on both datasets helps for \foreverdreamshort but hurts for \tvmegasiteshort.} 
    \label{summscreen-tab:effect_combine}
\end{table}

We study the effect of transfer learning using these two resources. When doing so, we use the training and development sets constructed from both resources, and at test time, we evaluate models on the official test splits. We experiment with the oracle hybrid model and report results in \cref{summscreen-tab:effect_combine}. In general, we find that extra training data helps \foreverdreamshort. We hypothesize that this is due to the relatively small size of \foreverdreamshort. However, for \tvmegasiteshort, training on \foreverdreamshort harms performance, which is likely because of the larger training set size for \tvmegasiteshort and the differences between the two resources. It is interesting to see that the anonymized \foreverdream benefits greatly from additional training data, supporting our previous hypothesis that the copy mechanism helps to reduce the amount of required supervision.

\paragraph{Human Evaluation.}
\begin{table}
    \centering\small
    \begin{tabular}{|l|c|c|}\hline
& Predicates & Character Relation \\\hline
NNM-t2t & 1.6$\pm$0.8 & 2.1$\pm$1.1 \\
Hybrid model & 2.3$\pm$0.9 & 2.0$\pm$1.0 \\
Hybrid model (oracle) & 2.4$\pm$1.0 & 2.4$\pm$1.0 \\\hline
\end{tabular}
    \caption{Human evaluation results. We report the average scores and their corresponding standard deviations for questions on predicate match and character relation similarity. }
    \label{summscreen-tab:human_eval}
\end{table}

We conduct human evaluations for three models: NNM-t2t, hybrid model, and hybrid model (oracle). To evaluate two key aspects of \summscreen, namely events and characters relationships, we ask human annotators two questions.  The first is ``Do the predicates in the generation match the predicates in the reference?''\footnote{By ``predicate'' here we mean the part of a sentence or clause containing a verb and stating something about the subject (e.g., ``went home'' in ``John went home'').} The second is ``When multiple characters are mentioned as being related in some way in the generated recap, are those same characters mentioned as being related in some way in the reference?'' We disregard the subjects in the first question because the second question involves evaluating characters and we want the two questions to focus on different aspects to maximize the efficiency of human annotations. Ratings are on a 1-5 scale with 5 indicating a perfect match. We randomly picked instances from the \foreverdreamshort test set. We (the authors) annotated 120 instances in total for each question. 

After dropping 2 invalid annotations for the second question (as there may not be multiple characters mentioned), we summarize results in \cref{summscreen-tab:human_eval}. While trends for the model performance on character relations are generally similar to our observations in \cref{summscreen-tab:result}, the results for predicate match are very different for NNM-t2t. This is likely because the first question is about predicates disregarding the correctness of the participants.
We also want to highlight that compared to the oracle hybrid model, the non-oracle one shows competitive performance on predicate match but is less close in terms of generating correct character relations, showing future opportunities for improving this model.

\paragraph{Generation Samples.}

\begin{table*}
    \centering\small
    \begin{tabular}{|p{0.1\textwidth}|p{0.8\textwidth}|}\hline
Reference & The remains of two witches , one of which is from the Salem witch trials from the 1600s and the other a modern day Wiccan , are discovered in the remains of a burnt out cabin . Booth and Brennan investigate the world of Wicca , including discovering the Wiccan group of which the victim was a part . Hodgins and Angela wind up in jail after some reckless driving and have to work the case from the jail cell . After spending quality time together , they realize they are still in love . Hodgins subsequently proposes to Angela and they are married by the judge who hears their case . \\\hline
Neural model & Booth and Brennan are called to investigate when they are found dead in the death of a young woman who is found in to investigate . Meanwhile , Brennan and Booth are found at the victim 's death of an old friend , but the team must find out to investigate the team up with the case . The team investigates a young man who was killed when they discover that the victim was killed . The victim was not have been retrieve her father , Booth and Angela and Booth 's father 's death . \\\hline
Hybrid model & While the team investigates the death of a 40-year - old woman , who was found buried in a rock quarry . They discover that the woman 's feet were curled after she was buried , and that the bones were de - fleshed prior to her death . Meanwhile , Hodgins and Angela are in jail . Hodgins tells Angela that he 's a witch , but Angela tells Hodgins that she 's not a witch . The team finds out that the victim 's sister , Cheri , was also buried in the quarry . \\\hline
Hybrid model (oracle) & While the team investigates the death of a woman found buried in the woods . They discover that the victim was a Wiccan , and that she may have been a victim of a ritual that was used during the Salem Witch Trials . They also find that the woman was wearing red slippers and that her feet curled up after she was dead . Meanwhile , Hodgins and Angela are in a jail cell , and they are having a hard time adjusting to their new life in the city . The case is complicated by the fact that the body of the woman who was found is a young girl . \\\hline
\end{tabular}
    \caption{Generation samples from \foreverdream. The instance is from the TV show ``Bones''.}
    \label{summscreen-tab:gen_example}
\end{table*}

In \cref{summscreen-tab:gen_example}, we show generation samples for the following models: the neural model, the hybrid model, and the oracle hybrid model. The neural model manages to fit most of the character names from the reference into the generation. The generation shares similar topics with the reference, but compared to the hybrid models it lacks specifics. This matches our observations from the automatics metrics where the neural model performs better on the generic metrics but worse on the entity metrics on the non-anonymized datasets. We hypothesize that this is caused by the difficulty of modeling long-form text. 

In the output of the non-oracle hybrid model, many facts that are not mentioned in the reference are actually from the transcript. For example, ``40-year-old woman'' and ``de-fleshed prior to her death'' are in the transcript. Despite containing many specifics, the generation misses a few important details, such as the absence of mentioning main characters involved (i.e., Brennan and Booth). It also has incorrect facts. For example, according to the transcript, there are rocks at the scene, but the model describes the setting as a rock quarry. Compared to the other three models, the generation from the oracle hybrid model is the most faithful,
although there are still incorrect facts (e.g., ``... and they are having a hard time adjusting to their new life in the city.''). The differences between the oracle and non-oracle hybrid model suggest that future research can focus on improving models' capabilities of doing content selection. As both oracle and non-oracle hybrid models suffer from generating incorrect facts,
faithfulness in generation is also an important future research direction.

\section{Story Generation with Constraints}
\label{sec:tvstorygen}
\subsection{Introduction}

Story generation is the task of generating a coherent narrative. Due to its open-ended nature, increasing efforts have been devoted to  constrained settings to facilitate reliable evaluation of computational models, such as generating stories from short prompts \citep{fan-etal-2018-hierarchical} and story continuations \citep{mostafazadeh-etal-2016-corpus} with various constraints \citep{akoury-etal-2020-storium}. In this work, we are interested in generating stories that accord with descriptions about the characters involved. The task is akin to writing stories based on true events or historical figures. For example, when writing historical fiction, writers use facts in biographies of historical figures (i.e., character descriptions) \citep{brown1998historical}. In a similar vein, cognitive psychologists observed that in order for narrative text to be compelling, it has to base its characters on real-world details such that readers can form emotional attachments to them even if the events occurring in the text are not realistic \citep{oatley1999fiction,green2003power}. In either case, computational models for this task can offer assistance in proposing possible stories constrained by relevant documents.

To this end, we create a story generation dataset \tvrecap that generates detailed TV show episode recaps from a brief summary of the episode and a set of lengthy character descriptions.
We construct \tvrecap from fan-contributed websites, which allows us to collect 26k episode recaps covering a variety of genres. An example from \tvrecap is shown in \cref{tvrecap-fig:dataset_example}. The dataset is challenging in that it requires drawing relevant information from the lengthy character description documents based on the brief summary. Since the detailed episode recaps are constrained by character descriptions, it also can evaluate neural models’ ability to maintain consistent traits or goals of particular characters during generation.

In addition, by considering generating the brief summary from the detailed recap, we show that \tvrecap is a challenging testbed for abstractive summarization. To evaluate the faithfulness of the generated stories to the brief summaries, we propose a metric that uses the perplexities from the summarization model trained on our dataset.

Empirically, we characterize the dataset with several nearest neighbour methods and oracle models, finding that the use of the brief summaries and the character descriptions generally benefits model performance. We find that our non-oracle models are competitive compared to nearest neighbour models, suggesting promising future directions. We also benchmark several large pretrained models on the summarization version of our dataset, finding that they perform worse than an extractive oracle by a large margin despite the fact that the dataset favors abstractive approaches. Human evaluation reveals that without character descriptions, models tend to dwell on each event separately rather than advancing the plot, whereas using character descriptions improves the interestingness of the generated stories. Qualitatively, we show that  models are able to generate stories that share similar topics with the summaries, but they may miss events in the summaries, leading to unfaithful generations.

We summarize our contributions below:
\begin{enumeratesquish}
\item We construct a story generation dataset of 26k instances and show (both qualitatively and quantitatively) that it has several unique challenges.

\item We show that inverting our dataset provides a challenging testbed for abstractive summarization. Models trained on the inverted dataset can be used in evaluation for the original  dataset, namely to determine whether generated stories are faithful to their input summaries.

\item We empirically characterize the story generation dataset and the summarization version of our dataset with several nearest neighbour methods, oracle models, and pretrained models, showing the challenges of these tasks and suggesting future research directions.

\end{enumeratesquish}
\subsection{Related Work}
Early methods in computational modeling for story generation rely on handwritten rules \citep{meehan1977tale,liu2002makebelieve} to structure narrative. Recent work has explored different approaches to improve the quality of story generation systems, including commonsense knowledge \citep{mao-etal-2019-improving,guan-etal-2020-knowledge}, automatically extracted key words \citep{peng-etal-2018-towards} and key phrases \citep{orbach-goldberg-2020-facts2story,rashkin-etal-2020-plotmachines}, event-based representations \citep{martin2018event}, and plot graphs \citep{li2013story}.

As our model involves plot generation and character modeling, it is related to work on plot planning \citep{riedl2010narrative,li2013story,martin2018event,yao2019plan,jhamtani-berg-kirkpatrick-2020-narrative}, character modeling \citep{clark-etal-2018-neural,liu2020character}, and the interplay between the two \citep{riedl2010narrative}. Our work is different in that it explicitly requires performing inference on lengthy documents about characters.

There have been other datasets built from TV shows, such as summarizing TV show character descriptions \citep{shi-etal-2021-descgen}, constructing knowledge bases \citep{chu2021knowfi}, summarizing TV show screenplays \citep{chen-etal-2022-summscreen}, entity tracking \citep{chen-choi-2016-character,choi-chen-2018-semeval}, entity linking \citep{logeswaran-etal-2019-zero}, coreference resolution \citep{chen-etal-2017-robust,zhou-choi-2018-exist}, question answering \citep{ma-etal-2018-challenging,yang-choi-2019-friendsqa}, speaker identification \citep{ma-etal-2017-text}, sarcasm detection \citep{joshi-etal-2016-harnessing}, emotion detection \citep{zahiri2017emotion,hsu-ku-2018-socialnlp}, and character relation extraction \citep{yu-etal-2020-dialogue}.

\subsection{\tvrecap}

\begin{figure*}
    \centering
    \includegraphics[scale=0.5]{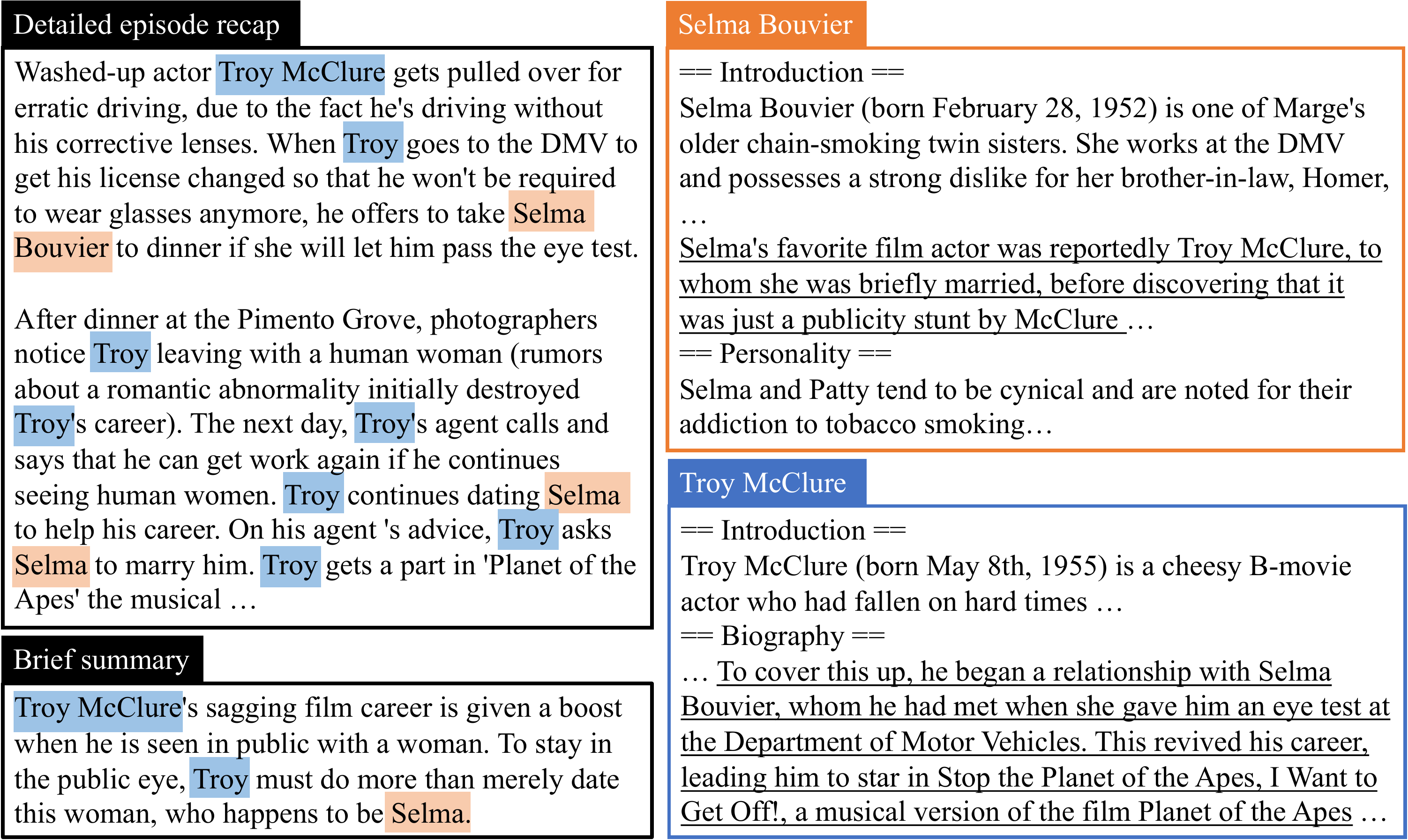}
    \caption{Excerpts from a \tvrecap instance that corresponds to the episode ``A Fish Called Selma'' in the TV show ``The Simpsons''. Colored texts are mentions of  characters. Texts surrounded by ``=='' in the character descriptions are section titles. We underline texts in the character descriptions that are closely related to the detailed recap. The shown part of the story includes complex interactions between \colorbox{atomictangerine}{Selma\vphantom{y} Bouvier} and \colorbox{babyblueeyes}{Troy McClure}, which requires models to use relevant information from the lengthy character descriptions. There are six characters involved in this story but we only show parts of the detailed recap and character descriptions for brevity.}
    \label{tvrecap-fig:dataset_example}
\end{figure*}

\begin{table*}
    \centering
    \small
    \begin{tabular}{|c|r|r|r|r|r|c|}\hline
         & \# stories & \# tps & \# cps & \# tuc & \# tpcd \\\hline
        ROCStories & 98.2k & 88.0 & - & - & -   \\\hline
        \writing & 303.4k & 735.0 & - & - & -  \\\hline
        \storium & 5.7k & 19.3k & 4.6 & 26.2k & 269.0  \\\hline
        \tvrecap & 29.0k & 1868.7 & 16.7 & 34.3k & 1553.4  \\\hline
    \end{tabular}
    \caption{Statistics for story generation datasets with prompts as part of inputs. \tvrecap has moderate numbers of stories, moderate lengths of stories, long character descriptions, and a large number of total unique characters and characters per story. tps: tokens per story. cps: characters per story. tuc: total unique characters. tpcd: tokens per character description.}
    \label{tvrecap-tab:story_generation_dataset_compare}
\end{table*}

In this section, we describe how we construct \tvrecap and compare it to other story generation datasets. 
An instance in \tvrecap is comprised of three components: (1) a detailed episode recap, (2) a brief summary of the episode, and (3) character descriptions, i.e., a set of documents describing the characters involved in the episode. The detailed episode recap delineates the events that occurred in the corresponding episode, which is usually written by fans after watching the episode. The documents about the characters contain biographical details and possibly personality traits. The summary either summarizes the whole episode or talks about the setup of the episode (to avoid spoilers). 

An example instance is shown in \cref{tvrecap-fig:dataset_example}, which comes from an episode of the TV show ``The Simpsons''. As there are relevant details mentioned in the character descriptions, generating the detailed recap requires drawing information from the lengthy character descriptions about the two characters. Moreover, due to the fact that the brief summary only depicts the setup of the episode, completing the story also necessitates using information in the character descriptions. That is, the character description information is expected to be useful for both filling in details that are not present in the brief summary as well as, for some of the instances, generating a plausible ending for the story.

\paragraph{Dataset Construction.}

We construct \tvrecap from two fan-contributed websites: \fandom\footnote{\url{https://www.fandom.com/}} (\fd) and \tvmegasite\footnote{\url{http://tvmegasite.net/}} (\tms). We collect brief summaries and detailed episode recaps for several long-running soap operas from \tvmegasite and other TV shows from \fandom. We collect character descriptions from \fandom.\footnote{Data from Fandom is available under Creative Commons licenses and we have received permission from the owners of \texttt{tvmegasite.net} to publish their data for use with attribution.}  Since the pages on \fandom have hyperlinks pointing to the character pages, we use the hyperlinks to connect episodes to the characters involved. For \tvmegasite, where there are no such hyperlinks, we use string matching to find the characters.
To ensure the quality of this dataset, we filter out episodes based on several criteria. See the appendix for more details on the criteria and the string matching algorithm.

\begin{table}
    \centering\small%
\begin{tabular}{|l|r|r|}\hline
& \fd & \tms \\\hline
number of shows & 106 & 9 \\
number of episodes & 13583 & 15430 \\
min. \# episodes per show & 2 & 17 \\
max. \# episodes per show & 574 & 2665 \\
median \# episodes per show & 14.0 & 300.0 \\
avg. \# episodes per show & 43.0 & 670.9 \\
\hline
avg. \# tokens in summaries & 56.7 & 366.6 \\
avg. \# tokens in detailed recaps & 1291.7 & 2375.3 \\
avg. \# tokens in char. desc. & 702.7 & 2300.9 \\
avg. \# characters & 15.5 & 17.8 \\\hline
\end{tabular}
    \caption{Dataset statistics for \tvrecap.}
    \label{tvrecap-tab:detailed_dataset_stats}
\end{table}

We report detailed statistics about \tvrecap in \cref{tvrecap-tab:detailed_dataset_stats}. 
As shown in the table, there are systematic differences between \fd and \tms in terms of length of detailed episode recaps, summaries, and character descriptions, among others. We note that the character descriptions for \tms also come from \fd. Considering the differences, we train and evaluate models on the two splits separately in experiments. Since the summaries in \fandom are shorter and likely only depict the setups of the detailed recaps, we conduct a human evaluation to check the fraction of setups in the summaries, finding that 61.7\% of the summaries are setups.\footnote{We sample 60 episodes with 2 episodes per show.}

\begin{table}
    \centering\small
    \begin{tabular}{|l|r|}\hline
Genre & Count (Fraction) \\\hline
Action & 5385 (15.0\%) \\
Comedy & 5214 (14.5\%) \\
Drama & 4793 (13.4\%) \\
Adventure & 4220 (11.8\%) \\
Children & 3280 (9.1\%) \\
Science-Fiction & 2524 (7.0\%) \\
Anime & 2239 (6.2\%) \\
Fantasy & 1949 (5.4\%) \\
Romance & 1408 (3.9\%) \\
Family & 1355 (3.8\%) \\
Crime & 1139 (3.2\%) \\
Supernatural & 729 (2.0\%) \\
Medical & 480 (1.3\%) \\
Horror & 309 (0.9\%) \\
Mystery & 299 (0.8\%) \\
Thriller & 240 (0.7\%) \\
Music & 220 (0.6\%) \\
History & 47 (0.1\%) \\
Legal & 19 (0.1\%) \\\hline
    \end{tabular}
    \caption{Genres in \fandom and their corresponding numbers and percentages of episodes.}
    \label{tvrecap-tab:fandom_genre_freq}
\end{table}

\begin{table}
    \centering\small
    \begin{tabular}{|l|r|}\hline
Genre & Count (Fraction) \\\hline
Drama & 15430 (44.6\%) \\
Romance & 8341 (24.1\%) \\
Family & 7686 (22.2\%) \\
Medical & 3144 (9.1\%) \\\hline
    \end{tabular}
    \caption{Genres in \tvmegasite and their corresponding numbers and percentages of episodes.}
    \label{tvrecap-tab:tvmegasite_genre_freq}
\end{table}

\begin{table}
    \centering\small
\begin{tabular}{|l|ccc|}
\hline
\fandom & Train & Dev & Test \\ \hline
\# shows & 106 & 104 & 106 \\
\# episodes & 10833 & 1320 & 1430 \\
\hline
\tvmegasite & Train & Dev & Test \\ \hline
\# shows & 7 & 7 & 9 \\
\# episodes & 11586 & 1452 & 2392 \\
\hline
\end{tabular}
    \caption{Statistics of train/dev/test splits for \fandom and \tvmegasite. }
    \label{tvrecap-tab:split_stat}
\end{table}

In \cref{tvrecap-tab:fandom_genre_freq,tvrecap-tab:tvmegasite_genre_freq}, we verify the diversity of topics covered in \tvrecap, finding that \fd covers far more genres than \tms with the most frequent occupying only 15\% of episodes. We randomly split the datasets into train/dev/test sets. For \tms, we additionally filter out instances if the overlap ratio of TV show characters appearing in the summary and the detailed recap is lower than 85\%. This extra filtering step ensures alignment between the summaries and detailed recaps. See \cref{tvrecap-tab:split_stat} for the train/dev/test sizes for \fd and \tms.

\paragraph{Dataset Comparison.}
\label{tvrecap-sec:compare_to_other_datasets}
We compare \tvrecap to other story generation datasets in \cref{tvrecap-tab:story_generation_dataset_compare}. Unlike ROCStories \citep{mostafazadeh-etal-2016-corpus} and \writing \citep{fan-etal-2018-hierarchical} where the inputs to models are either the first few sentences or short prompts, \tvrecap has character descriptions as extra constraints, making the task of generating the reference stories from the inputs less open-ended and therefore more feasible.

Since \storium \citep{akoury-etal-2020-storium} has character descriptions and other information as constraints, it is the most comparable resource to \tvrecap. Below we compare our dataset to \storium in detail.

\begin{table}
    \centering\small
    \begin{tabular}{|l|r|r|r|r|}\hline
         & uni. & bi. & tri. & four.  \\\hline
        \multicolumn{5}{|c|}{\tvrecap (\fandom)}\\\hline
        summ. & 34.3 & 3.4 & 0.8 & 0.3 \\
        char. desc. & 88.1 & 48.3 & 16.7 & 6.2 \\
        char. desc. $\setminus$ summ. & 54.3 & 45.4 & 16.3 & 6.1 \\
        summ. $\setminus$ char. desc. & 0.5 & 0.6 & 0.4 & 0.2 \\
        char. desc. $\cup$ summ. & 88.7 & 48.9 & 17.1 & 7.4\\\hline
        \multicolumn{5}{|c|}{\tvrecap (\tvmegasite)}\\\hline
        summ. & 61.7 & 14.7 & 3.0 & 1.2 \\
        char. desc. & 93.4 & 56.9 & 17.3 & 3.2 \\
        char. desc. $\setminus$ summ. & 32.7 & 44.2 & 16.1 & 3.1 \\
        summ. $\setminus$ char. desc. & 0.9 & 2.0 & 1.8 & 1.0 \\
        char. desc. $\cup$ summ. & 94.3 & 58.9 & 19.1 & 4.2 \\\hline
        \storium & 72.5 & 24.7 & 5.4 & 1.2 \\\hline
    \end{tabular}
    \caption{Fraction (\%) of n-grams in the \textbf{output stories} that also appear in the source inputs. Higher fraction of overlapping n-grams indicates that the two are more directly related. For \tvrecap, we vary different kinds of inputs. 
    }
    \label{tvrecap-tab:overlap_ratio_story_generation_dataset_compare}
\end{table}

\begin{enumeratesquish}
\item Our dataset has more stories, more characters, and longer character descriptions.
\item The stories in \storium often have detailed descriptions about environments and character utterances, whereas the stories in \tvrecap mostly narrate events that happened without these details. While this leads to shorter stories in \tvrecap, it also prevents the task from conflating generating events and generating other kinds of details in story generation.
\item Due to the fact that the plots in \storium are gamified and crafted by amateur writers, 89.8\% of stories in \storium are unfinished.\footnote{We label a story as unfinished if it has no completion date.} The stories in our dataset are created and refined by professional screenwriters (though the prose is written by fans, who are presumably amateurs).
\item  Stories in \storium are turn-based, where each turn is written from the perspective of a particular character and is composed by one player, so the stories often lack direct interactions among characters, unlike \tvrecap.
\item Unlike other story generation datasets, there is an episodic structure among the stories in \tvrecap, which can potentially be used to improve the modeling of characters.
\item Source inputs and output stories are more closely related in \tvrecap than \storium. To quantitatively illustrate the extent of relatedness between the source inputs and the output stories, we compute the n-gram overlap ratio (i.e., fraction of n-grams in the \textbf{output stories} that also appear in the source inputs) between the inputs and outputs where higher ratios indicate that the two are more directly related. When computing the results for \storium, we use the best setting, i.e., the setting that maximizes the automatic and human evaluation scores in the original paper. We report results in \cref{tvrecap-tab:overlap_ratio_story_generation_dataset_compare}. From the table, we see that for both \fd and \tms, using both character descriptions and summaries leads to an overlap ratio higher than \storium, suggesting that the reference stories are more reachable. Also, we observe there are more overlapping n-grams in the character descriptions than the summaries, suggesting that there is useful information that can be extracted from the character descriptions.
\item \storium lacks direction interactions among characters. We quantify this phenomenon in \storium by computing the frequency of occurrences of characters in each turn excluding the character that owns the turn, 
and the frequency is 0.8 on average with 50.4\% of the turns absent such occurrences.\footnote{We use string matching to detect the occurrences of characters as in the way we construct our dataset.} In contrast, TV shows advance plots by interactions among characters. 
\end{enumeratesquish}

Moreover, models trained on \tvrecap can potentially complement those from \storium by merging all the characters' turns in \storium into a coherent narrative.

\begin{table*}
    \centering\small
    \begin{tabular}{|l|r|r|r|c|}\hline
         & \# inst. & inp. len. & out. len. & Domain \\\hline
        \multicolumn{5}{|c|}{Long-form text summarization datasets}\\\hline
        Multi-News \citep{fabbri-etal-2019-multi} & 56.2k & 2103.5 & 264.7  & News \\
        RottenTomatoes \citep{wang-ling-2016-neural}  & 3.7k & 2124.7 & 22.2 & Reviews \\
        arXiv \citep{cohan-etal-2018-discourse}  & 215k & 4938.0 & 220.0 & Science \\
        PubMed \citep{cohan-etal-2018-discourse} & 113k & 3016.0 & 203.0 & Science \\
        GovReport \citep{huang-etal-2021-efficient} & 19.5k & 9409.4 & 553.4 & Government Reports \\
        \tvrecap & 29.0k & 1868.7 & 221.6 & Television Series \\
        \hline
        \multicolumn{5}{|c|}{Dialogue-related summarization datasets}\\\hline
        SAMSum \citep{gliwa-etal-2019-samsum} & 16.4k & 83.9 & 20.3 & Chitchat \\
        QMSum \citep{zhong-etal-2021-qmsum} & 1.8k & 9069.8 & 69.6 & Meetings \\
        MediaSum \citep{zhu-etal-2021-mediasum} & 463.6k & 1553.7 & 14.4 & News Interviews \\
        ForumSum \citep{khalman-etal-2021-forumsum-multi} & 4.1k & 303.5 & 36.0 & Forum Messages \\
        SummScreen \citep{chen-etal-2022-summscreen} & 26.9k & 6612.5 & 337.4 & Television Series  \\\hline
    \end{tabular}
    \caption{Statistics for datasets focusing on abstractive summarization for long-form text or dialogue. The numbers are averaged over instances.
    }
    \label{tvrecap-tab:other_summarization_dataset_compare}
\end{table*}

\begin{table}
    \centering\small
    \begin{tabular}{|l|r|r|r|r|r|r|}\hline
         & uni. & bi. & tri. & four. & src. & tgt. \\\hline
        \multicolumn{7}{|c|}{\tvrecapsum}\\\hline
        \fd & 73.0 & 26.7 & 8.1 & 3.0 & 1.3k & 56.7 \\
        \tms & 85.0 & 40.1 & 13.6 & 5.8 & 2.4k & 366.6 \\\hline
        \multicolumn{7}{|c|}{Other summarization datasets}\\\hline
        XSum\textsuperscript{\dag} & 64.2 & 16.6 & 4.5 & 1.5 & 431.1 & 23.3 \\
        CNNDM\textsuperscript{\S} & 80.5 & 43.1 & 25.6 & 17.2 & 810.6 & 56.2\\
        MNews\textsuperscript{\S} & 82.2 & 42.9 & 24.3 & 17.7 & 2.1k & 264.7 \\\hline
    \end{tabular}
    \caption{Fraction (\%) of n-grams in the \textbf{output summaries} that also appear in the inputs, and the average numbers of tokens for the inputs and outputs. Datasets with smaller fractions of overlapping n-grams tend to favor abstractive summarization approaches. Results marked by \dag\xspace and \S\xspace are from \citet{narayan-etal-2018-dont} and \citet{fabbri-etal-2019-multi} respectively.}
    \label{tvrecap-tab:overlap_ratio_summarization_dataset_compare}
\end{table}

\paragraph{Summarization.} By considering generating the brief summary from the detailed episode recap, we can view \tvrecap as an abstractive summarization dataset, which we call \tvrecapsum. We simply use the detailed episode recap as the source input and the summary as the target output and leave the integration of character descriptions to future work. We briefly compare \tvrecapsum to  three summarization datasets: CNNDM \citep{teaching2015hermann}, XSum \citep{narayan-etal-2018-dont}, and MNews \citep{fabbri-etal-2019-multi}. For \tvrecapsum, we simply use the detailed episode recap as the source input and the summary as the target output and leave the integration of character descriptions to future work. We report n-gram overlap ratio (i.e., fraction of n-grams in the \textbf{output stories} that also appear in the source inputs) and length statistics in \cref{tvrecap-tab:overlap_ratio_summarization_dataset_compare}. The n-gram overlap ratio is usually used as an indicator of the abstractiveness of a summarization dataset. Lower ratio indicates a higher degree of abstraction. CNNDM favors extractive approaches, whereas XSum is known for it is abstractiveness. We also compare to MNews because it shares similar input and output lengths as our dataset. As shown in the table, our dataset tends to be more abstractive. In addition, unlike other summarization datasets, our dataset focuses on stories. These two characteristics make our dataset a potentially valuable contribution for the summarization community. Comparison to other abstractive summarization datasets is in \cref{tvrecap-tab:other_summarization_dataset_compare}. The fact that \tvrecapsum favors abstractive approaches and focuses on stories make it a potentially valuable contribution for the summarization community.

\paragraph{Dataset Challenges.}

\begin{table}
    \centering\small
    \begin{tabular}{|p{0.7\textwidth}|}\hline
         ...Meanwhile, Patty and {\bf Selma} have received a promotion at the DMV and have more disposable income. As a last resort, {\bf Homer} asks the two if they will lend him the money. They agree, but he must become their loyal servant. Patty and {\bf Selma} make {\bf Homer}'s life a living hell...\\\hline\hline
        ...The next day, news of {\bf Homer}'s ``death'' spreads across Springfield, and Marge starts getting condolences from prominent Springfieldians. Patty and {\bf Selma} offer their condolences in the form of a tombstone celebrating {\bf Homer}’s death...\\\hline
    \end{tabular}
    \caption{Two excerpts in detailed recaps from \tvrecap that correspond to different episodes in the TV show ``The Simpsons''. The excerpts involve interactions between Homer and Selma where Selma consistently shows a strong dislike for Homer, matching the character description in \cref{tvrecap-fig:dataset_example}. }
    \label{tvrecap-tab:challenge_example}
\end{table}

\tvrecap poses several challenges for story generation models. The first challenge stems from the long lengths of the inputs and outputs. Specifically, the average instance in \tvrecap has a story of 1.8k tokens and character descriptions of more than 10k tokens (see \cref{tvrecap-tab:detailed_dataset_stats}). In contrast, story generation and multi-document summarization datasets may have lengthy inputs or outputs, but rarely both (see \ref{tvrecap-tab:story_generation_dataset_compare} and \cref{tvrecap-sec:compare_to_other_datasets} for detailed statistics). The long inputs and outputs make it challenging to design models that can effectively integrate lengthy character descriptions into the generation process of long and coherent stories.

The other set of challenges relates to consistency in character modeling. Since the episode recaps are constrained by character descriptions, the dataset provides opportunities to evaluate neural models' ability to maintain consistent personalities or goals of particular characters during generation. The consistency of personalities and goals is related to the notion of ``character believability'' \citep{bates1994role,riedl2010narrative}, which has been deemed important for composing convincing stories. We illustrate this challenge with two excerpts in \cref{tvrecap-tab:challenge_example}: the strong dislike that Selma has shown for Homer matches her description and is consistent across episodes. Solving this challenge requires models to first identify related information in the lengthy character descriptions based on the plot and integrate it into the generated narrative. We aim to incorporate this idea into the design of our models.

\subsection{Method}

We follow \citet{fan-etal-2019-strategies} to take a hierarchical story generation approach. The generation process is broken into two steps that use two separately parameterized models: a text-to-plot model and a plot-to-text model. The text-to-plot model first generates detailed plots based on the inputs, and then conditioned on the plots the plot-to-text model generates detailed stories. In this paper, we define the plots as linearized semantic role labeling (SRL) structures. More details on SRL are in the appendix.
For example, a plot may be as follows:
\begin{align*}
    &\langle\text{VERB}\rangle\text{spots}\langle\text{ARG0}\rangle\text{Mummy Pi} \langle\text{ARG1}\rangle\text{how messy the car is}\langle\text{SEP}\rangle\\ &\langle\text{VERB}\rangle\text{clean}\langle\text{ARG0}\rangle\text{they}
    \langle\text{ARG1}\rangle\text{the car}
\end{align*}
\noindent where $\langle\text{SEP}\rangle$ is a special token used to separate SRL structures for different sentences.

\paragraph{Text-to-Plot Model.} During training, we use the oracle plots, i.e., the SRL tags extracted from the reference recaps. During test time, we use BM25 to find the most similar plot in the training set from the same show based on either the summaries or the detailed recaps (as an oracle baseline).\footnote{We find the plots generated by neural models to be of lower quality.} If a show is not present in the training set, we search over the whole training set.

\paragraph{Plot-to-Text Model.} Our models are based on the sequence-to-sequence transformer architecture \citep{attention_is_all_you_need}. 
Similar to \citet{rothe-etal-2020-leveraging} that uses pretrained BERT-like models to initialize sequence-to-sequence models, we use the pretrained RoBERTa-base model \citep{liu2019roberta} as the decoder.\footnote{We chose RoBERTa over GPT-2 \citep{radford2019language} because BERT-style models outperform GPT-2 in the encoder-decoder setting in the results reported by  \citet{liu2019roberta}.} For the encoder, we choose to use a one-layer randomly initialized Longformer \citep{beltagy2020longformer} due to the lengthy inputs and computational constraints. We randomly initialize other parameters and finetune the whole model during training. 

Given a plot, we use the neural models to generate sentence by sentence as we find this yields better performance than generating the whole detailed recap. When doing so, we concatenate the SRL tags for the adjacent sentence of the target sentence with the SRL tags for the target sentence. This gives similar performance to showing the SRL tags for the entire detailed recap but is much more efficient (due to shorter sequence lengths). Because the character descriptions are lengthy, we use BM25 to retrieve the most salient information from character descriptions (i.e., one sentence) for each sentence in the detailed recap. We note that during test time, when the text-to-plot model retrieves plots from the training set, we also use the corresponding selected character descriptions.

The pipeline that retrieves relevant character information and then adapts it based on the plot is the first step that we take to simulate a writing system that can dynamically update its belief about particular characters based on the given relevant documents. This differs from prior work on entity representations for story generation \citep{clark-etal-2018-neural} that does not consider character descriptions as we do.

The inputs to plot-to-text models contain two sources: plots and character descriptions. Since there could be multiple entries corresponding to different characters in the character descriptions, we include a type embedding to differentiate different entries and sources in the input. Similar approaches have been used to represent table entries in neural models \citep{dhingra-etal-2019-handling,herzig-etal-2020-tapas,yin-etal-2020-tabert}. For example, for \cref{tvrecap-fig:dataset_example} the inputs are
\begin{align*}
    \langle\text{SEP}\rangle_0\text{Troy McClure's ...}_0\langle\text{SEP}\rangle_1\text{Selma Bouvier}\\
    \langle\text{SEP}\rangle_1\text{Selma's favorite film ...}_1\langle\text{SEP}\rangle_2\text{...}
\end{align*}
where the subscripts indicate the ID of the type embedding and we always prepend the character names to the corresponding selected character descriptions. The final vector representation of the input is the summation of subword unit embeddings, positional embeddings, and the type embeddings. Conditioned on the input representations, we train the RoBERTa decoder on the reference recaps using a cross-entropy loss.

Due to computational constraints, for the Longformer encoder, we use the global attention on the $\langle\text{SEP}\rangle$ tokens, and use the encoded representations for the summary, the SRL tags, and the $\langle\text{SEP}\rangle$ tokens in character descriptions as the input to decoders.

\subsection{Experiments}
We perform experiments for both story generation and summarization. 

\paragraph{Experimental Setup.}

For both story generation and summarization, we use a batch size of 200, beam search of size 5 with n-gram blocking where probabilities of repeated trigrams are set to 0 during beam search,\footnote{We did not find nucleus sampling \citep{Holtzman2020The} leading to better generation quality (i.e., fluency and faithfulness to the summaries and the recaps) than beam search with n-gram blocking, possibly due to the fact that our models generate at  the sentence level.} and report BLEU (BL), ROUGE-1 (R1), ROUGE-2 (R2), and ROUGE-L (RL) scores. For story generation, we additionally report perplexities of the summaries given the generated stories using the summarization models. We will refer to this metric as ``PL''. This metric evaluates the faithfulness of the generated stories to the summaries. Lower PL suggests better faithfulness. When computing PL, we use the Pegasus model \cite{pegasus-zhang20ae} finetuned on our dataset as it has the best test set perplexities. More details on hyperparameters are in the appendix. The dataset is available at \url{https://github.com/mingdachen/TVStoryGen}.

\paragraph{Experimental Result.}

\begin{table*}[t]\footnotesize
    \centering\setlength{\tabcolsep}{6pt}
    \begin{tabular}{|l|r|r|r|r|r|}\hline
         & BL ($\uparrow$) & R1 ($\uparrow$) & R2 ($\uparrow$) & RL ($\uparrow$) & PL ($\downarrow$)  \\\hline
         \multicolumn{6}{|c|}{Development set results} \\\hline
        (NN) Nearest neighbour plot + summary + char. desc. & 7.1 & 40.7 & 11.0 & 39.6 & 32.3 \\
        (NN) Oracle plot & 21.2 & 52.8 & 24.1 & 51.8 & 23.1 \\
        (NN) Oracle plot + summary & 24.5 & 54.3 & 25.6 & 55.2 &  20.8 \\
        (NN) Oracle plot + summary + oracle char. desc. & \bf 28.4 & \bf 63.0 & \bf 32.8 & \bf 61.2 & \bf 17.9 \\
        \hline
         \multicolumn{6}{|c|}{Test set results} \\\hline
        (Return) reference & 100.0 & 100.0 & 100.0 & 100.0 & 12.9 \\
        (Return) oracle plot  & 3.6 & 43.9 & 19.7 & 41.9 & - \\
        (Return) oracle plot + summary & 5.4 & 48.5 & 20.5 & 46.2 &  - \\
        (Return) oracle plot + oracle char. desc. & 1.2 & 11.0 & 4.6 & 10.6 & - \\
        (Return) oracle plot + oracle char. desc. + summary & 1.2 & 11.0 & 4.7 & 10.6 & - \\\hline
        (Return) Nearest neighbour detailed recap & 5.1 & 41.1 & 9.3 & 39.6 & 31.2 \\
        (Return) Oracle nearest neighbour detailed recap & 4.8 & 41.2 & 10.8 & 39.9 & 28.5 \\\hline
        (NN) Nearest neighbour plot + summary + char. desc. & 6.0 & 41.7  & 10.7 & 40.3 & 28.0 \\
        (NN) Oracle plot + summary + oracle char. desc. & \bf 28.4  & \bf 63.2 & \bf 32.9 & \bf 61.5 & \bf 18.2  \\\hline
    \end{tabular}
    \caption{ \fandom results. The results for the return-input baselines and the neural models are indicated by``(Return)'' and  ``(NN)'' respectively. The best result in each column for each split (excluding the references) is boldfaced.}
    \label{tvrecap-tab:fandom_main_result}
\end{table*}

\begin{table*}\footnotesize
    \centering\setlength{\tabcolsep}{6pt}
    \begin{tabular}{|l|r|r|r|r|r|}\hline
         & BL ($\uparrow$) & R1 ($\uparrow$) & R2 ($\uparrow$) & RL ($\uparrow$) & PL ($\downarrow$)  \\\hline
         \multicolumn{6}{|c|}{Development set results} \\\hline
        (NN) Nearest neighbour plot + summary + char. desc. & 10.7 & 43.5 & 14.9 & 42.9 & 20.7 \\
        (NN) Oracle plot & 26.4 & 60.5 & 34.0 & 60.0  & 17.0 \\
        (NN) Oracle plot + summary & 28.3 & 64.3 & 36.1 & 63.9 & 16.4  \\
        (NN) Oracle plot + summary + oracle char. desc. & \bf 30.9 & \bf 68.3 & \bf 44.0 & \bf 67.5 & \bf 15.7 \\
        \hline
         \multicolumn{6}{|c|}{Test set results} \\\hline
        (Return) Reference & 100.0 & 100.0 & 100.0 & 100.0 & 13.9 \\
        (Return) Oracle plot  & 7.1 & 53.1 & 22.3  & 52.4  & - \\
        (Return) Oracle plot + summary & 13.6 & 62.9 & 25.6 & 62.0 &  - \\
        (Return) Oracle plot + oracle char. desc. & 1.1 & 12.3 & 5.1  & 11.9 & - \\
        (Return) Oracle plot + oracle char. desc. + summary & 1.2 & 12.4 & 5.1 & 12.0 & - \\\hline
        (Return) Nearest neighbour detailed recap & 6.6 & 49.8 & 16.0 & 49.2 & 26.3 \\
        (Return) Oracle nearest neighbour detailed recap & 7.5 & 49.9 & 18.5 & 49.3 & 25.8 \\\hline
        (NN) Nearest neighbour plot + summary + char. desc. & 7.3 & 50.6 & 17.6 & 49.8 & 25.3 \\
        (NN) Oracle plot + summary + oracle char. desc. &  \bf 28.1 & \bf 67.0 &\bf 40.9  & \bf 66.2 & \bf 18.3  \\\hline
    \end{tabular}
    \caption{\tvmegasite results. The results for the return-input baselines and the neural models are indicated by ``(Return)'' and ``(NN)'' respectively. The best result in each column for each split (excluding the references) is boldfaced.}
    \label{tvrecap-tab:tvmegasite_main_result}
\end{table*}

We report results for \fd and \tms in \cref{tvrecap-tab:fandom_main_result,tvrecap-tab:tvmegasite_main_result}, respectively. We report several return-input baselines on the test sets to show the benefits of using neural models as plot-to-text models. We report PL on the test sets as an approximated lower bound of this metric. We do not report PL on return-input baselines as the output detailed recaps involve SRL sequences, which are not natural language, and therefore the results are not comparable to others.

On the development sets, adding summaries and oracle character descriptions generally improves performance by a significant margin, showing that the extra information aids generation.

Regarding the test set results, we find that (1) the return-input baselines show that the performances of our neural models are non-trivial; (2) while the oracle nearest neighbour baselines achieve competitive performance to our non-oracle neural models, the non-oracle neural models are consistently better than the non-oracle baselines, showing promising results for future research on our datasets. We note that the return-input baselines involving character descriptions display much worse results than other return-input baselines because they are lengthy, which leads to low precision.

\begin{table}
    \centering\small\setlength{\tabcolsep}{6pt}
    \begin{tabular}{|l|r|r|r|r|}\hline
         & BL & R1 & R2 & R3  \\\hline
         \multicolumn{5}{|c|}{\fandom} \\\hline
        Extractive oracle & 8.3 & 37.0 & 11.3 & 30.9 \\
        BART-base & 5.2 & 31.2 & 7.3 & 25.5 \\
        BART-large & 5.4 & 30.7 & 7.6 & 25.3 \\
        Pegasus & \bf 5.7 & \bf 31.3 & \bf 7.7 & \bf 25.6 \\\hline
         \multicolumn{5}{|c|}{\tvmegasite} \\\hline
        Extractive oracle & 16.9 & 55.8 & 20.9 & 53.6 \\
        BART-base & \bf 8.3 & \bf 43.8 & \bf 12.6 & \bf 42.3 \\
        BART-large & 8.1 & 43.2 & 12.3 & 41.8 \\
        Pegasus & 7.7 & 43.5 & \bf 12.6 & 42.1 \\\hline
    \end{tabular}
    \caption{Test results for summarizing detailed episode recaps. The best result in each column for each domain (excluding the oracle) is boldfaced.}
    \label{tvrecap-tab:summarization_main_result}
\end{table}

We report results in \cref{tvrecap-tab:summarization_main_result}. We report the performance of an extractive oracle where for each sentence in the reference summary, we pick a sentence in the detailed episode recap that maximizes the average of the three ROUGE scores compared against the summary sentence. While recent pretrained models, such as Pegasus, have outperformed the oracle extractive approaches by a large margin on datasets with a high degree of abstractiveness (e.g., XSum \citep{narayan-etal-2018-dont}), the results in the table show that our dataset is still challenging for these pretrained models.

It is also interesting to see that while Pegasus is best for \fandom, it is not best on \tvmegasite. This may be because \tvmegasite has longer summaries than \fandom. Also, the performance of BART-base is comparable to that of BART-large on \fandom and is better than Pegasus and BART-large on \tvmegasite. This is likely because there is a limited amount of data with similar writing style in pretraining, resulting in little benefit of using larger models for this downstream task. We provide this abstractive summarization task to the community as a challenging dataset for future work.
\subsection{Analysis}

\paragraph{Human Evaluation.}

\begin{table}
    \centering\small
    \begin{tabular}{|l|r|r|}\hline
        & \multicolumn{1}{|c|}{Relevancy} & \multicolumn{1}{|c|}{Interesting} \\\hline
        \multicolumn{3}{|c|}{Expert annotations}\\\hline
        Prefer summary & 60.0\% (30/50) & 40.0\% (20/50) \\
        Prefer char. desc. & 54.0\% (27/50) & 70.0\% (35/50) \\\hline
        \multicolumn{3}{|c|}{Crowdsourced annotations}\\\hline
        Prefer summary & 50.0\% (11/20) &  55.0\% (10/20)\\
        Prefer char. desc. & 55.0\% (11/20) & 55.0\% (11/20)\\\hline
    \end{tabular}
    \caption{Human annotation results analyzing the effect of including different components in the inputs. The percentage is the fraction of annotations that favor the models to include the corresponding component. The numbers in parentheses are the number of positive annotations divided by the total number of annotations.}
    \label{tvrecap-tab:human_annoation_result}
\end{table}

To measure the impact of including different components in \tvrecap, we conduct a human evaluation. We show two generated stories from different models along with the corresponding brief summary and ask annotators to choose which story they prefer according to two aspects: (1) which generation is more relevant to the summary; (2) which story is more interesting.

We make two comparisons: ``oracle plot'' vs. ``oracle plot+summary'' for studying the benefits of using summaries (``Prefer summary''), and ``oracle plot+summary'' vs. ``oracle plot+summary+oracle char.~desc.'' for studying the benefits of using character descriptions (``Prefer char.~desc.''). We sample instances from the \fd development set because the average lengths in \fd are shorter, and we only show annotators the first 100 tokens of the texts as we expect it to be challenging to annotate lengthy texts. We use Amazon Mechanical Turk (AMT) and collect 20 annotations per question for each comparison with 6 workers involved (shown in \cref{tvrecap-tab:human_annoation_result} as ``crowdsourced annotations'').\footnote{To ensure annotation quality, we hire workers with master qualification and pay them with a target hourly wage of \$12.} We (the authors) also annotate 50 instances per comparison using the same interface as AMT (shown in the table as ``expert annotations''). While the crowdsourced annotations do not suggest clear benefits of using summaries and character descriptions, the expert annotations show that including the summary helps to improve relevancy but hurts the interestingness of the stories, whereas including character descriptions improves the interestingness despite the marginal benefits of improving the relevancy. Recent work \citep{karpinska-etal-2021-perils} also found that compared to experts, AMT workers produce lower quality annotations for tasks like story generation.

When examining annotations, we find that models without using character descriptions tend to generate sentences that use the word ``but'' to negate what has been said in the earlier part of the sentence, leaving the sentence dwelling on each event separately rather than advancing the plot (see \cref{tvrecap-sec:gen_examples} for examples). To quantify the observation that the models tend to generate sentences that use the word ``but'' to negate what has been said in the earlier part of the sentence, we compute the frequency of the word ``but'' per sentence for the reference stories, ``oracle plot+summary'', and ``oracle plot+summary+oracle char.~desc.''. The results are  0.13, 0.53, and 0.24, respectively.

\begin{table}
    \centering\small
    \begin{tabular}{|l|r|r|}\hline
         &  \multicolumn{1}{|c|}{Acc.} & \multicolumn{1}{|c|}{$F_1$} \\\hline
        BL & 55.9 & 35.1 \\
        BL (trunc. gen.) & 50.0 & 32.4 \\
        PL & 55.0 & 34.5 \\ 
        PL (trunc. gen.) & \bf 61.4 & \bf 49.6 \\\hline
    \end{tabular}
    \caption{Accuracies and $F_1$ scores when evaluating the automatic metrics against human annotations. The best performance in each column is in bold.}
    \label{tvrecap-tab:eval_pl_metric}
\end{table}

To verify the efficacy of our proposed metric PL, we compute accuracies and F1 scores between the PL metric and the human annotations (we use the expert relevancy annotations from both comparisons in human evaluation results). 
We consider BL as a baseline metric by computing the generation against the brief summaries. When reporting results for PL and BL, we consider two variants: one that uses the truncated generation and the other one that uses all the tokens in the generation. We show the results in \cref{tvrecap-tab:eval_pl_metric}. While PL and BL show similar performance in the non-truncated setting, we find that in the truncated setting PL outperforms BL significantly, showing that PL is a promising metric for evaluating the faithfulness of generated story. We speculate that the discrepancy is likely caused by the fact that the annotations were collected based on the truncated generations.

\paragraph{Generation Examples.}
\label{tvrecap-sec:gen_examples}

\begin{table*}
    \centering
    \footnotesize
\begin{tabular}{|p{0.25\textwidth}|p{0.21\textwidth}|p{0.22\textwidth}|p{0.3\textwidth}|}\hline
\multicolumn{1}{|c|}{Input summary} & \multicolumn{1}{|c|}{Reference} & \multicolumn{1}{|c|}{Oracle plot+summary} & \multicolumn{1}{c|}{Oracle plot+summ.+oracle char.} \\\hline
\ul{Elfman and Evergreen , after much struggle , lose to Rustyrose and his imagination - based Magic .} In the meantime , Gray , Lucy , Cana and Loke have been overpowered by Caprico alone . Loke decides to take him on by himself because of his mysterious Magic .
& \ul{Elfman and Evergreen encounter a cliff as they run away from Rustyrose 's Belcusas the Thunderclap .} Rustyrose appears shortly afterwards and expands on the idea of " The Ultimate World of Magic " , saying that all those who can not use Magic and the trash in the guilds are useless ...
& Elf episode begins with Elfman and Evergreen encounter a cliff in the middle of the forest , and the two begin to fight . All , all those those are going to use Magic to defeat them , and they will be able to defeat all of them . Ruth , however , has n't been able to outsmart them , and the two of them begin to fight ... 
& \ul{Elfman and Evergreen encounter a cliff as they run away from the danger of the Rustyrose 's attacks .} Evergreen explains that " all those who use Magic " will use the power of the Magic of the Seven . However , Rustyrose outsmarts them , stating that their Magic is useless against them , as they have already been defeated . He then finishes the two with Tower of Dingling .. \\\hline
\ul{Chuck is preparing for his new club opening and enlists Serena 's help} , but Blair begins to feel left out . Jenny , the new Queen at Constance , struggles between proving herself and her friendship with Eric , and Dan feels inferior after watching one of Olivia 's movies . Meanwhile , Lily tries to respect Rufus ' Halloween traditions .
& ... Blair is confident her plan will work , until \ul{Chuck calls and asks for Serena 's help in conducting the club opening .} He tells her he wants to open the next day on Halloween and that he does n't want Blair anywhere near the planning ...
& ... Chuck tells Blair that he 's not going to let her go , {\bf but that she 's going to be there for him} . He he opens the club and tells her that he 's going to see Serena at the loft , where she 's staying . She suggests an 80-year party party , {\bf but she says that 's not what she 's looking for} ...
& ... Blair goes to see Serena at the gallery , and she explains to her that she went to see Chuck . She tells her that she hired a party planner to help her hang out with Chuck , and Serena offers to help . \ul{At the VDW 's , Serena is conducting the club opening and Blair asks her to help .} He tells her that he does n't want Blair anywhere near the planning , and he wants her to be at the party for the night ...\\\hline
\end{tabular}
\caption{Excerpts from generation examples, which come from the TV shows ``Fairy Tail'' and ``Gossip Girl'' respectively. The highlighted texts are meaningless negations. We underline texts that describe similar events.}
\label{tvrecap-tab:gen_examples}
\end{table*}

We display the generation examples in \cref{tvrecap-tab:gen_examples} where we find that generations from both models generally share similar topics and character names with the summaries and the references. For example, for the first instance, both generations are about a battle that concerns Elfman, Evergreen, and Rustyrose. However, as observed in the human evaluations, the ``oracle plot+summary'' model suffers from meaningless negation. For example, see the second generation example, where the highlighted texts keep negating the earlier plot development. While the ``Oracle plot+summ.+oracle char.'' model does not have this problem, it is still not faithful to the summary. Specifically, both the summary and the reference mention that Chuck needs Serena's help for his new club opening, but the generation states that ``Serena is conducting the club opening'' and ``Blair asks her to help''. This is likely caused by the model's inability to understand the states of each character (possibly due to the fact that our models generate at the sentence level) and to  effectively integrate multiple sources of information into a coherent narrative.

\section{Summary}
In this chapter, we showed that naturally-occurring textual resources can be tailored to build datasets for long-form data-to-text generation, long-form text summarization, and story generation with constraints. For each dataset, we conducted experiments to characterize the challenges in these new datasets. We also proposed new (either automatic or human-evaluation) metrics and models for these tasks to promote research in these directions.
\chapter{Conclusion}

This thesis has made the following contributions:

\begin{itemize}
    \item We improved self-supervised training objectives for large-scale pretrained language models in \cref{CHAPTER:SELFSUPERVISION}. In \cref{sec:sentence-order-prediction}, we replaced the next sentence prediction loss with a novel sentence ordering prediction loss in language model pretraining and showed that the change led to a series of state-of-the-art pretrained encoders. In \cref{sec:incontext-learning}, in contrast to previous work, which finetuned pretrained decoders on human-annotated datasets, we showed that self-supervised tasks with proper designs could also lead to similar gains in the in-context few-shot learning setting, promoting models' ability in cross-task generalization.
    \item We converted various naturally-occurring data structures on Wikipedia into supervision for various NLP tasks in \cref{CHAPTER:WIKIPEDIA}. In \cref{sec:wikipedia-entity-representations}, we leveraged hyperlinks as supervision for pretraining entity representations, leading to models that can encode arbitrary entities. In \cref{sec:wikipedia-discourse-sentence-representations}, we used article structures, such as section and document titles, to train sentence representations. Evaluation results on discourse-related tasks showed that such training helped model performance. In \cref{sec:wikipedia-concept-hierarchies}, we extracted training data from article category graphs and demonstrated that the extracted data improved model performance on textual entailment tasks. These results revealed the advantages of structure-aware model pretraining.
    \item We defined novel tasks that disentangled semantics and syntax and tackled the tasks by designing training objectives and neural architectures in \cref{CHAPTER:DISENTANGLE}. In \cref{section:vgvae-representation}, we built the first neural models to disentangle semantics and syntax in sentence representations. The models use the fact that for a paraphrase pair, the semantics is shared, but syntax varies. In addition to semantic evaluation metrics, we proposed evaluation metrics for syntactic representations, finding that the best performance for both metrics is achieved when there is maximal disentanglement between the two latent representations. In \cref{section:vgvae-generation}, we adapted this framework for controlled paraphrasing, where we seek to control the output text with a syntactic, sentential exemplar. To formally define this controlled generation task, we annotated evaluation sets and proposed evaluation metrics. In a later work, we extended this framework and task setting to machine translation \citep{chen2020exemplar}, showing the potential that this idea could generalize to arbitrary data with the pair data structure.
    \item In \cref{CHAPTER:EVALUATION}, we built challenging datasets from fan-contributed websites. We also proposed evaluation metrics and possible solutions and conducted thorough experiments to characterize the new challenges. In \cref{sec:wikitablet}, we generate arbitrary Wikipedia section text from various tabular data by casting the task as long-form data-to-text generation and creating a large-scale dataset. The task is challenging as models need to generate a coherent passage connecting all the entities in the tabular data, and the story also needs to fit the background knowledge in the tabular data. In \cref{sec:summscreen}, we summarize lengthy transcripts for TV shows. The task has several challenges: e.g., plot information is not stated explicitly but rather only implied in the dialogue and the need to draw information from a wide range of the input transcripts. As characters are fundamental to TV show plots, we also proposed two character-centric evaluation metrics. In \cref{sec:tvstorygen}, we generate long-form stories from character descriptions and summaries. The task poses several challenges for story generation models, including lengthy inputs and outputs and consistency in character modeling.
\end{itemize}

Below we discuss several possible future directions.

\begin{itemize}
    \item  \textbf{Disentangling Latent Factors.} \cref{CHAPTER:DISENTANGLE} introduced neural models for improving interpretability and controllability using implicit yet natural supervision from paraphrases and bilingual text. Future work could generalize this idea to any resources that are formed by data pairs, such as dialogues or summarization. They could be used for disentangling the factors that are shared between pairs of inputs and those that are not shared, such as intentions and the personalized styles in dialogues, sentence-level fluency and document-level discourse in sentence modeling, or important events and irrelevant details in summarization.
    
    Another possibility is to disentangle task supervision when a task can be decomposed into two sub-tasks, e.g., cross-lingual summarization can be thought of as a combined task of summarization and translation. Disentangling task supervision could help us improve the models' ability in cross-task generalization and tease out the valuable intermediate supervision that is usually unavailable.
    
    In general, disentangling latent factors is an appealing research direction in that, although large pretrained models have yielded superhuman performance, researchers still lack an understanding of the behaviors of these models. Outside this thesis, I have also completed work benefiting from interpretable latent variables, leading to efficient neural models \citep{chen-gimpel-2018-smaller} and effective semi-supervised learning \citep{chen-etal-2018-variational}.
    Better interpretability could also help us improve their robustness and worst-case behaviors to be better applied in user-facing applications.

    \item \textbf{Natural Supervision for Text Generation.} \cref{CHAPTER:WIKIPEDIA} presented approaches to leverage various natural supervision for representation learning. For future work, it would be interesting to see whether we could apply the same thing to text generations. In particular, future work may consider using hyperlinks to improve the entity tracking performance in their text generation systems and using article structures to enhance the discourse coherence in the generated texts.

    \item \textbf{Unified Models for Various Language Supervision.} While \cref{CHAPTER:WIKIPEDIA} described modeling choices to take different language knowledge (e.g., entities and discourse) into account, it is still unclear as to the best design for a unified model that can incorporate all of these learning signals. Future work may find a unified model to show superior performance as humans rely on multiple language properties simultaneously to solve tasks. In addition, future work may consider combining discourse, linking, and paraphrase objectives with BERT-like models, as well as other types of natural supervision, such as naturally-occurring bold/italics/underlining annotations in web text, and long-distance discourse cues like two paragraphs in two chapters, among others.
    
    \item \textbf{Learning Commonsense Knowledge from Natural Supervision.} Future work could also consider learning commonsense knowledge from naturally occurring data. For example, learning domain-specific commonsense from dialogues in certain subreddits\footnote{Subreddits are on the social media website Reddit and dedicated to a particular topic that people write about.} (e.g., technical or social) or distilling commonsense knowledge from existing pretrained models. Commonsense understanding is ubiquitous in language. As humans often assume the knowledge is well-known to anything they interact with, it is seldom explicitly described. Also, due to the assumption, humans tend to believe any intelligent system should understand such knowledge. The two properties of commonsense knowledge make it a challenging and imperative capability. In practice, when deploying models in real-life applications, the knowledge can also make them more reliable due to the improved language understanding.

    \item \textbf{Text Generation with Rich Descriptions.} Future work could explore text generation with rich, detailed descriptions about the world in which the task is situated. This direction is related to the work in \cref{CHAPTER:EVALUATION} as the descriptions can be either tabular data about certain background knowledge (\cref{sec:wikitablet}) or lengthy documents about fictional characters (\cref{sec:tvstorygen}). These detailed descriptions explicitly describe knowledge that the generated text should follow, so they are like ``controlled environments'' that simulate the real world, offering opportunities to improve evaluations for text generations and enhance the faithfulness of neural models.

\end{itemize}

\begin{appendices}
\crefalias{section}{appendix}

\chapter{Appendix to Chapter \ref{CHAPTER:SELFSUPERVISION}}

\section{Adding a Backward Language Modeling Loss to GPT-Style Pretraining}\label{appendix-sec:bidir}

Previous work \citep{NEURIPS2019_c20bb2d9} combines various pretraining losses, including the forward and backward language modeling losses and the masked/prefix language modeling loss. However, little work has been done to identify whether the forward and backward language modeling losses are sufficient for these pretrained models to achieve strong performance in downstream tasks. This idea is in part inspired by Electra \citep{Clark2020ELECTRA:} where they found having training losses at every position helps the training efficiency and we want to see whether concatenating forward and backward language modeling losses can achieve similar outcome. In addition, if we could achieve ELMo-style bidirectionality by simply manipulating the attention masks without introducing additional parameters, it could allow us to adapt the pretrained GPT checkpoints to be bidirectional by finetuning with the backward language modeling. The models could also have advantages when used for computing the probabilities of text to re-rank outputs due to its approach of achieving bidirectionality.

In experiments, we train BERT models with 128 hidden dimensions on Wikipedia using both forward and backward language modeling losses (BiDir). For baselines, we consider a BERT model trained with forward only language modeling loss (UniDir) and a BERT model trained with the masked language modeling loss (MLM). We also consider Electra and XLNet \citep{NEURIPS2019_dc6a7e65} of similar sizes to compare different ways to achieve bidirectionality. In particular, Electra uses a discriminative objective and XLNet uses permutation language modeling. For all these experiments, we train the models from scratch and use a batch size of 256, a learning rate of 1e-4, and 300k training steps. For evaluating downstream performance, we perform the BERT-style evaluation on the GLUE benchmark and finetune the whole model. We note that when evaluating Unidir on GLUE tasks, we use the last embedding. For BiDir, we concatenate the last and first embeddings.

\begin{table}
    \centering\small
\begin{tabular}{|l|ccccccccc|}\hline
& cola & mnli & mrpc & qnli & qqp & rte & sst2 & stsb & avg \\\hline
Electra & 51.8 & 79.1 & 90.9 & 86.1 & 84.5 & 63.9 & 87.5 & 84.6 & 78.6 \\
XLNet & 24.7 & 75.7 & 87.5 & 80.9 & 84.0 & 56.7 & 86.1 & 81.7 & 72.2 \\\hline
\multicolumn{10}{c}{BERT Models} \\\hline
MLM & 35.0 & 76.4 & 85.2 & 83.9 & 83.6 & 57.4 & 85.4 & 82.0 & 73.6 \\
UniDir & 25.8 & 76.5 & 81.5 & 81.4 & 82.3 & 57.4 & 86.9 & 66.0 & 69.7 \\
BiDir & 25.3 & 75.3 & 81.6 & 80.9 & 82.4 & 58.5 & 86.5 & 75.7 & 70.8 \\\hline
\end{tabular}
    \caption{GLUE dev set results. MLM, UniDir, and BiDir represents baselines that use different training objectives.}
    \label{appendix-tab:bidir-compare}
\end{table}

We report results in \cref{appendix-tab:bidir-compare}. We find that Electra shows best performance likely due to its efficient training objective. XLNet shows slightly worse performance than the MLM-trained BERT model. Interestingly, BiDir achieves better performance than UniDir on average, but it performs worse than MLM, likely due to the fact that while compared to UniDir, BiDir has an extra training signal (i.e., backward language modeling loss), it still falls behind the MLM as the models never use full attention and full attention is a superior choice than causal attention for GLUE tasks.

\begin{table}
    \centering\small
\begin{tabular}{|lccccccccc|}\hline
& cola & mnli & mrpc & qnli & qqp & rte & sst2 & stsb & avg \\\hline
BiDir & 25.3 & 75.3 & 81.6 & 80.9 & 82.4 & 58.5 & 86.5 & 75.7 & 70.8 \\
BiDir full attention & 21.5 & 69.8 & 82.2 & 74.8 & 77.9 & 55.2 & 84.2 & 40.3 & 63.2 \\\hline
UniDir & 25.8 & 76.5 & 81.5 & 81.4 & 82.3 & 57.4 & 86.9 & 66.0 & 69.7 \\
UniDir full attention & 14.9 & 69.9 & 80.9 & 67.7 & 78.1 & 56.3 & 83.4 & 41.3 & 61.6 \\\hline
\end{tabular}
    \caption{GLUE dev set results. We compare the use of full attention to causal attention in GLUE evaluation.}
    \label{tab:compare_full_causal_attention}
\end{table}

We investigate the effect of using full attention on UniDir and BiDir in \cref{tab:compare_full_causal_attention}. Generally using different attention patterns during test time leads to much worse performance. However, it is interesting to see that BiDir models still maintain the improvement with the full attention pattern, showing that the models learn to better leverage information from both sides than UniDir.

\section{Details about Downstream Evaluation Tasks for ALBERT}
\label{albert-appendix:downstream_detailed_description}
\paragraph{GLUE.} GLUE is comprised of 9 tasks, namely Corpus of Linguistic Acceptability (CoLA;~\citealp{warstadt-etal-2019-neural}), Stanford Sentiment Treebank (SST;~\citealp{socher-etal-2013-recursive}), Microsoft Research Paraphrase Corpus
(MRPC;~\citealp{dolan-brockett-2005-automatically}), Semantic Textual Similarity Benchmark (STS;~\citealp{cer-etal-2017-semeval}),
Quora Question Pairs (QQP;~\citealp{qqp2016url}), Multi-Genre NLI (MNLI;~\citealp{williams-etal-2018-broad}), Question NLI (QNLI;~\citealp{rajpurkar-etal-2016-squad}), Recognizing Textual
Entailment (RTE;~\citealp{dagan2005pascal,bar2006second,giampiccolo-etal-2007-third,bentivogli2009fifth}) and
Winograd NLI (WNLI;~\citealp{10.5555/3031843.3031909}). It focuses on evaluating model capabilities for natural language understanding. 
When reporting MNLI results, we only report the ``match'' condition (MNLI-m). We follow the finetuning procedures from prior work \citep{devlin-etal-2019-bert,liu2019roberta,NEURIPS2019_dc6a7e65} and report the held-out test set performance obtained from GLUE submissions. For test set submissions, we perform task-specific modifications for WNLI and QNLI as described by \citet{liu2019roberta} and \citet{NEURIPS2019_dc6a7e65}. 

\paragraph{\squad.} \squad is an extractive question answering dataset built from Wikipedia. The answers are segments from the context paragraphs and the task is to predict answer spans. We evaluate our models on two versions of SQuAD: v1.1 and v2.0. \squad v1.1 has 100,000 human-annotated question/answer pairs. \squad v2.0 additionally introduced 50,000 unanswerable questions. For \squad v1.1, we use the same training procedure as \bert, whereas for \squad v2.0, models are jointly trained with a span extraction loss and an additional classifier for predicting answerability~\citep{NEURIPS2019_dc6a7e65,liu2019roberta}. We report both development set and test set performance.

\paragraph{RACE.} RACE is a large-scale dataset for multi-choice reading comprehension, collected from English examinations in China with nearly 100,000 questions. Each instance in RACE has 4 candidate answers. Following prior work~\citep{NEURIPS2019_dc6a7e65,liu2019roberta}, we use the concatenation of the passage, question, and each candidate answer as the input to models. Then, we use the representations from the \texttt{[CLS]} token for predicting the probability of each answer. The dataset consists of two domains: middle school and high school. We train our models on both domains and report accuracies on both the development set and test set.

\section{Additional Details for \lpp and Classification Tasks}
\label{incontext-appendix:lpp_function_words}

The label strings we used for LPP are as follows: Yes and No, Y and N, True and False, and T and F. We randomly choose from Yes, Y, True, and T as the label string for the positive label and use the other one in the selected pair as the negative label.

The label strings we used for the binary classification task are the same as the classification \lpp task. For the three-way classification task, we use the following label strings: Positive and Negative and Neutral, True and False and Neither, T and F and N, Yes and No and Unknown, Y and N and U.

\paragraph{List of Function Words for \lpp.} We used the following function words for identifying the last phrase: the, a, an, for, including, and, in, is, are, were, was, neither, or, nor, be, at, in, on, by, to, would, will, before, after, of, about, from, excluding, except, during, under, above, then, into, onto, should, shall, must, may, might, than, with, using, can, could, about, as, from, within, without, have, had, been.

\section{Dataset Statistics for SuperGLUE and Natural-Instructions}
\label{incontext-appendix:dataset_statistics}
\begin{table}
    \centering\small
    \begin{tabular}{|l|l|l|l|l|l|}\hline
        \bf Dataset & \bf Task Category & \bf Metrics & \bf\#Train & \bf\#Test & \bf\#Class \\\hline
        BoolQ &  Question Answering & Accuracy & 9427 & 3270 & 2 \\\hline
        MultiRC & Question Answering & F1$_a$/EM & 5100 & 953 & 2 \\\hline
        COPA & Question Answering & Accuracy & 400 & 100 & 2 \\\hline
        RTE & Natural Language Inference & Accuracy & 2500 & 278 & 2 \\\hline
        CB & Natural Language Inference & Accuracy/F1 & 250 & 57 & 3 \\\hline
    \end{tabular}
    \caption{Dataset statistics for SuperGLUE. We use the official development sets as test sets.}
    \label{incontext-appendix-tab:superglue_dataset_stat}
\end{table}

\begin{table*}
    \centering\small
    \begin{tabular}{|l|l|l|l|}\hline
        \bf Dataset & \bf Task Category & \bf \#Train & \bf\#Test \\\hline
        subtask003\_mctaco\_question\_generation\_event\_duration & Question Generation & 330 & 100  \\\hline
        subtask040\_qasc\_question\_generation & Question Generation & 6400 & 100 \\\hline
        subtask002\_quoref\_answer\_generation & Answer Generation & 6400 & 100 \\\hline
        subtask033\_winogrande\_answer\_generation & Answer Generation & 6400 & 100 \\\hline
        subtask034\_winogrande\_question\_modification\_object & Minimal Modification & 6400 & 100 \\\hline
        subtask045\_miscellaneous\_sentence\_paraphrasing & Minimal Modification & 93 & 100 \\\hline
        subtask039\_qasc\_find\_overlapping\_words & Verification & 6400 & 100 \\\hline
        subtask044\_essential\_terms\_identifying\_essential\_words & Verification & 2138 & 100 \\
        \hline
    \end{tabular}
    \caption{Dataset statistics for Natural-Instructions.}
    \label{incontext-appendix-tab:naturalinstructions_dataset_stat}
\end{table*}

We report dataset statistics for SuperGLUE and Natural-Instructions in \cref{incontext-appendix-tab:superglue_dataset_stat} and \cref{incontext-appendix-tab:naturalinstructions_dataset_stat}, respectively.

\subsection{More Details about Natural-Instructions}

\paragraph{Dataset Sources.}  CosmosQA \citep{huang-etal-2019-cosmos}, DROP \citep{dua-etal-2019-drop}, EssentialTerms \citep{khashabi-etal-2017-learning}, MCTACO \citep{zhou-etal-2019-going}, MultiRC \citep{khashabi-etal-2018-looking},
QASC \citep{Khot2020QASCAD}, Quoref \citep{dasigi-etal-2019-quoref}, ROPES \citep{lee-etal-2021-rope} and Winogrande \citep{Sakaguchi2020WINOGRANDEAA}.

\paragraph{Training Datasets.}
We used the following 8 datasets when training models in the cross-task setting: subtask026\_drop\_question\_generation, subtask060\_ropes\_question\_generation, subtask028\_drop\_answer\_generation, subtask047\_misc\_answering\_science\_questions, subtask061\_ropes\_answer\_generation, subtask059\_ropes\_story\_generation, subtask027\_drop\_answer\_type\_generation, subtask046\_miscellaenous\_question\_typing.

\section{Templates for SuperGLUE}
\label{incontext-appendix:superglue_templates}
\begin{table}
    \centering\small
    \begin{subtable}{1\textwidth}
    \centering
    \begin{tabular}{|l|p{0.8\textwidth}|}\hline
        \bf GPT3 & \$\{Context\}$\langle$newline$\rangle$ question: \$\{Question\}$\langle$newline$\rangle$answer:\bf\textcolor{red}{\$\{Answer\}} \\\hline
        \bf Ours & Input: \$\{Context\} question: \$\{Question\} answer: True$\langle$newline$\rangle$Output: \bf\textcolor{red}{\$\{Answer\}} \\ \hline
    \end{tabular}
    \caption{BoolQ Template.}
    \end{subtable}\vspace{1em}
    \begin{subtable}{1\textwidth}
    \centering
    \begin{tabular}{|l|p{0.8\textwidth}|}\hline
        \bf GPT3 & \$\{Context\}$\langle$newline$\rangle$ question: \$\{Question\} True or False?$\langle$newline$\rangle$answer:\bf\textcolor{red}{\$\{Answer\}} \\\hline
        \bf Ours & Input: \$\{Context\} question: \$\{Question\} answer: True$\langle$newline$\rangle$Output: \bf\textcolor{red}{\$\{Answer\}} \\ \hline
    \end{tabular}
    \caption{RTE Template.}
    \end{subtable}\vspace{1em}
    \begin{subtable}{1\textwidth}
    \centering
    \begin{tabular}{|l|p{0.8\textwidth}|}\hline
        \bf GPT3 & \$\{Context\}$\langle$newline$\rangle$\bf\textcolor{red}{\$\{Answer\}} \\\hline
        \bf Ours & Input: \$\{Context\}$\langle$newline$\rangle$Output:\bf\textcolor{red}{\$\{Answer\}} \\ \hline
    \end{tabular}
    \caption{COPA Template.}
    \end{subtable}\vspace{1em}
    \begin{subtable}{1\textwidth}
    \centering
    \begin{tabular}{|l|p{0.8\textwidth}|}\hline
        \bf GPT3 & \$\{Context\}$\langle$newline$\rangle$ question: \$\{Question\} true, false, or neither?$\langle$newline$\rangle$answer:\bf\textcolor{red}{\$\{Answer\}} \\\hline
        \bf Ours & Input: \$\{Context\} question: \$\{Question\} true, false, or neither?$\langle$newline$\rangle$Output: \bf\textcolor{red}{\$\{Answer\}} \\ \hline
    \end{tabular}
    \caption{CB Template.}
    \end{subtable}
    \caption{Evaluation templates for SuperGLUE. \$\{$\cdot$\} represents values drawn from a particular data field. We alter the GPT3 templates for these tasks to share similar formats with one of our self-supervised tasks. The red, boldfaced texts are used to compute the language modeling perplexities for ranking the labels. We note that the shown templates are for a single example, and there could be multiple examples within an instance.}
    \label{incontext-appendix-tab:superglue_template}
\end{table}

We show the SuperGLUE templates in \cref{incontext-appendix-tab:superglue_template}.

\chapter{Appendix to Chapter \ref{CHAPTER:EVALUATION}}

\section{Details about Dataset Construction for \wikitablet}
\label{wikitablet-appendix-sec:dataset-construction}

When collecting data, we consider five resources: \wikidata tables, infoboxes in Wikipedia pages, hyperlinks in the passage, named entities in the passage obtained from named entity recognition (NER), and Wikipedia article structure.
For each article in Wikipedia, we use the same infobox and \wikidata table for all sections. These tables can serve as background knowledge for the article.
For each section in the article, we create a second table corresponding to section-specific data, i.e., section data. 
The section data contains records constructed from hyperlinks and entities identified by a named entity recognizer. Section data contributes around 25\% of the records in \wikitablet.

We filter out several entity types related to numbers\footnote{List of filtered entity types: PERCENT, TIME, QUANTITY, ORDINAL, CARDINAL.} as the specific meanings of these numbers in the section of interest are difficult to recover from the information in the tables. After filtering, we use the identified entities as the values and the entity types as the attributes. This contributes roughly 12\% of the records in our final dataset.

We also create records from hyperlinks in the section of interest. We first expand the hyperlinks available for each section with hyperlinks available in the parent categories. %
We first group hyperlinks across all Wikipedia articles with those same categories, and then we perform string matching between these hyperlinks and the text in the section. If there are exact matches, we will include those hyperlinks as part of the hyperlinks in this section. 

Details for constructing a record with attribute $a$ and value $v$ for a hyperlink with surface text $t$ and hyperlinked article $\ell$ are as follows. To set $a$, we use the value of the ``instance of'' or ``subclass of'' tuple in the \wikidata table for $\ell$. If $\ell$ does not have a \wikidata table or no appropriate tuple, we consider the parent categories of $\ell$ as candidates for $a$. If there are multiple candidates for $a$, we first embed these candidates and $a$ using GloVe embeddings and then choose the one that maximizes cosine similarity between the document titles or section titles and the candidates for $a$.  
For the value $v$ of the tuple, we use the document title of $\ell$ rather than the actual surface text $t$ to avoid giving away too much information in the reference text.  The records formed by hyperlinks contribute approximately 13\% of the records in \wikitablet. 

We shuffle the ordering of the records from NER and the hyperlinks to prevent models from relying on the ordering of records in the reference text.

The records from the section data can be seen as section-specific information that can make the task more solvable. Complementary to the article data, we create a title table that provides information about the position in which the section is situated, which includes the article title and the section titles for the target section. As the initial sections in Wikipedia articles do not have section titles, we use the section title ``Introduction'' for these.\footnote{Among millions of section titles in Wikipedia, there are only 4672 sections, including nested sections, that are called ``Introduction''. Therefore, we believe this process will not introduce much noise into the dataset.} 

As the records in our data tables come from different resources, we perform extra filtering to remove duplicates in the records. In particular, we give \wikidata the highest priority as it is a human-annotated well-structured data resource (infoboxes are human-annotated but not well-structured due to the way they are stored on Wikipedia) and the entities from NER the lowest priority as they are automatically constructed. That is, when we identify duplicates across different resources, we will keep the records from the higher priority resource and drop those from the lower one. More specifically, the duplicates between \wikidata records and infoboxes are determined by whether there are duplicate values or duplicate attributes: for hyperlinks and infoboxes or \wikidata, they are judged by duplicate values; for NER and hyperlinks, they are based on whether there is any token overlapping between values.

After table collection, we have the following criteria for filtering out the texts: (1) we limit the text length to be between 50 and 1000 word tokens; (2) to ensure that there is sufficient information in the table, we only keep data-text pairs that contain more than 2 records per sentence and more than 15 records per 100 tokens from \wikidata and infoboxes; (3) to avoid texts such as lists of hyperlinks, we filter out texts where more than 50\% of their word tokens are from hyperlink texts.

\section{Human Evaluation for \wikitablet}
\label{wikitablet-appendix-sec:human-eval}

\begin{table}[t]
    \centering\small
    \begin{tabular}{|p{0.45\textwidth}|}\hline
        1 = it is completely ungrammatical, as it is impossible to understand the text. \\\hline
        2 = it has many grammatical errors, and these errors make the text very difficult to understand. \\\hline
        3 = it has grammatical errors, and some of them make part of the text difficult to understand. \\\hline
        4 = it has some grammatical errors, but they are minor errors that do not affect reading. \\\hline
        5 = it is completely grammatical, as it does not have any grammatical errors.\\\hline
    \end{tabular}
    \caption{Rating explanations %
    for grammaticality.}
    \label{wikitablet-appendix-tab:human_evaluation_options_grammar}
\end{table}

\begin{table}[t]
    \centering\small
    \begin{tabular}{|p{0.45\textwidth}|}\hline
        1 = it is completely incoherent, as it is impossible to piece together information in the text. \\\hline
        2 = it is incoherent in most places. You can only understand part of the story. \\\hline
        3 = it is incoherent in many places, but if you spend time reading it, you still can understand the whole story. \\\hline
        4 = it is mostly coherent. Although the text is incoherent in some places, it does not affect reading. \\\hline
        5 = it is completely coherent.\\\hline
    \end{tabular}
    \caption{Rating explanations for coherence.} 
    \label{wikitablet-appendix-tab:human_evaluation_options_coherence}
\end{table}

\begin{table}[t]
    \centering\small
    \begin{tabular}{|p{0.45\textwidth}|}\hline
        1 = it is completely contradictory to what is described in the table. \\\hline
        2 = it has some facts contradictory to what is described in the table. \\\hline
        3 = it is not supported by the table, and it does not contradict the table. \\\hline
        4 = some of the text is supported by the facts in the table, and the rest of it does not contradict the facts in the table. \\\hline
        5 = it is completely supported by the table.\\\hline
    \end{tabular}
    \caption{Rating explanations for faithfulness.}
    \label{wikitablet-appendix-tab:human_evaluation_options_faithful}
\end{table}

\begin{table}[t]
    \centering\small
    \begin{tabular}{|p{0.45\textwidth}|}\hline
        1 = the text is completely irrelevant to the reference. \\\hline
        2 = most of the text is irrelevant to the reference. \\\hline
        3 = some of the text is relevant to the reference. \\\hline
        4 = most of the text is relevant to the reference. \\\hline
        5 = the text is talking about the same thing as the reference.\\\hline
    \end{tabular}
    \caption{Rating explanations for relevance.}
    \label{wikitablet-appendix-tab:human_evaluation_options_relevance}
\end{table}

\begin{table}[t]
    \centering\small
    \begin{tabular}{|p{0.45\textwidth}|}\hline
        1 = it has quite a few facts contradictory to what is described in the reference. \\\hline
        2 = it has some facts contradictory to what is described in the reference. \\\hline
        3 = it is not supported by the reference, and it does not contradict the reference. \\\hline
        4 = some of the text is supported by the facts in the reference, and the rest of it does not contradict the reference. \\\hline
        5 = it is completely supported by the reference.\\\hline
    \end{tabular}
    \caption{Rating explanations for supportedness.}
    \label{wikitablet-appendix-tab:human_evaluation_options_support}
\end{table}

The selected topics for human evaluations are: human (excluding the introduction and biography section), film, single (song), song, album, television series. When evaluating grammaticality and coherence, only the generated text is shown to annotators. The question for grammaticality is ``On a scale of 1-5, how much do you think the text is grammatical? (Note: repetitions are grammatical errors.)'' (option explanations are shown in \cref{wikitablet-appendix-tab:human_evaluation_options_grammar}), and the question for coherence is ``On a scale of 1-5, how much do you think the text is coherent? (Coherence: Does the text make sense internally, avoid self-contradiction, and use a logical ordering of information?)'' (rating explanations are in \cref{wikitablet-appendix-tab:human_evaluation_options_coherence}).

When evaluating faithfulness, we show annotators the article data and the generation. The question is ``On a scale of 1-5, how much do you think the text is supported by the facts in the following table?'' (rating explanations are in \cref{wikitablet-appendix-tab:human_evaluation_options_faithful}).

When evaluating coherence and relevance, annotators were shown the reference text and the generation, as well as the Wikipedia article title and section titles for ease of understanding the texts. Annotators were asked two questions, with one being ``On a scale of 1-5, how much do you think the text is relevant to the reference'' (\cref{wikitablet-appendix-tab:human_evaluation_options_relevance}), and the other being ``On a scale of 1-5, how much do you think the text is supported by the facts in the reference?'' (\cref{wikitablet-appendix-tab:human_evaluation_options_support}).

\subsection{Details about Dataset Construction for \tvrecap}

\paragraph{String Matching Algorithm.} For example, for the character name ``John Doe'', valid mentions are itself, ``John'', ``J.D.'', and ``JD'' due to the writing style on \tvmegasite. While this matching algorithm may lead to extra characters aligned to particular episodes, it at least includes all characters that are actually involved in the episode.

\paragraph{Episode Filtering Criteria.} We filter out episodes if (1) an episode contains fewer than 3 characters (to avoid stories that do not involve many character interactions); or (2) the detailed recap has fewer than 200 word tokens (ensuring that stories have enough details); or (3) the brief summary has fewer than 20 word tokens (to ensure that there is sufficient information given as the input).

\section{Details about Semantic Role Labeling for \tvrecap}
\label{tvrecap-appendix-sec:srl}

We use a pretrained model from \citet{shi2019simple} to generate SRL tags of the detailed episode recaps. We eliminate the SRL tags for sentences that do not contain $\langle\text{ARG0}\rangle$ or only contain pronouns to avoid ambiguity. For each sentence, we also only keep the SRL tags that correspond to the first verb that appears in the sentence to avoid the SRL tags being too specific, so there will be a balanced burden between the text-to-plot model and the plot-to-text model. In addition, following \citet{goldfarb-tarrant-etal-2020-content}, we discard SRL tags of generic verbs.

The list of verbs we discard is as follows: ``is'', ``was'', ``were'', ``are'', ``be'', ``'s'', ``'re'', ``'ll'', ``can'', ``could'', ``must'', ``may'', ``have to'', ``has to'', ``had to'', ``will'', ``would'', ``has'', ``have'', ``had'', ``do'', ``does'', ``did''.

We also eliminate arguments that are longer than 5 tokens.

\section{Details about Decoding Algorithms and Hyperparameters for \tvrecap}
\paragraph{Decoding}
For both story generation and summarization, we use a batch size of 200, beam search of size 5 with n-gram blocking \citep{paulus2018a} where probabilities of repeated trigrams are set to 0 during beam search. We did not find nucleus sampling \citep{Holtzman2020The} leading to better generation quality (i.e., fluency and faithfulness to the summaries and the recaps) than beam search with n-gram blocking, possibly due to the fact that our models generate at the sentence level.

\paragraph{Hyperparamters}
Because our plot-to-text models work at the sentence level, leading to many training instances for both \fd and \tms (i.e., 0.5 million and 1.5 million sentences respectively), we train these plot-to-text models for 10 epochs without early stopping. During generation, we set the minimum number of decoding steps to 24 and the maximum number of decoding steps to 48.

As for summarization, we benchmark pretrained BART-base, BART-large \citep{lewis-etal-2020-bart}, and Pegasus \citep{pegasus-zhang20ae}. As the average length of the detailed recaps is much longer than the default maximum sequence length of these pretrained models, we extend the maximum sequence length to 4096. When doing so, we randomly initialize new positional embeddings for BART. Since Pegasus uses Sinusoidal positional embeddings, we simply change the default value of maximum sequence length. We train the models for 15 epochs and perform early stopping on the dev set perplexities. During generation, we limit the minimum decoding step to be 50 and 300, and the maximum decoding step to be 100 and 600 for \fd and \tms respectively. The minimum decoding steps roughly match the average length of the summaries in \fd and \tms.

\end{appendices}

\bibliographystyle{acl_natbib}

\setlength{\bibsep}{2pt}
\bibliography{custom,anthology}
\cleardoublepage
\restoregeometry

\end{document}